\pdfoutput=1
\documentclass[10pt,twocolumn,letterpaper]{article}
\usepackage{pifont}
\usepackage[table]{xcolor}

\PassOptionsToPackage{table}{xcolor}
\IfFileExists{cvpr.sty}{
  \usepackage[pagenumbers]{cvpr}
  \def\confName{CVPR}
  \def\confYear{2027}
  \def\paperID{*****}
}{
  \usepackage[margin=0.75in]{geometry}
  \usepackage{amsmath,amssymb,graphicx,booktabs,xcolor,enumitem}
}

\usepackage{algorithm}
\usepackage{algpseudocode}
\usepackage{pifont}
\usepackage[T1]{fontenc}
\usepackage{upquote}
\usepackage{mathtools}
\usepackage{fancyvrb}
\usepackage{longtable}
\usepackage{array}
\usepackage{float}
\makeatletter
\newcommand{\supplementcontents}{\begingroup
  \def\numberline##1\hskip-1em.~{\hb@xt@\@tempdima{##1\hfil}}%
  \@starttoc{toc}\endgroup}
\makeatother
\usepackage[pagebackref,breaklinks=true,colorlinks,citecolor=blue!60!black,linkcolor=blue!60!black,urlcolor=blue!60!black]{hyperref}
\hypersetup{pdftitle={Back to Basics: A Skill-Centric Approach to Generalist Web Agents},
  pdfauthor={Minh Pham-Dinh, Md Farhan Ishmam, Hernan Yupanqui, Reaz Hassan Joarder, Kenneth Marino}}
\makeatletter\g@addto@macro\bibfont{\raggedright}\makeatother

\graphicspath{{figures/}}

\newcommand{\miniweb}{\mbox{\textup{\textsc{\textbf{MiniWeb}}}}}
\newcommand{\miniwebgen}{\mbox{\textup{\textsc{\textbf{MiniWebGen}}}}}
\newcommand{\miniwebagent}{\mbox{\textup{\textsc{\textbf{MiniWebAgent}}}}}
\newcommand{\logolink}[3]{\href{#2}{\makebox[1.35em][l]{\raisebox{-0.2em}{\includegraphics[height=1.1em]{figures/logos/#1}}}\textcolor{blue!60!black}{\texttt{#3}}}}
\newcommand{\websitelink}{\logolink{globe-solid-full.pdf}{https://minhphd.github.io/miniweb/}{Website}}
\newcommand{\codelink}{\logolink{github.pdf}{https://github.com/sparklabutah/MiniWeb}{Code}}
\newcommand{\datalink}{\logolink{huggingface.pdf}{https://huggingface.co/datasets/miniweb/MiniWeb}{Dataset}}

\definecolor{famdrag}{HTML}{E27098}
\definecolor{famsel}{HTML}{60A5E2}
\definecolor{famtext}{HTML}{6DBB7C}
\definecolor{famform}{HTML}{EF8B55}
\definecolor{famnav}{HTML}{58B7C6}
\definecolor{famio}{HTML}{9B7FEF}
\definecolor{famreas}{HTML}{C8A44D}
\newcommand{\famhl}[2]{{\setlength{\fboxsep}{1.1pt}\colorbox{#1!22}{\textbf{#2}\vphantom{Ag}}}}
\newcommand{\famdrag}[1]{\famhl{famdrag}{#1}}
\newcommand{\famsel}[1]{\famhl{famsel}{#1}}
\newcommand{\famtext}[1]{\famhl{famtext}{#1}}
\newcommand{\famform}[1]{\famhl{famform}{#1}}
\newcommand{\famnav}[1]{\famhl{famnav}{#1}}
\newcommand{\famio}[1]{\famhl{famio}{#1}}
\newcommand{\famreas}[1]{\famhl{famreas}{#1}}

\definecolor{mwhead}{HTML}{BE0029}
\definecolor{mwrule}{HTML}{BE0029}
\definecolor{mwmech}{HTML}{EEF3FA}
\definecolor{mwreas}{HTML}{FAEEEE}
\definecolor{miniwebblue}{HTML}{D9ECFF}
\definecolor{atomgenorange}{HTML}{FFE4C7}
\definecolor{atomagentgreen}{HTML}{DDF5E7}

\newcounter{defn}
\newenvironment{macrodefn}[1]{%
  \begin{quote}\small
  \refstepcounter{defn}%
  \noindent\textbf{Definition \thedefn}\ (\textbf{#1})\textbf{.}\ \itshape}%
  {\end{quote}}

\newenvironment{promptbox}[1]{\VerbatimEnvironment\par\addvspace{\baselineskip}\begin{Verbatim}[frame=single,framesep=2mm,fontsize=\scriptsize,
  rulecolor=\color{black!40},label={\small\textbf{#1}}]}{\end{Verbatim}}
\newenvironment{definition}[2]{\begin{macrodefn}{#1}}{\end{macrodefn}}

\title{Back to Basics: A Skill-Centric Approach to Generalist Web Agents}

\author{%
Minh Pham-Dinh$^1$, Md Farhan Ishmam$^1$, Hernan Yupanqui$^1$,
\\Reaz Hassan Joarder$^2$, Kenneth Marino$^1$\\
$^1$University of Utah\quad$^2$Islamic University of Technology \\
{\tt\small \{\textbf{m.pham-dinh},farhan.ishmam,kenneth.marino\}@utah.edu}\\[8pt]
{\small\websitelink\qquad\codelink\qquad\datalink}
}

\begin{document}
\addtocontents{toc}{\protect\setcounter{tocdepth}{-1}}
\maketitle

\begin{abstract}

Browser agents do not just need more practice. They need to practice the right things. We introduce \miniweb{}, a realistic miniature of the web comprising 65 interconnected websites across 15 domains, where agents navigate, search, and transact as people do. Its 369 human-annotated tasks cover everyday workflows and label 37 reusable interaction primitives grouped into 7 skill families, allowing us to diagnose which skills fail rather than only whether a task succeeds. Benchmarking ten open-weight agents, we find that drag \& gesture is consistently the weakest family, with no agent solving more than 24\% of its instances. To address this gap, we introduce \miniwebgen{}, a framework that autonomously generates verified, skill-centric training data from \miniweb{}. Fine-tuning \mbox{Qwen3.5-4B} on just 3.1K verified trajectories improves single-primitive success by 24.9pp across 13 unseen websites, with gains across every skill family. However, these trajectories teach short-horizon interactions, which do not directly translate to full-task success. We build on this result with \miniwebagent{}, which pairs the trained executor with a planner that decomposes full tasks into the skill-level steps it learned to perform. This approach improves task success by 19.4pp on unseen \miniweb{} sites and transfers to external benchmarks, with gains of 9.3pp on WebArena-Lite, 16.1pp on Online-Mind2Web, and 20.0pp on WebVoyager, without training on their trajectories. We provide a simple recipe for compact browser agents: practice reusable skills directly, then compose them to solve unfamiliar tasks.

\end{abstract}

\begin{figure}[!t]
    \centering
    \includegraphics[width=0.95\linewidth]{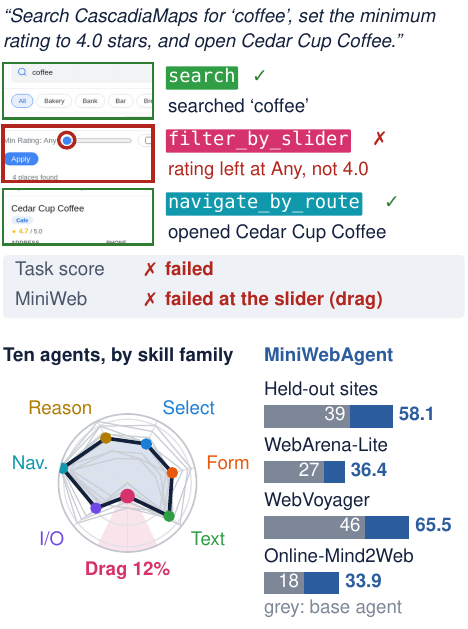}
    \caption{\textbf{\miniweb{} grades every primitive of a task.} Top: \mbox{Fara1.5-4B} opens the right caf\'e but leaves the rating filter at Any. Bottom left: Overall skill profile of 10 evaluated agents. Bottom right: task success of \mbox{Qwen3.5-4B} before and after our training. }
    \label{fig:teaser}
\end{figure}

\section{Introduction}
What makes humans generalists? We rarely learn a complex behavior from scratch. Before learning to ride a bike, we have already spent years learning to balance, coordinate our limbs, recover from mistakes, and navigate through space. New tasks become easier because they reuse skills we have practiced in many other settings. Generalization comes not only from seeing more tasks, but from building a repertoire of reusable skills and learning how to compose them.

Web agents should be able to learn in the same way.
Current approaches to training general web agents focus on collecting more task trajectories on diverse websites, difficulty, and solution strategies~\cite{Awadallah2025Fara7BAE, Awadallah2026Fara15SL, Wu2026AutoWebWorldSI, Gupta2026MolmoWebOV}. While large models can benefit from this scale, compact open-weight agents have limited capacity and training budgets, making the choice of what to practice especially important~\cite{pmlr-v70-andreas17a}. We therefore ask: \emph{should a small web agent spend its training budget on more complete tasks, or on the reusable interaction skills those tasks are built from?} We study the latter, training on short-horizon, skill-centric trajectories and composing the learned skills at inference time.

\begin{figure*}[t]
    \centering
    \includegraphics[width=0.95\linewidth]{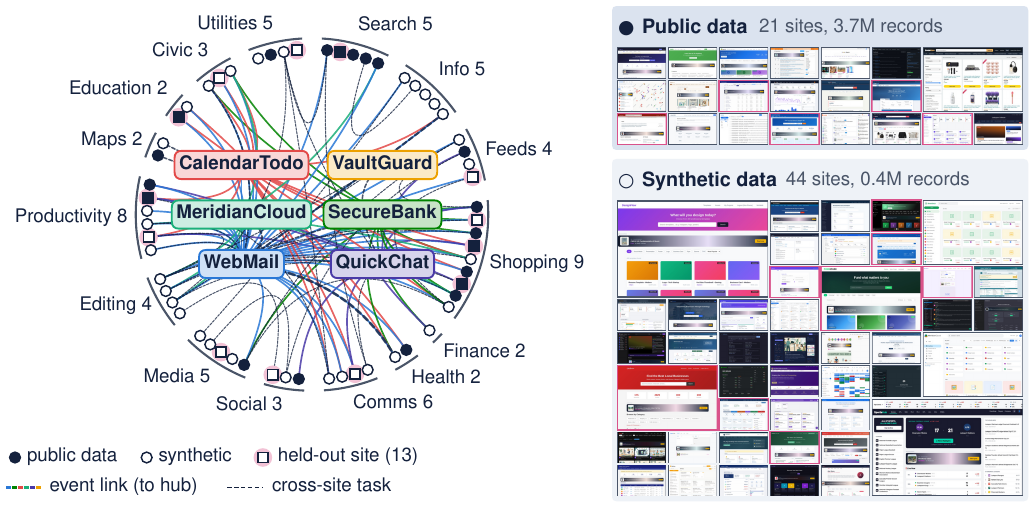}
    \caption{\textbf{The 65 websites.} Left: sites by domain (squares: held-out), linked to hubs at the center (colored edges) and by cross-site tasks (dashed). Right: the 65 \miniweb{} sites, divided by data source. }
    \label{fig:sites}
\end{figure*}

We introduce \textbf{\miniweb{}}, a collection of 65 websites (Fig.~\ref{fig:sites}) built around 37 reusable interaction primitives. Rather than starting from tasks and labeling their skills afterward, we build the environment from the primitives upward (Fig.~\ref{fig:construction}) and compose them into 369 human-authored tasks with skill-level verification. This gives us a controlled place to both measure what agents know and practice what they do not.

Across ten open-weight agents, we found that drag \& gesture is the weakest family for every model (Fig.~\ref{fig:teaser}), with only 12.4\% average success compared with 38.0\% for the next-weakest family. Web- and computer-use-specialized agents perform even worse on dragging than general VLMs despite being stronger on common interactions such as typing. We believe that scaling end-to-end task trajectories can systematically favor frequently encountered interactions while leaving rarer but reusable skills underpracticed.

To test whether skill-specific practice closes these gaps, we introduce \miniwebgen{}, a multi-agent pipeline that generates verified trajectories for skills in \miniweb{}. Training \mbox{Qwen3.5-4B} on 3.1K generated trajectories improves success on held-out single-primitive tasks by 24.9 points, with gains across all skill families. These gains do not automatically transfer to full tasks, so we introduce \miniwebagent{}, which pairs the trained executor with a planner that decomposes tasks into practiced interactions. This approach improves success on held-out \miniweb{} sites by 19.4 points and transfers to three external benchmarks without target-benchmark training: +9.3 points on WebArena-Lite, +20.0 on WebVoyager, and +16.1 on Online-Mind2Web. Overall, we contribute the following:

\begin{itemize}[label=\ding{118}]
    \item \textbf{\miniweb{}}, a benchmark of 65 websites and 369 tasks built from 37 reusable interaction primitives with skill-level verification, revealing that drag \& gesture is the weakest skill family for all ten open-weight agents we evaluate.
    \item \textbf{\miniwebgen{}}, a multi-agent pipeline for generating verified, skill-specific trajectories, which at matched training size outperforms task-centric trajectories from InSTA~\cite{trabucco2025instainternetscaletrainingagents} on all 4 benchmarks.
    \item \textbf{\miniwebagent{}}, a compact browser agent that composes practiced skills via a planner and transfers to WebArena-Lite, WebVoyager, and Online-Mind2Web without target-benchmark training.
\end{itemize}

\begin{table}[t]
\centering
\footnotesize
\setlength{\tabcolsep}{4pt}
\definecolor{mwpurple}{HTML}{5B2E91}
\definecolor{mwtint}{HTML}{F1ECF8}
\definecolor{mwgreen}{HTML}{2E7D32}
\newcommand{\yes}{\textcolor{black}{\ding{51}}}
\newcommand{\no}{\textcolor{black}{\ding{55}}}
\includegraphics{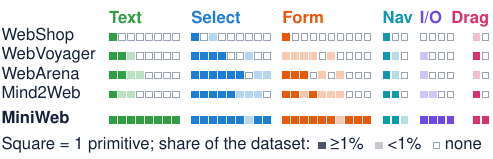}\\[3pt]
\begin{tabular}{@{}lccc>{\columncolor{green!18}[\tabcolsep][0pt]}c@{}}
\toprule
 & \textbf{WebArena} & \textbf{WebVoyager} & \textbf{OM2W} & {\miniweb{}} \\
\midrule
\multicolumn{5}{@{}l}{\textit{\textbf{Scale}}}  \\
Tasks                & 812   & 643  & 300  & 369 \\
Websites             & 5     & 15   & 136  & 65 \\
Hosting              & Local & Live & Live & Local \\
\midrule
\multicolumn{5}{@{}l}{\textit{\textbf{Skill-level evaluation}}} \\
Skill-labeled steps  & \no   & \no  & \no  & \yes \\
Segmented demos      & \no   & \no  & \no  & \yes \\
Per-skill verdicts   & \no   & \no  & \no  & \yes \\
\bottomrule
\end{tabular}
\caption{\textbf{\miniweb{} attributes task failures to skills.} Top: our 36 primitives (reasoning excluded) by family; benchmark rows use our labels of 15.7K tasks, and \miniweb{} uses \miniwebgen{}'s trajectories (Supp.~\ref{app:mining-to-primitives}). Bottom: \miniweb{} labels, segments, and verifies every step by skill; MiniWoB++ \cite{Shi2017WorldOB, Liu2018ReinforcementLO} and ComponentBench \cite{Guan2026ComponentBench} test interactions only outside tasks. OM2W: Online-Mind2Web \cite{Xue2025AnIO}. }
\label{tab:benchmark-comparison}
\end{table}

\section{Related Work}\label{sec:relatedWork}
\paragraph{Web benchmarks and synthetic environments.}
Web benchmarks aim for realism, reproducibility, or coverage. MiniWoB and MiniWoB++~\citep{Shi2017WorldOB, Liu2018ReinforcementLO} pioneered synthetic environments for low-level web interaction, with small self-contained pages that exercise individual widgets such as sliders and drag targets. Self-hosted environments such as WebShop~\citep{Yao2022WebShopTS}, WebArena~\citep{Zhou2023WebArenaAR}, WorkArena~\citep{Drouin2024WorkArenaHC}, and MyPCBench~\citep{Jang2026MyPCBenchAB} provide realistic locally hosted websites, but are limited in term of coverage, while Mind2Web~\citep{Deng2023Mind2WebTA} and live-web benchmarks~\citep{He2024WebVoyagerBA, Xue2025AnIO, jang2026odysseys, gill2026knows} broaden coverage at the cost of task stability and reproducibility. Environment-generation pipelines address this by compiling real-site interaction structures into finite-state machines~\citep{Wu2026AutoWebWorldSI}, generating environments with verifiable tasks~\citep{zhou2026wainf, Awadallah2026Fara15SL}, or synthesizing versions of webpages~\citep{ishmam2026timewarpevaluatingwebagents}. Most of these benchmarks score whole tasks: step-level metrics~\citep{Deng2023Mind2WebTA}, progress rates~\citep{Ma2024AgentBoardAA}, and key-node evaluation~\citep{Pan2024WebCanvasBW}. They measure how far an agent gets but do not attribute a failure to a specific skill. Concurrent with our work, ComponentBench~\citep{Guan2026ComponentBench} evaluates agents on 97 canonical UI components, including sliders and date pickers, using programmatically verified tasks across three React component libraries. \miniweb{} shares this focus on interaction skills but embeds them in full multi-page websites with persistent backend interactions. It also composes primitives into multi-step, cross-site tasks and provides skill-segmented demonstrations with per-primitive verdicts.

\paragraph{Training data and skill composition for web agents.}
Web agents are traditionally trained on end-to-end task trajectories~\citep{He2024WebVoyagerBA, Qin2025UITARSPA, Gupta2026MolmoWebOV, Wu2024OSATLASAF, Wang2025UITARS2TR}. Recent work has scaled these training corpora by converting tutorials into demonstrations~\citep{Ou2024SynatraTI, Xu2024AgentTrekAT}, relabeling or reverse-synthesizing tasks from exploration~\citep{Murty2024NNetNavUL, Sun2024OSGenesisAG}, or collecting trajectories on the live web~\citep{trabucco2025instainternetscaletrainingagents}. These pipelines inherit the coverage of their sources, whereas \miniwebgen{} uses privileged knowledge of each site to direct generation at predefined primitives, including interactions that end-to-end data rarely contains (dragging, searching, \dots). While other work reuses skills at inference time as induced workflows~\citep{Wang2024AgentWM} or distilled APIs~\citep{Zheng2025SkillWeaverWA}, we instead train skills into an executor's weights through practice. To compose them, \miniwebagent{} adopts the established separation of planning from execution found in planner--grounder designs~\citep{Zheng2024GPT4VisionIA, Gou2024NavigatingTD}, manager--worker frameworks~\citep{Agashe2024AgentSA, Agashe2025AgentS2}, and plan-and-act agents~\citep{erdogan2025planandact, piet2026web}. Our contribution is not this architecture but the experience used to train the executor: we examine whether targeted practice improves individual interactions on unseen websites, and whether a planner can compose those improvements into more successful multi-step task execution.

\section{Methodology}\label{sec:method}
We want to study the primitive skills, or macros, of current web agents. Given a task that requires a specific set of skills, can an agent reliably perform each one? When the task fails, can we identify which skill it struggled with? To answer these questions, we need tasks whose component skills are explicit and can be evaluated separately, which current web-agent benchmarks do not provide \cite{Zhou2023WebArenaAR, Deng2023Mind2WebTA, He2024WebVoyagerBA, Yao2022WebShopTS, Jang2026MyPCBenchAB, ishmam2026timewarpevaluatingwebagents, gill2026knows} (Tab.~\ref{tab:benchmark-comparison}).

\subsection{\miniweb{}}\label{sec:miniweb}
\begin{figure}[t]
    \centering
    \includegraphics[width=0.95\linewidth]{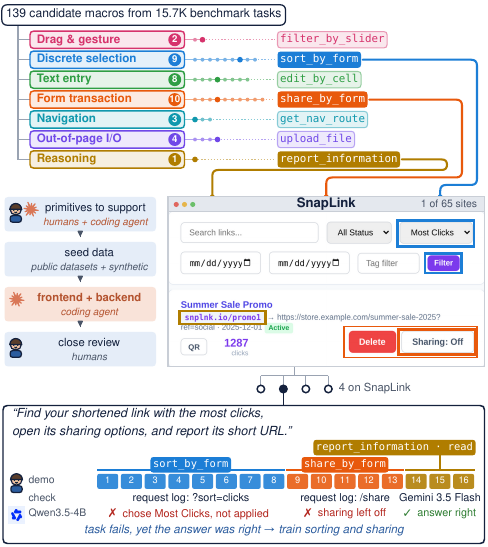}
    \caption{\textbf{Building \miniweb{} bottom up.} (1)~Tasks of existing benchmarks are mined into 37 primitives in 7 families. (2)~Sites are built around their primitives. (3)~Tasks chain primitives, with a verdict per primitive. \textbf{Takeaway:} every step of a task maps to one skill that can be graded.}
    \label{fig:construction}
\end{figure}

We introduce \miniweb{}, a collection of 65 synthetic websites built around 37 interaction primitives. We build it from the bottom up (Fig.~\ref{fig:construction}): First the primitives, then supporting websites, and finally tasks that chain them together.

\paragraph{Primitive selection and schema.}
The set of primitives a web agent needs is hard to enumerate from first principles, so we start from existing data. With Claude Opus~4.8, we label 15.7K tasks from WebArena \cite{Zhou2023WebArenaAR}, WebShop \cite{Yao2022WebShopTS}, WebVoyager \cite{He2024WebVoyagerBA}, and Mind2Web \cite{Deng2023Mind2WebTA} with the recurring interaction patterns that each requires, reusing an existing label whenever one fits. This first pass yields 139 candidate macros, which are dominated by searching, filtering, and extracting information, and only 0.3\% of their 52K labels involve dragging and none drawing, uploading a file, or editing a spreadsheet cell. In group brainstorming sessions, we merge near-duplicate and site-specific candidates, add the interactions the source benchmarks miss, and separate reasoning from the physical interaction. The result is an initial set of 46 \emph{primitives}, which we refine after annotation. Each is a site-agnostic interaction with a single purpose that may take several browser actions, such as filtering items with a slider or editing a spreadsheet cell. We group them into 7 skill families (Fig.~\ref{fig:construction}). Formally, each primitive $m \in \mathcal{M}$ ($|\mathcal{M}|=37$ after refinement) has a family $\phi(m) \in \mathcal{F}$ ($|\mathcal{F}|=7$), a span rule that marks where it starts and ends in a trajectory, and a verifier $V_m$ that checks its outcome. Reasoning is an optional operation $\rho \in \mathcal{O}$ (\eg, \texttt{count}) attached to a primitive, whose verifier then also checks the result: $V_{m\cdot\rho} = V_m \wedge V_\rho$.

\paragraph{Website construction.}
We select 65 websites across 15 representative domains to cover a broad range of interaction primitives and cross-site workflows. Half of these domains are absent from WebArena \cite{Zhou2023WebArenaAR}, Mind2Web \cite{Deng2023Mind2WebTA}, WebVoyager \cite{He2024WebVoyagerBA}, and WebShop \cite{Yao2022WebShopTS}. We then have annotators specify each site's supported primitives, features, and connections to other sites. These links form an information flow graph, which allow for realistic cross-site features such as payment confirmations or cross-site sharing. For realistic content, 21 sites are seeded from public datasets\footnote{We note the license of each data source we used in Tab.~\ref{tab:supp-licenses}.}, which we join through a synthetic layer of users and accounts. The remaining 44 sites use deterministic synthetic data, for about 4M records in total. More details can be seen in Fig.~\ref{fig:sites}. A coding agent\footnote{We used Claude Code with Opus 4.8.} implements each site's frontend and backend from its specification, and all the authors and annotators review every site and fix it until it supports its primitives. The entire process took roughly a month with 6 annotators working 8 hours a week. The backends log every request in a session and write all changes to a per-session overlay, so verifiers can inspect what an agent did, and many agents can run in parallel from the same initial state. 

\begin{figure*}[ht]
    \centering
    \includegraphics[width=\linewidth]{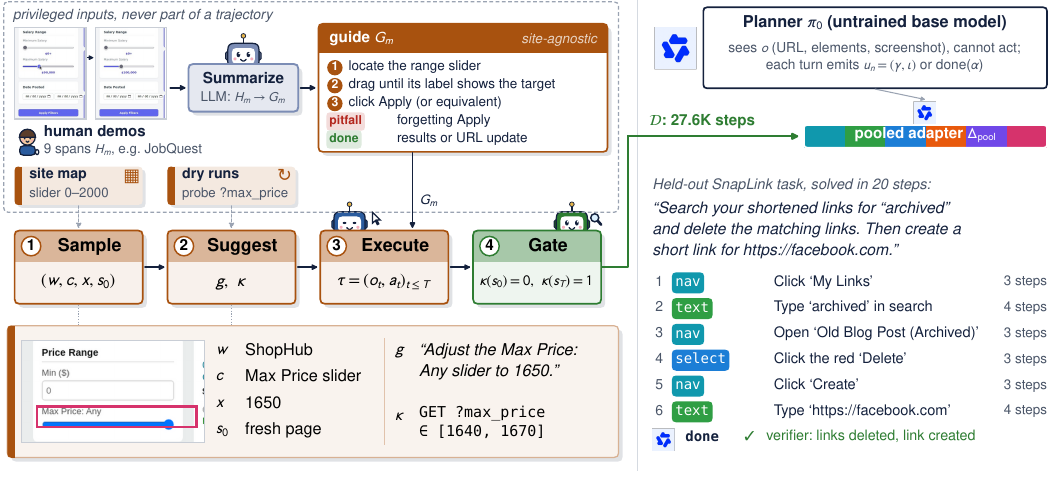}
    \caption{\textbf{\miniwebgen{}} (left) generates and gates single-primitive trajectories (real ShopHub example). \textbf{\miniwebagent{}} (right): an untrained \mbox{Qwen3.5-4B} planner hands each step to one trained LoRA executor (real held-out trace).}
    \label{fig:method}
\end{figure*}

\paragraph{Task annotation and verification.}
An episode is a trajectory $\tau = (o_0, a_1, o_1, \dots, a_T, o_T)$ of observations and browser actions, during which the backend records a request log $L(\tau)$. A task $\mathcal{T} = (g, \langle m_1, \dots, m_k \rangle)$ pairs an instruction $g$ with a chain of primitives. Annotators compose tasks from supported primitives, specify their dependencies, and write realistic instructions. They then perform each task while our tool records browser interactions and server requests, marking the span $[b_j, e_j]$ of each $m_j$ to produce a skill-segmented demonstration $\tau^\star$. Six annotators produced 369 tasks over three months, including 43 cross-site tasks. At the end, the first author manually reviewed every task together with its verifier and demonstration, correcting or removing those with errors. As a final refinement of the primitive set, we drop the 9 hyper-specific primitives that annotators used fewer than three times. The remaining 37 primitives in 7 families form the final set $\mathcal{M}$.

We use two graders. Parameterized verifiers \cite{zhou2026wainf} check observable interactions against the backend request log, while a VLM judge \cite{Xue2025AnIO} evaluates semantic and reasoning outcomes using the primitive's rubric, request log, and screenshots. A task succeeds only if every required primitive $j \in R$ passes,
\begin{equation}
    \mathrm{Succ}(\tau) = \textstyle\bigwedge_{j \in R} V_{m_j}\big(\tau, L(\tau)\big),
    \label{eq:succ}
\end{equation}
and the per-primitive verdicts show where the agent struggled 
(Fig.~\ref{fig:construction}).

\paragraph{Grader validity.} We validate both graders on 183 human demonstrations that remain valid on the current sites and pass the task verifier. For each, we construct two negatives: one truncated before the final required primitive and one empty episode. Across these 549 cases, the deterministic verifiers achieve 98.9\% accuracy and the judge 98.2\%. On 3,689 episodes from the ten untrained agents, the two graders agree on 93.1\% of task outcomes (Cohen's $\kappa=0.85$). A human audit of 120 agent episodes, stratified by the graders' verdicts and weighted back to all episodes, finds the judge right on 92.2\% of task outcomes ($\kappa=0.84$) and the verifiers on 88.9\% ($\kappa=0.77$), close to the agreement between two human labelers ($\kappa=0.86$). Supp.~\ref{app:graders} details the validation protocol, the recordings we leave out, the grader disagreements of each agent, and the audit.

\subsection{\miniwebgen{}}\label{sec:miniwebgen}
Annotating enough human demonstrations to cover each primitive can be challenging, so \miniwebgen{} scales them into verified single-primitive trajectories on the training sites $\mathcal{W}_{\text{train}}$ (Fig.~\ref{fig:method}, left), using privileged knowledge of the benchmark: a crawler maps the controls $\mathcal{C}_m(w)$ that realize primitive $m$ on site $w$ and probes the request each control sends. Generation has five steps.
\begin{enumerate}[(1)]
    \item \emph{\textbf{Summarize.}} Inspired by TimeWarp \cite{ishmam2026timewarpevaluatingwebagents}, an LLM first summarizes the human demonstration spans $H_m$ of each primitive $m$ on training sites (up to 24, each with brief context) into a site-agnostic guide $G_m$: preconditions, steps, pitfalls, and a completion condition.
    \item \emph{\textbf{Sample.}} A task configuration is drawn as
\begin{equation}
    (w, c, z, s_0) \sim P_m, \quad w \in \mathcal{W}_{\text{train}},\; c \in \mathcal{C}_m(w),
    \label{eq:sample}
\end{equation}
where $z$ is an argument of control $c$ and $s_0$ a start state that may already carry another filter or follow another primitive. $P_m$ spreads samples evenly over sites and controls.
    \item \emph{\textbf{Suggest.}} An LLM phrases the instruction $g$. The check $\kappa: \mathcal{S} \to \{0,1\}$ is never written by the LLM: it is the probed request of $(c, z)$, or the data changes that two dry runs agree on.
    \item \emph{\textbf{Execute.}} A coding LLM, given $g$, $G_m$, and read access to the DOM, writes a short script against a restricted wrapper that acts only through mouse and keyboard events on visible elements. Every step is recorded, giving $\tau = (o_0, a_1, \dots, a_T, o_T)$ with screenshots $o_t$ and pixel actions $a_t$.
    \item \emph{\textbf{Gate.}} An attempt that ends in state $s_T$ is kept only if
\begin{equation}
    \kappa(s_0) = 0 \;\wedge\; \kappa(s_T) = 1,
    \label{eq:gate}
\end{equation}
so no attempt passes without acting. The kept set ${\mathcal{D}_{\text{gen}} = \{(g, \tau)\}}$ contains only what a student model will see; privileged information decides only what is generated and how. With Gemini~3.5 Flash, $\mathcal{D}_{\text{gen}}$ holds 3,128 trajectories at about \$0.04 each: 2,687 for 36 of the 37 primitives, and 441 for primitives later dropped in the refinement, which we keep for training. The answer-reporting primitive, \texttt{report\_information}, gets no generated data; in \miniwebagent{}, reasoning steps run on the base model (Sec.~\ref{sec:miniwebagent}).
\end{enumerate}

\providecommand{\se}[1]{\,{\scriptsize(#1)}}
\begin{table*}[t]
\centering
\small
\setlength{\tabcolsep}{3pt}
{%
\begin{tabular}{@{}llcccccccccccc@{}}
\toprule
 & & Task & \multicolumn{7}{c}{Skill family: instances passed} & \multicolumn{4}{c}{Reasoning operation} \\
\cmidrule(lr){4-10}\cmidrule(lr){11-14}
Agent & Size & success & \famnav{Nav.} & \famtext{Text} & \famsel{Select} & \famform{Form} & \famio{I/O} & \famdrag{Drag} & \famreas{Reason} & read & verify & extremum & count \\
\midrule
\multicolumn{14}{@{}l}{\textit{General VLMs, browser-use harness}} \\
Qwen3.5-4B & 4B & 40.1\se{2.5} & 55\se{6} & 39\se{4} & 42\se{4} & 37\se{4} & 32\se{6} & \cellcolor{red!18}12\se{8} & \cellcolor{green!18}60\se{4} & 56\se{5} & 43\se{11} & 37\se{7} & 31\se{8} \\
Qwen3.5-9B & 9B & \textbf{54.7}\se{2.5} & \cellcolor{green!18}71\se{5} & 55\se{4} & \textbf{60}\se{4} & \textbf{62}\se{4} & \textbf{61}\se{6} & \cellcolor{red!18}\textbf{24}\se{10} & 60\se{4} & \textbf{65}\se{5} & 71\se{10} & 46\se{8} & 44\se{8} \\
Qwen3-VL-8B & 8B & 43.8\se{2.6} & \cellcolor{green!18}63\se{6} & 43\se{4} & 42\se{4} & 53\se{4} & 50\se{7} & \cellcolor{red!18}12\se{8} & 52\se{5} & 53\se{5} & 62\se{10} & 41\se{8} & 43\se{9} \\
InternVL3.5-8B & 8B & 39.0\se{2.5} & \cellcolor{green!18}61\se{6} & 44\se{4} & 41\se{4} & 42\se{4} & 32\se{6} & \cellcolor{red!18}\textbf{24}\se{10} & 45\se{5} & 46\se{5} & 48\se{11} & 32\se{7} & 33\se{8} \\
\midrule
\multicolumn{14}{@{}l}{\textit{Web and computer-use agents, own agent loop}} \\
Fara1.5-4B & 4B & 53.1\se{2.7} & \cellcolor{green!18}\textbf{79}\se{5} & \textbf{63}\se{4} & 54\se{4} & 58\se{4} & 55\se{7} & \cellcolor{red!18}18\se{10} & \textbf{64}\se{4} & 60\se{5} & \textbf{76}\se{9} & \textbf{56}\se{8} & \textbf{47}\se{9} \\
Fara-7B & 7B & 29.3\se{2.3} & 54\se{6} & \cellcolor{green!18}54\se{4} & 26\se{3} & 40\se{4} & 41\se{7} & \cellcolor{red!18}0\se{0} & 34\se{4} & 41\se{5} & 29\se{10} & 24\se{7} & 19\se{7} \\
MolmoWeb-4B & 4B & 35.0\se{2.4} & \cellcolor{green!18}61\se{6} & 54\se{4} & 40\se{4} & 38\se{4} & 27\se{5} & \cellcolor{red!18}18\se{10} & 45\se{4} & 44\se{5} & 38\se{11} & 27\se{7} & 28\se{7} \\
MolmoWeb-8B & 8B & 39.3\se{2.4} & \cellcolor{green!18}62\se{5} & 55\se{4} & 47\se{4} & 42\se{4} & 32\se{7} & \cellcolor{red!18}6\se{6} & 51\se{4} & 49\se{5} & 52\se{11} & 41\se{8} & 28\se{8} \\
UI-TARS-1.5-7B & 7B & 25.5\se{2.4} & \cellcolor{green!18}54\se{6} & 47\se{5} & 25\se{3} & 27\se{3} & 20\se{6} & \cellcolor{red!18}0\se{0} & 26\se{4} & 25\se{5} & 33\se{10} & 32\se{7} & 11\se{5} \\
OpenCUA-7B & 7B & 29.5\se{2.3} & \cellcolor{green!18}68\se{5} & 47\se{4} & 31\se{3} & 32\se{3} & 30\se{6} & \cellcolor{red!18}12\se{8} & 33\se{4} & 29\se{5} & 52\se{11} & 34\se{7} & 22\se{7} \\
\bottomrule
\end{tabular}}
\caption{\textbf{Skill profiles of untrained agents on all 369 \miniweb{} tasks.} Task success, then the share of primitive instances passed (\%) per skill family and reasoning operation, all judged by the VLM judge (Gemini 3.5 Flash). One run per agent; in brackets: standard error over 1,000 bootstrap resamples of the tasks. Red: each agent's weakest skill family; green: its strongest. Bold: best per column. }
\label{tab:rq1}
\end{table*}

\subsection{\miniwebagent{}}\label{sec:miniwebagent} 
A model fine-tuned on \miniwebgen{}'s single-primitive trajectories inherits their short horizon and their primitive-level instructions (Sec.~\ref{sec:exp-practice}), so we add a planner that chains primitives. To that end, we introduce \miniwebagent{}, a planner-executor agent over one base model $\pi_0$ (Fig.~\ref{fig:method}, right). Its planner is $\pi_0$ itself, without training. The planner observes $o$ (the URL, the interactive elements, and a screenshot) but cannot act on the page; at each turn $n$ it emits
\begin{equation}
    u_n \in (\mathcal{G} \times \mathcal{I}) \cup \{\mathtt{done}(\alpha)\},
    \label{eq:planner}
\end{equation}
either a delegation, a family group $\gamma \in \mathcal{G} = \{$drag, select, io, rest, reasoning$\}$ (rest covers text entry, forms, and navigation) with a short instruction $\iota \in \mathcal{I}$, or a final answer $\alpha$. A delegation runs the executor $\pi_{\text{pool}}$, which is $\pi_0$ with one LoRA adapter $\Delta_{\text{pool}}$ trained on every skill family, on the same browser for at most 15 actions, $a_t \sim \pi_{\text{pool}}(\cdot \mid \iota, o_t)$, and returns what it did and whether the page changed. Reasoning steps, which only read the page and report, run on $\pi_0$, since \miniwebgen{} generates no reasoning data. One served model hosts the planner and the executor.

\paragraph{Training.} We replay each $(g, \tau) \in \mathcal{D}_{\text{gen}}$ through the agent's browser-use harness, which turns step $t$ into the exact test-time prompt $x_t$ and a target $y_t$ (the action and a short memory, which the base model \mbox{Qwen3.5-4B} writes). We add variants that inject and then correct a mistake (629) or chain two trajectories on one site (857). With the 3,128 plain trajectories, this gives $\mathcal{D}$, with 27.6K steps. The adapter $\Delta_{\text{pool}}$, a rank-64 LoRA on the language model's attention and MLP projections, is fit on all of $\mathcal{D}$ for one epoch (details in Supp.~\ref{app:results}) with the token-averaged cross-entropy
\begin{equation}
    \mathcal{L}_{\text{CE}}(\Delta; \mathcal{D}) = -\mathbb{E}_{(x,y) \sim \mathcal{D}}\, \tfrac{1}{|y|} \textstyle\sum_i \log \pi_\Delta(y_i \mid x, y_{<i}).
    \label{eq:ce}
\end{equation}
No trajectory from the held-out sites or from an external benchmark is used for training.

\section{Experiments and Analyses}\label{sec:experiments}
We use \miniweb{} to ask four questions. Which skills do current web agents lack (Sec.~\ref{sec:exp-diagnosis})? Does practicing skills in isolation improve them, and the tasks that chain them (Sec.~\ref{sec:exp-practice})? Does an agent that composes trained skills transfer to websites and benchmarks it never saw (Sec.~\ref{sec:exp-compose})? And does our skill-centric data transfer better than a task-centric corpus of the same size (Sec.~\ref{sec:exp-data})?

\paragraph{Setup.}\label{sec:exp-setup}
We profile ten open-weight agents (Tab.~\ref{tab:rq1}): four general models in the browser-use harness~\cite{browser_use2024} and six web or computer-use agents in their native agent loops~\cite{Awadallah2025Fara7BAE, Awadallah2026Fara15SL, Gupta2026MolmoWebOV, Qin2025UITARSPA, wang2025opencuaopenfoundationscomputeruse}. For training experiments, we use \mbox{Qwen3.5-4B} and compare an untrained single agent, a single agent trained on all \miniwebgen{} trajectories, an untrained planner-executor, and \miniwebagent{}, which pairs the planner with the trained executor (Tab.~\ref{tab:rq3}). To isolate the effect of training data, we also train the same executor on a size-matched set of task-centric trajectories collected on our training sites with InSTA's pipeline~\cite{trabucco2025instainternetscaletrainingagents}.

Training uses only the 52 \miniweb{} training sites. We evaluate on 62 human-authored tasks and 103 generated single-primitive tasks from 13 held-out sites, as well as WebArena-Lite, WebVoyager, and Online-Mind2Web. For the live benchmarks, we exclude tasks that are impossible or blocked by our network; full evaluation details, action budgets, repeated-run protocol, and statistical tests appear in Supp.~\ref{app:results} and Supp.~\ref{app:liveweb}. We report our key findings below:

\subsection{\texorpdfstring{\famdrag{Drag \& gesture}}{Drag \& gesture} is the weakest family in all web agent models.}\label{sec:exp-diagnosis}
As shown in Tab.~\ref{tab:rq1}, \famdrag{drag \& gesture} is the weakest skill family for every untrained agent. Web and computer-use agents perform worse on drag than general models (9.0\% vs.\ 18.0\% on average), despite performing better on typing, suggesting that end-to-end web training underrepresents rarer interactions such as dragging. Although drag is our smallest family, the gap is large: pooled across agents, drag reaches only 12.4\% success, compared with 38.0\% for the next-weakest family, \famio{out-of-page I/O}.

\begin{figure}[t]
    \centering
    \includegraphics[width=0.8\linewidth]{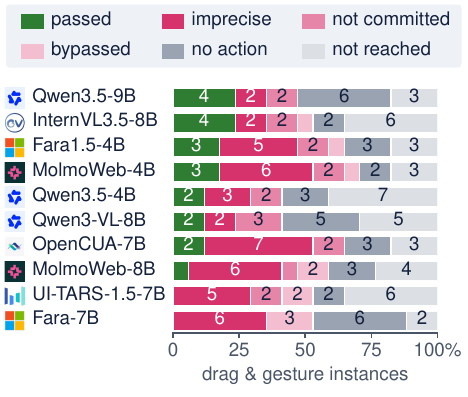}
    \caption{\textbf{Most drag failures happen at the gesture itself.} Outcomes of the ten untrained agents on every \famdrag{drag \& gesture} instance (11 slider and 6 signature tasks), from the VLM judge's explanations; \emph{bypassed} means, \eg, typing a name instead of drawing it.}
    \label{fig:drag}
\end{figure}

The failures are also highly localized. In 72\% of failed drag instances, the agent reaches the correct slider or canvas but fails during the gesture or immediately afterward (Fig.~\ref{fig:drag}). Dragging is typically planned open-loop from a single screenshot, while the resulting value only becomes visible after release, so small spatial errors can directly produce an incorrect result. When agents do set sliders exactly, they often avoid dragging altogether by using arrow keys or directly entering the value. This suggests that the bottleneck is not reaching the right interface element, but executing and verifying the gesture itself.

More broadly, the ten agents exhibit distinct skill profiles even when their overall task success is similar. Task success correlates strongly with \famsel{discrete selection}, \famform{form transactions}, and \famreas{reasoning} (Spearman $\rho$ of 0.96, 0.83, and 0.95), but only weakly with \famtext{text entry} ($\rho = 0.21$). For example, \mbox{Qwen3.5-9B} and \mbox{Fara1.5-4B} differ by only 1.6 points in task success, yet \mbox{Fara1.5-4B} is 8 points better at \famnav{navigation} and 6 points worse at \famsel{discrete selection}. \famreas{Reasoning} skills show similar variation: aggregation is consistently difficult, with counting harder than reading a value for every agent and finding an extremum harder for eight of ten, likely because both require maintaining state across scrolls or pages.

\begin{figure}[t]
    \centering
    \includegraphics[width=0.9\linewidth]{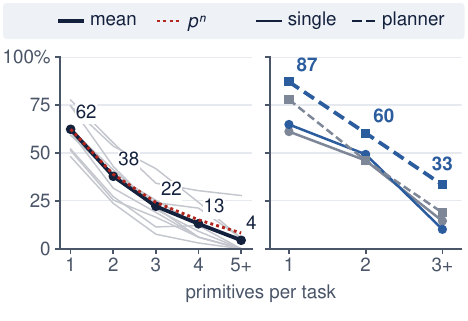}
    \caption{\textbf{Failures compound along a task's chain of primitives.} Left: the ten untrained agents of Tab.~\ref{tab:rq1} on all 369 tasks (grey), with $p$ their mean success on one-primitive tasks. Right: \mbox{Qwen3.5-4B} untrained (grey) or trained (blue) on the 62 held-out tasks (Tab.~\ref{tab:rq3}).}
    \label{fig:chain}
\end{figure}

These primitive-level weaknesses matter because they compound over longer tasks. As task length increases, success falls steeply (Fig.~\ref{fig:chain}, left): averaged across the ten agents, an $n$-primitive task succeeds at roughly $p^n$, within 3.8 points at every length, where $p=62\%$ is mean success on one-primitive tasks. Since every primitive in a chain must succeed, even modest gains at the primitive level compound across the task. This is precisely what skill-specific practice targets.

\providecolor{mwtuntrained}{HTML}{7D8798}
\providecolor{mwtinsta}{HTML}{A8520F}
\providecolor{mwtours}{HTML}{2B5C9E}
\providecommand{\se}[1]{\,{\scriptsize(#1)}}
\begin{table}[tb]
\centering
\small
\setlength{\tabcolsep}{3.4pt}
\begin{tabular}{@{}lcccc}
\toprule
 & \miniweb{} & WA-Lite & WV & OM2W \\
\textcolor{gray}{\footnotesize tasks} & \textcolor{gray}{\footnotesize 62} & \textcolor{gray}{\footnotesize 154} & \textcolor{gray}{\footnotesize 536} & \textcolor{gray}{\footnotesize 280} \\
\midrule
\multicolumn{5}{@{}l}{\textit{\textbf{Single agent, trained on}}} \\
\hspace{0.6em}\textcolor{mwtuntrained}{nothing} & 38.7\se{5.5} & 27.1\se{3.0} & 45.5\se{2.2} & 17.9\se{2.2} \\
\hspace{0.6em}\textcolor{mwtours}{our data} & 39.2\se{5.7} & \cellcolor{red!18}20.3\se{3.0} & \cellcolor{red!18}32.8\se{2.0} & \cellcolor{red!18}8.6\se{1.6} \\
\addlinespace[3pt]
\multicolumn{5}{@{}l}{\textit{\textbf{Planner, executor trained on}}} \\
\hspace{0.6em}\textcolor{mwtuntrained}{nothing} & 45.2\se{5.9} & 30.7\se{3.3} & 59.9\se{2.2} & 21.8\se{2.4} \\
\hspace{0.6em}\textcolor{mwtours}{task-centric data} & 47.8\se{5.7} & 34.2\se{3.4} & 62.1\se{2.1} & 23.2\se{2.6} \\
\hspace{0.6em}\textcolor{mwtours}{skill-centric data} & \cellcolor{green!18}\textbf{58.1}\se{6.0} & \cellcolor{green!18}\textbf{36.4}\se{3.3} & \cellcolor{green!18}\textbf{65.5}\se{2.1} & \cellcolor{green!18}\textbf{33.9}\se{2.8} \\
\bottomrule
\end{tabular}
\caption{\textbf{Composing trained skills raises task success.} Task success (\%) of Qwen3.5-4B on \miniweb{}'s held-out sites, WebArena-Lite (WA-Lite), WebVoyager (WV), and Online-Mind2Web (OM2W), as a single agent or as a planner whose executor is untrained or trained at matched size on task-centric trajectories or on \miniwebgen{}'s skill-centric trajectories. In brackets: standard error over 1,000 bootstrap resamples of the tasks. Green (red): significantly above (below) the untrained row of the group, $p<0.05$, by a paired test over the same tasks. }%
\label{tab:rq3}
\end{table}

\subsection{\miniweb{} practice improves single-primitive task generation, but not task success.}\label{sec:exp-practice}
\begin{figure}[t]
    \centering
    \includegraphics[width=0.9\linewidth]{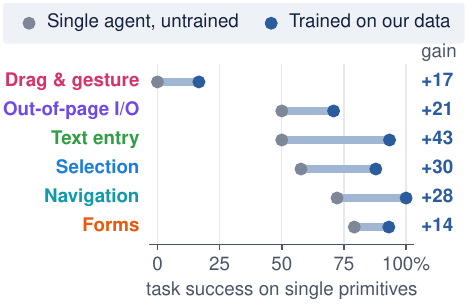}
    \caption{\textbf{Practice lifts every skill family.} Success per skill family on the 103 generated single-primitive tasks of the held-out sites, for a single \mbox{Qwen3.5-4B} agent untrained (grey) and trained on our data (blue), with the gain at the right; \famreas{reasoning}, which is not trained, is left out.}%
    \label{fig:skill_improve}
\end{figure}

Since task success compounds over a task's primitives, \miniwebgen{} trains the agent directly on those primitives. This works at the skill level: on 103 generated single-primitive tasks from held-out sites, \mbox{Qwen3.5-4B} improves from 55.3\% to 80.3\% success, with gains across every skill family (Fig.~\ref{fig:skill_improve}). But these gains do not transfer to full tasks. On 62 human-authored held-out tasks, primitive success remains essentially unchanged (44.1\% to 45.1\%), and so does task success (38.7\% to 39.2\%); performance also drops on all three external benchmarks (Tab.~\ref{tab:rq3}).

The problem is composition. \miniwebgen{} trains mostly on one-primitive episodes with instructions that explicitly name the skill, while real tasks chain several primitives under a single high-level request. The trained single agent therefore often stops before reaching the skills it learned. Supp.~\ref{app:results} analyzes this further. \miniwebagent{} addresses both mismatches with a planner: an untrained planner decomposes the task into primitive-level instructions, and the trained executor carries out each one in the setting it practiced.
  
\subsection{Composing trained skills raises task success on unseen websites and benchmarks.}\label{sec:exp-compose}
With a planner, the practiced primitives translate into end-to-end task success (Tab.~\ref{tab:rq3}). Relative to the untrained single agent, \miniwebagent{} gains 19.4 points on held-out \miniweb{} sites and improves by 9.3, 20.0, and 16.1 points on WebArena-Lite, WebVoyager, and Online-Mind2Web, respectively, without training on target-benchmark trajectories.

The planner and trained executor contribute differently. Planning alone gives its largest gain on WebVoyager, where long information-seeking tasks benefit from decomposition, but helps little elsewhere. Adding the trained executor consistently improves over the untrained planner across all four evaluations. On held-out \miniweb{} tasks, \miniwebagent{} also raises primitive success from 49.8\% to 63.7\%, with gains across every skill family. As expected from compounding primitive failures, the advantage grows on longer tasks: for tasks with three or more primitives, \miniwebagent{} more than doubles the success of the untrained single agent (33.3\% vs.\ 14.5\%) (Fig.~\ref{fig:chain}, right).

\subsection{At matched size, our skill data matches or beats task-centric data.}\label{sec:exp-data}
Does the gain come from skill-specific practice, or simply from training on more web trajectories? To test this, we reimplemented InSTA's trajectory collection pipeline on \miniweb{} \cite{trabucco2025instainternetscaletrainingagents} to collect a size-matched corpus of task-centric trajectories of \miniweb{}, which, like our skill data, \mbox{Qwen3.5-4B} writes. We then train the same executor on such corpus. The task-centric executor does not significantly improve over the untrained planner on any of the four benchmarks, while our skill-trained executor does on all four (Tab.~\ref{tab:rq3}). Head to head, our executor is ahead everywhere, and the gap is significant on Online-Mind2Web (+10.7 points, $p<0.001$) and held-out \miniweb{} (+10.2, $p=0.027$; Holm $p=0.081$). At this size, skill-centric data is at least as useful as task-centric data, and it is the one that reliably adds to planning.

\section{Conclusion}\label{sec:conclusion}
In this paper, we introduce \miniweb{}, a synthetic benchmark centered on 37 interaction primitives across 65 locally hosted websites, providing per-primitive evaluation and skill-segmented data generation for web agents. Its skill profiles show that \famdrag{drag \& gesture} is the weakest skill family of every agent we tested. Using this infrastructure, \miniwebgen{} expands limited seed demonstrations into 3.1K verified trajectories for targeted skill practice. Fine-tuning a compact base model on them improves the execution of single primitives, but on its own it does not raise success on multi-step tasks and lowers it on external benchmarks, because the model learns single primitives from single-primitive episodes, not how to chain them: it stops after one primitive, and removing that stop alone does not help. \miniwebagent{} separates task planning from practiced skill execution; without training on any target-benchmark trajectories, it improves over the base model on the held-out sites, WebArena-Lite, WebVoyager, and Online-Mind2Web, and its trained executor adds to what planning alone achieves, most on the held-out sites and Online-Mind2Web. Our findings suggest that targeted skill practice on synthetic websites, combined with an explicit separation of planning and execution, helps compact open-weight agents on websites they were not trained on.   
\section{Limitations \& Future Works}
We acknowledge several limitations in our study. First, \miniweb{} contains fewer human-authored tasks than Mind2Web~\cite{Deng2023Mind2WebTA} and WebArena~\cite{Zhou2023WebArenaAR}. Although our annotation interface (Supp.~\ref{app:annotation}) streamlines task creation and trajectory recording, devising diverse tasks still requires substantial effort and creativity. \miniwebgen{} alleviates training-data scarcity but does not replace broader human-authored evaluation. Second, although parameterized verifiers reduce grading costs, running closed-source frontier agents remains expensive. We therefore leave these agents unevaluated, reflecting both resource constraints and our focus on small open-weight models. Third, we do not fully resolve the \famdrag{drag \& gesture} deficiencies identified by our benchmark. Our focus is on showing that improvements across skill families, when composed, can improve task performance on unseen websites, rather than achieving mastery of any individual skill. We leave specialized methods for mastering \famdrag{drag \& gesture} interactions to future work. Finally, this study does not include scaling experiments due to compute and evaluation costs. Future work should investigate how skill-centric training performs with larger datasets, greater website diversity, and larger models.
\section{Acknowledgements}
We sincerely thank all annotators who dedicated their time to contributing to and reviewing this benchmark. We especially thank Omar Rizman, an undergraduate researcher at the University of Utah, as well as members of the Spark Lab, the Utah NLP group, and the LUNA Lab at the University of Utah for their valuable feedback and review. This material is based upon work supported by the National Science Foundation under Grant No. 2535195

\bibliographystyle{plainnat}
\bibliography{ref}

\appendix
\onecolumn
\addtocontents{toc}{\protect\setcounter{tocdepth}{2}}
\section*{Supplementary Material}
The supplement follows the paper: the benchmark (Supp.~\ref{app:induction}--\ref{app:graders}), the agents and their training (Supp.~\ref{app:spaces}--\ref{app:specialists}), the evaluation (Supp.~\ref{app:liveweb}--\ref{app:compute}), additional results (Supp.~\ref{app:results}), and the prompts (Supp.~\ref{app:prompts}).

\subsection*{Contents}
\begingroup\hypersetup{linkcolor=black}
\supplementcontents
\endgroup
\clearpage

\section{Mining candidate macros from web-agent benchmarks}\label{app:induction}
This section expands the first step of Sec.~\ref{sec:miniweb}: it defines a macro (Supp.~\ref{subsec:macro-def}), describes how candidate macros were mined from the tasks of four web-agent benchmarks (Supp.~\ref{subsec:macro-bank}), reports what the mining found (Supp.~\ref{app:mining}), and explains how the candidates became the 37 primitives (Supp.~\ref{app:mining-to-primitives}).

\subsection{What is a macro?}
\label{subsec:macro-def}

We define a \emph{macro} as the smallest unit of web interaction that carries a
distinct, recurring \textbf{intent}. A macro occupies an intermediate level of
granularity between an \emph{atomic action}
(\texttt{click}, \texttt{keystroke}, or \texttt{scroll}) and
a \emph{task}, a high-level natural-language goal that may require extended
planning to fulfil. A macro is \textbf{short-horizon}: it realizes a single coherent
intent through a short, contiguous chain of atomic actions and requires no further
decomposition into sub-goals. Crucially, macros are written
\textbf{site-agnostically}.

Structurally, each macro is a pair of a \emph{verb} and a \emph{modifier},
combined in \texttt{snake\_case} as \texttt{verb\_modifier}. The verb denotes the
primary goal of the interaction (e.g., \texttt{navigate}, \texttt{search},
\texttt{extract}, \texttt{submit}) and is drawn from a closed, human-curated
vocabulary~$\mathcal{V}$; the modifier specifies how the target of that goal is
defined or structured (e.g., \texttt{by\_semantic}, \texttt{from\_free\_text},
\texttt{by\_attribute}) and is drawn from a closed vocabulary~$\mathcal{M}$. A
macro is therefore an element of a curated subset of $\mathcal{V}\times\mathcal{M}$.

\begin{definition}{Macro}{}
Let $\mathcal{A}$ denote the space of atomic web actions, and let a \emph{task} be
a sequence over $\mathcal{A}$ that fulfils a natural-language goal on some website.
A \emph{macro} $m=(v,\mu)$, with $v\in\mathcal{V}$ and $\mu\in\mathcal{M}$, is a
site-agnostic unit of web interaction satisfying:
\begin{enumerate}
  \item \textbf{Single intent.} $m$ expresses exactly one intent;
  \item \textbf{Verifiable terminal state.} Executing $m$ reaches a single
        terminal state whose success is checkable by a single evaluator; and
  \item \textbf{Atomic minimality.} Decomposing $m$ yields only atomic actions in
        $\mathcal{A}$---that is, $m$ cannot be split into a chain of smaller
        macros.
\end{enumerate}
\end{definition}

These conditions induce a strict granularity hierarchy,
$\text{atomic action} \prec \text{macro} \prec \text{task}$: a task decomposes
into a chain of one or more macros, whereas a macro decomposes into a chain of
atomic actions but admits no intermediate decomposition into smaller macros.
\subsection{Macro bank creation}
\label{subsec:macro-bank}

We construct the macro bank through a combination of LLM-driven induction and
human supervision. The procedure first elicits a prior over the macro space by
mining the tasks of four existing web-agent benchmarks (WebArena, WebVoyager,
Mind2Web, and WebShop), and then extends this seed bank to the domains of \miniweb{}.
The induction is orchestrated by prompting Claude Opus~4.8 to operate over a
relational store with four tables: a verb bank~$\mathcal{V}$, a modifier
bank~$\mathcal{M}$, a macro bank, and the pooled task set. The process proceeds in
three stages.

\paragraph{First pass: reference-bank construction.}
We pool the tasks from all four datasets into a single corpus and process them
iteratively. For each task, the agent reasons about the sub-goals required to
accomplish it and proposes the macros, and where necessary the constituent verbs
and modifiers, needed to express those sub-goals. When an existing entry in
$\mathcal{V}$ or $\mathcal{M}$ suffices, the agent reuses it; otherwise it proposes
a new verb or modifier and marks it as proposed. Throughout, the agent consults the
current state of the bank before each addition to avoid introducing redundant
entries.

\paragraph{Deduplication.}
Because the first pass admits proposed verbs and modifiers greedily, the resulting
vocabularies contain near-duplicates. The agent therefore reviews $\mathcal{V}$ and
$\mathcal{M}$ and flags candidate duplications, which are then adjudicated by a
human annotator who either merges each entry into an existing verb or modifier or
keeps it as distinct. This step yields the curated vocabularies that fix the macro
bank.

\paragraph{Second pass: task--macro assignment.}
The macro assignments produced during the first pass are confounded by the order in
which tasks were processed, since the bank was still growing as tasks were seen. To
remove this dependence, we discard the first-pass assignments, fix the curated bank
from the deduplication step, and prompt the agent to reassign macros to every task
from scratch, disregarding the prior assignments. Fixing the reference bank ensures
that each task is labeled against the same macro vocabulary, independent of its
position in the corpus.

This procedure yields a seeded macro bank grounded in existing benchmarks. We then
assign the seeded macros to each website domain of \miniweb{} and brainstorm
additional macros, again with LLM assistance under human review, to cover intents
that the four source datasets do not exercise.

Supp.~\ref{app:prompt-induction} gives the orchestrator's full prompt.

\subsection{Mining statistics}\label{app:mining}
The final labeling covers 15,727 tasks: all 812 of WebArena, 12,251 WebShop instructions, all 643 of WebVoyager, and 2,021 of Mind2Web's 2,350, with 52,207 macro labels in total (Tab.~\ref{tab:supp-mining}). The labels use 136 distinct candidate macros, built from 18 verbs and 42 modifiers; 6 WebShop entries whose instruction states no task carry the label \texttt{OFF\_VOCAB} instead. A task needs 2.2 to 3.5 macros on average, depending on the benchmark. Most candidates recur across benchmarks: 22 appear in one benchmark only, 13 of them in Mind2Web, whose tasks span the most websites.

\begin{table}[H]
\centering
\small
\begin{tabular}{@{}lrrrrr@{}}
\toprule
Benchmark & Tasks & Labels & Labels per task & Candidates used & Only in this benchmark \\
\midrule
WebArena & 812 & 1,813 & 2.23 & 99 & 5 \\
WebShop & 12,251 & 42,859 & 3.50 & 36 & 0 \\
WebVoyager & 643 & 1,865 & 2.90 & 85 & 4 \\
Mind2Web & 2,021 & 5,670 & 2.81 & 124 & 13 \\
\midrule
All & 15,727 & 52,207 & 3.32 & 136 & 22 \\
\bottomrule
\end{tabular}
\caption{\textbf{The mined tasks.} Tasks labeled per benchmark, their macro labels, and the candidate macros they use. \emph{Only in this benchmark}: candidates that no other benchmark's tasks use.}
\label{tab:supp-mining}
\end{table}

\paragraph{What the benchmarks exercise.} Tab.~\ref{tab:supp-verbs} gives the share of labels per verb. WebShop contributes 78\% of the tasks, nearly all of them product search and option choice, so the pooled shares are dominated by selecting and searching; without WebShop, searching, filtering, and extracting information account for 61\% of the labels. The ten most frequent candidates are \texttt{search\_by\_keyword} (12,919), \texttt{filter\_by\_attribute} (6,037), \texttt{select\_to\_purchase} (3,880), \texttt{add\_to\_cart} (3,871), \texttt{add\_to\_purchase} (3,622), \texttt{select\_by\_color} (2,993), \texttt{select\_by\_size} (2,389), \texttt{select\_by\_attribute} (2,332), \texttt{submit\_to\_purchase} (1,477), \texttt{select\_by\_variant} (1,454). No label names dragging, drawing, uploading a file, setting a slider, signing, or editing a spreadsheet cell. The closest are the 157 price filters (\texttt{filter\_by\_price}), which a slider sets (Supp.~\ref{app:mining-to-primitives}), and only 28 labels write a file.

\begin{table}[H]
\centering
\small
\begin{tabular}{@{}lrrrl@{}}
\toprule
Verb & Labels (\%) & Without WebShop (\%) & Candidates & Most frequent candidate \\
\midrule
select & 31.3 & 9.4 & 27 & \texttt{select\_to\_purchase} \\
search & 26.4 & 16.5 & 5 & \texttt{search\_by\_keyword} \\
filter & 17.1 & 27.8 & 20 & \texttt{filter\_by\_attribute} \\
add & 14.6 & 3.1 & 8 & \texttt{add\_to\_cart} \\
submit & 4.4 & 8.4 & 23 & \texttt{submit\_to\_purchase} \\
extract & 3.0 & 16.9 & 14 & \texttt{extract\_by\_question\_answer} \\
navigate & 1.9 & 10.8 & 11 & \texttt{navigate\_by\_category} \\
create & 0.5 & 2.6 & 7 & \texttt{create\_by\_form} \\
edit & 0.2 & 1.3 & 3 & \texttt{edit\_by\_form} \\
compute & 0.2 & 1.1 & 3 & \texttt{compute\_by\_quantity} \\
react & 0.1 & 0.6 & 3 & \texttt{react\_to\_post} \\
compare & 0.1 & 0.6 & 2 & \texttt{compare\_by\_attribute} \\
play & 0.0 & 0.3 & 4 & \texttt{play\_media} \\
share & 0.0 & 0.2 & 1 & \texttt{share\_to\_post} \\
export & 0.0 & 0.2 & 1 & \texttt{export\_to\_file} \\
delete & 0.0 & 0.1 & 2 & \texttt{delete\_from\_table} \\
verify & 0.0 & 0.1 & 1 & \texttt{verify\_by\_category} \\
follow & 0.0 & 0.1 & 1 & \texttt{follow\_to\_follow} \\
\bottomrule
\end{tabular}
\caption{\textbf{Labels per verb.} Share of the mined labels by the verb of their macro, over all four benchmarks and without WebShop; the number of candidate macros with that verb; and the most frequent of them.}
\label{tab:supp-verbs}
\end{table}

\subsection{From candidates to primitives}\label{app:mining-to-primitives}
The candidates describe what existing tasks ask for at the granularity of their instructions, so many differ only in the value they target. In group brainstorming sessions we merged candidates that perform the same interaction on different values (\eg, the 27 \texttt{select} candidates, such as \texttt{select\_by\_color} and \texttt{select\_by\_size}, are choices from a set of options), removed site-specific ones, and separated reasoning from the physical interaction: the \texttt{extract}, \texttt{compute}, \texttt{compare}, and \texttt{verify} candidates correspond to the reasoning operations of Supp.~\ref{app:primitives}, which are attached to the interaction they follow. We then added interactions the benchmarks rarely or never exercise, such as dragging a slider, drawing a signature, uploading a file, and editing a spreadsheet cell. This gave the initial set of 46 primitives in 7 skill families (Sec.~\ref{sec:miniweb}). After annotation, we kept the 37 primitives that annotators used at least 3 times (Supp.~\ref{app:primitives}).

\paragraph{Mapping the candidates to the final primitives.} To measure how often the benchmarks need each final primitive, we map each of the 136 candidates to the final primitive that performs its interaction, or to none. Gemini~3.6 Flash proposes each mapping from the candidate's name, its number of labels, and three example tasks from the benchmarks that use it most, and one of the authors checks every candidate. The author keeps 131 of the 136 mappings (96.3\%, $\kappa=0.96$; 99.7\% of the labels). The 5 changes cover 176 labels: posts and replies move from the free-text primitives to \texttt{create\_by\_form}, comparisons of on-page values from \texttt{compare\_by\_form} to \texttt{report\_information}, and a search form to \texttt{sort\_by\_form}. No candidate maps to none. The price filter is the only candidate in the drag \& gesture family, so dragging accounts for 0.3\% of the mined labels (at most 2.1\%, in Mind2Web), against 7.6\% of the primitive instances in \miniwebgen{}'s training episodes. Tab.~\ref{tab:benchmark-comparison} (top) shows which primitives each dataset exercises. How evenly it exercises them is the effective number of primitives, the exponential of the entropy of the primitive shares, which equals the number of primitives in use when all are used equally. Excluding the reasoning primitive, which every information task ends with, the benchmarks' labels spread over 3.4 (WebShop) to 9.6 (WebArena) of the 36 primitives and \miniwebgen{}'s trajectories over 32.8; on the 25 primitives that some benchmark uses, they still spread over 23.1.

\section{Interaction primitives and reasoning operations}\label{app:primitives}
Every primitive instance in a \miniweb{} task carries two tags: a \emph{base primitive}, the physical interaction (\eg, \texttt{filter\_by\_slider}), and an optional \emph{reasoning operation} the agent applies to what it sees (\eg, \texttt{extremum}). An answer the agent must return to the user is tagged with the operation-only primitive \texttt{report\_information}; reasoning that only serves a later step is attached to the primitive it belongs to rather than tagged on its own. Primitives are called \emph{macros} in our code and in the induction pipeline of Supp.~\ref{app:induction}.

A single registry defines each primitive by a description, a \emph{span} (the action where an instance starts and the one where it ends in a demonstration, used to cut demonstrations into segments), and a rubric for the VLM judge (Supp.~\ref{app:prompts}). The seven skill families group primitives by the control they operate; the family-specialist variant of \miniwebagent{} trains one specialist per group of families: drag \& gesture, discrete selection, out-of-page I/O, reasoning, and one for the rest (text entry, form transactions, navigation; Sec.~\ref{sec:miniwebagent}). Tab.~\ref{tab:supp-primitives} lists the 37 primitives of the final set with their coverage in the human tasks and the generated pool, and Tab.~\ref{tab:supp-ops} the reasoning operations. The table caption names the 9 hyper-specific primitives dropped in the final refinement (Sec.~\ref{sec:miniweb}). Operations with fewer than 15 instances are omitted from the skill profiles (Tab.~\ref{tab:rq1}); the caption of Tab.~\ref{tab:supp-ops} gives their counts.

\begin{table}[H]
\centering
\footnotesize
\setlength{\tabcolsep}{4pt}
\begin{tabular}{@{}l>{\raggedright\arraybackslash}p{0.42\linewidth}>{\raggedright\arraybackslash}p{0.36\linewidth}r@{}}
\toprule
Operation & Definition & Check & Instances \\
\midrule
\texttt{read} & the info is present on the page --- locate and read it & answer appears verbatim on a visited page & 100 \\
\texttt{extremum} & get the max/min among on-page items without using a sort control & chosen item is the argmax/argmin of the candidate set & 41 \\
\texttt{count} & count the on-page items that match a condition & count matches the true number of matching items & 36 \\
\texttt{verify} & compare a value on the page against a value given in the instruction & the on-page value satisfies the instruction's stated value/condition & 21 \\
\bottomrule
\end{tabular}
\caption{\textbf{Reasoning operations} attached to instances of the 37 primitives in the human tasks. A primitive without an operation is a pure interaction. Not listed (fewer than 15 instances): \texttt{compute} (7), \texttt{spatial} (4), \texttt{compare} (3).}
\label{tab:supp-ops}
\end{table}

{\footnotesize\setlength{\tabcolsep}{4pt}
\begin{longtable}{@{}>{\raggedright\arraybackslash}p{0.235\linewidth}>{\raggedright\arraybackslash}p{0.565\linewidth}rrr@{}}
\caption{\textbf{The 37 interaction primitives}, by skill family. \emph{Tasks}: instances in the 369 human tasks; \emph{Sites}: sites those tasks use; \emph{Gen.}: trajectories generated by \miniwebgen{} in the training pool (2,687; the pool's other 441 are of dropped primitives, and \texttt{report\_information} is learned from the base model's own successes instead). The span marks where an instance starts and ends in a demonstration. Dropped in the final refinement (fewer than 3 human-task instances): \texttt{reposition\_by\_drag} (2), \texttt{save\_by\_form} (2), \texttt{search\_by\_pan\_zoom} (2), \texttt{reveal\_by\_2fa} (1), \texttt{search\_by\_playback} (1), \texttt{carry\_info\_cross\_site} (0), \texttt{count\_entries} (0), \texttt{edit\_by\_ranking} (0), \texttt{edit\_by\_textbox} (0).}\label{tab:supp-primitives}\\
\toprule Primitive & Description & Tasks & Sites & Gen. \\ \midrule \endfirsthead
\caption[]{(continued)} \\ \toprule Primitive & Description & Tasks & Sites & Gen. \\ \midrule \endhead
\bottomrule \endlastfoot
\multicolumn{5}{@{}l}{\textit{Drag \& gesture}} \\*
\texttt{filter\_by\_slider} & Narrow results with a range slider. \emph{Span:} first drag $\rightarrow$ filter applied & 11 & 9 & 74 \\
\texttt{sign\_by\_freeformdrawing} & Sign a document by DRAWING your signature freeform on a canvas (``Sign here'' field). \emph{Span:} start drawing in the signature field $\rightarrow$ signature adopted/submitted & 6 & 4 & 100 \\
\multicolumn{5}{@{}l}{\textit{Discrete selection}} \\*
\texttt{toggle\_relationship} & Locate a user/post/item, then toggle a relationship (follow / subscribe / save / join-group / block). \emph{Span:} locate the target $\rightarrow$ toggle clicked & 46 & 28 & 97 \\
\texttt{filter\_by\_dropdown} & Narrow results by selecting a dropdown value (incl. category search). \emph{Span:} open the dropdown $\rightarrow$ option selected / results update & 41 & 28 & 97 \\
\texttt{sort\_by\_form} & Sort a listing by operating a sort control. \emph{Span:} first sort-control interaction $\rightarrow$ state change (sorted view) & 35 & 26 & 91 \\
\texttt{filter\_by\_date\_range} & Narrow results to a date range. \emph{Span:} first date input $\rightarrow$ last date / submit & 24 & 22 & 100 \\
\texttt{filter\_by\_options} & Narrow results by clicking option fields (chips / checkboxes). \emph{Span:} first option $\rightarrow$ last option / filter applied & 21 & 16 & 68 \\
\texttt{delete\_from\_table} & Delete an item from a list/table. \emph{Span:} delete clicked $\rightarrow$ confirmed & 17 & 15 & 99 \\
\texttt{play\_by\_playback} & Operate a media player (play/pause/seek; + at timestamp/quality/subtitle). \emph{Span:} open media page $\rightarrow$ play clicked & 14 & 8 & 8 \\
\texttt{feedback\_by\_react} & React to content (like / upvote). \emph{Span:} react control $\rightarrow$ reacted & 10 & 6 & 100 \\
\texttt{feedback\_by\_star} & Give a star/slider rating. \emph{Span:} first star/drag $\rightarrow$ rating set & 5 & 4 & 82 \\
\multicolumn{5}{@{}l}{\textit{Text entry}} \\*
\texttt{search} & Find content by typing a text/natural-language query. \emph{Span:} click search box $\rightarrow$ enter / submit & 71 & 43 & 100 \\
\texttt{message\_from\_free\_text} & Send a message by typing text. \emph{Span:} open the message field $\rightarrow$ send clicked & 20 & 11 & 85 \\
\texttt{edit\_by\_cell} & Enter or change data directly in a grid/table cell (incl. adding a row or column to extend the table). \emph{Span:} focus a grid cell $\rightarrow$ cell value committed (save or autosave) & 8 & 5 & 32 \\
\texttt{translate\_by\_query} & Translate entered text. \emph{Span:} entering text $\rightarrow$ translation received & 6 & 2 & 52 \\
\texttt{compute\_by\_tool} & Use an on-page calculator/tool to compute a result. \emph{Span:} open the tool $\rightarrow$ result shown & 5 & 2 & 99 \\
\texttt{feedback\_by\_text} & Leave a written review/comment (NEW). \emph{Span:} open the review box $\rightarrow$ submit & 5 & 3 & 62 \\
\texttt{write\_executable\_program} & Write and run code. \emph{Span:} open editor $\rightarrow$ run & 4 & 2 & 59 \\
\texttt{sign\_by\_text} & Sign a document by TYPING your name into the signature field (adopt a typed signature). \emph{Span:} focus the typed-signature field $\rightarrow$ signature adopted/submitted & 3 & 2 & 100 \\
\multicolumn{5}{@{}l}{\textit{Form transaction}} \\*
\texttt{create\_by\_form} & Create/register/apply/invite/post a record by filling and submitting a form. \emph{Span:} open the form $\rightarrow$ submit the form & 80 & 41 & 99 \\
\texttt{authenticate\_by\_form} & Log in by entering credentials. \emph{Span:} open the form $\rightarrow$ submit & 26 & 20 & 57 \\
\texttt{edit\_by\_form} & Edit an existing record through a form. \emph{Span:} open the form $\rightarrow$ close/submit the form & 14 & 11 & 100 \\
\texttt{configure\_by\_form} & Change one or more settings via a form. \emph{Span:} open the form $\rightarrow$ submit/save & 13 & 11 & 77 \\
\texttt{pay\_by\_form} & Make a payment via a form. \emph{Span:} open the form $\rightarrow$ submit the form & 12 & 9 & 35 \\
\texttt{share\_by\_form} & Share an item via a share form/dialog. \emph{Span:} share is clicked $\rightarrow$ share form is submitted & 8 & 7 & 100 \\
\texttt{book\_by\_form} & Book/reserve by filling a form (incl. dates). \emph{Span:} open the form $\rightarrow$ submit the form & 7 & 5 & 60 \\
\texttt{checkout\_by\_form} & Complete checkout (incl. promo/redeem) via a payment form. \emph{Span:} open the form $\rightarrow$ submit the form & 7 & 3 & 66 \\
\texttt{cancel\_by\_form} & Cancel a booking or order via a form. \emph{Span:} open the form $\rightarrow$ submit the form & 6 & 5 & 11 \\
\texttt{compare\_by\_form} & Compare items by selecting them through form controls (e.g. dropdowns) that render a side-by-side view. \emph{Span:} open/enter the compare view $\rightarrow$ both items selected and the comparison is visible & 3 & 3 & 100 \\
\multicolumn{5}{@{}l}{\textit{Navigation}} \\*
\texttt{navigate\_by\_route} & Go to a page by clicking a link/menu/tab. \emph{Span:} link/menu click $\rightarrow$ page changed & 61 & 40 & 99 \\
\texttt{get\_nav\_route} & Get directions by entering origin/destination. \emph{Span:} first point typed $\rightarrow$ submit & 5 & 2 & 56 \\
\texttt{join\_meeting} & Join a meeting (by code or link). \emph{Span:} open meeting $\rightarrow$ joined & 5 & 1 & 11 \\
\multicolumn{5}{@{}l}{\textit{Out-of-page I/O}} \\*
\texttt{upload\_file} & Upload a file via the file picker. \emph{Span:} file picker opens $\rightarrow$ file attached & 26 & 18 & 68 \\
\texttt{export} & Export data (choose a format and download). \emph{Span:} export clicked $\rightarrow$ downloaded & 15 & 11 & 43 \\
\texttt{copy\_content} & Copy content via a copy control. \emph{Span:} copy clicked $\rightarrow$ copied & 8 & 6 & 100 \\
\texttt{edit\_by\_image} & Edit an image via canvas tools. \emph{Span:} select tool $\rightarrow$ save & 5 & 3 & 100 \\
\multicolumn{5}{@{}l}{\textit{Reasoning}} \\*
\texttt{report\_information} & Report the answer back to the human. Always the terminal tag --- span the last action(s) where the answer is on screen; carries the reasoning op. Intermediate reasoning folds onto its own macro instead. \emph{Span:} the answer is located/surfaced on the page $\rightarrow$ the answer is reported & 121 & 59 & -- \\
\end{longtable}}

\section{Websites}\label{app:sites}
\miniweb{} has 65 websites in 15 domains, served offline by one local server. They share one fictional world (a city, Lakeport, WA, and a company, Meridian Systems, recur across sites), so one task can move between sites. Six hubs carry the effects of actions across sites: WebMail receives receipts, notifications and two-factor codes, SecureBank records payments and trades, CalendarTodo receives bookings, MeridianCloud stores created files, QuickChat receives messages, and VaultGuard stores passwords. Twenty-one sites are seeded from public datasets; the other 44 hold synthetic data generated with fixed seeds. Thirteen sites are held out: no training data comes from them, and they are used only for evaluation (Sec.~\ref{sec:experiments}).

The gallery below shows the landing page of every site at the agents' viewport (1280$\times$800), grouped by domain. Each entry gives the site's name and identifier, its domain, split, data source, the number of human tasks that use the site, and the number of primitives the site supports, followed by the description from the site's documentation. The code release documents every page and API route of each site.

Tab.~\ref{tab:supp-licenses} lists the public sources and the licenses their publishers state; for a dataset taken from Kaggle, the license on its Kaggle page. Several sources carry share-alike or attribution duties, and some restrict use to research.

\begin{table}[H]
\centering
\footnotesize
\setlength{\tabcolsep}{4pt}
\begin{tabular}{@{}>{\raggedright\arraybackslash}p{0.27\linewidth}>{\raggedright\arraybackslash}p{0.4\linewidth}>{\raggedright\arraybackslash}p{0.27\linewidth}@{}}
\toprule
Site & Source & Stated license \\
\midrule
\texttt{academic-paper-db} & arXiv metadata & CC0 1.0 \\
\texttt{flights-hotels} & US airline routes and fares 1993--2024 (Kaggle; US DOT data); TBO hotels (Kaggle) & Apache 2.0; MIT \\
\texttt{health-fitness-tracking} & USDA FoodData Central; Fitbit data (Furberg et al., Zenodo) & CC0 1.0; CC BY 4.0 \\
\texttt{job-sites} & Job Description Dataset (Kaggle; synthetic postings) & CC0: public domain \\
\texttt{wikis} & English Wikipedia & CC BY-SA 4.0 \\
\texttt{news} & Wikinews & CC BY 2.5 / 4.0 \\
\texttt{dictionaries-language-tools} & Wiktionary (via wiktextract) & CC BY-SA 4.0 \\
\texttt{qa-knowledge} & Stack Exchange data dump (2024-04) & CC BY-SA 2.5--4.0 \\
\texttt{map-services} & OpenStreetMap & ODbL 1.0 \\
\texttt{books-comics} & Common Pile v0.1 (Pressbooks) & CC BY, CC BY-SA, CC0, or public domain, per book \\
\texttt{project-mgmt-issue-tracking} & Jira social repository & Apache 2.0 \\
\texttt{conference-review-submission} & PeerRead (ICLR 2017) & reviews CC BY 4.0; papers: author copyright \\
\texttt{visual-how-to-guides} & wikiHow via VisualHow & CC BY-NC-SA 3.0, research use \\
\texttt{email} & Enron email corpus (CMU) & released publicly by CMU \\
\texttt{e-commerce}, \texttt{auctions-p2p-} \texttt{marketplaces} & WebShop product data & none stated for the data \\
\texttt{forums}, \texttt{version-control} & WebArena Reddit and GitLab environments & none stated for the data \\
\texttt{software-marketplace} & Google Play Store apps (Kaggle) & CC BY-SA 4.0 \\
\texttt{real-estate-buy-rent} & USA Real Estate Dataset (Kaggle; realtor.com) & other (Kaggle description): educational use \\
\texttt{comparison-aggregators} & GSMArena phone specifications (GitHub dump) & GPL-3.0 (repository) \\
\bottomrule
\end{tabular}
\caption{\textbf{Public data sources} of the 21 seeded sites and their stated licenses. The other 44 sites use only synthetic data.}
\label{tab:supp-licenses}
\end{table}

\medskip
\noindent{\small A site's task count includes every task that uses it, so the 13 held-out sites' counts sum to 65: the 62 held-out tasks (those that start on a held-out site; 3 also use a training site) and 3 tasks that start on a training site and also use a held-out site, which neither split uses.}

\medskip
\noindent
\begin{minipage}[t]{0.32\linewidth}\raggedright
\includegraphics[width=\linewidth]{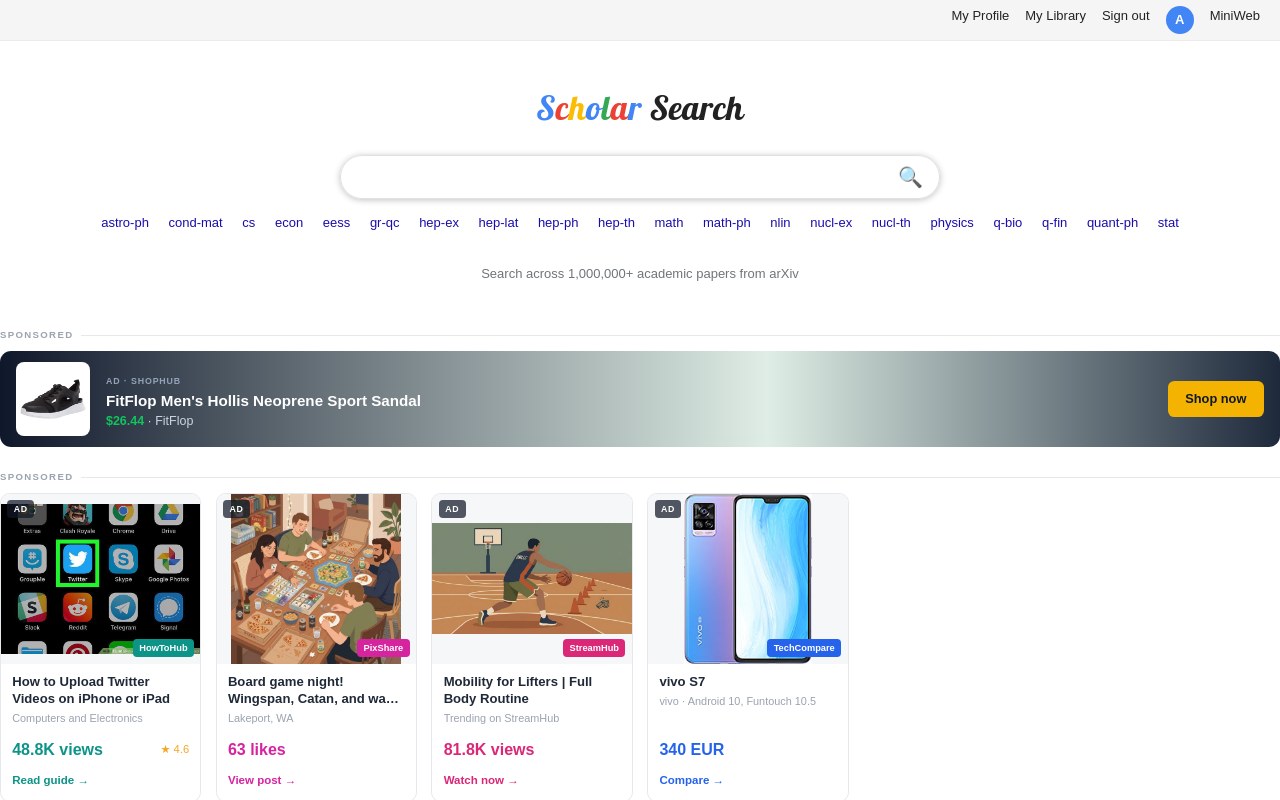}\\[1pt]
{\small\textbf{Scholar Search}\par}
{\scriptsize\color{black!65} \texttt{academic-paper-db} $\cdot$ Search \& reference $\cdot$ train $\cdot$ data: arXiv $\cdot$ 10 tasks $\cdot$ 7 primitives\par}\vspace{1pt}
{\scriptsize An academic paper search engine over arXiv metadata, modeled on Google Scholar, Semantic Scholar and arXiv.org. Users search and browse papers, open paper and author pages, compare two papers side by side, and keep a library of saved papers and followed authors.\par}
\end{minipage}\hfill\begin{minipage}[t]{0.32\linewidth}\raggedright
\includegraphics[width=\linewidth]{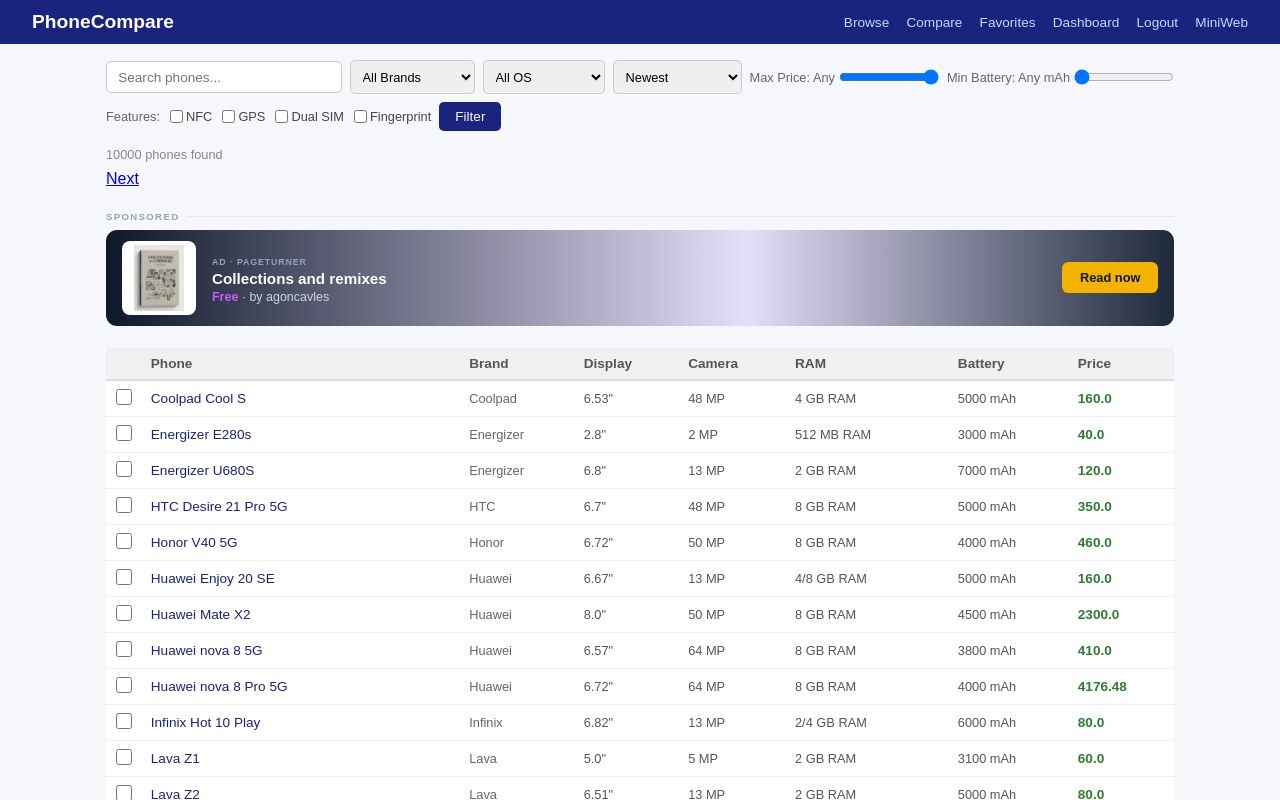}\\[1pt]
{\small\textbf{PhoneCompare}\par}
{\scriptsize\color{black!65} \texttt{comparison-aggregators} $\cdot$ Search \& reference $\cdot$ \textbf{held-out} $\cdot$ data: GSMArena $\cdot$ 3 tasks $\cdot$ 8 primitives\par}\vspace{1pt}
{\scriptsize A GSMArena-style phone catalog and comparison site. Users filter and sort phones by specs, open spec sheets, compare two phones side by side, and keep a favorites list and a compare list.\par}
\end{minipage}\hfill\begin{minipage}[t]{0.32\linewidth}\raggedright
\includegraphics[width=\linewidth]{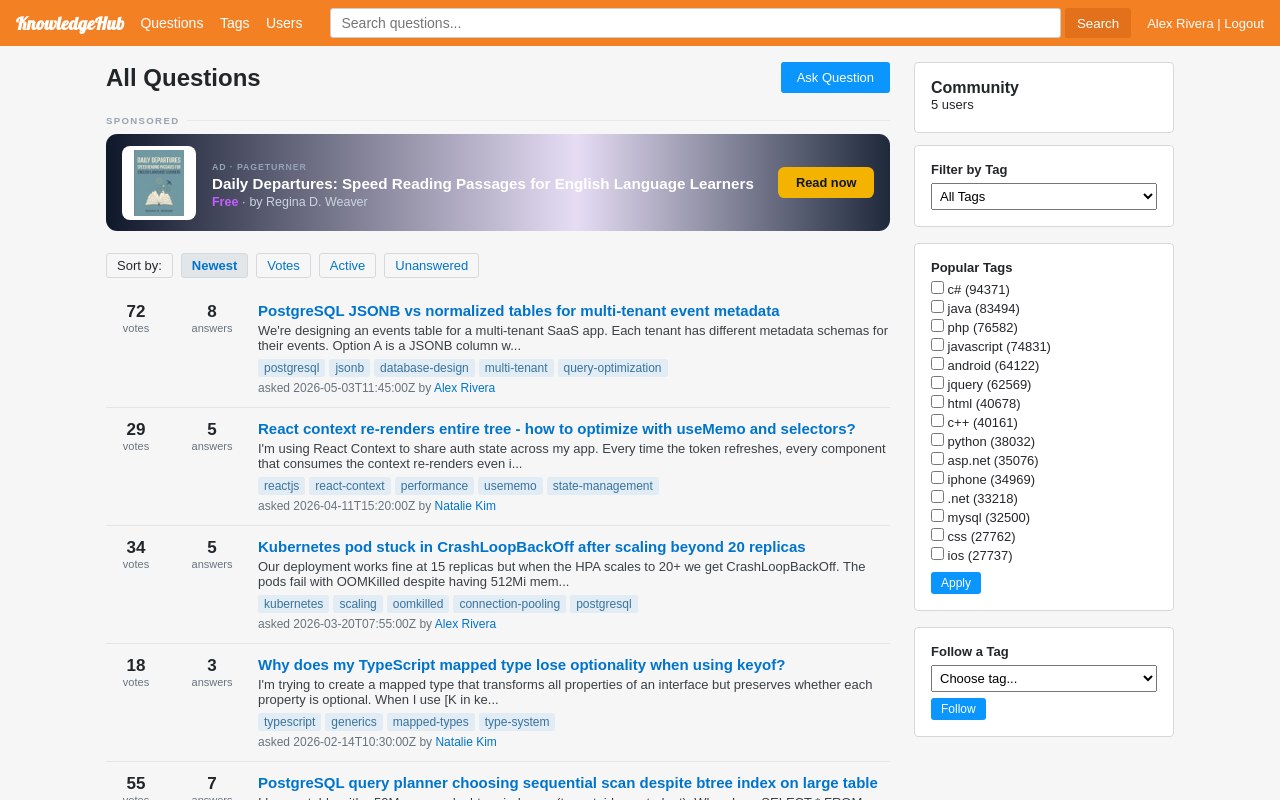}\\[1pt]
{\small\textbf{KnowledgeHub}\par}
{\scriptsize\color{black!65} \texttt{qa-knowledge} $\cdot$ Search \& reference $\cdot$ train $\cdot$ data: StackExchange $\cdot$ 6 tasks $\cdot$ 11 primitives\par}\vspace{1pt}
{\scriptsize A Stack Overflow-style programming Q\&A site built from a Stack Exchange data dump. Users search and browse questions by tag, read answer threads, vote, answer, comment, accept answers, save questions, follow tags, share and report content, and ask or edit their own questions.\par}
\end{minipage}
\par\bigskip\noindent
\begin{minipage}[t]{0.32\linewidth}\raggedright
\includegraphics[width=\linewidth]{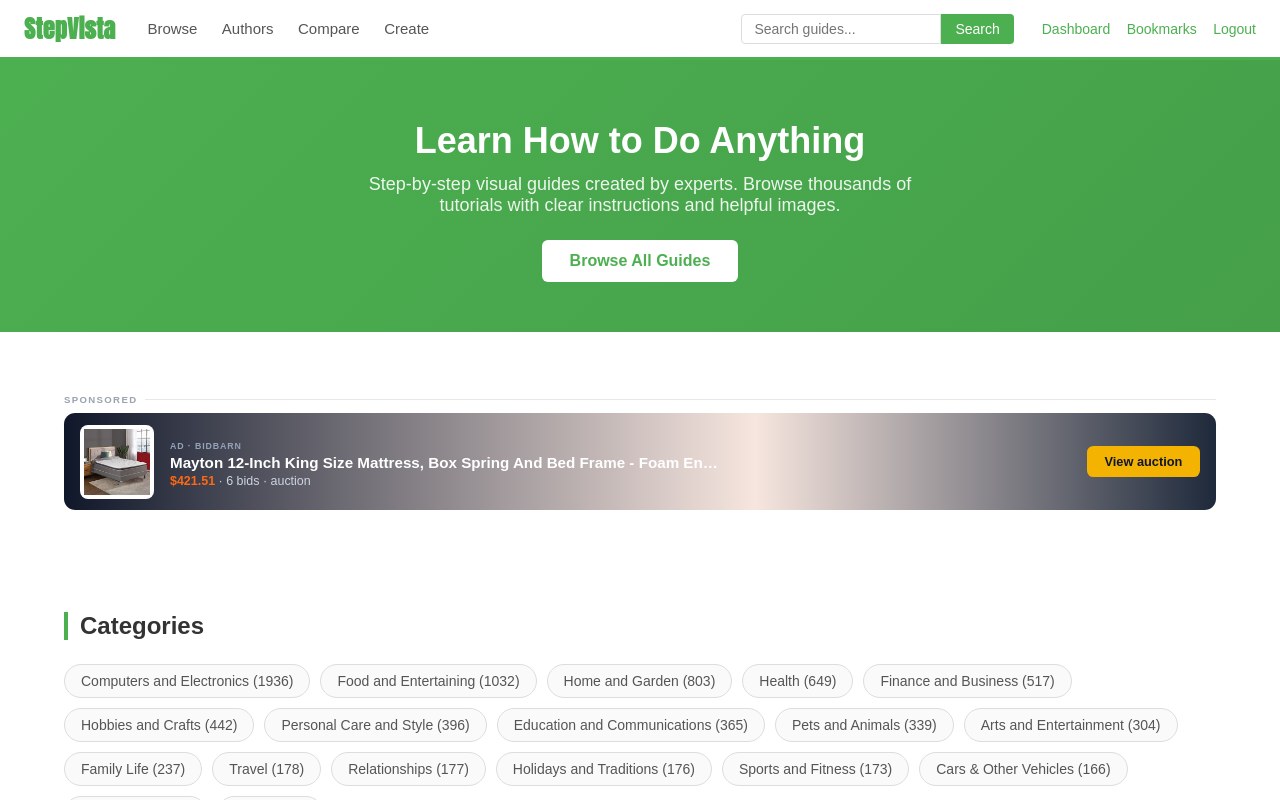}\\[1pt]
{\small\textbf{StepVista How-To Guides}\par}
{\scriptsize\color{black!65} \texttt{visual-how-to-guides} $\cdot$ Search \& reference $\cdot$ train $\cdot$ data: WikiHow $\cdot$ 4 tasks $\cdot$ 9 primitives\par}\vspace{1pt}
{\scriptsize A wikiHow / Instructables-style site of illustrated step-by-step guides (home, cooking, tech and more). Users browse, filter and compare guides, step through them one step at a time, rate and comment, bookmark guides and follow authors.\par}
\end{minipage}\hfill\begin{minipage}[t]{0.32\linewidth}\raggedright
\includegraphics[width=\linewidth]{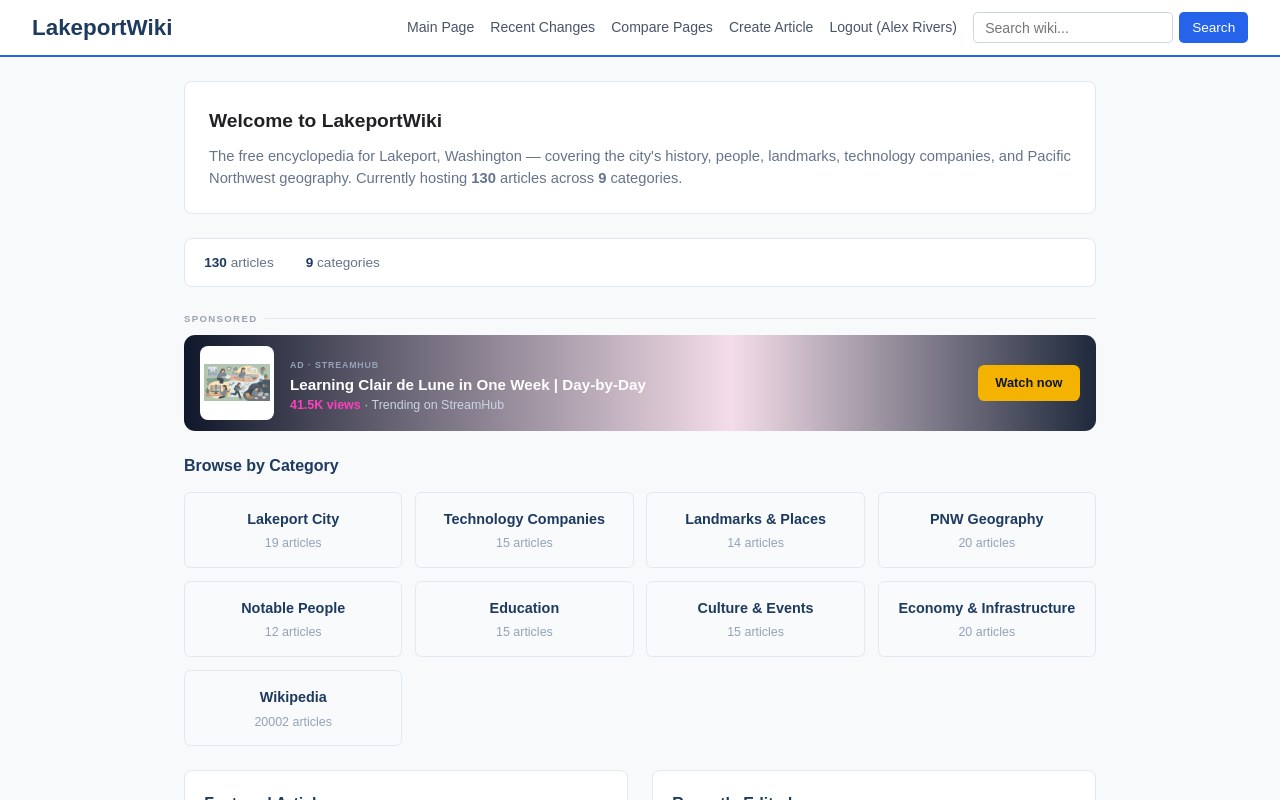}\\[1pt]
{\small\textbf{LakeportWiki}\par}
{\scriptsize\color{black!65} \texttt{wikis} $\cdot$ Search \& reference $\cdot$ train $\cdot$ data: Wikipedia $\cdot$ 4 tasks $\cdot$ 5 primitives\par}\vspace{1pt}
{\scriptsize A Wikipedia/MediaWiki-style encyclopedia about the fictional city of Lakeport, WA and the Pacific Northwest, mixed with real Wikipedia articles. Users read, search, compare, create and edit articles and browse revision history, diffs and recent changes.\par}
\end{minipage}\hfill\begin{minipage}[t]{0.32\linewidth}\raggedright
\includegraphics[width=\linewidth]{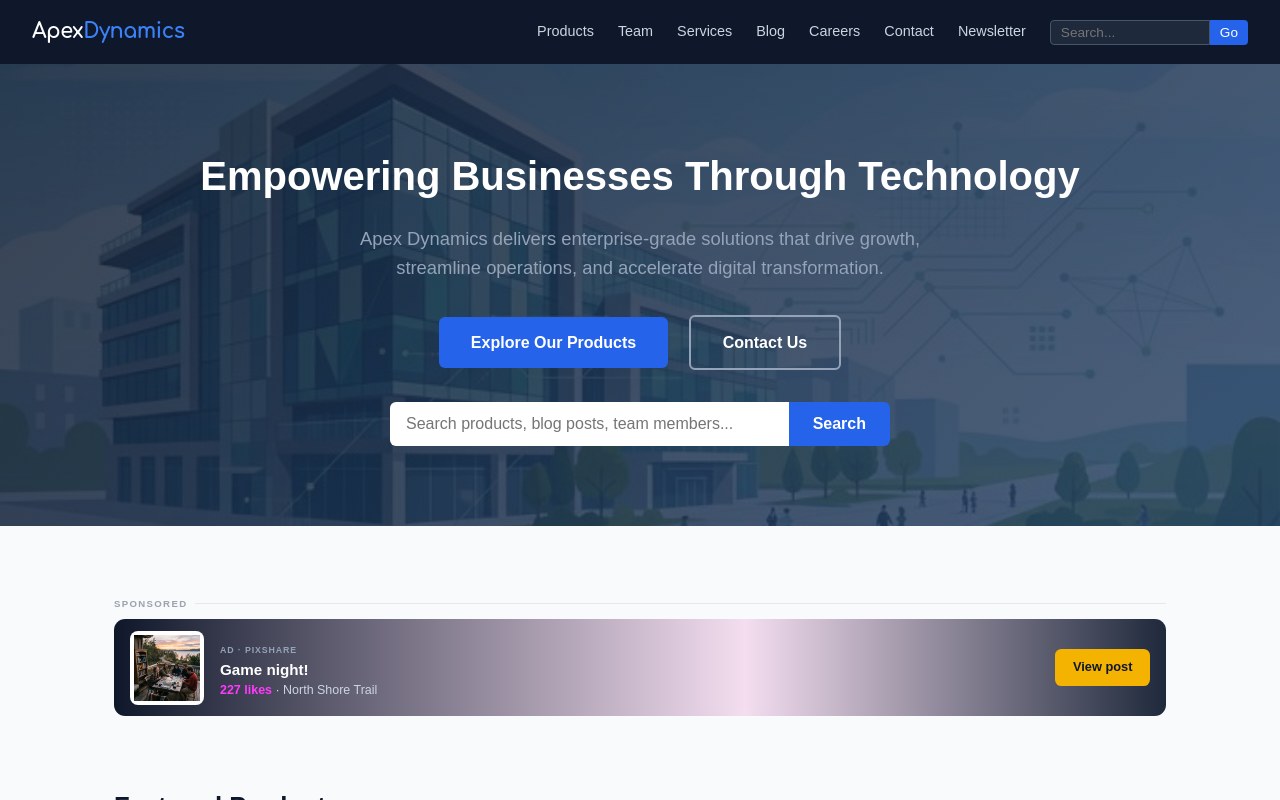}\\[1pt]
{\small\textbf{Apex Dynamics Corp}\par}
{\scriptsize\color{black!65} \texttt{business-company} $\cdot$ Informational $\cdot$ train $\cdot$ synthetic data $\cdot$ 5 tasks $\cdot$ 5 primitives\par}\vspace{1pt}
{\scriptsize A corporate marketing website for a fictional software company: products, services, team, blog, careers, contact form and newsletter. Its product line is the ``Meridian'' suite (MeridianFlow, MeridianVault and others).\par}
\end{minipage}
\par\bigskip\noindent
\begin{minipage}[t]{0.32\linewidth}\raggedright
\includegraphics[width=\linewidth]{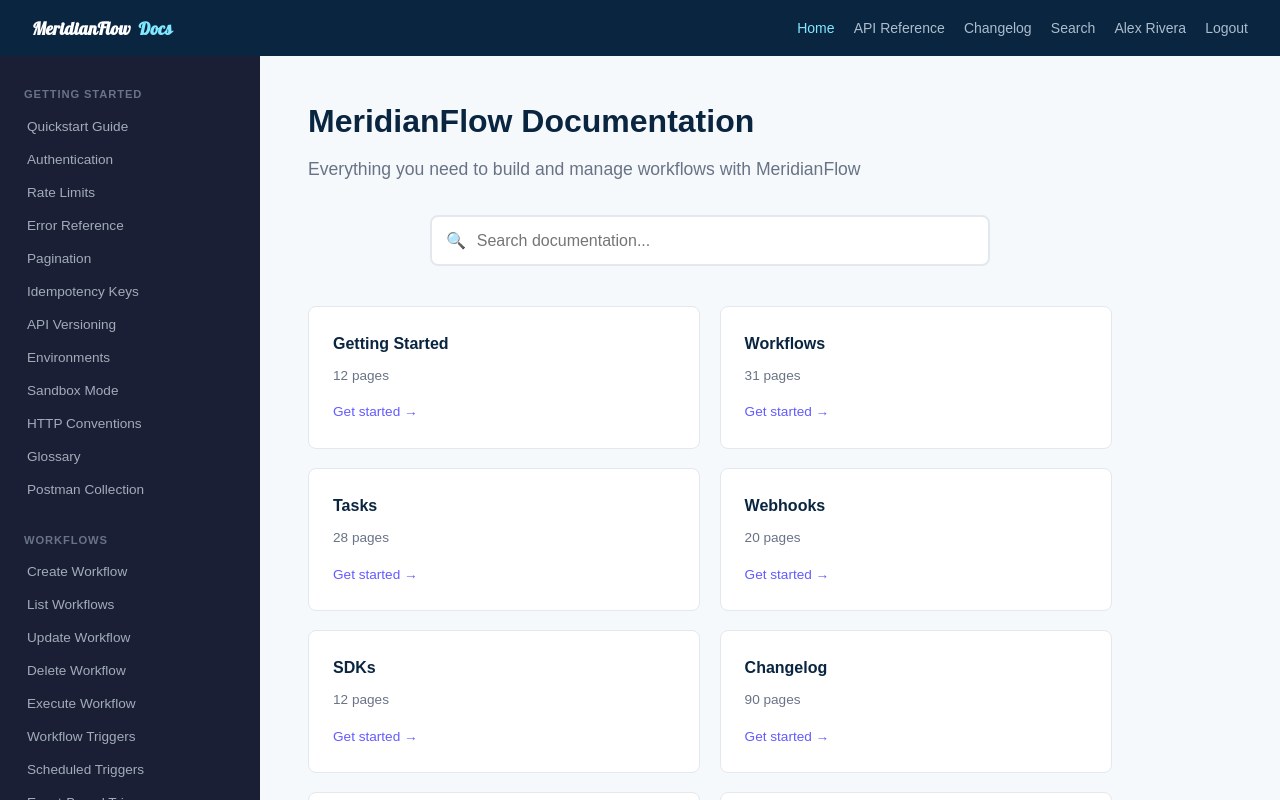}\\[1pt]
{\small\textbf{MeridianFlow Docs}\par}
{\scriptsize\color{black!65} \texttt{documentation-api-docs} $\cdot$ Informational $\cdot$ train $\cdot$ synthetic data $\cdot$ 3 tasks $\cdot$ 4 primitives\par}\vspace{1pt}
{\scriptsize A developer documentation site modeled after Stripe Docs, for the fictional MeridianFlow workflow API. It has guides, an API reference with method badges, a changelog, full-text search, and bookmarks for signed-in users.\par}
\end{minipage}\hfill\begin{minipage}[t]{0.32\linewidth}\raggedright
\includegraphics[width=\linewidth]{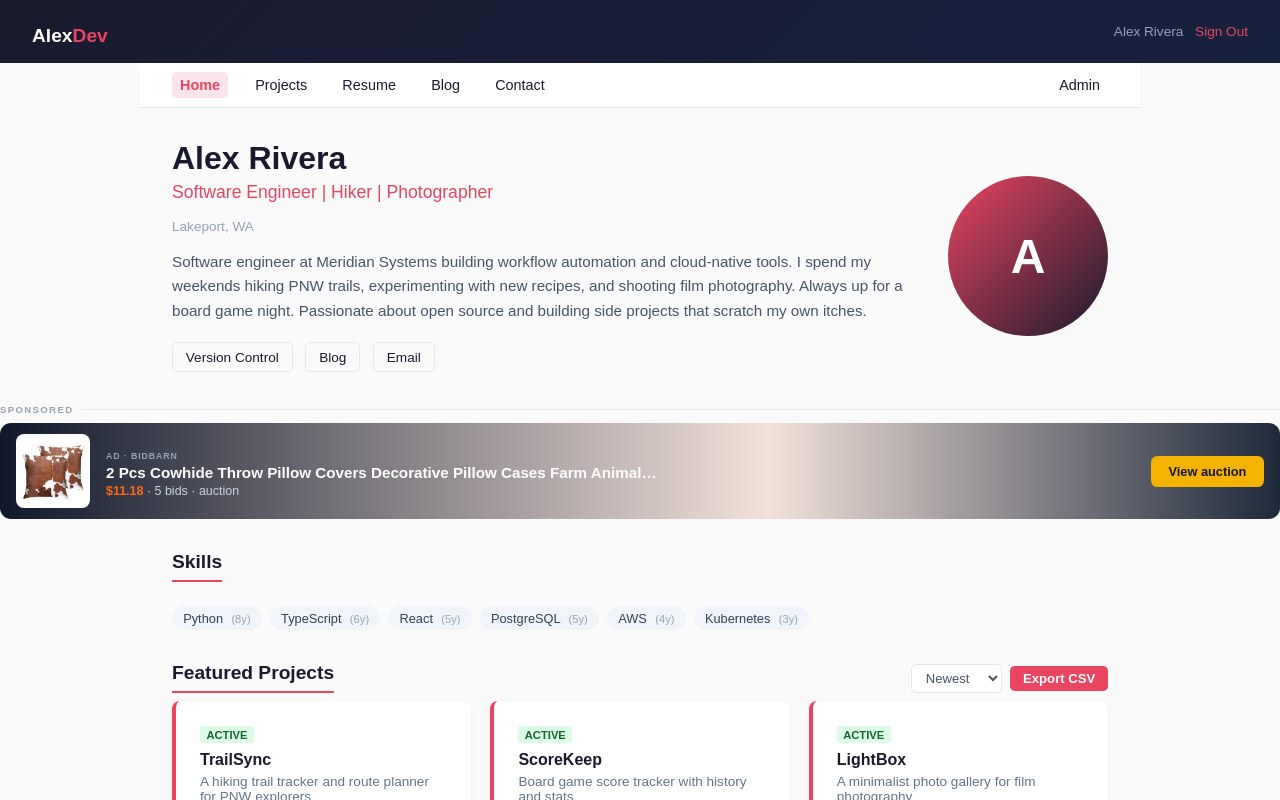}\\[1pt]
{\small\textbf{Alex Rivera}\par}
{\scriptsize\color{black!65} \texttt{personal-portfolio} $\cdot$ Informational $\cdot$ train $\cdot$ synthetic data $\cdot$ 2 tasks $\cdot$ 6 primitives\par}\vspace{1pt}
{\scriptsize The personal developer portfolio of Alex Rivera, the shared persona, in the style of a GitHub Pages / Vercel personal site. It shows a profile, project gallery, structured resume, skills table and blog links, with a contact form and site-wide search.\par}
\end{minipage}\hfill\begin{minipage}[t]{0.32\linewidth}\raggedright
\includegraphics[width=\linewidth]{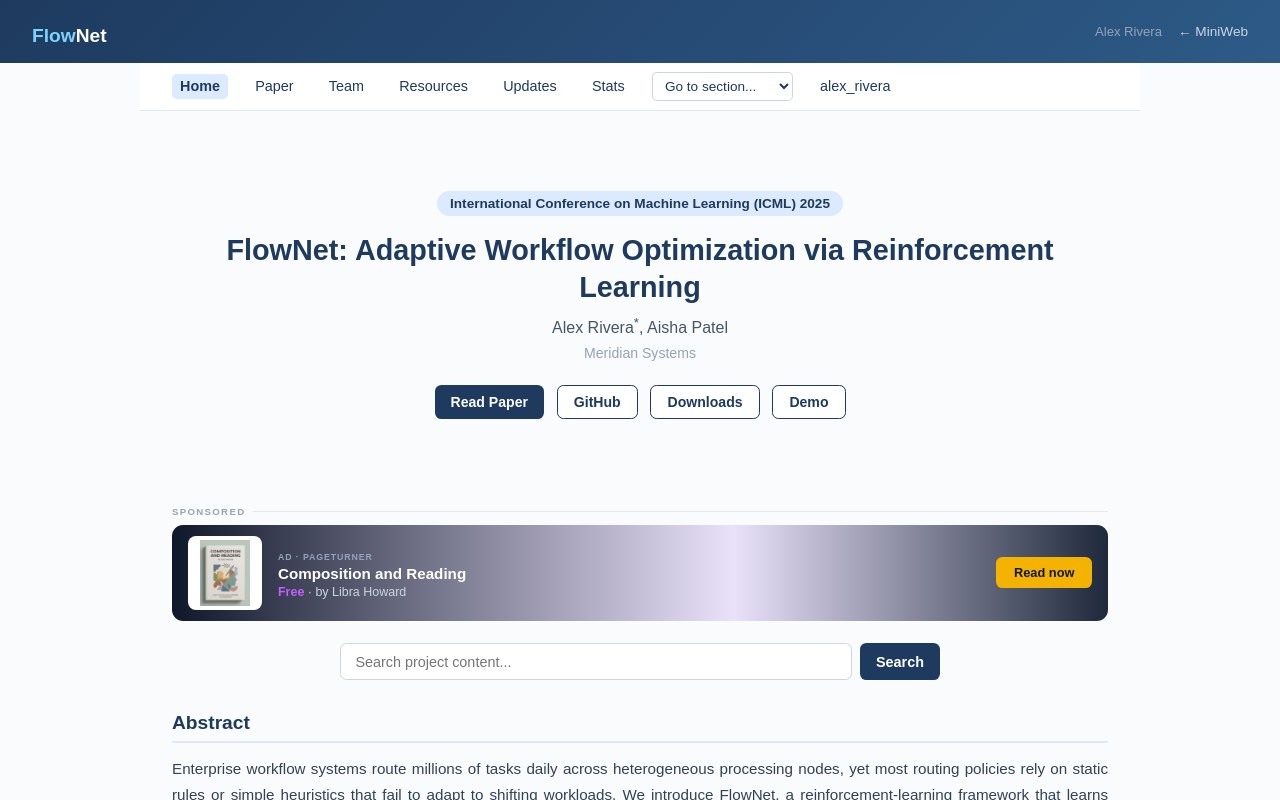}\\[1pt]
{\small\textbf{FlowNet Project Homepage}\par}
{\scriptsize\color{black!65} \texttt{project-homepages} $\cdot$ Informational $\cdot$ train $\cdot$ synthetic data $\cdot$ 3 tasks $\cdot$ 4 primitives\par}\vspace{1pt}
{\scriptsize An academic project landing page for the FlowNet paper (ICML 2025), in the style of nerfies.github.io and other paper homepages. It shows the paper sections, team, downloadable resources, news updates, statistics and citation export.\par}
\end{minipage}
\par\bigskip\noindent
\begin{minipage}[t]{0.32\linewidth}\raggedright
\includegraphics[width=\linewidth]{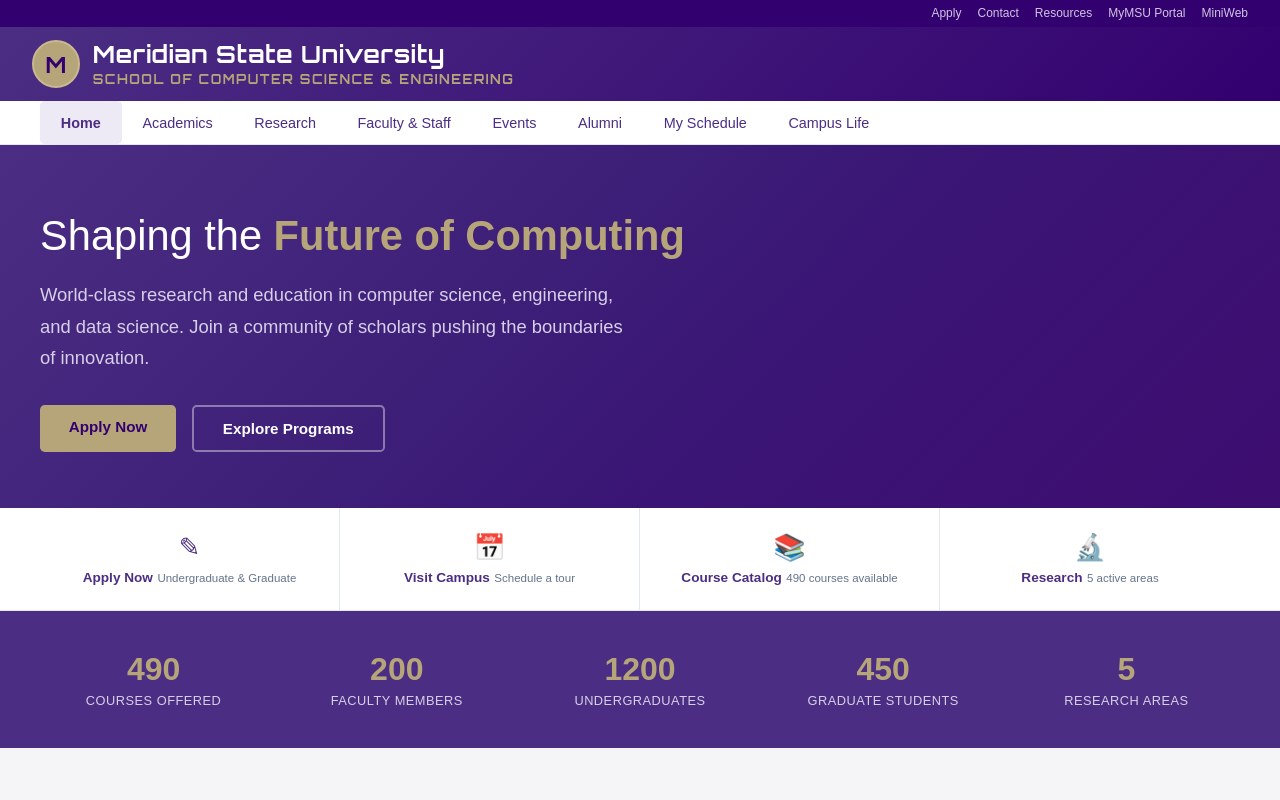}\\[1pt]
{\small\textbf{Meridian State University}\par}
{\scriptsize\color{black!65} \texttt{university-academic} $\cdot$ Informational $\cdot$ train $\cdot$ synthetic data $\cdot$ 6 tasks $\cdot$ 8 primitives\par}\vspace{1pt}
{\scriptsize A university website and student portal: course catalog, faculty directory, research areas, events, alumni network, admissions application and a ``My Schedule'' course registration view, plus an editable course gradebook.\par}
\end{minipage}\hfill\begin{minipage}[t]{0.32\linewidth}\raggedright
\includegraphics[width=\linewidth]{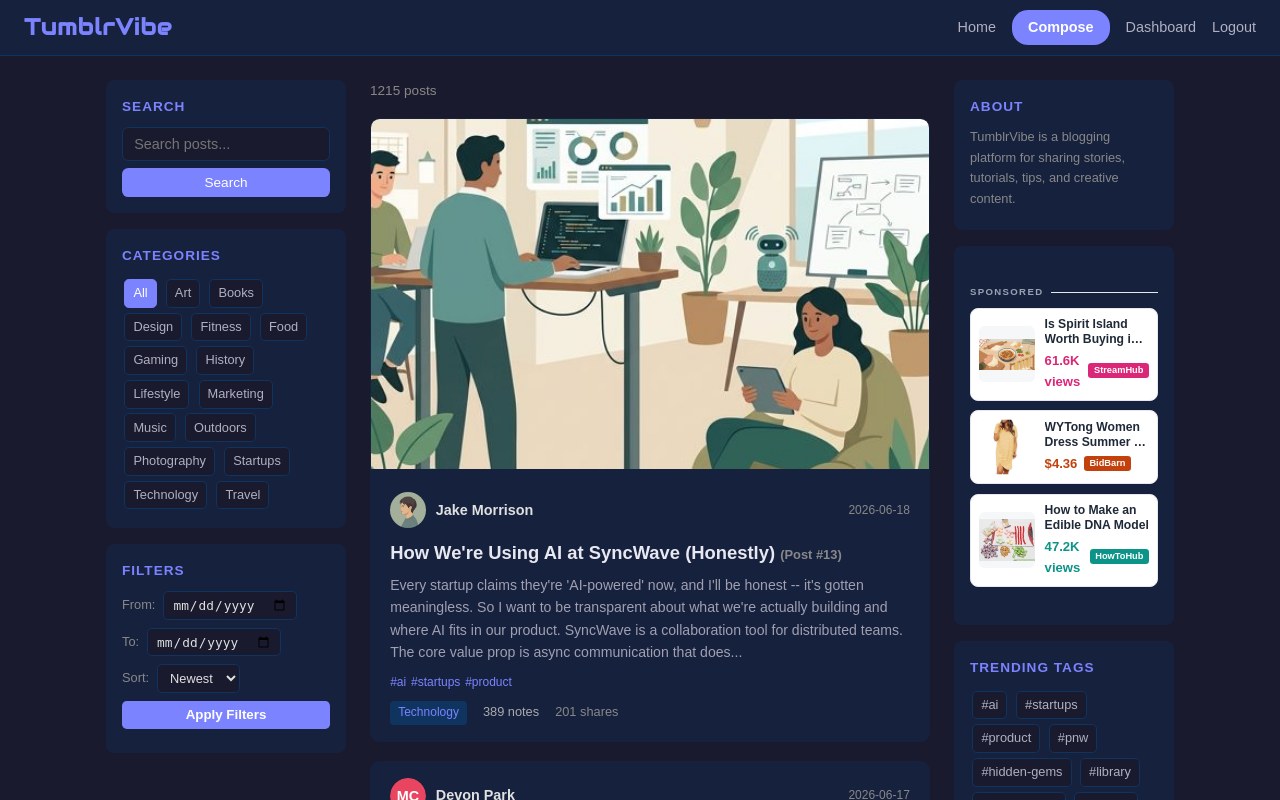}\\[1pt]
{\small\textbf{TumblrVibe Blogging Platform}\par}
{\scriptsize\color{black!65} \texttt{blogs} $\cdot$ News \& feeds $\cdot$ train $\cdot$ synthetic data $\cdot$ 8 tasks $\cdot$ 9 primitives\par}\vspace{1pt}
{\scriptsize A Tumblr-style blogging platform with posts (articles, tips, stories, tutorials), comments, tags, likes and reblogs. Users browse and filter the feed, write posts, comment, follow authors, save posts and subscribe to tags.\par}
\end{minipage}\hfill\begin{minipage}[t]{0.32\linewidth}\raggedright
\includegraphics[width=\linewidth]{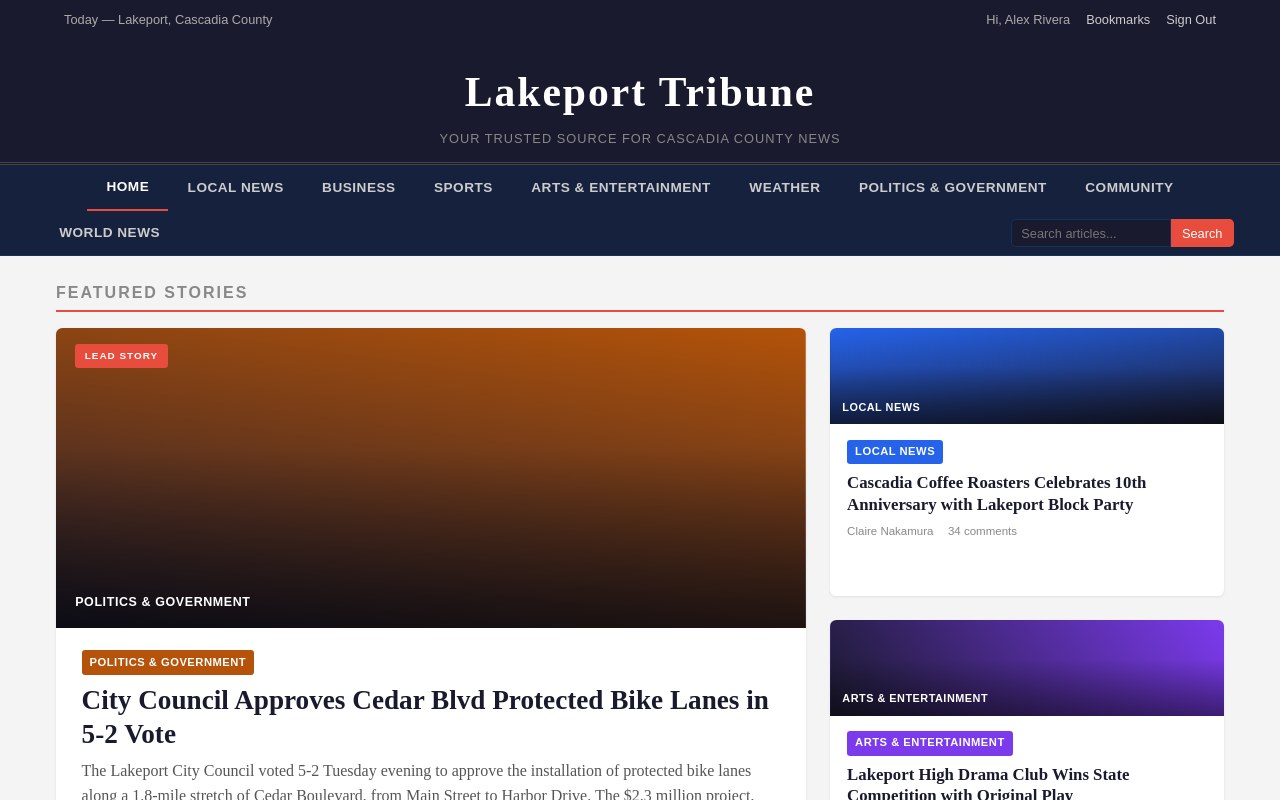}\\[1pt]
{\small\textbf{Lakeport Tribune}\par}
{\scriptsize\color{black!65} \texttt{news} $\cdot$ News \& feeds $\cdot$ train $\cdot$ data: Wikinews $\cdot$ 4 tasks $\cdot$ 9 primitives\par}\vspace{1pt}
{\scriptsize A local newspaper site for the fictional city of Lakeport in Cascadia County, modeled on Patch.com and small-city news portals. Readers browse category pages, search articles, listen to articles, comment, bookmark, share, follow authors or categories, and report articles.\par}
\end{minipage}
\par\bigskip\noindent
\begin{minipage}[t]{0.32\linewidth}\raggedright
\includegraphics[width=\linewidth]{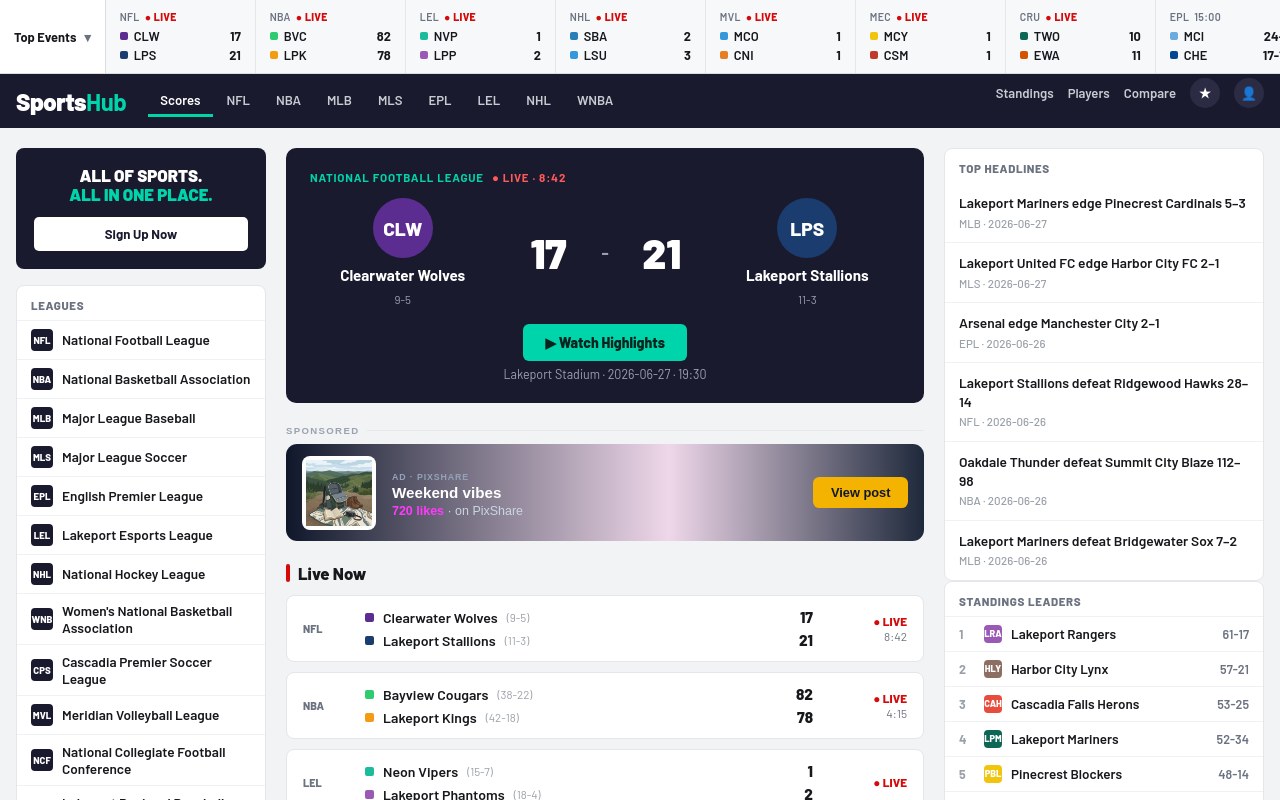}\\[1pt]
{\small\textbf{Lakeport Sports Hub}\par}
{\scriptsize\color{black!65} \texttt{sports-esports} $\cdot$ News \& feeds $\cdot$ train $\cdot$ synthetic data $\cdot$ 6 tasks $\cdot$ 10 primitives\par}\vspace{1pt}
{\scriptsize A multi-sport scoreboard in the style of ESPN / Flashscore covering the NFL, NBA, MLB, MLS, Premier League and a local esports league. It shows live, final and scheduled matches, standings, team and player pages, a team comparison tool, match highlights, comments and a favorites list.\par}
\end{minipage}\hfill\begin{minipage}[t]{0.32\linewidth}\raggedright
\includegraphics[width=\linewidth]{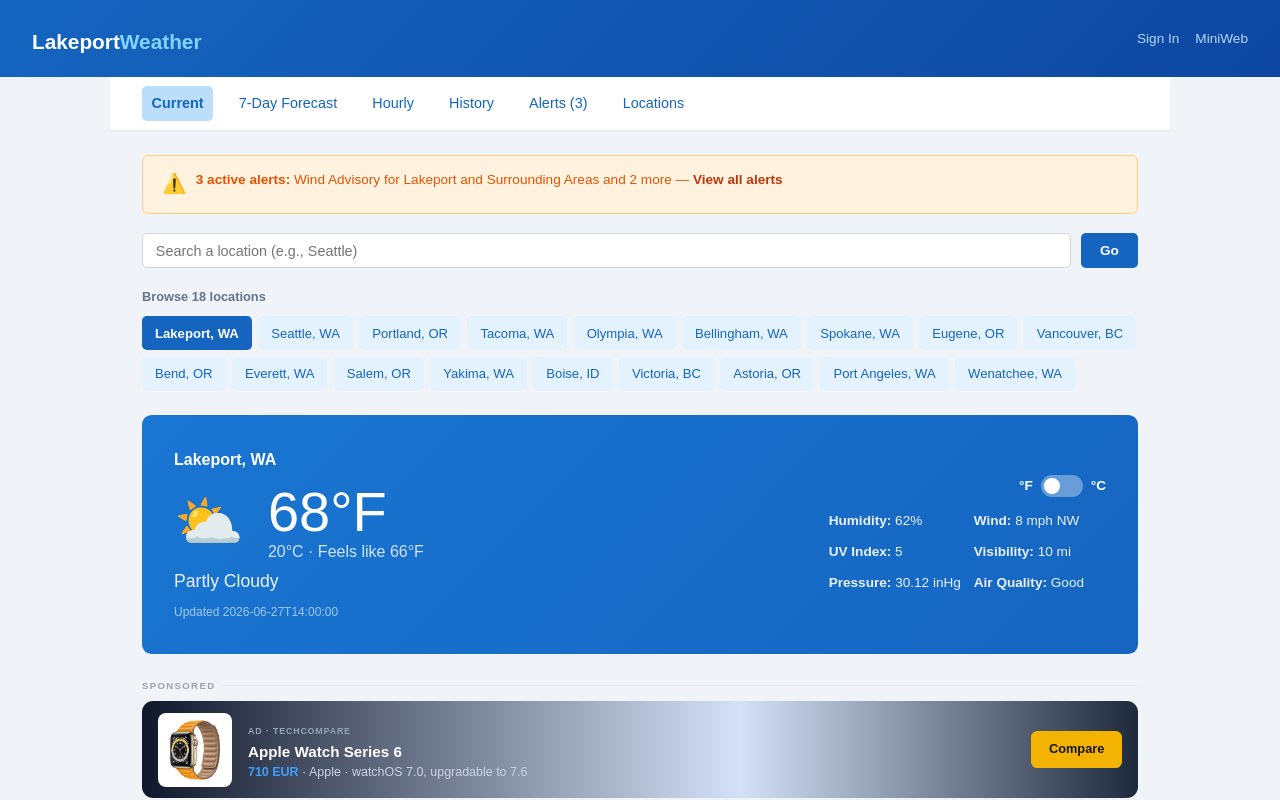}\\[1pt]
{\small\textbf{Lakeport Weather}\par}
{\scriptsize\color{black!65} \texttt{weather} $\cdot$ News \& feeds $\cdot$ \textbf{held-out} $\cdot$ synthetic data $\cdot$ 5 tasks $\cdot$ 6 primitives\par}\vspace{1pt}
{\scriptsize A local weather portal for the fictional city of Lakeport, WA and nearby Pacific Northwest stations, modeled on weather.gov and Weather Underground. It shows current conditions, forecasts, a multi-year history archive and active alerts, and lets a signed-in user keep a list of saved locations.\par}
\end{minipage}\hfill\begin{minipage}[t]{0.32\linewidth}\raggedright
\includegraphics[width=\linewidth]{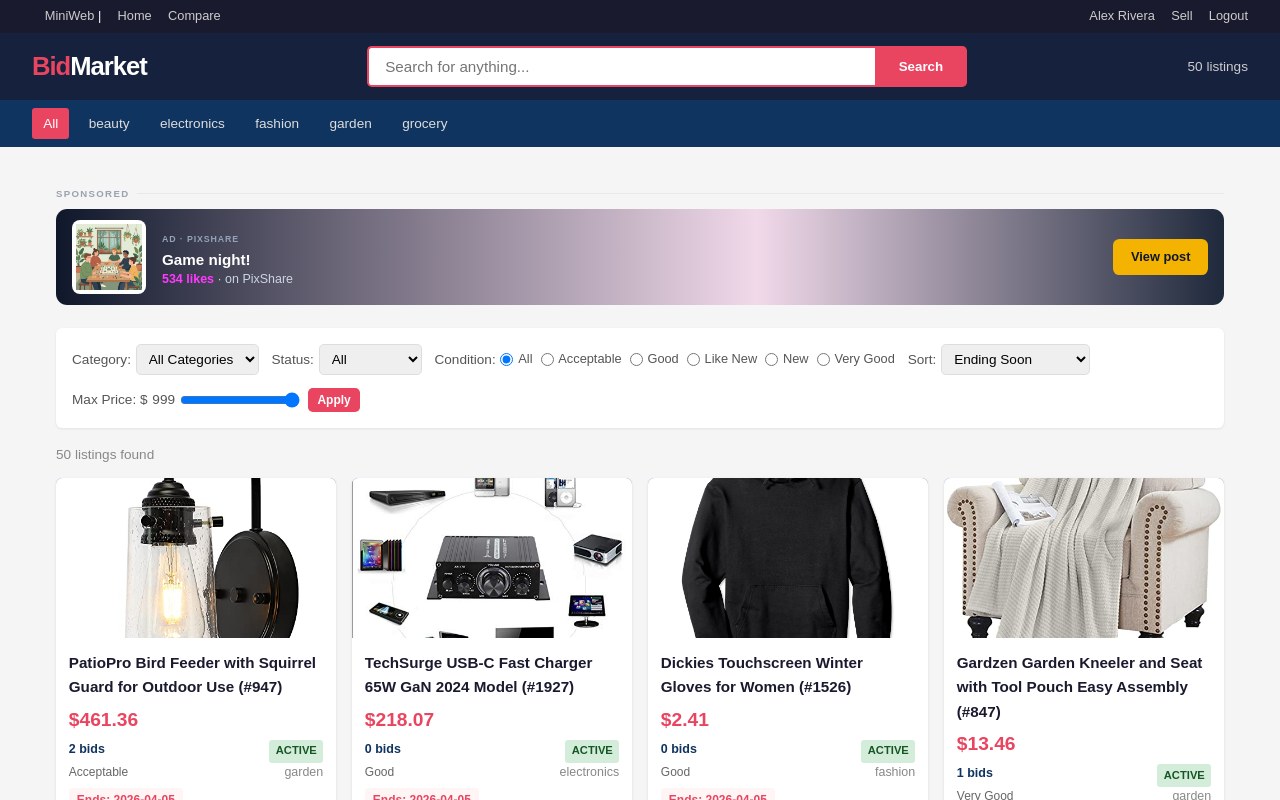}\\[1pt]
{\small\textbf{BidMarket}\par}
{\scriptsize\color{black!65} \texttt{auctions-p2p-marketplaces} $\cdot$ Shopping \& transactions $\cdot$ train $\cdot$ data: WebShop $\cdot$ 5 tasks $\cdot$ 16 primitives\par}\vspace{1pt}
{\scriptsize An eBay-style auction and peer-to-peer marketplace. Users browse and filter listings, bid or buy now, watch and save items, follow sellers, message sellers, and sell their own items.\par}
\end{minipage}
\par\bigskip\noindent
\begin{minipage}[t]{0.32\linewidth}\raggedright
\includegraphics[width=\linewidth]{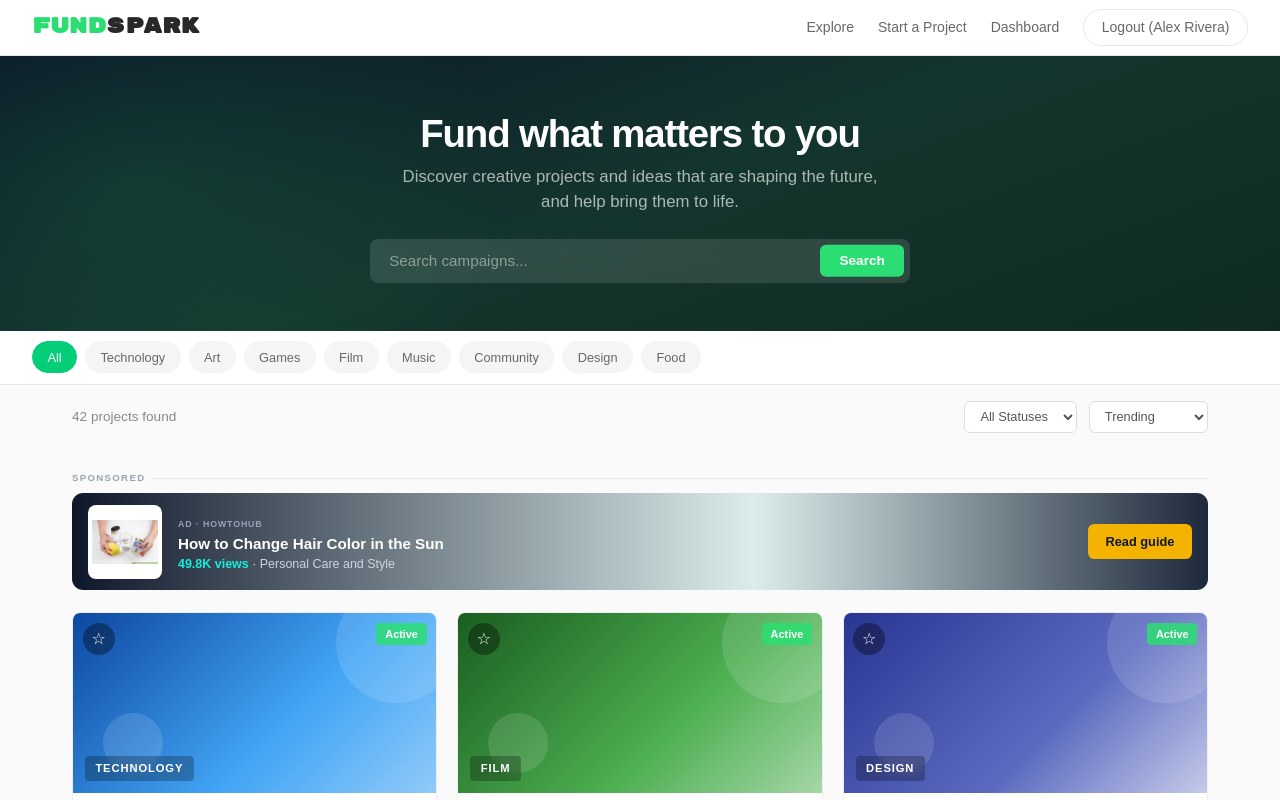}\\[1pt]
{\small\textbf{FundSpark}\par}
{\scriptsize\color{black!65} \texttt{crowdfunding-donations} $\cdot$ Shopping \& transactions $\cdot$ \textbf{held-out} $\cdot$ synthetic data $\cdot$ 6 tasks $\cdot$ 12 primitives\par}\vspace{1pt}
{\scriptsize A Kickstarter/GoFundMe-style crowdfunding site. Users browse and search campaigns, back a project through a reward-tier checkout, start their own campaign, post updates, and save, follow, subscribe to or share campaigns.\par}
\end{minipage}\hfill\begin{minipage}[t]{0.32\linewidth}\raggedright
\includegraphics[width=\linewidth]{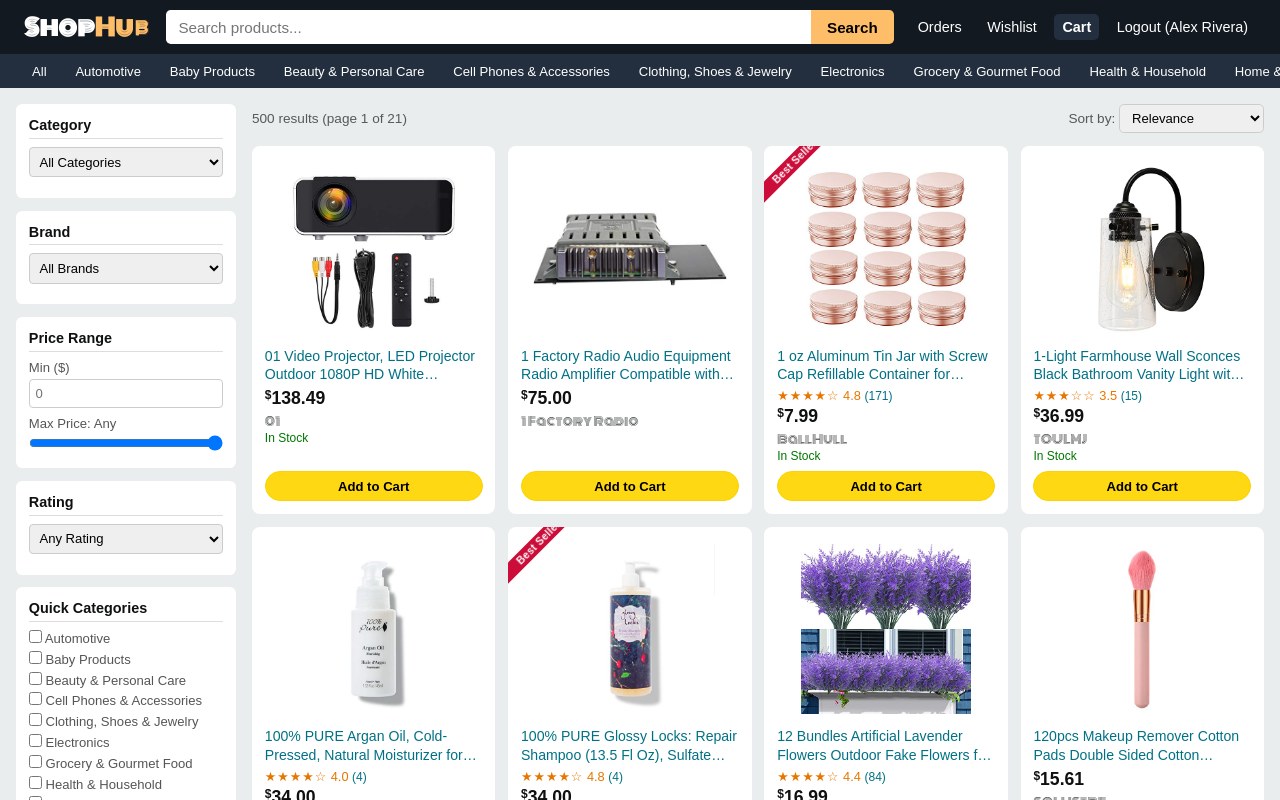}\\[1pt]
{\small\textbf{ShopHub}\par}
{\scriptsize\color{black!65} \texttt{e-commerce} $\cdot$ Shopping \& transactions $\cdot$ train $\cdot$ data: WebShop $\cdot$ 6 tasks $\cdot$ 13 primitives\par}\vspace{1pt}
{\scriptsize An Amazon or eBay-style online store. Users search and filter a large product catalog, read and write reviews, keep a cart and a wishlist, check out, and cancel or return orders.\par}
\end{minipage}\hfill\begin{minipage}[t]{0.32\linewidth}\raggedright
\includegraphics[width=\linewidth]{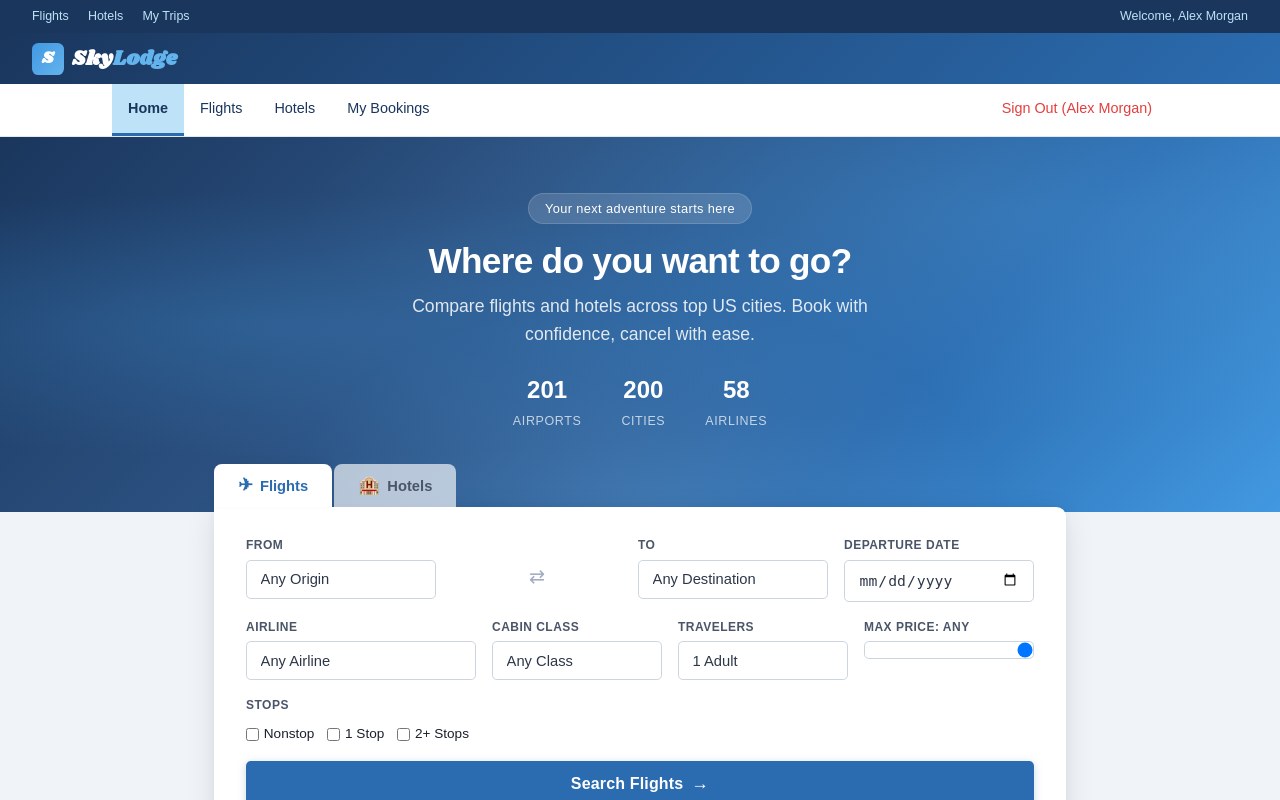}\\[1pt]
{\small\textbf{SkyLodge Travel}\par}
{\scriptsize\color{black!65} \texttt{flights-hotels} $\cdot$ Shopping \& transactions $\cdot$ \textbf{held-out} $\cdot$ data: US flights, TBO hotels $\cdot$ 5 tasks $\cdot$ 13 primitives\par}\vspace{1pt}
{\scriptsize A flight and hotel booking site in the style of Expedia / Kayak, covering major US cities. Users search and filter flights and hotels, book a flight through a multi-step flow (travelers, seat map, review, payment), book hotel stays, and view or cancel their bookings.\par}
\end{minipage}
\par\bigskip\noindent
\begin{minipage}[t]{0.32\linewidth}\raggedright
\includegraphics[width=\linewidth]{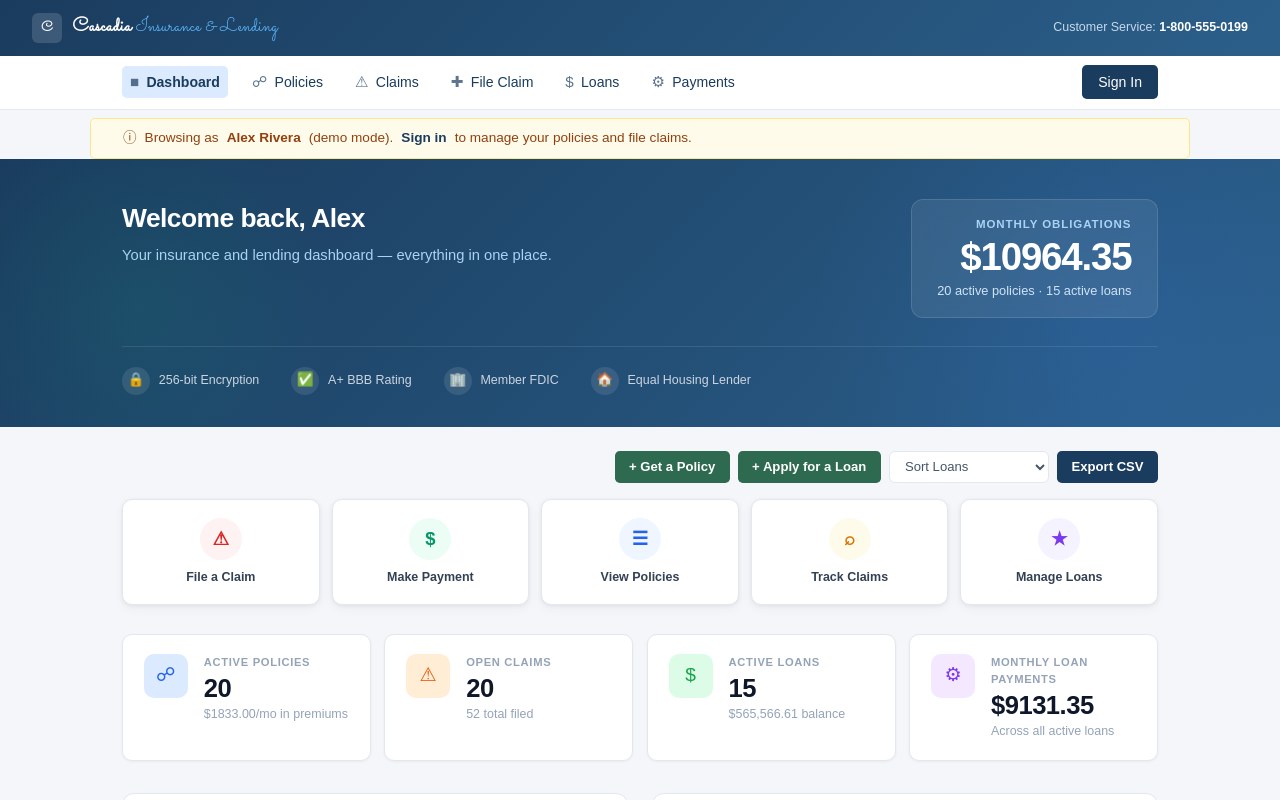}\\[1pt]
{\small\textbf{Cascadia Insurance \& Lending}\par}
{\scriptsize\color{black!65} \texttt{insurance-loans} $\cdot$ Shopping \& transactions $\cdot$ train $\cdot$ synthetic data $\cdot$ 8 tasks $\cdot$ 14 primitives\par}\vspace{1pt}
{\scriptsize An insurance and lending customer portal in the style of Progressive or SoFi. Users review policies, claims, loans and payments, file claims, buy a policy or apply for a loan, sign the agreement, make payments and download policy documents.\par}
\end{minipage}\hfill\begin{minipage}[t]{0.32\linewidth}\raggedright
\includegraphics[width=\linewidth]{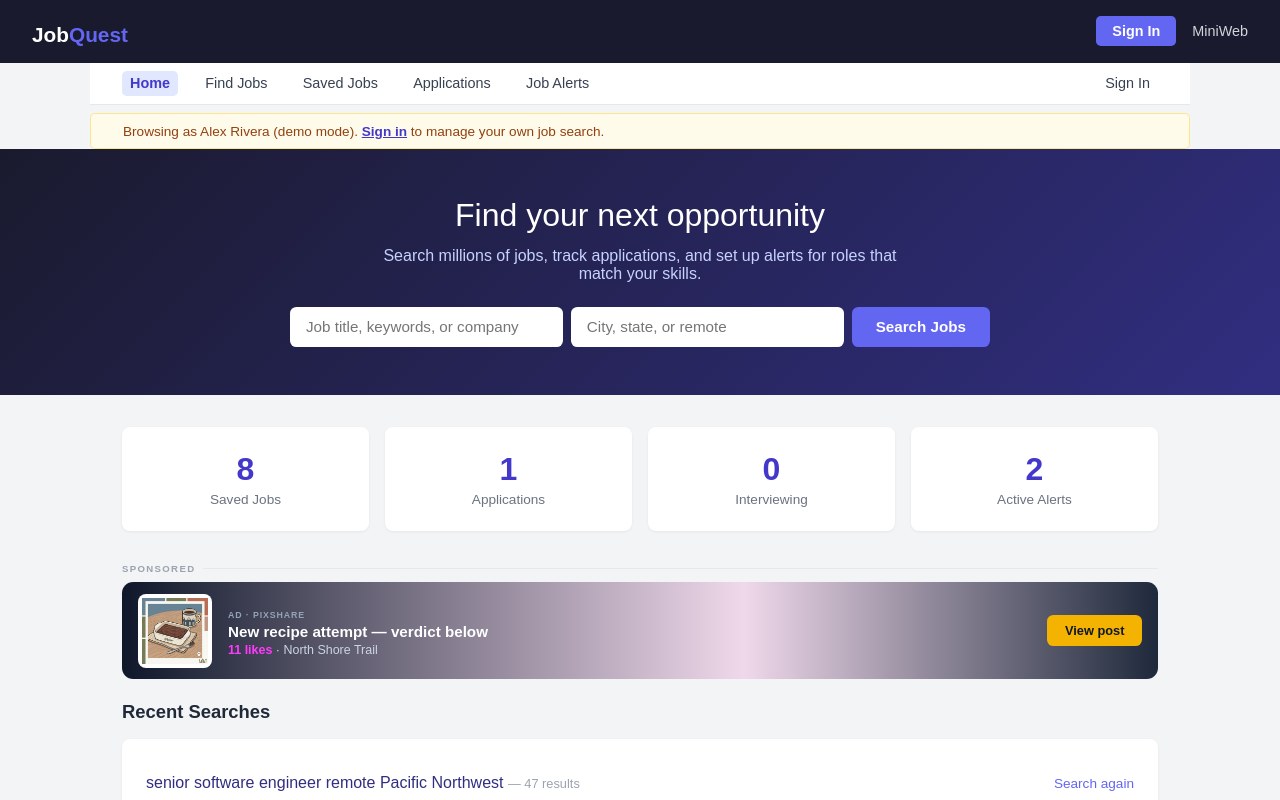}\\[1pt]
{\small\textbf{JobQuest}\par}
{\scriptsize\color{black!65} \texttt{job-sites} $\cdot$ Shopping \& transactions $\cdot$ train $\cdot$ data: Kaggle Job Dataset $\cdot$ 4 tasks $\cdot$ 11 primitives\par}\vspace{1pt}
{\scriptsize An Indeed or LinkedIn Jobs-style job board and application tracker. Users search and filter job listings, save jobs, follow companies, apply with a resume and cover letter, track applications and manage job alerts.\par}
\end{minipage}\hfill\begin{minipage}[t]{0.32\linewidth}\raggedright
\includegraphics[width=\linewidth]{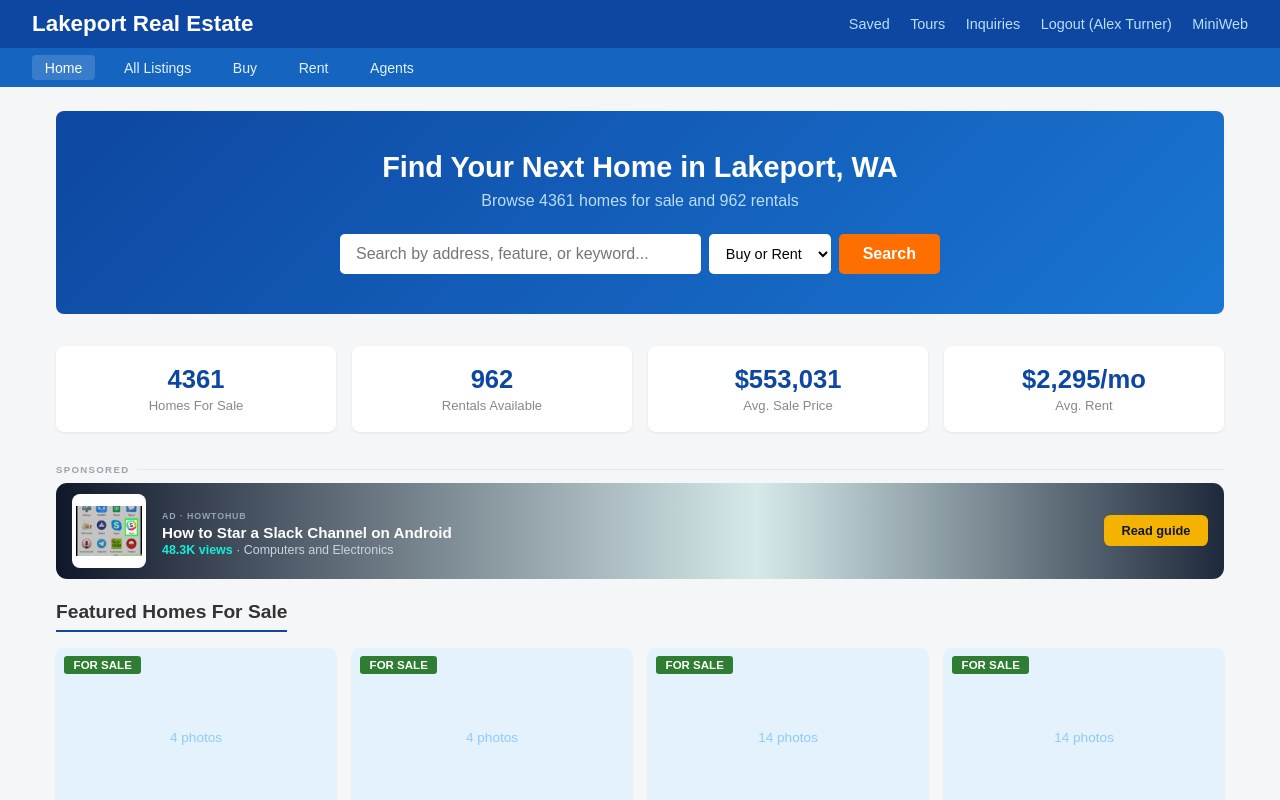}\\[1pt]
{\small\textbf{Lakeport Real Estate}\par}
{\scriptsize\color{black!65} \texttt{real-estate-buy-rent} $\cdot$ Shopping \& transactions $\cdot$ \textbf{held-out} $\cdot$ data: Zillow, Realtor.com $\cdot$ 5 tasks $\cdot$ 10 primitives\par}\vspace{1pt}
{\scriptsize A Zillow/Redfin-style home search site for Lakeport, WA. Users search and filter homes for sale or rent, save listings, contact agents, schedule tours and follow agents.\par}
\end{minipage}
\par\bigskip\noindent
\begin{minipage}[t]{0.32\linewidth}\raggedright
\includegraphics[width=\linewidth]{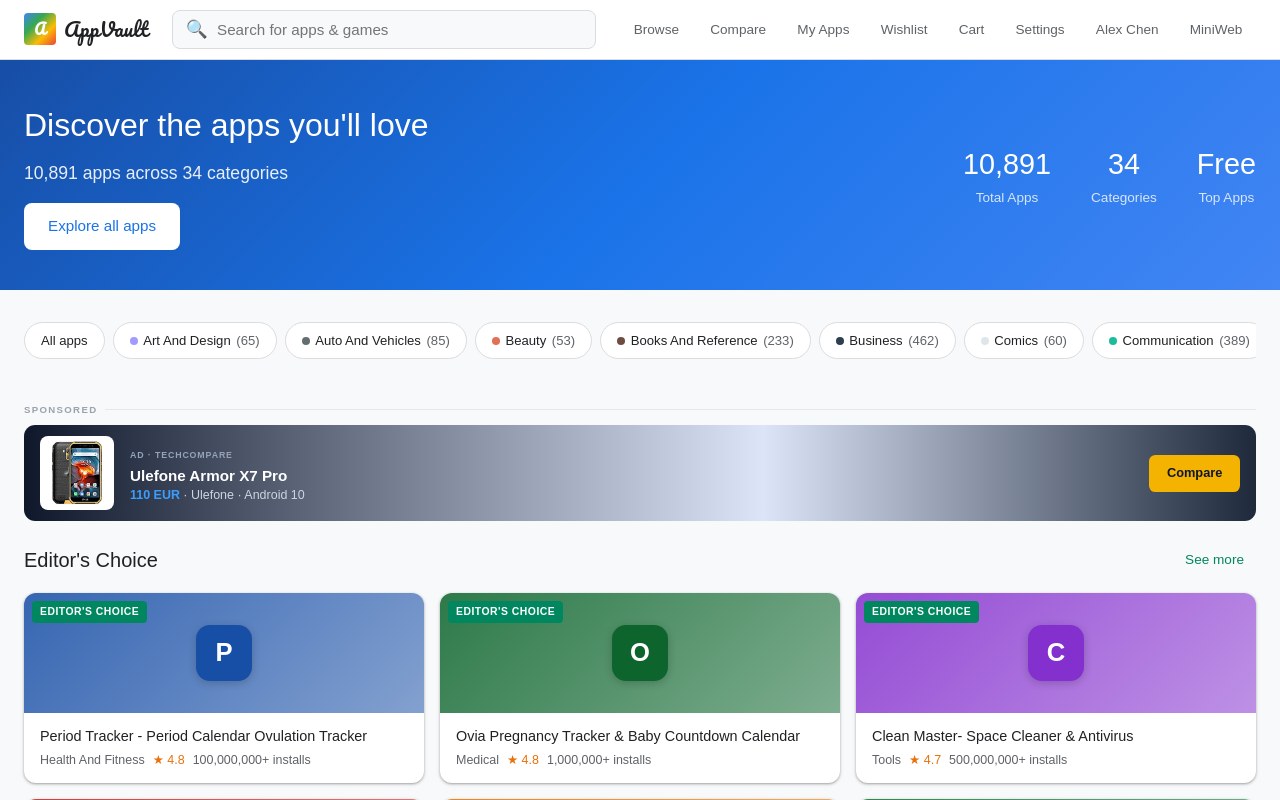}\\[1pt]
{\small\textbf{AppVault Software Marketplace}\par}
{\scriptsize\color{black!65} \texttt{software-marketplace} $\cdot$ Shopping \& transactions $\cdot$ train $\cdot$ data: Google Play $\cdot$ 6 tasks $\cdot$ 10 primitives\par}\vspace{1pt}
{\scriptsize A Google Play-style app store. Users browse, search and filter apps, compare two apps, install and review them, keep a wishlist, buy paid apps through a cart and card checkout with promo codes, and manage purchases and refunds.\par}
\end{minipage}\hfill\begin{minipage}[t]{0.32\linewidth}\raggedright
\includegraphics[width=\linewidth]{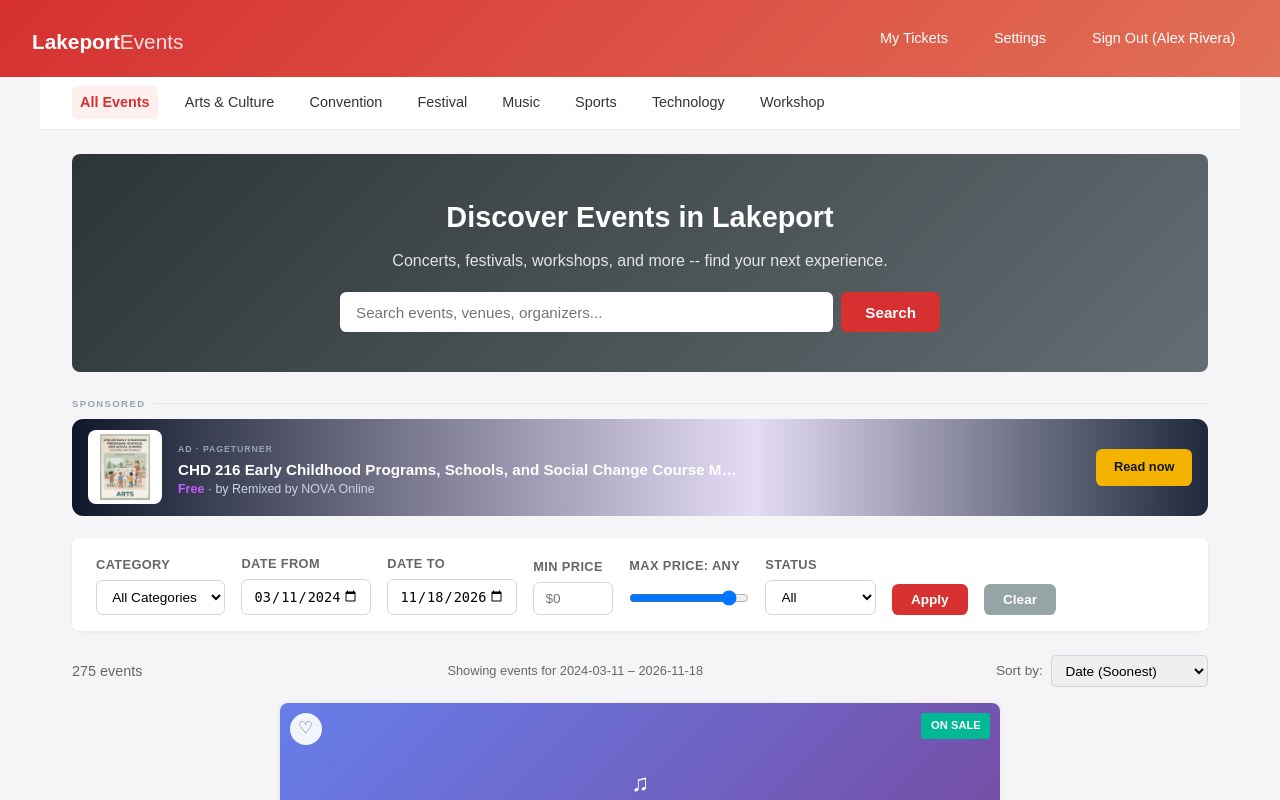}\\[1pt]
{\small\textbf{Lakeport Events}\par}
{\scriptsize\color{black!65} \texttt{ticketing-events} $\cdot$ Shopping \& transactions $\cdot$ train $\cdot$ synthetic data $\cdot$ 8 tasks $\cdot$ 15 primitives\par}\vspace{1pt}
{\scriptsize An Eventbrite/Ticketmaster-style ticketing site for events in Lakeport. Users discover events in an infinite-scroll feed, filter and sort them, compare two events, buy tickets (with reserved seating and promo codes), save events and cancel orders.\par}
\end{minipage}\hfill\begin{minipage}[t]{0.32\linewidth}\raggedright
\includegraphics[width=\linewidth]{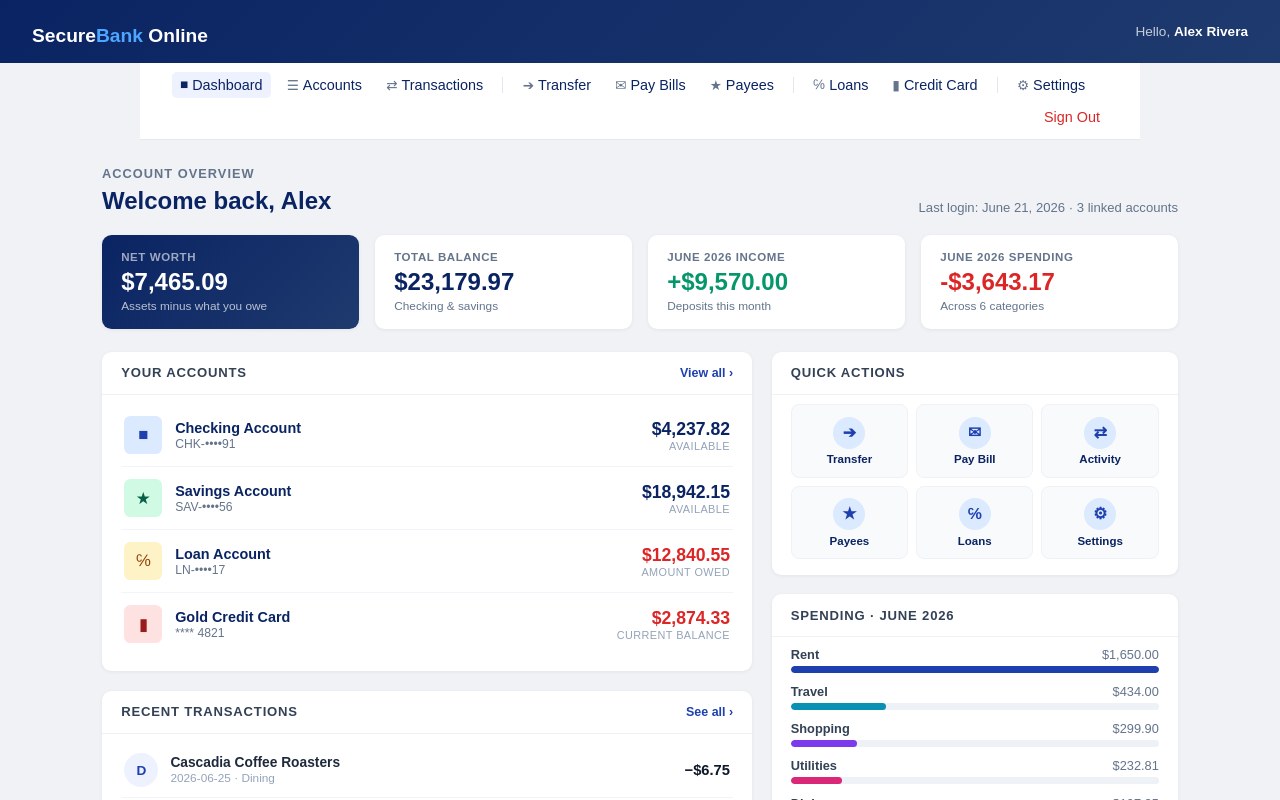}\\[1pt]
{\small\textbf{SecureBank Online}\par}
{\scriptsize\color{black!65} \texttt{banking} $\cdot$ Finance $\cdot$ train $\cdot$ synthetic data $\cdot$ 4 tasks $\cdot$ 14 primitives\par}\vspace{1pt}
{\scriptsize A retail online-banking portal in the style of Chase, TD or Amex. It covers deposit accounts, transaction history, transfers, bill pay, payees, loans and a credit-card section with statements, payments, rewards and disputes.\par}
\end{minipage}
\par\bigskip\noindent
\begin{minipage}[t]{0.32\linewidth}\raggedright
\includegraphics[width=\linewidth]{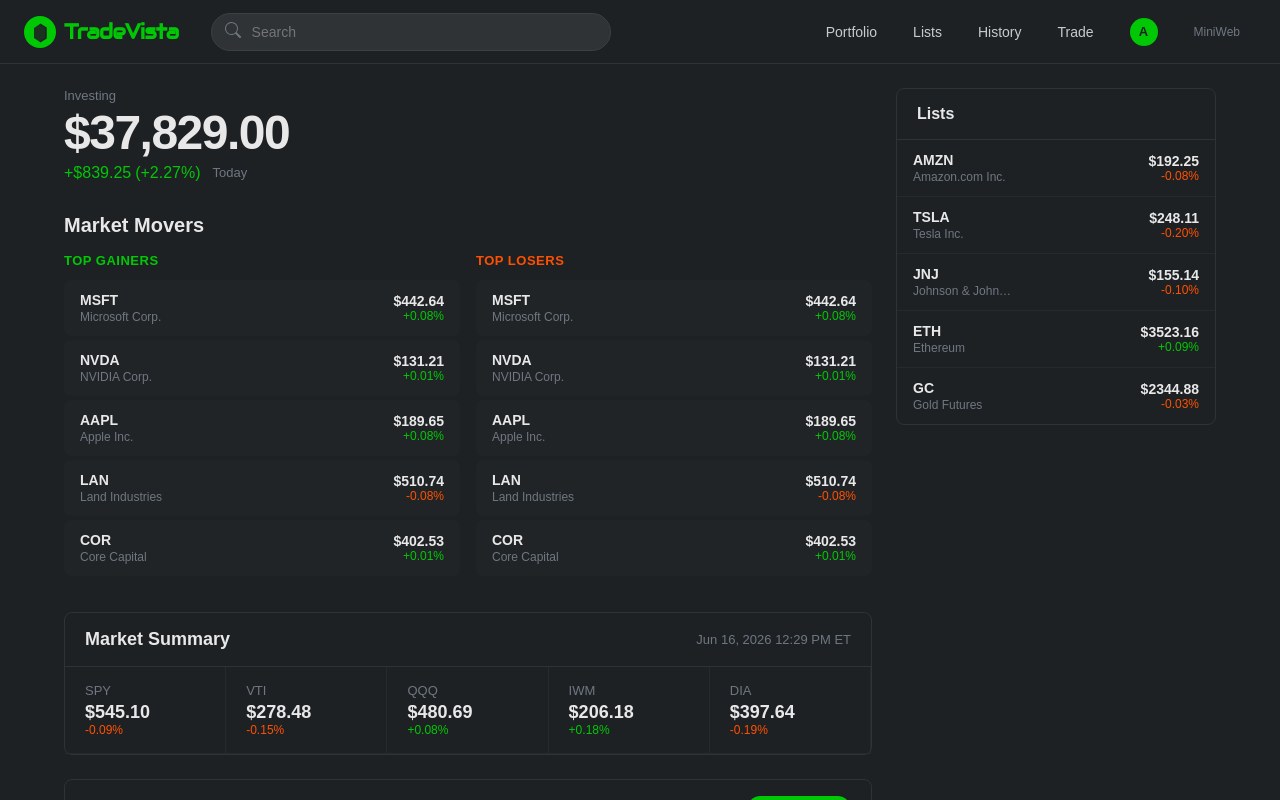}\\[1pt]
{\small\textbf{TradeVista}\par}
{\scriptsize\color{black!65} \texttt{brokerage} $\cdot$ Finance $\cdot$ train $\cdot$ synthetic data $\cdot$ 8 tasks $\cdot$ 12 primitives\par}\vspace{1pt}
{\scriptsize A Robinhood-style trading app for stocks, index funds, crypto, options and futures. Users browse securities, view ticker charts, place and cancel orders, manage a watchlist and move cash in and out.\par}
\end{minipage}\hfill\begin{minipage}[t]{0.32\linewidth}\raggedright
\includegraphics[width=\linewidth]{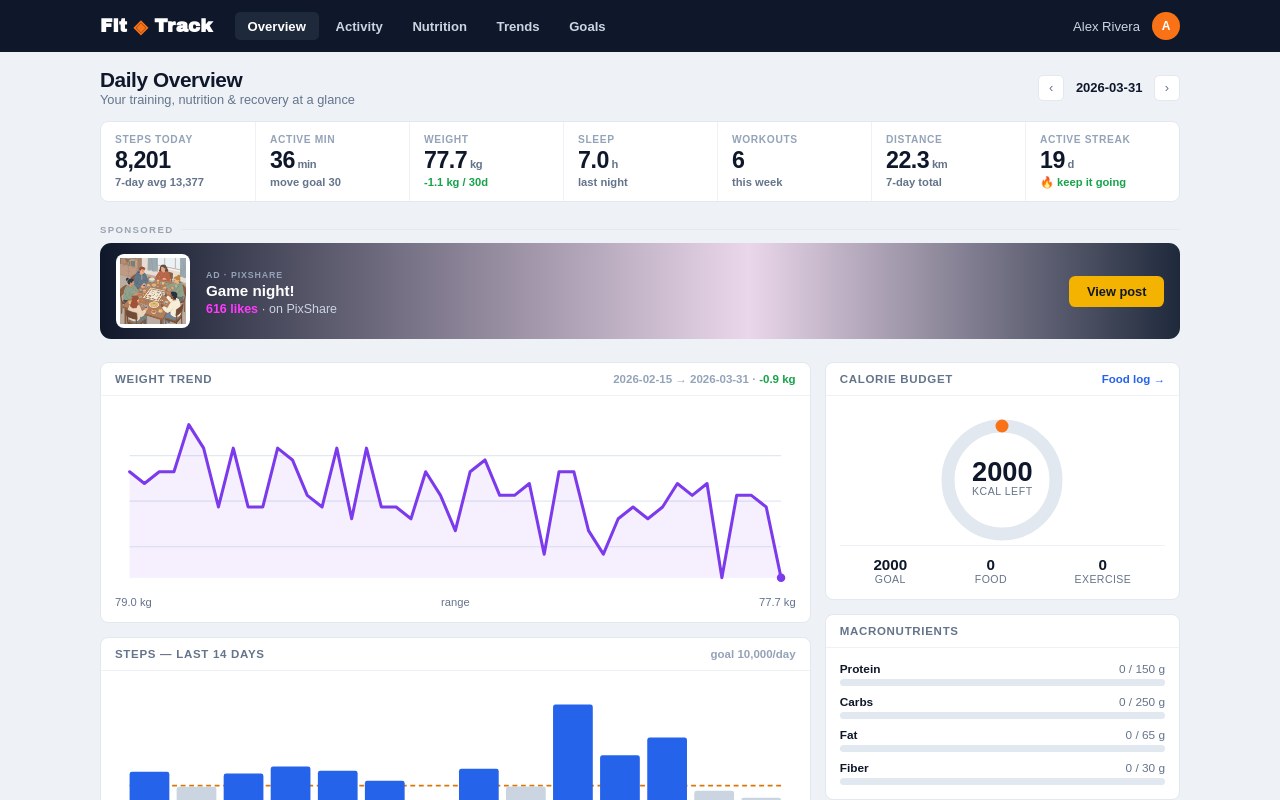}\\[1pt]
{\small\textbf{FitTrack}\par}
{\scriptsize\color{black!65} \texttt{health-fitness-tracking} $\cdot$ Health $\cdot$ train $\cdot$ data: USDA, Zenodo $\cdot$ 8 tasks $\cdot$ 11 primitives\par}\vspace{1pt}
{\scriptsize A personal health and fitness dashboard modeled on MyFitnessPal and Fitbit Web. Users log and review workouts, keep a daily food diary, track daily stats (steps, sleep, calories, water, weight) and manage fitness goals and targets.\par}
\end{minipage}\hfill\begin{minipage}[t]{0.32\linewidth}\raggedright
\includegraphics[width=\linewidth]{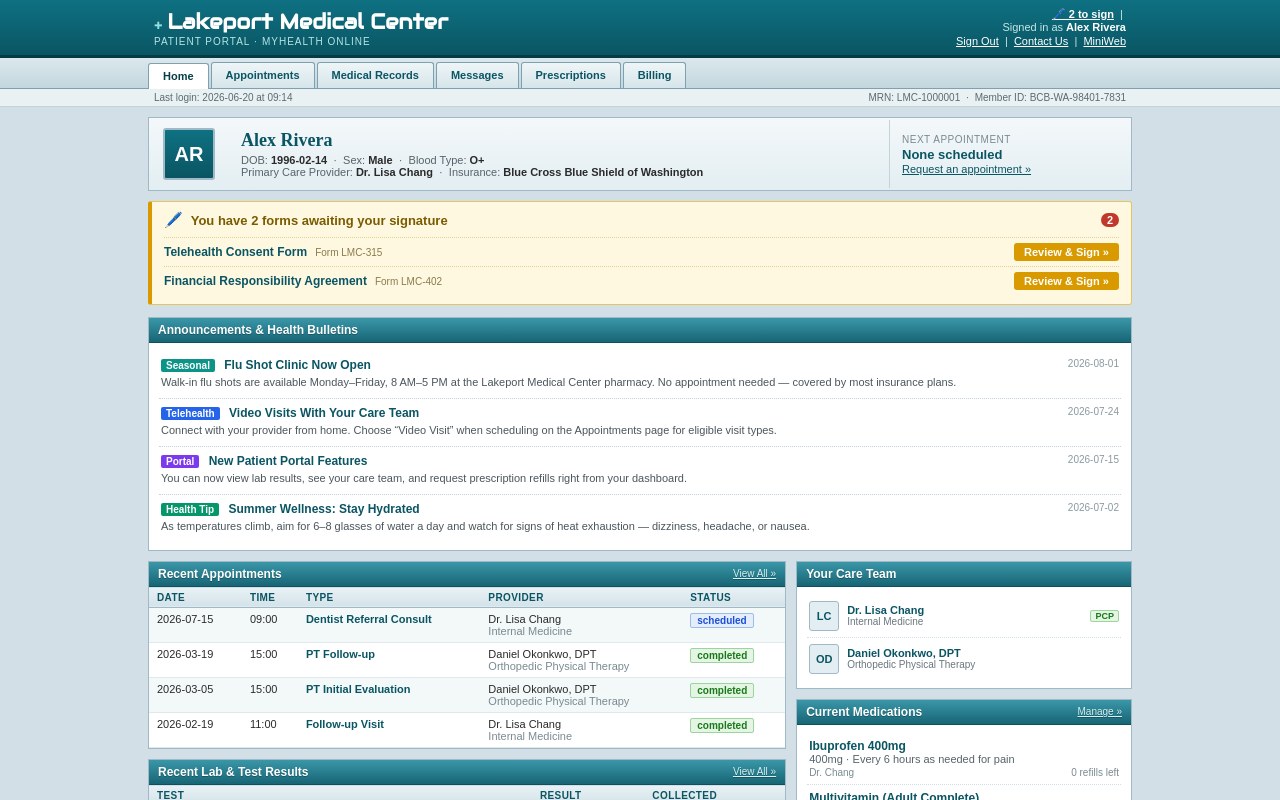}\\[1pt]
{\small\textbf{Lakeport Medical Center Patient Portal}\par}
{\scriptsize\color{black!65} \texttt{health-portals} $\cdot$ Health $\cdot$ train $\cdot$ synthetic data $\cdot$ 8 tasks $\cdot$ 15 primitives\par}\vspace{1pt}
{\scriptsize A MyChart-style patient portal for Lakeport Medical Center. Patients manage appointments, read medical records (vitals, labs, diagnoses), message their care team, request prescription refills, pay bills and e-sign consent forms.\par}
\end{minipage}
\par\bigskip\noindent
\begin{minipage}[t]{0.32\linewidth}\raggedright
\includegraphics[width=\linewidth]{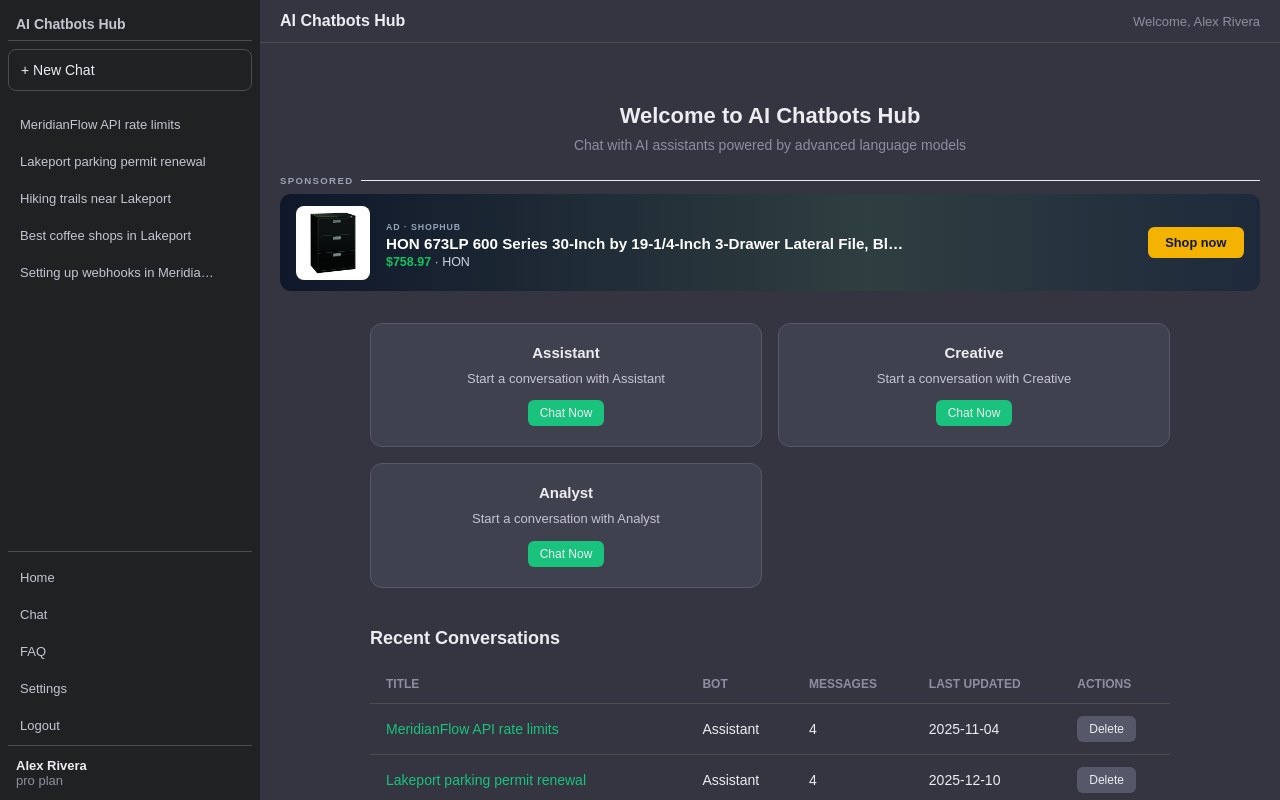}\\[1pt]
{\small\textbf{AI Chatbots Hub}\par}
{\scriptsize\color{black!65} \texttt{ai-chatbots} $\cdot$ Communication $\cdot$ train $\cdot$ synthetic data $\cdot$ 10 tasks $\cdot$ 12 primitives\par}\vspace{1pt}
{\scriptsize A ChatGPT-style chat app with three bot personas (Assistant, Creative, Analyst), a searchable knowledge base, an FAQ and a prompts library. Replies come from a local, deterministic rule engine (intent matching plus retrieval over the knowledge base and FAQ); no external LLM is ever called.\par}
\end{minipage}\hfill\begin{minipage}[t]{0.32\linewidth}\raggedright
\includegraphics[width=\linewidth]{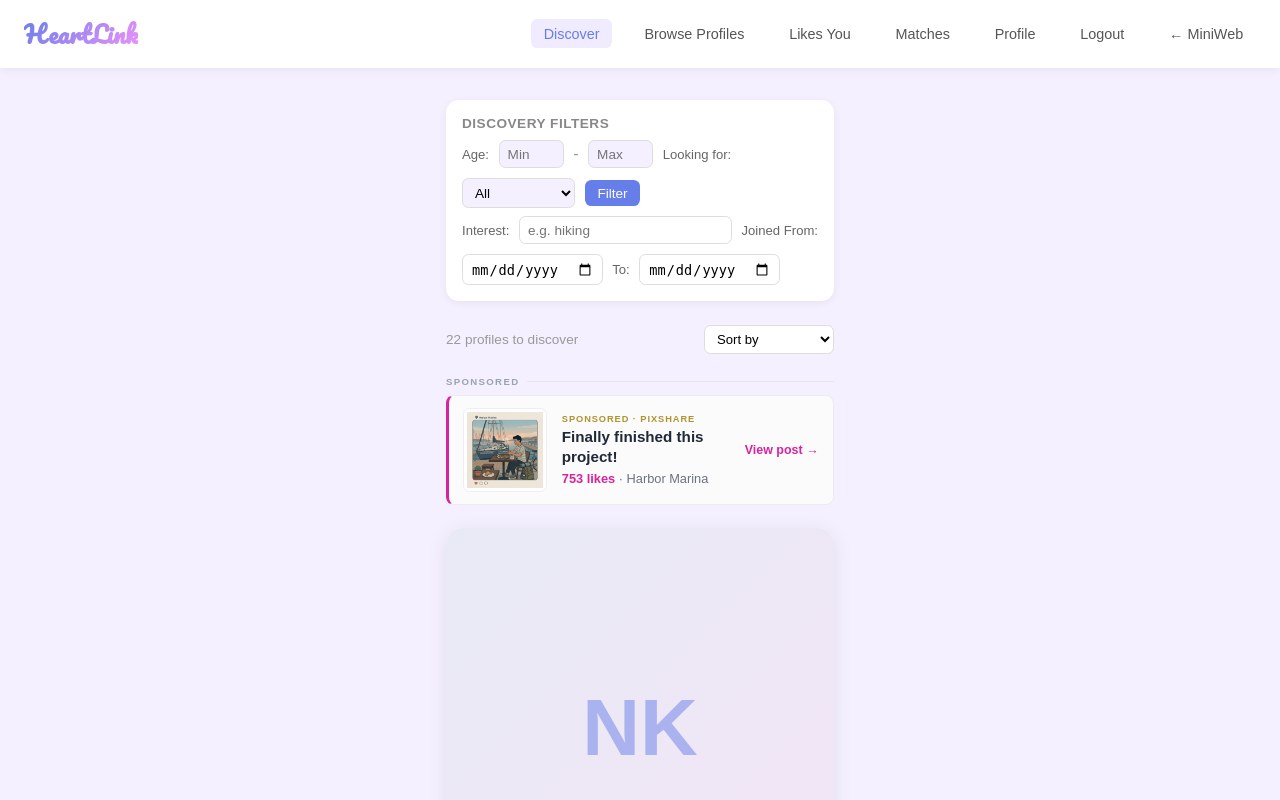}\\[1pt]
{\small\textbf{HeartLink}\par}
{\scriptsize\color{black!65} \texttt{dating} $\cdot$ Communication $\cdot$ \textbf{held-out} $\cdot$ synthetic data $\cdot$ 9 tasks $\cdot$ 15 primitives\par}\vspace{1pt}
{\scriptsize A Tinder, Bumble or Hinge-style dating app. Users swipe through discovery cards, browse and filter the profile directory, like or pass, chat with matches and edit their own profile and preferences.\par}
\end{minipage}\hfill\begin{minipage}[t]{0.32\linewidth}\raggedright
\includegraphics[width=\linewidth]{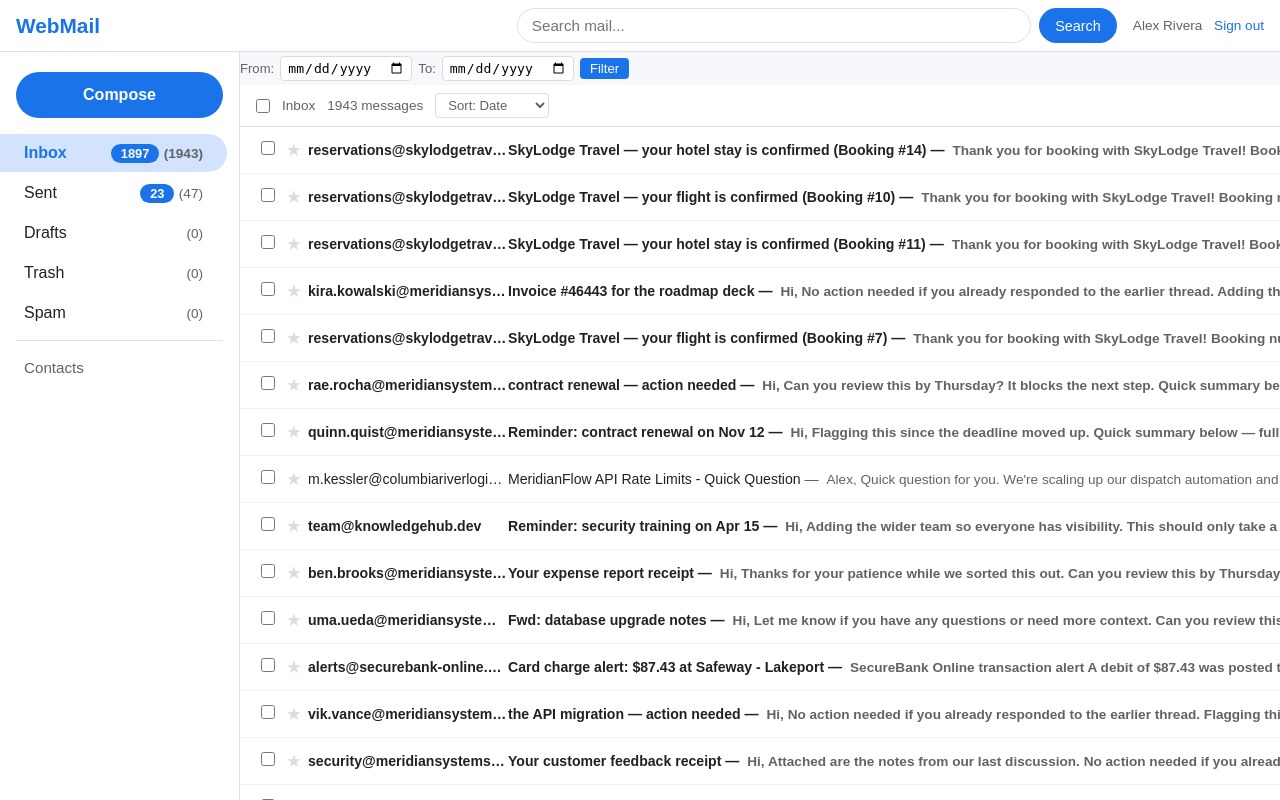}\\[1pt]
{\small\textbf{WebMail}\par}
{\scriptsize\color{black!65} \texttt{email} $\cdot$ Communication $\cdot$ train $\cdot$ data: Enron $\cdot$ 21 tasks $\cdot$ 14 primitives\par}\vspace{1pt}
{\scriptsize A Gmail/Outlook-style webmail client. Users read folders, search, compose with attachments, star, label, move, delete, block senders and report messages. It is also the inbox every other \miniweb{} site writes to: order confirmations, notifications and 2FA codes all arrive here.\par}
\end{minipage}
\par\bigskip\noindent
\begin{minipage}[t]{0.32\linewidth}\raggedright
\includegraphics[width=\linewidth]{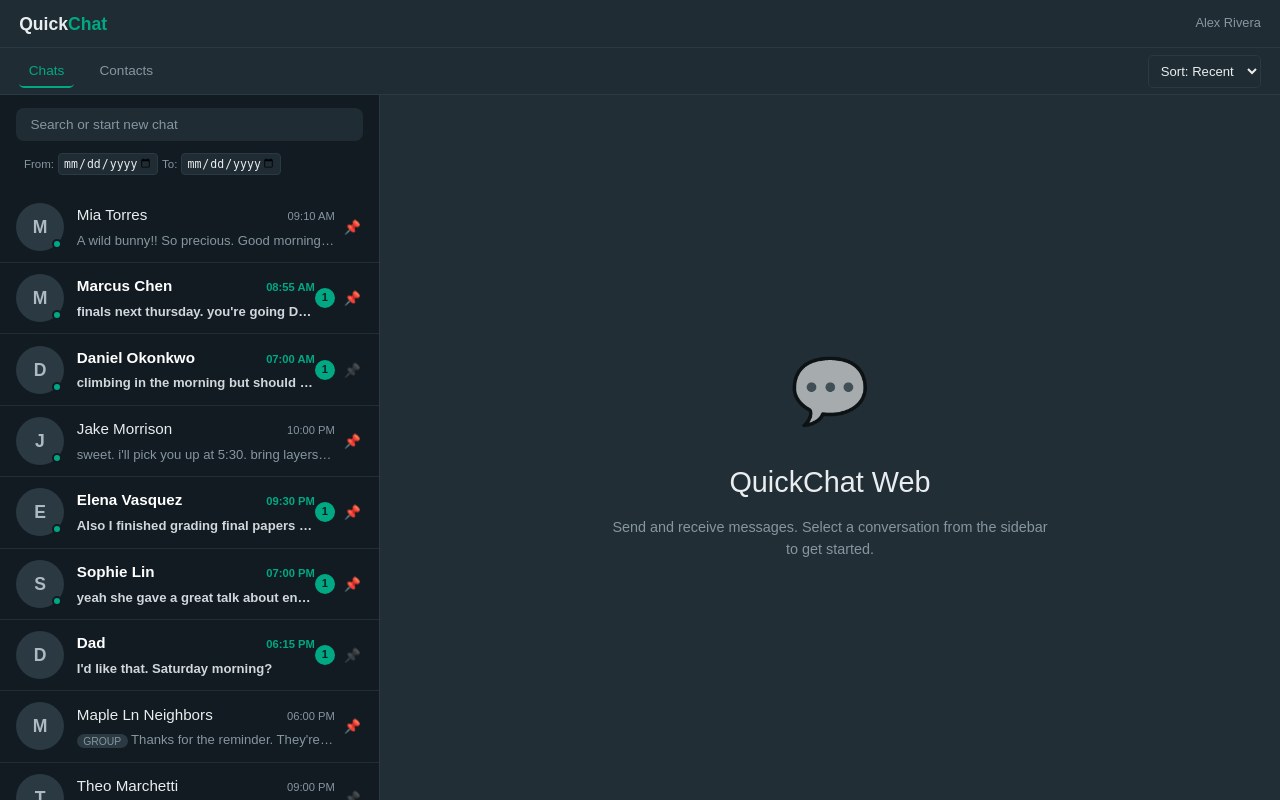}\\[1pt]
{\small\textbf{QuickChat}\par}
{\scriptsize\color{black!65} \texttt{instant-messaging} $\cdot$ Communication $\cdot$ train $\cdot$ synthetic data $\cdot$ 12 tasks $\cdot$ 15 primitives\par}\vspace{1pt}
{\scriptsize A WhatsApp Web / Telegram-style messenger: a dark split-pane view with the conversation list on the left and the chat on the right, plus a contacts page. The user is Alex Rivera, chatting with friends, family and a neighborhood group.\par}
\end{minipage}\hfill\begin{minipage}[t]{0.32\linewidth}\raggedright
\includegraphics[width=\linewidth]{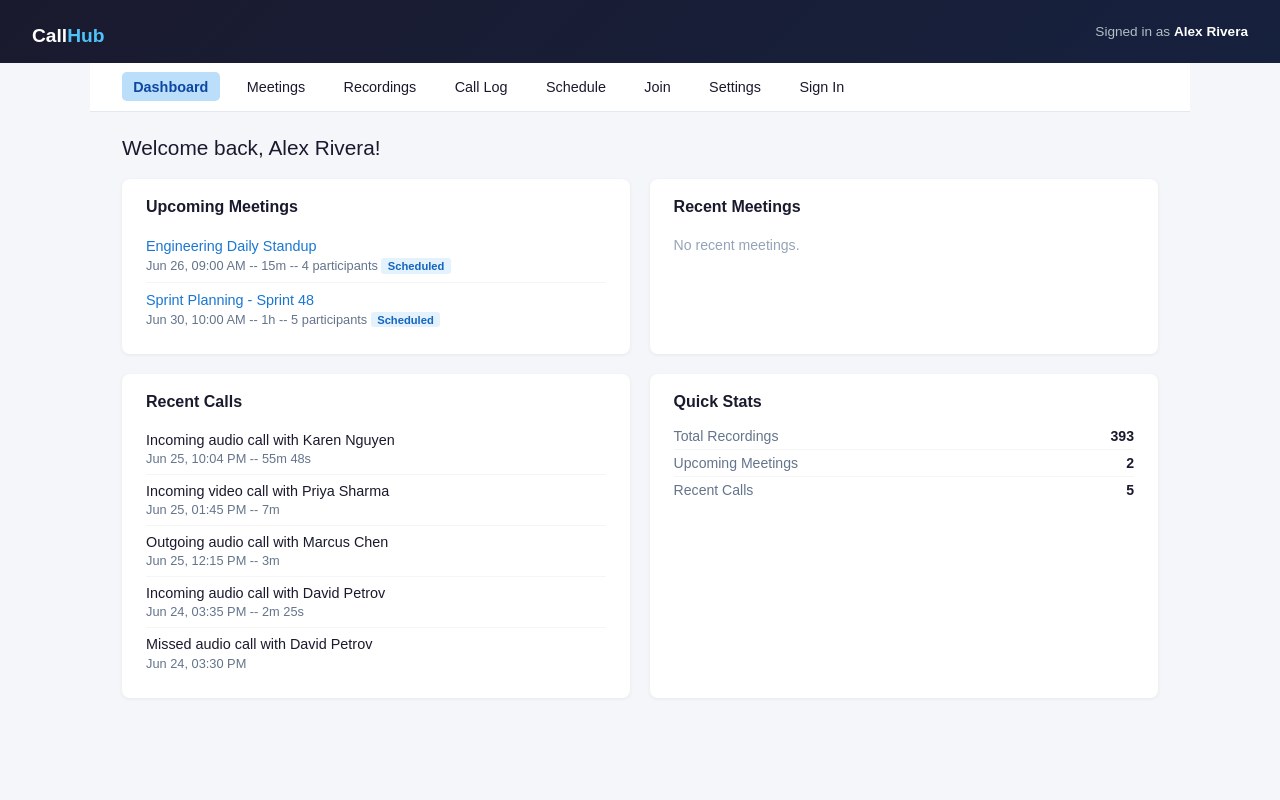}\\[1pt]
{\small\textbf{CallHub}\par}
{\scriptsize\color{black!65} \texttt{remote-calls} $\cdot$ Communication $\cdot$ train $\cdot$ synthetic data $\cdot$ 12 tasks $\cdot$ 15 primitives\par}\vspace{1pt}
{\scriptsize A Zoom, Teams or Google Meet-style video-calling platform for the fictional company Meridian Systems. Users schedule and join meetings, watch recordings and read transcripts, review their call log, and chat inside a simulated call room.\par}
\end{minipage}\hfill\begin{minipage}[t]{0.32\linewidth}\raggedright
\includegraphics[width=\linewidth]{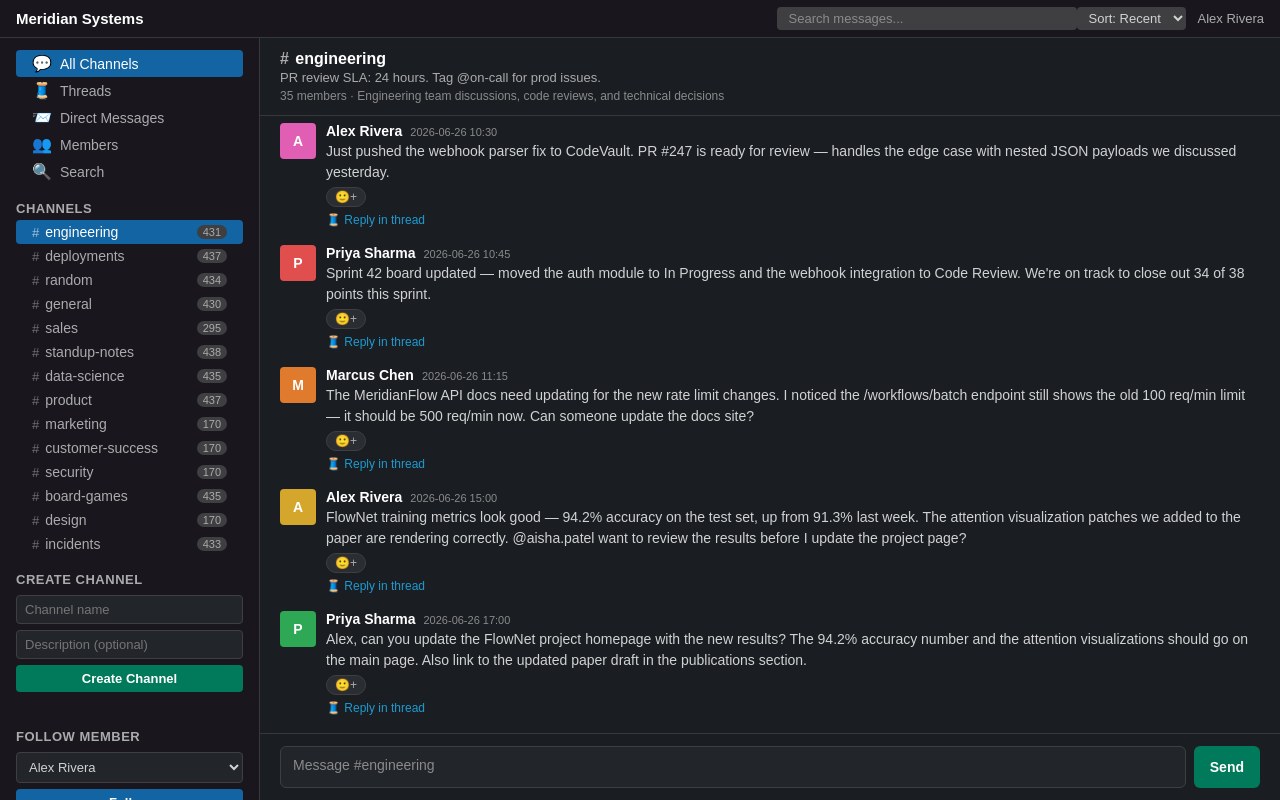}\\[1pt]
{\small\textbf{Meridian Chat}\par}
{\scriptsize\color{black!65} \texttt{team-chat-workspace} $\cdot$ Communication $\cdot$ train $\cdot$ synthetic data $\cdot$ 11 tasks $\cdot$ 14 primitives\par}\vspace{1pt}
{\scriptsize A Slack or Microsoft Teams-style workspace for the employees of the fictional Meridian Systems. Users read and post in channels, reply in threads, react, save and share messages, send direct messages, upload files and manage channel membership.\par}
\end{minipage}
\par\bigskip\noindent
\begin{minipage}[t]{0.32\linewidth}\raggedright
\includegraphics[width=\linewidth]{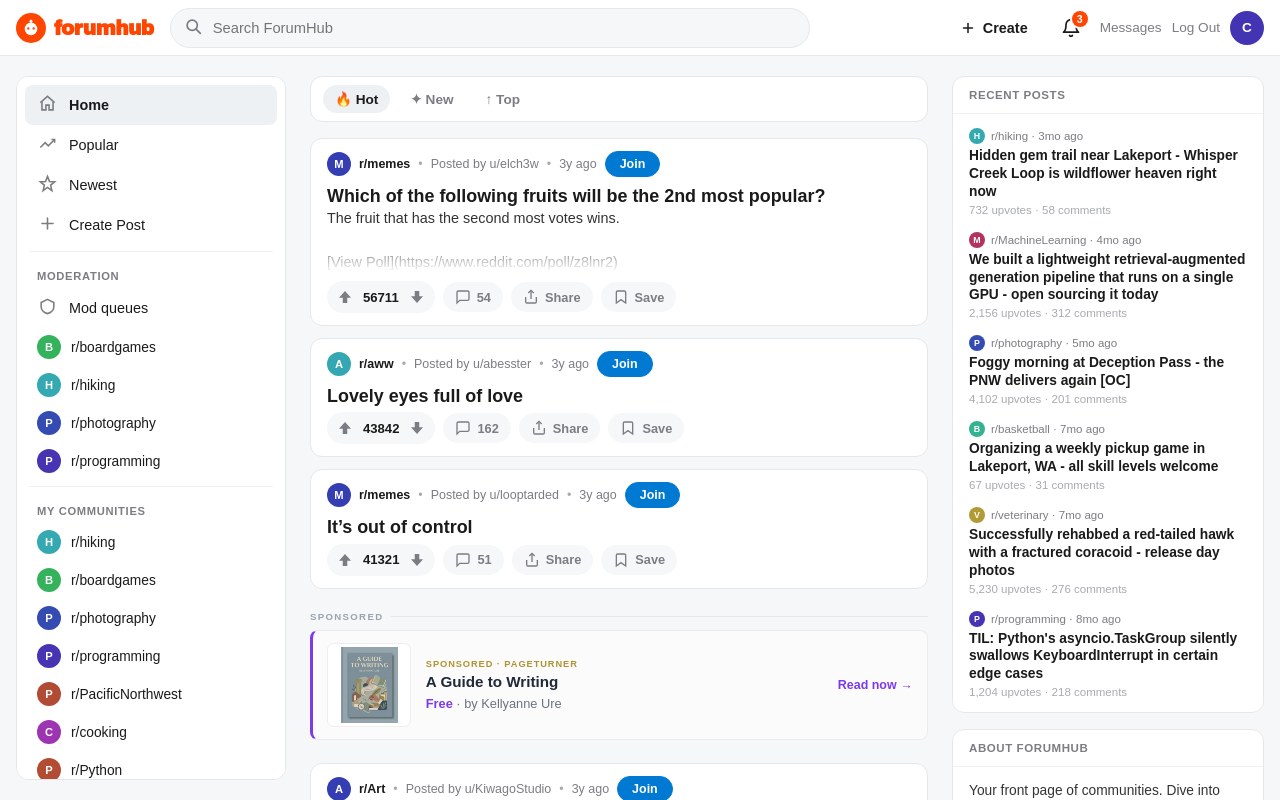}\\[1pt]
{\small\textbf{ForumHub}\par}
{\scriptsize\color{black!65} \texttt{forums} $\cdot$ Social media $\cdot$ train $\cdot$ data: WebArena Reddit $\cdot$ 4 tasks $\cdot$ 14 primitives\par}\vspace{1pt}
{\scriptsize A Reddit-style discussion forum. Users browse communities (subreddits), read and write posts and threaded comments, vote, save, share and report content, follow or block users, join communities, send direct messages and moderate the communities they run.\par}
\end{minipage}\hfill\begin{minipage}[t]{0.32\linewidth}\raggedright
\includegraphics[width=\linewidth]{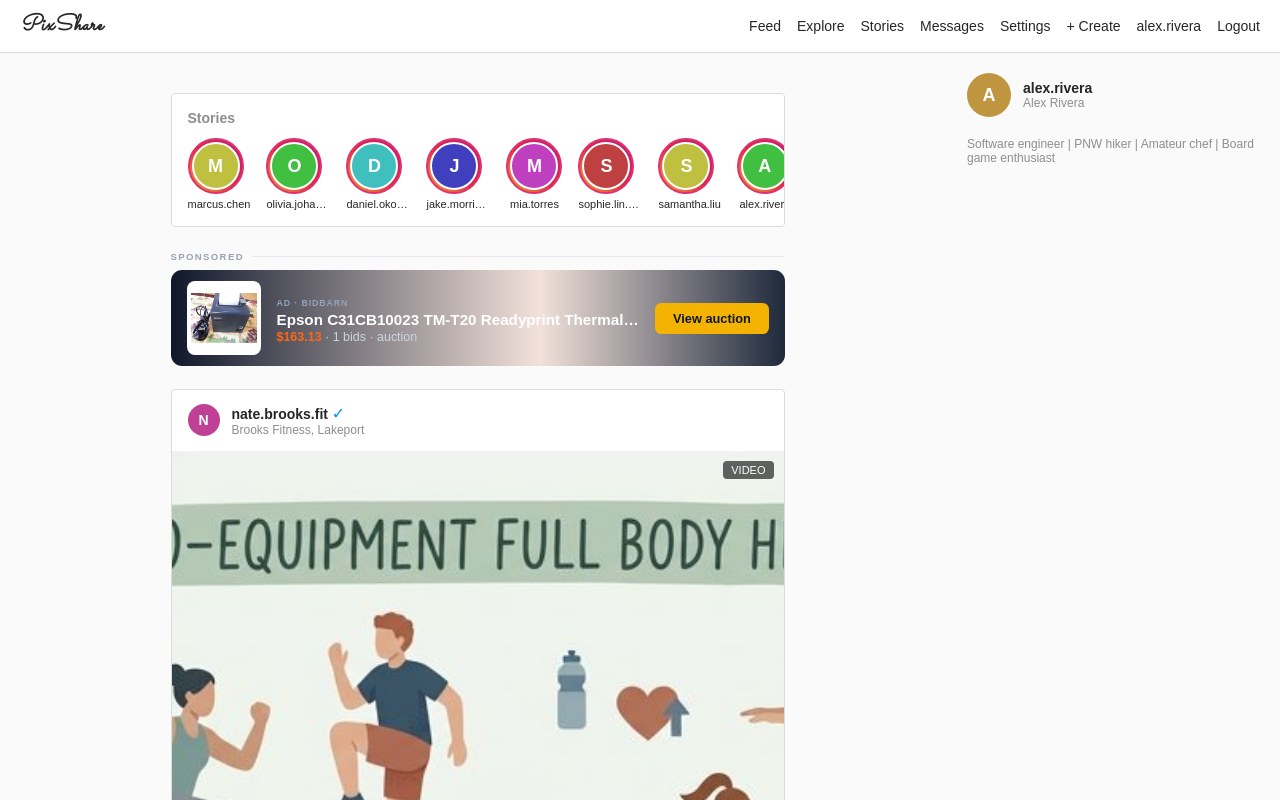}\\[1pt]
{\small\textbf{PixShare}\par}
{\scriptsize\color{black!65} \texttt{multimedia-posting} $\cdot$ Social media $\cdot$ train $\cdot$ synthetic data $\cdot$ 10 tasks $\cdot$ 15 primitives\par}\vspace{1pt}
{\scriptsize An Instagram/Twitter-style social network for photo, video and carousel posts. It has a feed of followed accounts, an explore page, stories, profiles, direct messages and the usual social actions (like, save, follow, block, share).\par}
\end{minipage}\hfill\begin{minipage}[t]{0.32\linewidth}\raggedright
\includegraphics[width=\linewidth]{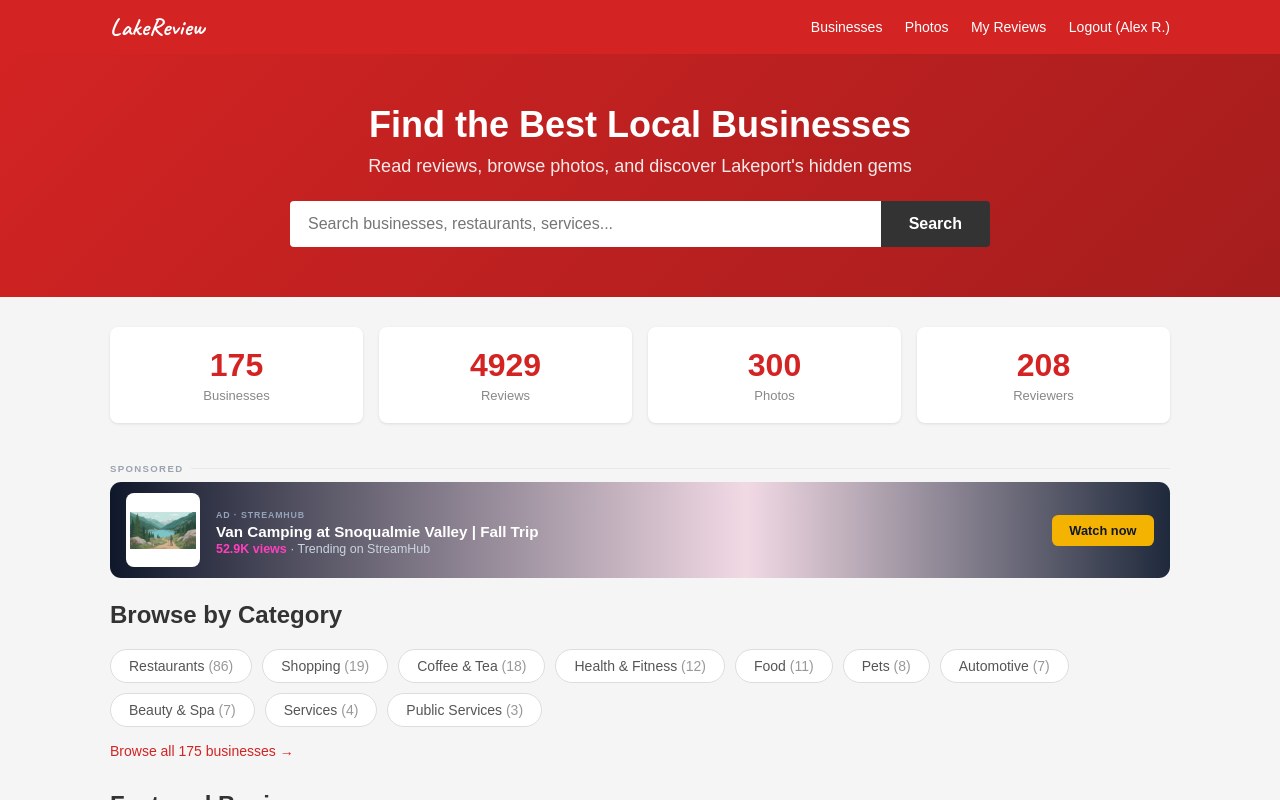}\\[1pt]
{\small\textbf{LakeReview}\par}
{\scriptsize\color{black!65} \texttt{rating-review} $\cdot$ Social media $\cdot$ \textbf{held-out} $\cdot$ synthetic data $\cdot$ 5 tasks $\cdot$ 9 primitives\par}\vspace{1pt}
{\scriptsize A Yelp-style review site for local businesses in the fictional town of Lakeport, WA. Users browse and filter businesses, read and write star-rated reviews, vote on reviews, follow reviewers and save businesses.\par}
\end{minipage}
\par\bigskip\noindent
\begin{minipage}[t]{0.32\linewidth}\raggedright
\includegraphics[width=\linewidth]{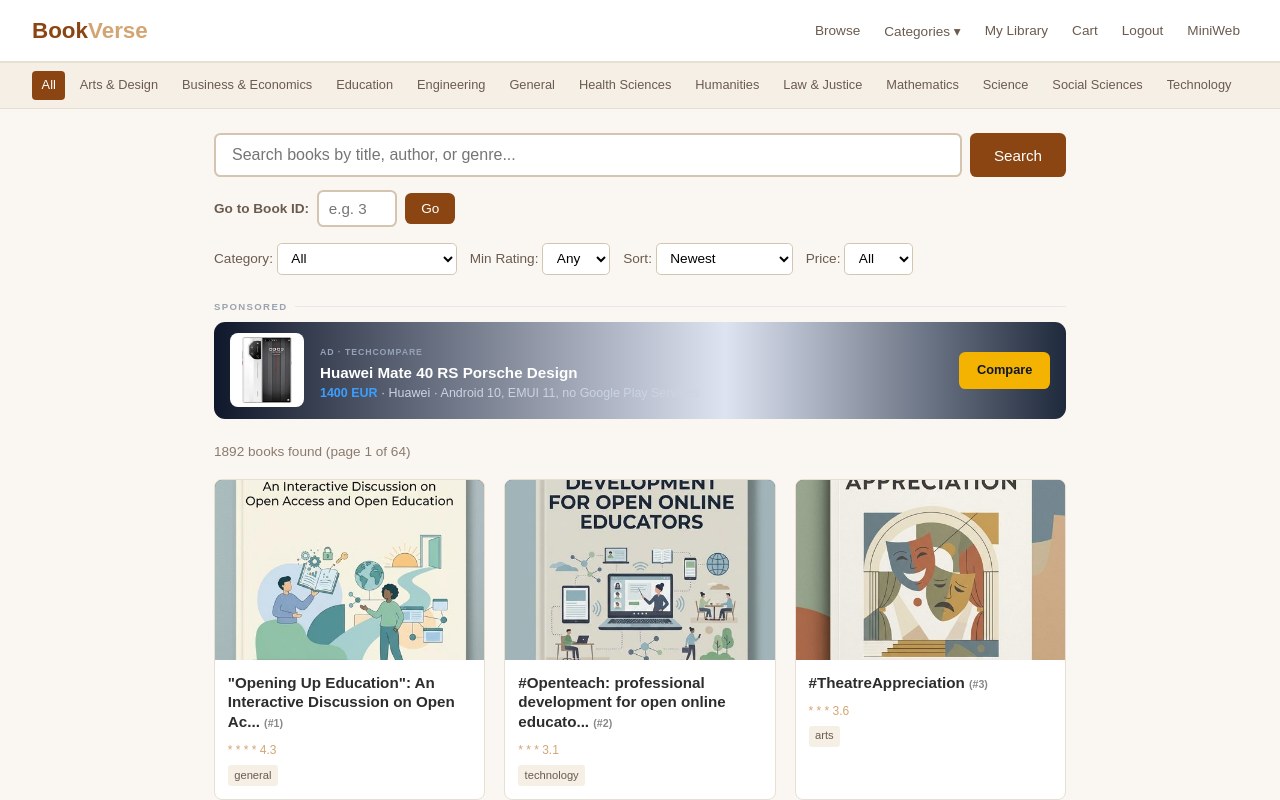}\\[1pt]
{\small\textbf{BookVerse}\par}
{\scriptsize\color{black!65} \texttt{books-comics} $\cdot$ Streaming \& media $\cdot$ train $\cdot$ data: Common Pile $\cdot$ 9 tasks $\cdot$ 10 primitives\par}\vspace{1pt}
{\scriptsize A digital bookstore and reading app in the style of Kindle / Google Play Books. Users discover books by category, read chapters in an in-browser reader, review and rate titles, follow authors, subscribe to categories and buy books through a cart and checkout.\par}
\end{minipage}\hfill\begin{minipage}[t]{0.32\linewidth}\raggedright
\includegraphics[width=\linewidth]{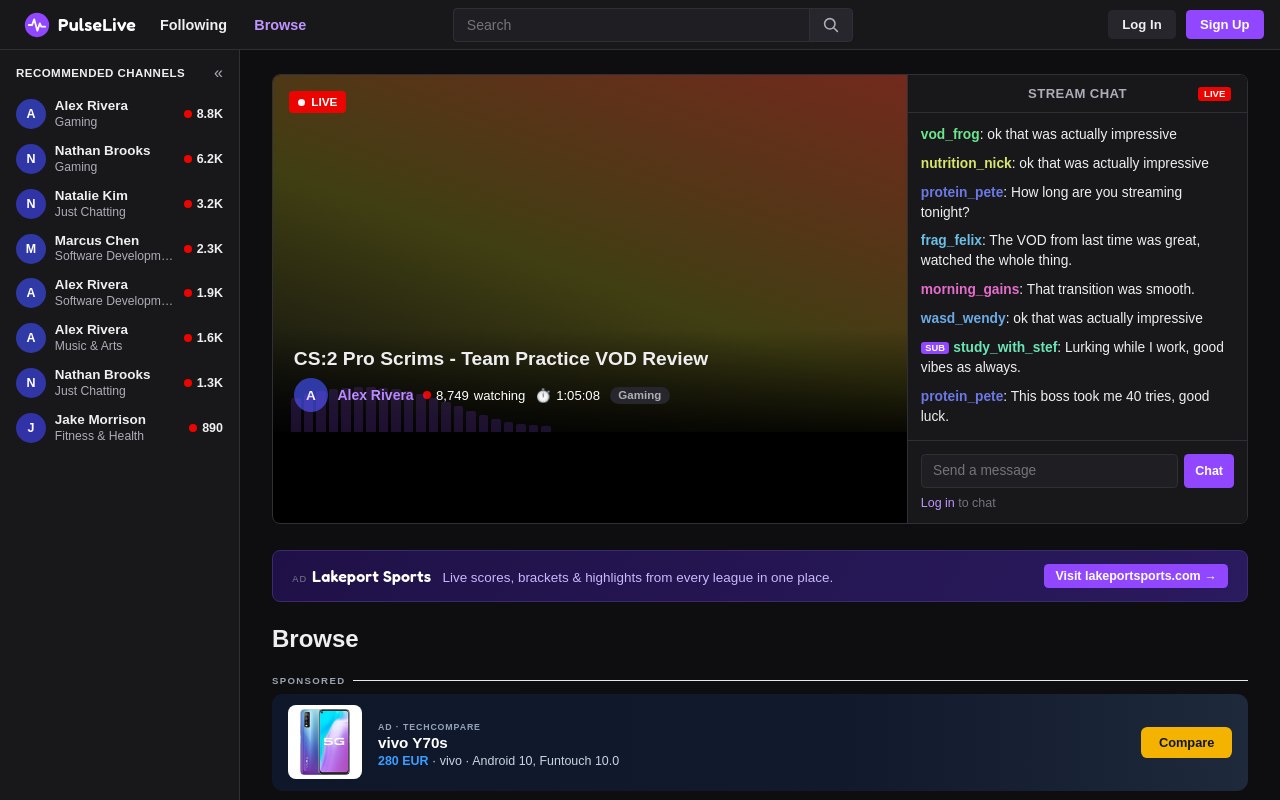}\\[1pt]
{\small\textbf{PulseLive}\par}
{\scriptsize\color{black!65} \texttt{live} $\cdot$ Streaming \& media $\cdot$ train $\cdot$ synthetic data $\cdot$ 4 tasks $\cdot$ 11 primitives\par}\vspace{1pt}
{\scriptsize A Twitch-style live-streaming platform. Viewers browse live and past streams, watch with a seekable player and live chat, follow and subscribe to channels, cheer and redeem channel points, and browse clips.\par}
\end{minipage}\hfill\begin{minipage}[t]{0.32\linewidth}\raggedright
\includegraphics[width=\linewidth]{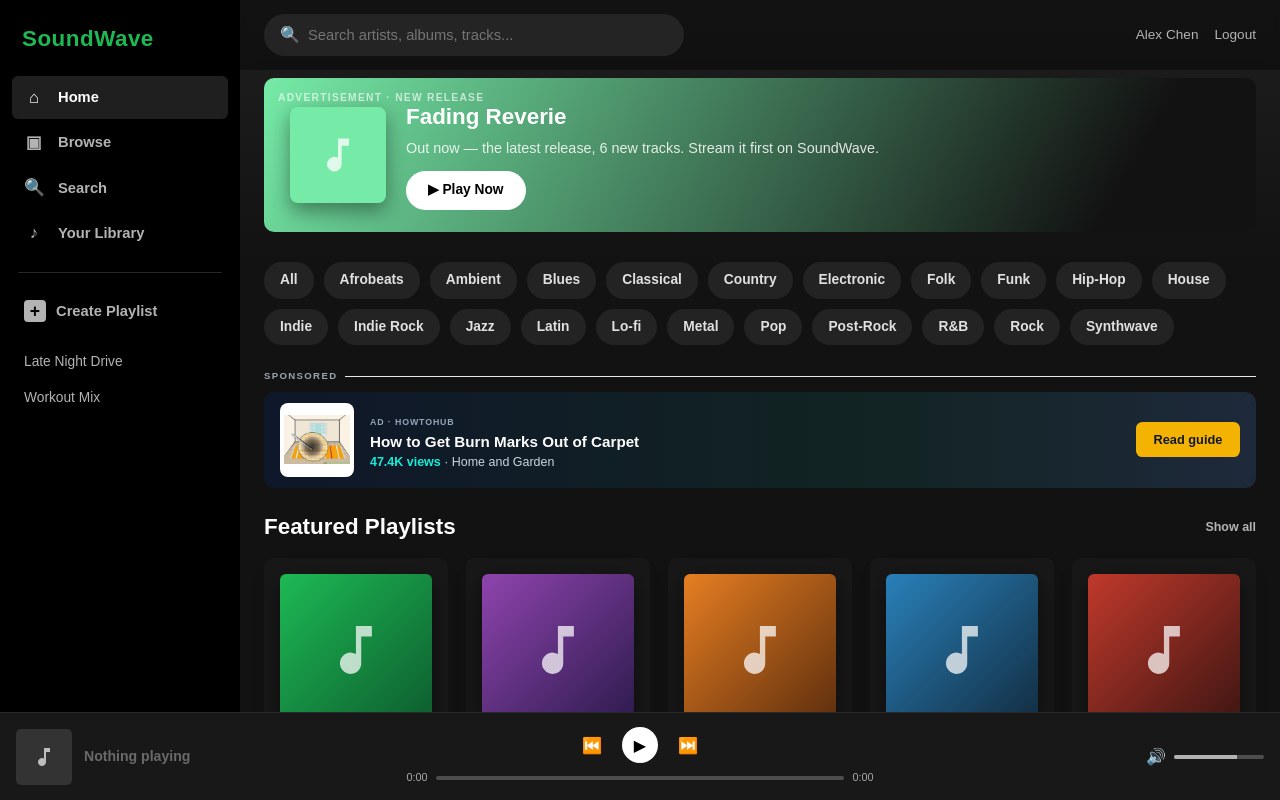}\\[1pt]
{\small\textbf{SoundWave}\par}
{\scriptsize\color{black!65} \texttt{music} $\cdot$ Streaming \& media $\cdot$ \textbf{held-out} $\cdot$ synthetic data $\cdot$ 5 tasks $\cdot$ 9 primitives\par}\vspace{1pt}
{\scriptsize A Spotify-style music streaming site with a dark theme and a persistent bottom player. Users browse artists, albums and tracks, play music, like tracks and albums, follow and subscribe to artists, build playlists and share.\par}
\end{minipage}
\par\bigskip\noindent
\begin{minipage}[t]{0.32\linewidth}\raggedright
\includegraphics[width=\linewidth]{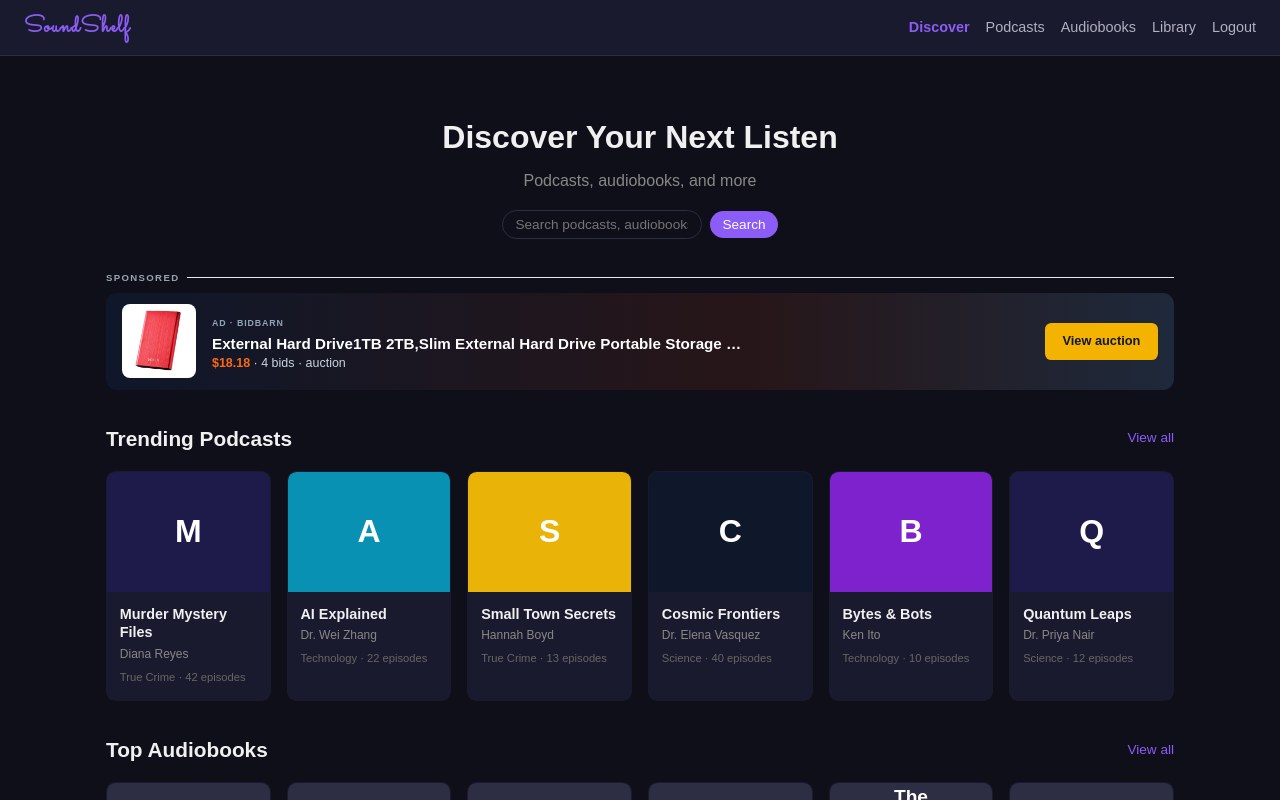}\\[1pt]
{\small\textbf{SoundShelf}\par}
{\scriptsize\color{black!65} \texttt{podcasts-audiobooks} $\cdot$ Streaming \& media $\cdot$ train $\cdot$ synthetic data $\cdot$ 5 tasks $\cdot$ 11 primitives\par}\vspace{1pt}
{\scriptsize A podcast and audiobook app in the style of Apple Podcasts and Audible. Users discover shows, play episodes, subscribe to and follow podcasts, buy audiobooks into their library, like and save items, and write reviews.\par}
\end{minipage}\hfill\begin{minipage}[t]{0.32\linewidth}\raggedright
\includegraphics[width=\linewidth]{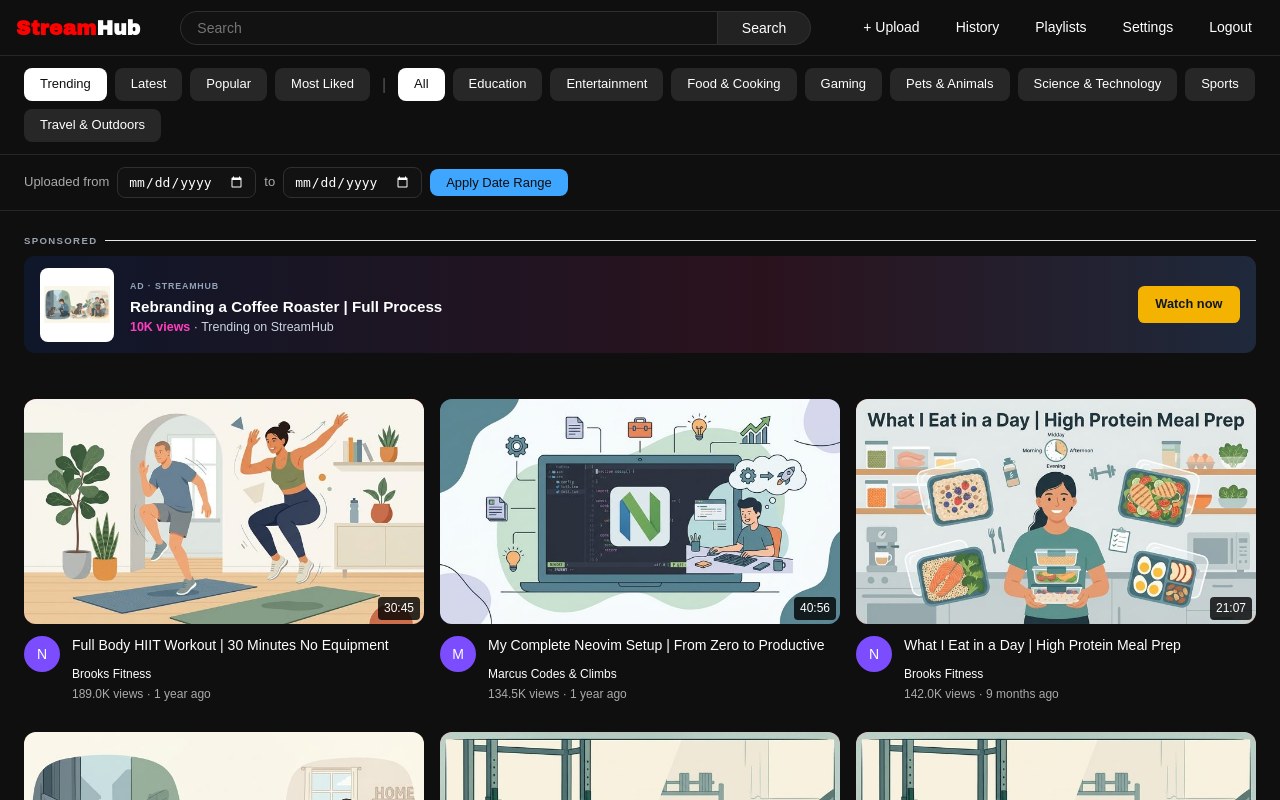}\\[1pt]
{\small\textbf{StreamHub}\par}
{\scriptsize\color{black!65} \texttt{video} $\cdot$ Streaming \& media $\cdot$ train $\cdot$ synthetic data $\cdot$ 8 tasks $\cdot$ 14 primitives\par}\vspace{1pt}
{\scriptsize A YouTube-style video sharing platform. Users browse trending and latest videos, search, filter by upload date, watch videos in the shared mini-player, like, comment, save, share and report videos, subscribe to or follow channels, manage playlists and upload new videos.\par}
\end{minipage}\hfill\begin{minipage}[t]{0.32\linewidth}\raggedright
\includegraphics[width=\linewidth]{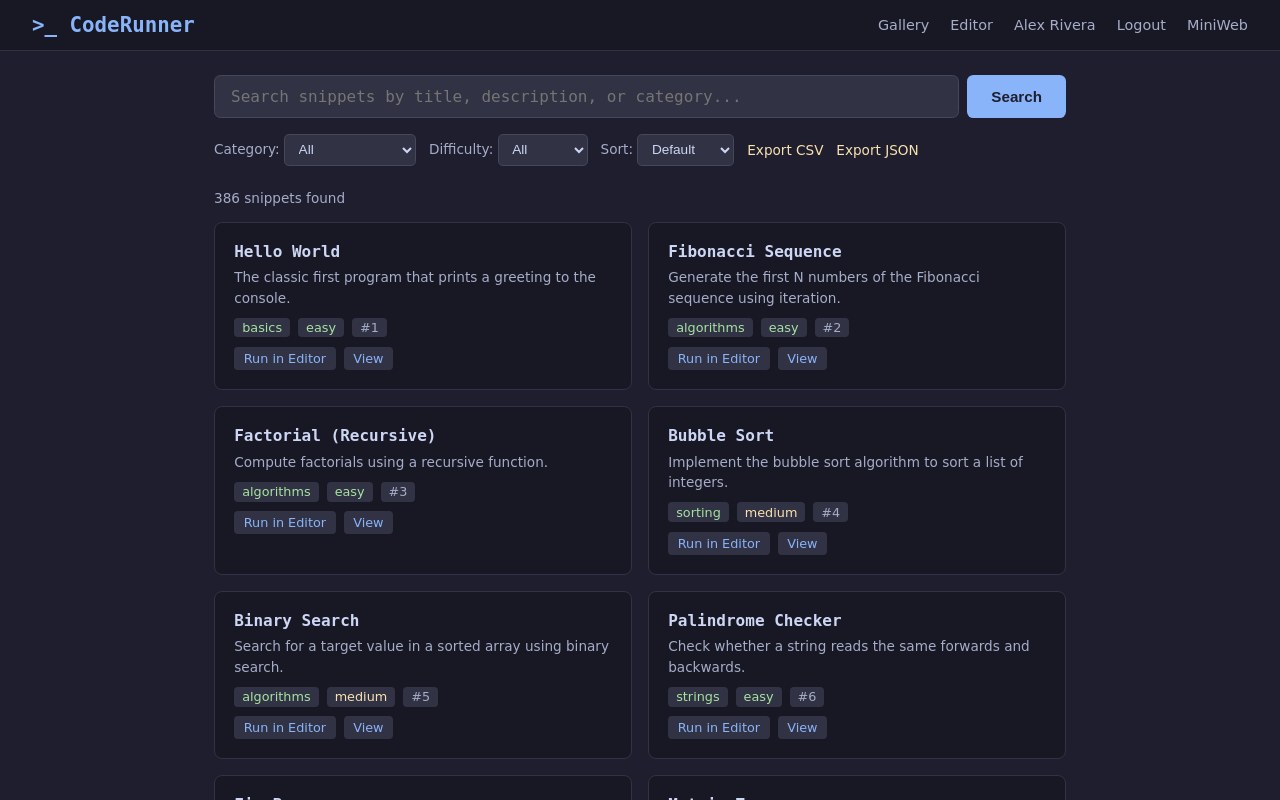}\\[1pt]
{\small\textbf{CodeRunner}\par}
{\scriptsize\color{black!65} \texttt{code-editor-execution} $\cdot$ Editing $\cdot$ train $\cdot$ synthetic data $\cdot$ 7 tasks $\cdot$ 7 primitives\par}\vspace{1pt}
{\scriptsize An online Python IDE modeled after the GeeksforGeeks IDE: a gallery of example snippets, an in-browser editor that really runs Python, and a dashboard of saved snippets.\par}
\end{minipage}
\par\bigskip\noindent
\begin{minipage}[t]{0.32\linewidth}\raggedright
\includegraphics[width=\linewidth]{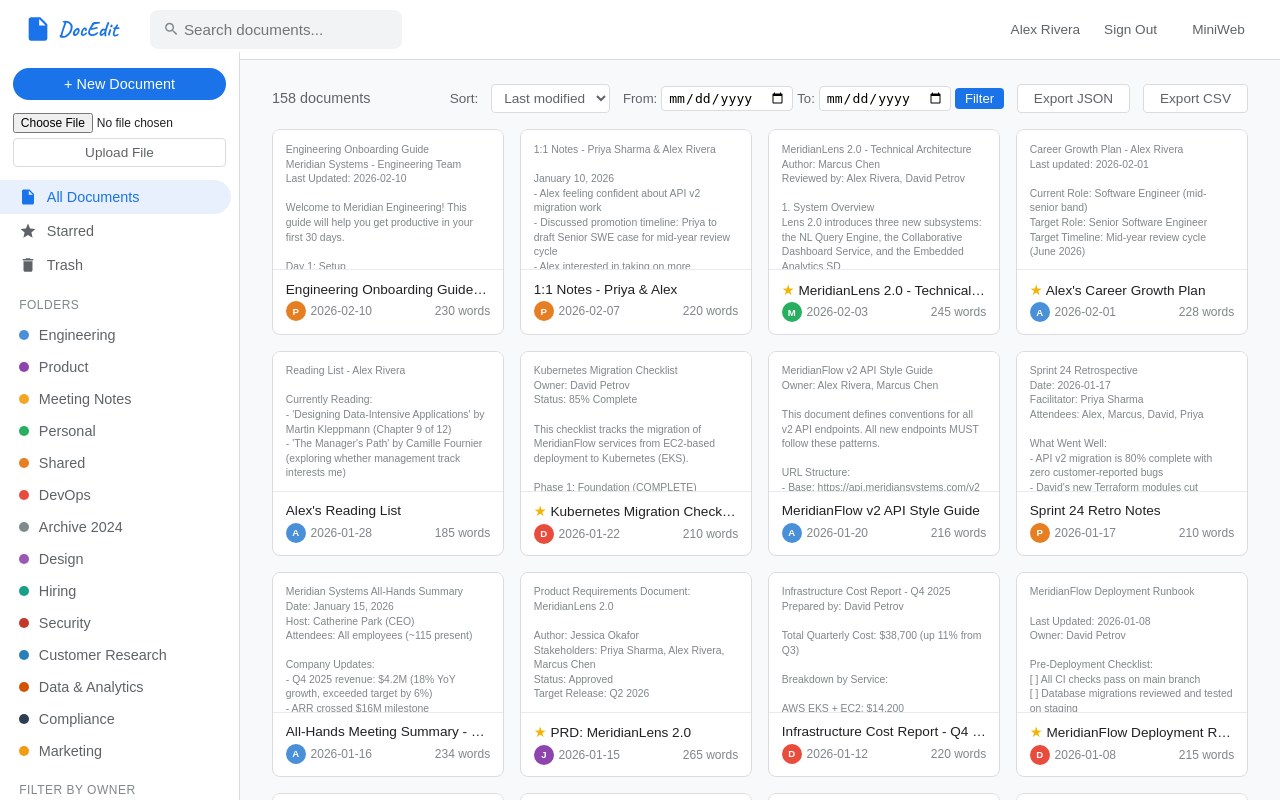}\\[1pt]
{\small\textbf{DocEdit}\par}
{\scriptsize\color{black!65} \texttt{documents} $\cdot$ Editing $\cdot$ train $\cdot$ synthetic data $\cdot$ 5 tasks $\cdot$ 10 primitives\par}\vspace{1pt}
{\scriptsize A Google Docs-style document editor. Users create, edit, share, star, move and trash documents, browse folders, view and restore past revisions, and edit a document's data grid cell by cell.\par}
\end{minipage}\hfill\begin{minipage}[t]{0.32\linewidth}\raggedright
\includegraphics[width=\linewidth]{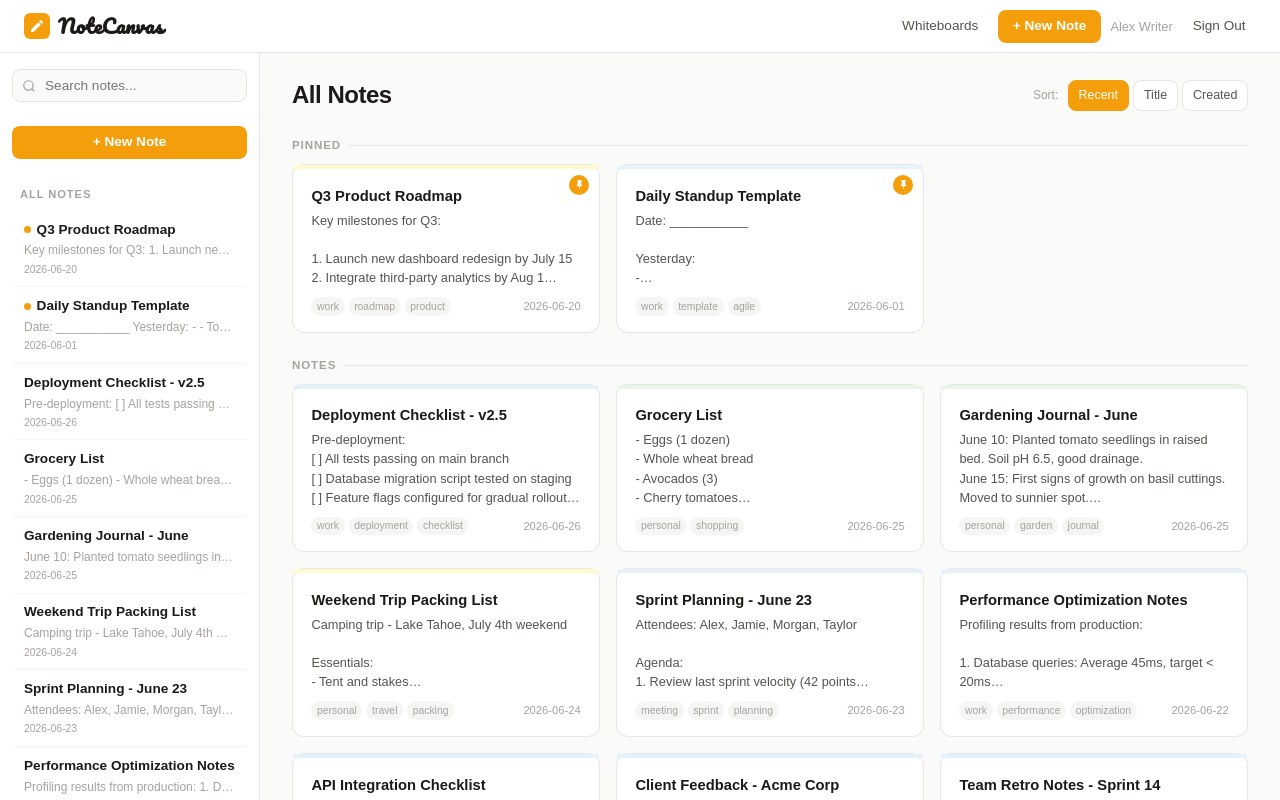}\\[1pt]
{\small\textbf{NoteCanvas}\par}
{\scriptsize\color{black!65} \texttt{handwritten-notes-whiteboards} $\cdot$ Editing $\cdot$ train $\cdot$ synthetic data $\cdot$ 7 tasks $\cdot$ 10 primitives\par}\vspace{1pt}
{\scriptsize A notes and whiteboard app in the spirit of Apple Notes, Notion and Miro. Every note has a text body and a freeform drawing canvas. Whiteboards hold sticky notes, shapes and text that can be dragged around.\par}
\end{minipage}\hfill\begin{minipage}[t]{0.32\linewidth}\raggedright
\includegraphics[width=\linewidth]{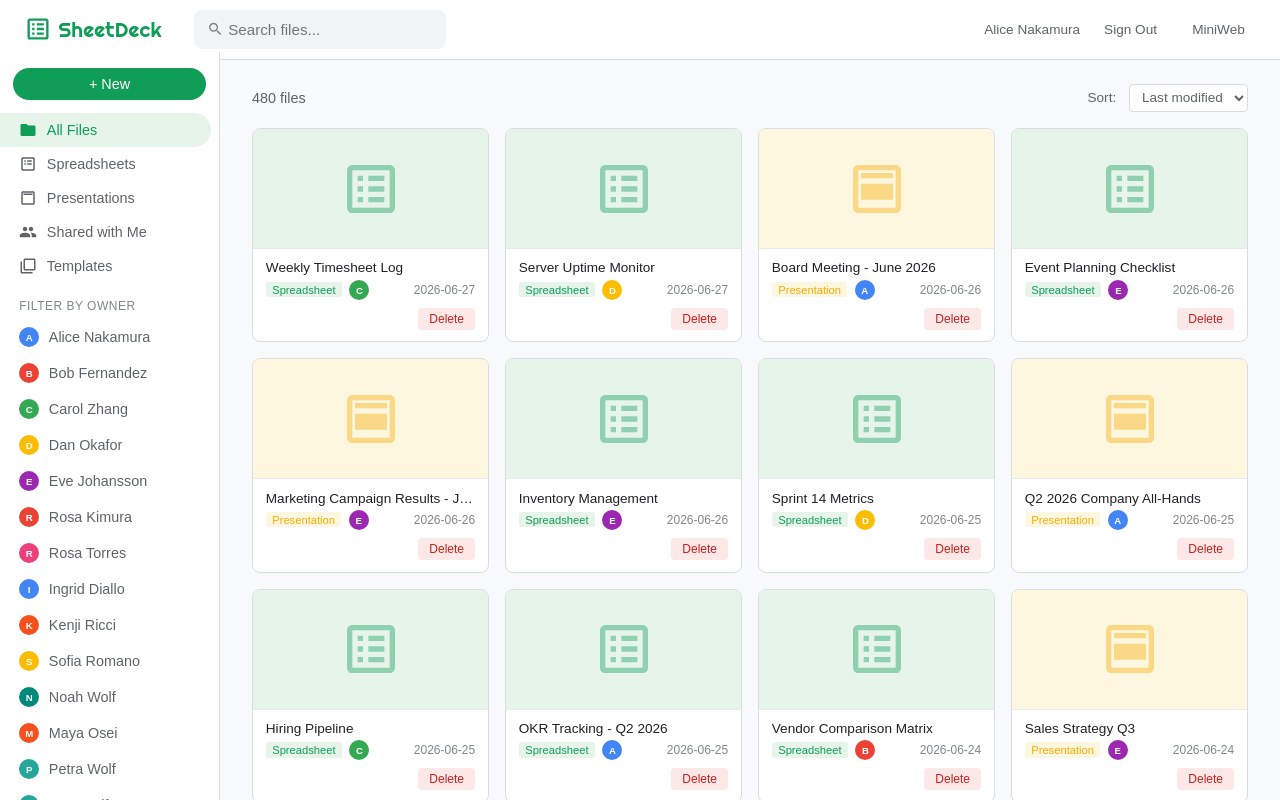}\\[1pt]
{\small\textbf{SheetDeck}\par}
{\scriptsize\color{black!65} \texttt{spreadsheets-slides} $\cdot$ Editing $\cdot$ train $\cdot$ synthetic data $\cdot$ 8 tasks $\cdot$ 8 primitives\par}\vspace{1pt}
{\scriptsize A Google Sheets / Google Slides-style workspace for a fictional company (Meridian Systems). Users browse files, edit spreadsheet cells in a grid, add and reorder slides, share files and export them.\par}
\end{minipage}
\par\bigskip\noindent
\begin{minipage}[t]{0.32\linewidth}\raggedright
\includegraphics[width=\linewidth]{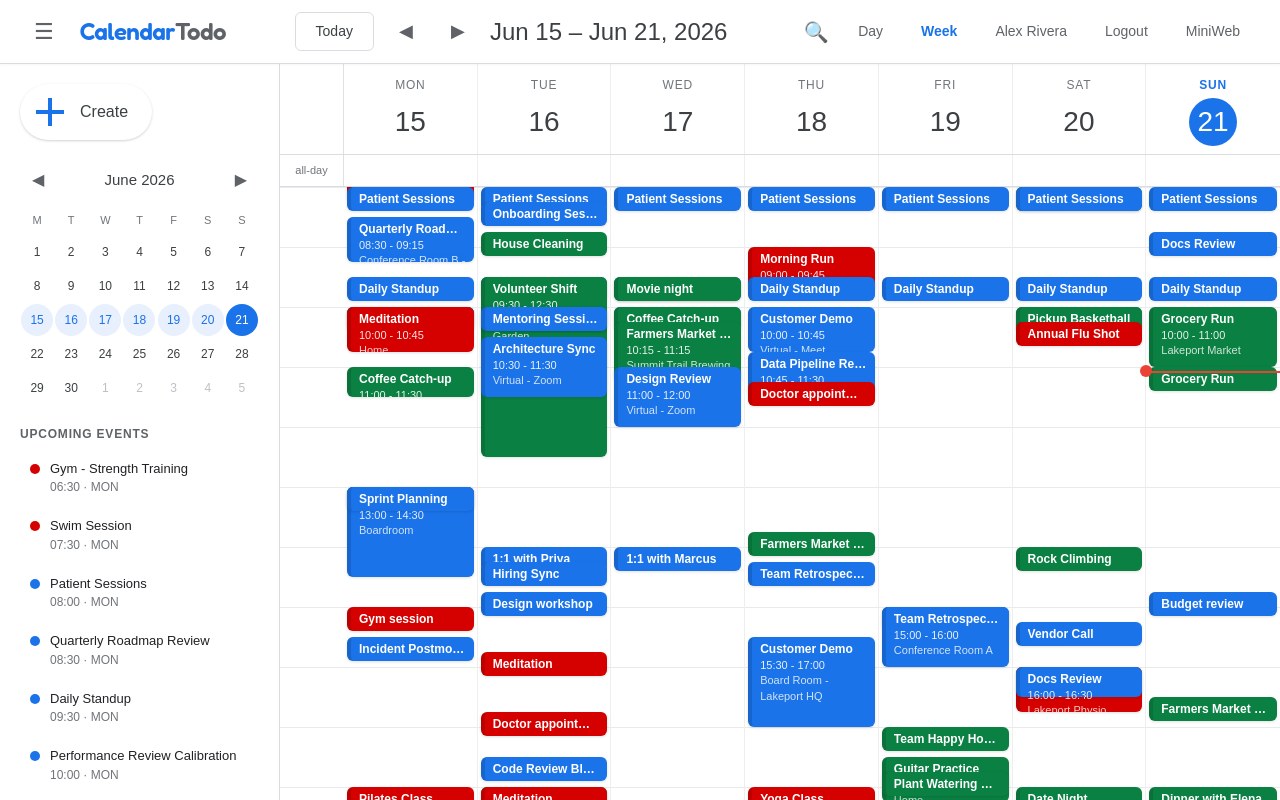}\\[1pt]
{\small\textbf{CalendarTodo}\par}
{\scriptsize\color{black!65} \texttt{calendar-todo} $\cdot$ Productivity $\cdot$ train $\cdot$ synthetic data $\cdot$ 6 tasks $\cdot$ 11 primitives\par}\vspace{1pt}
{\scriptsize A Google Calendar-style calendar and to-do app. Users browse events by week or day, filter them by category, priority, calendar owner or date range, and create, edit, share, invite to and RSVP to events.\par}
\end{minipage}\hfill\begin{minipage}[t]{0.32\linewidth}\raggedright
\includegraphics[width=\linewidth]{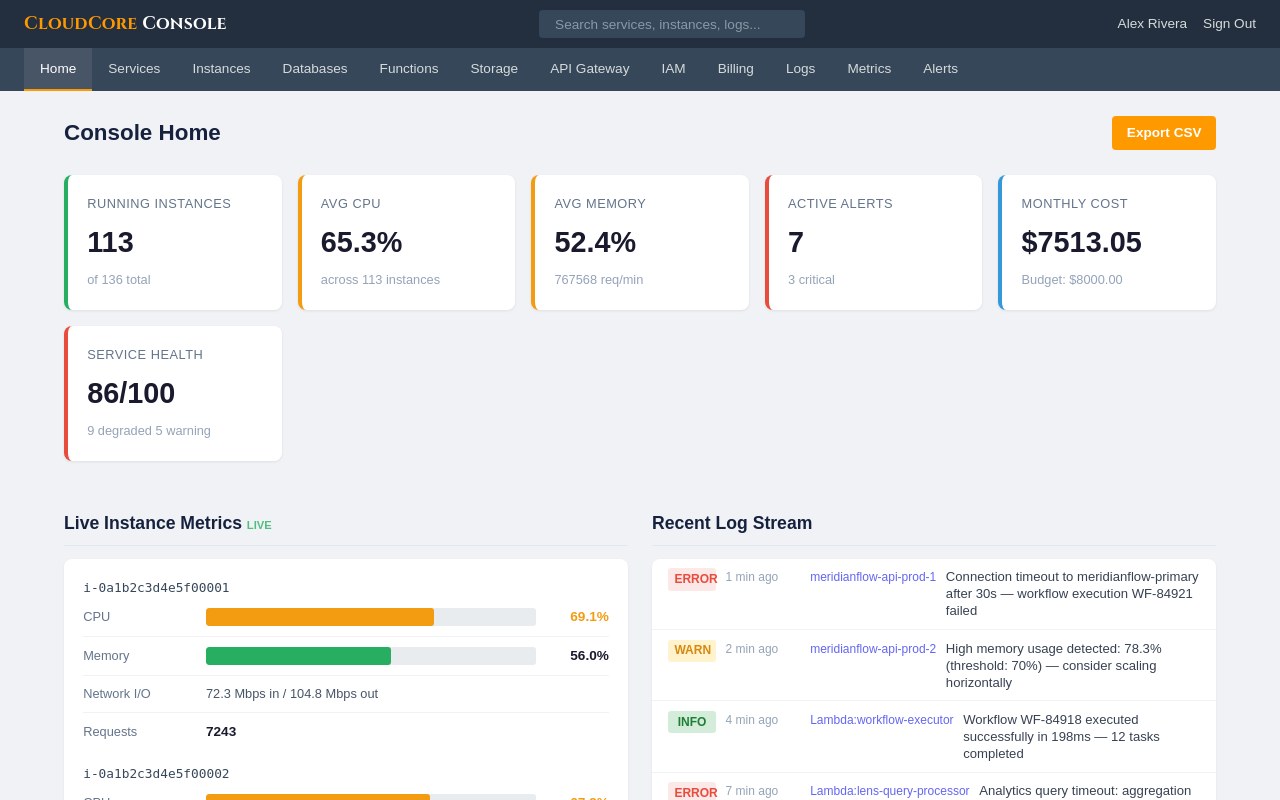}\\[1pt]
{\small\textbf{CloudCore Console}\par}
{\scriptsize\color{black!65} \texttt{cloud-dev-consoles} $\cdot$ Productivity $\cdot$ train $\cdot$ synthetic data $\cdot$ 10 tasks $\cdot$ 12 primitives\par}\vspace{1pt}
{\scriptsize An AWS/GCP-style cloud management console. Users browse services, launch, edit, power-cycle and terminate instances, provision functions, databases, storage buckets and IAM users, and inspect billing, metrics, logs, alerts and API gateway endpoints.\par}
\end{minipage}\hfill\begin{minipage}[t]{0.32\linewidth}\raggedright
\includegraphics[width=\linewidth]{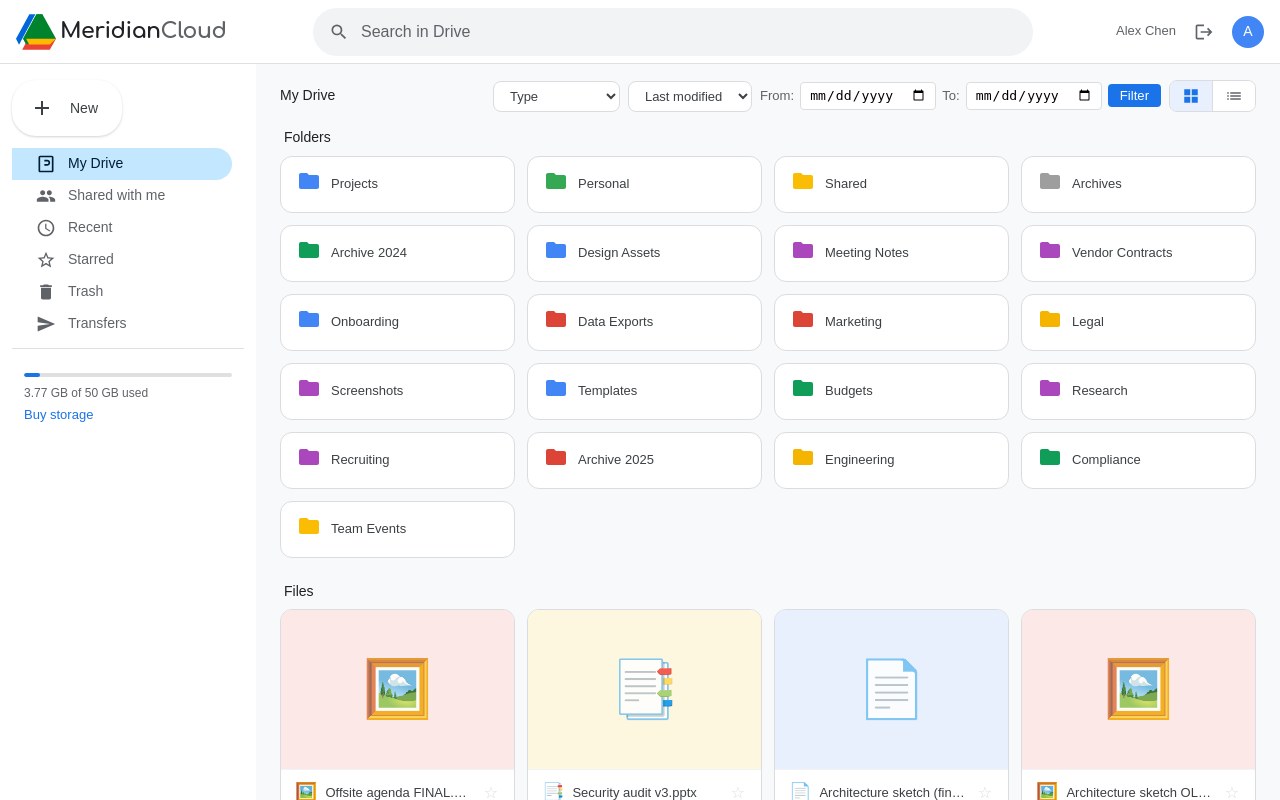}\\[1pt]
{\small\textbf{MeridianCloud}\par}
{\scriptsize\color{black!65} \texttt{cloud-storage-file-transfer} $\cdot$ Productivity $\cdot$ train $\cdot$ synthetic data $\cdot$ 7 tasks $\cdot$ 15 primitives\par}\vspace{1pt}
{\scriptsize A cloud-drive and file-transfer portal in the style of Google Drive / Dropbox, holding a software team's workspace at Meridian Systems. Users browse a folder tree, search and filter files, star, rename, move, trash and download them, share files with teammates and send file transfers via share links.\par}
\end{minipage}
\par\bigskip\noindent
\begin{minipage}[t]{0.32\linewidth}\raggedright
\includegraphics[width=\linewidth]{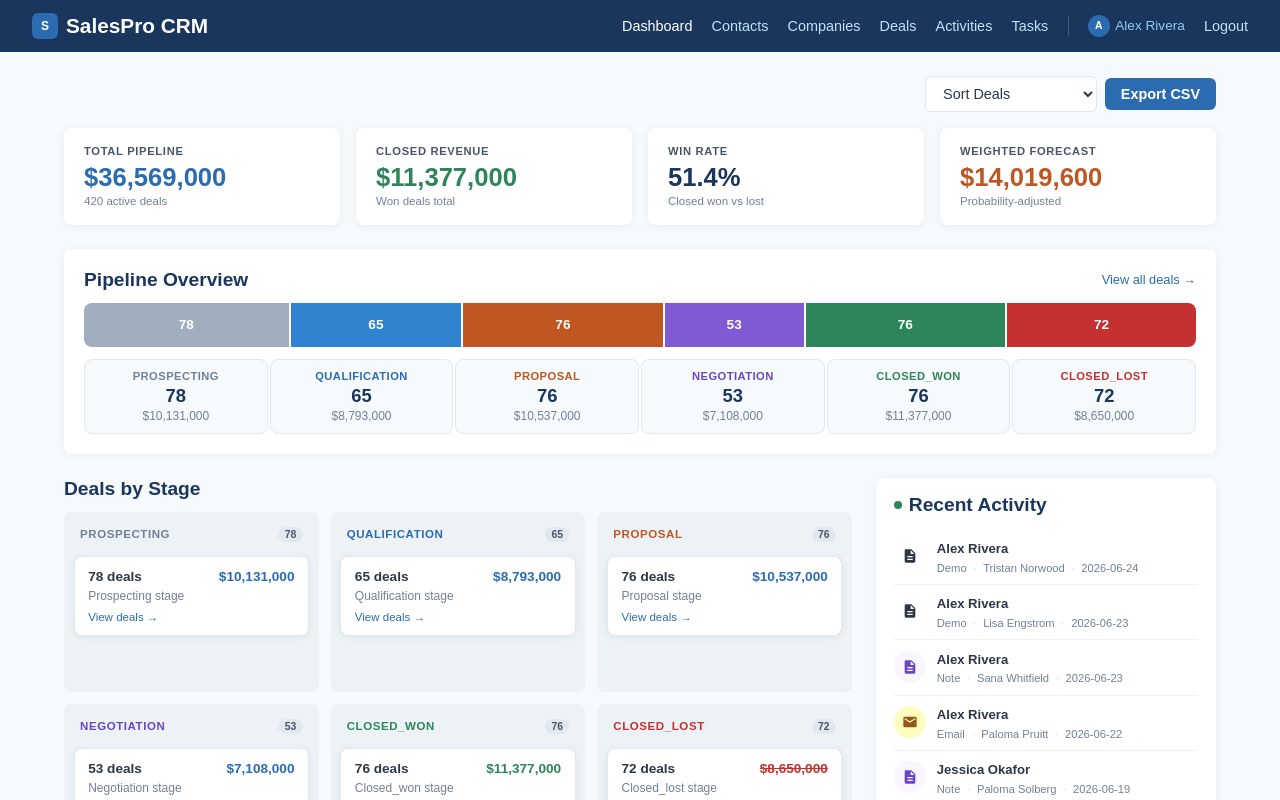}\\[1pt]
{\small\textbf{SalesPro CRM}\par}
{\scriptsize\color{black!65} \texttt{crm} $\cdot$ Productivity $\cdot$ \textbf{held-out} $\cdot$ synthetic data $\cdot$ 7 tasks $\cdot$ 11 primitives\par}\vspace{1pt}
{\scriptsize A Salesforce/HubSpot-style CRM for a sales team: contacts, companies, a deal pipeline, logged activities and a follow-up task queue.\par}
\end{minipage}\hfill\begin{minipage}[t]{0.32\linewidth}\raggedright
\includegraphics[width=\linewidth]{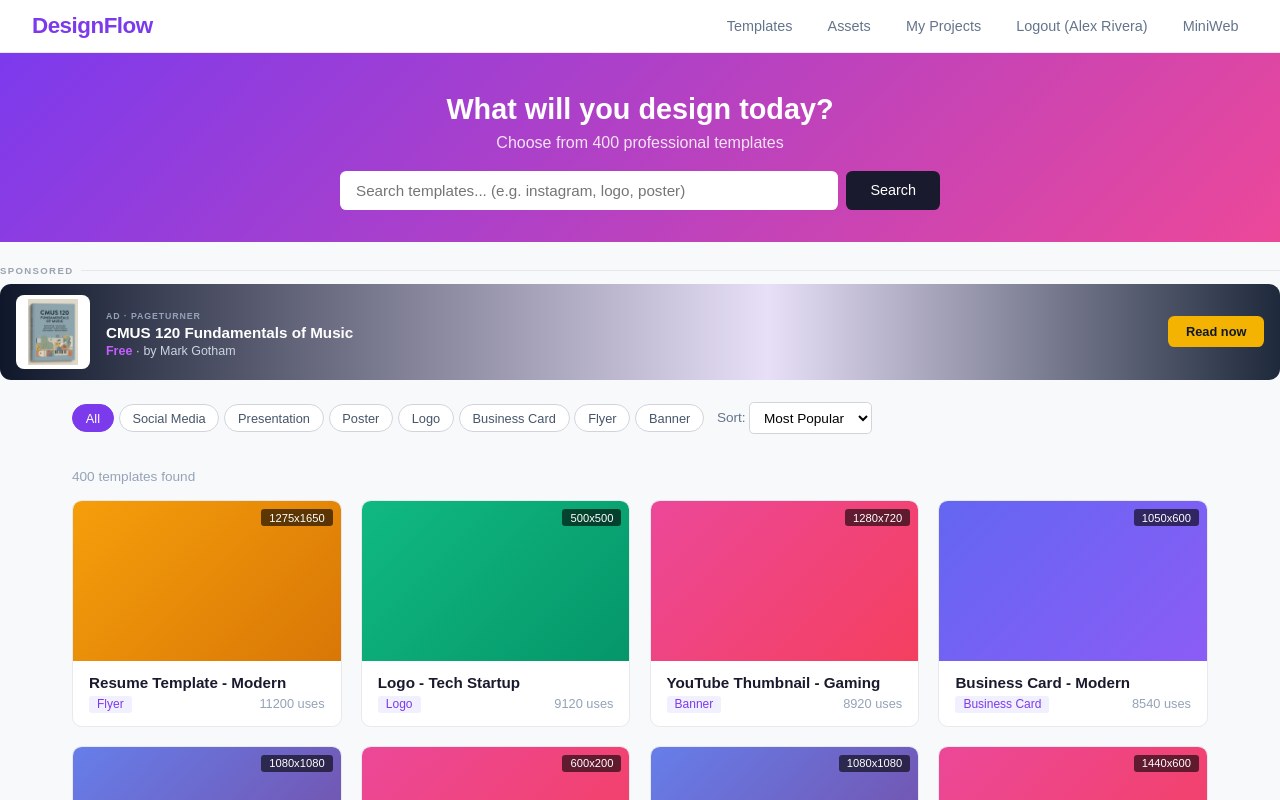}\\[1pt]
{\small\textbf{DesignFlow}\par}
{\scriptsize\color{black!65} \texttt{design-creative} $\cdot$ Productivity $\cdot$ train $\cdot$ synthetic data $\cdot$ 7 tasks $\cdot$ 14 primitives\par}\vspace{1pt}
{\scriptsize A Canva-style design tool. Users browse a template gallery, start projects from templates, edit designs in a simplified canvas editor (positioned HTML elements, not a real canvas), manage an asset library, invite collaborators and export designs as SVG.\par}
\end{minipage}\hfill\begin{minipage}[t]{0.32\linewidth}\raggedright
\includegraphics[width=\linewidth]{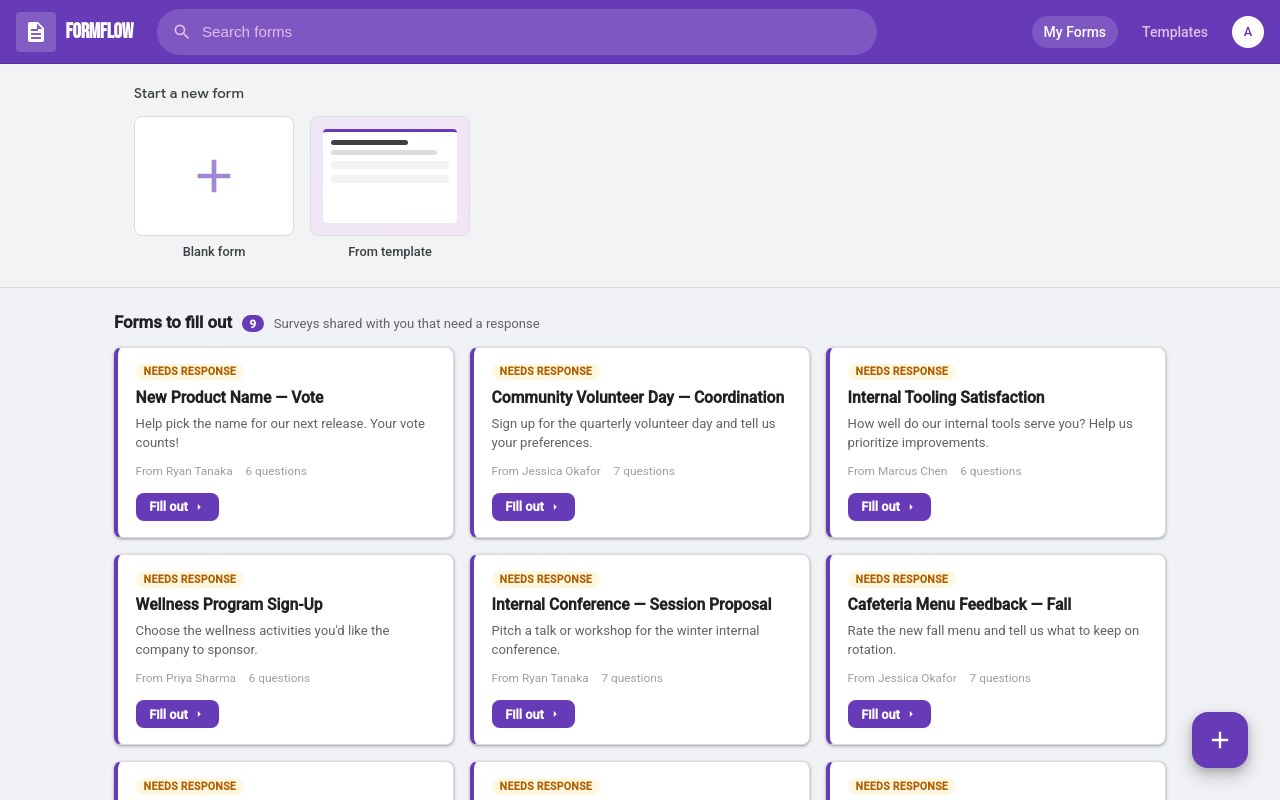}\\[1pt]
{\small\textbf{FormFlow}\par}
{\scriptsize\color{black!65} \texttt{forms-surveys} $\cdot$ Productivity $\cdot$ train $\cdot$ synthetic data $\cdot$ 4 tasks $\cdot$ 8 primitives\par}\vspace{1pt}
{\scriptsize A Google Forms / SurveyMonkey-style form builder. Users create forms from scratch or from templates, fill them in, review results and analytics, edit responses in a grid, attach files and share links.\par}
\end{minipage}
\par\bigskip\noindent
\begin{minipage}[t]{0.32\linewidth}\raggedright
\includegraphics[width=\linewidth]{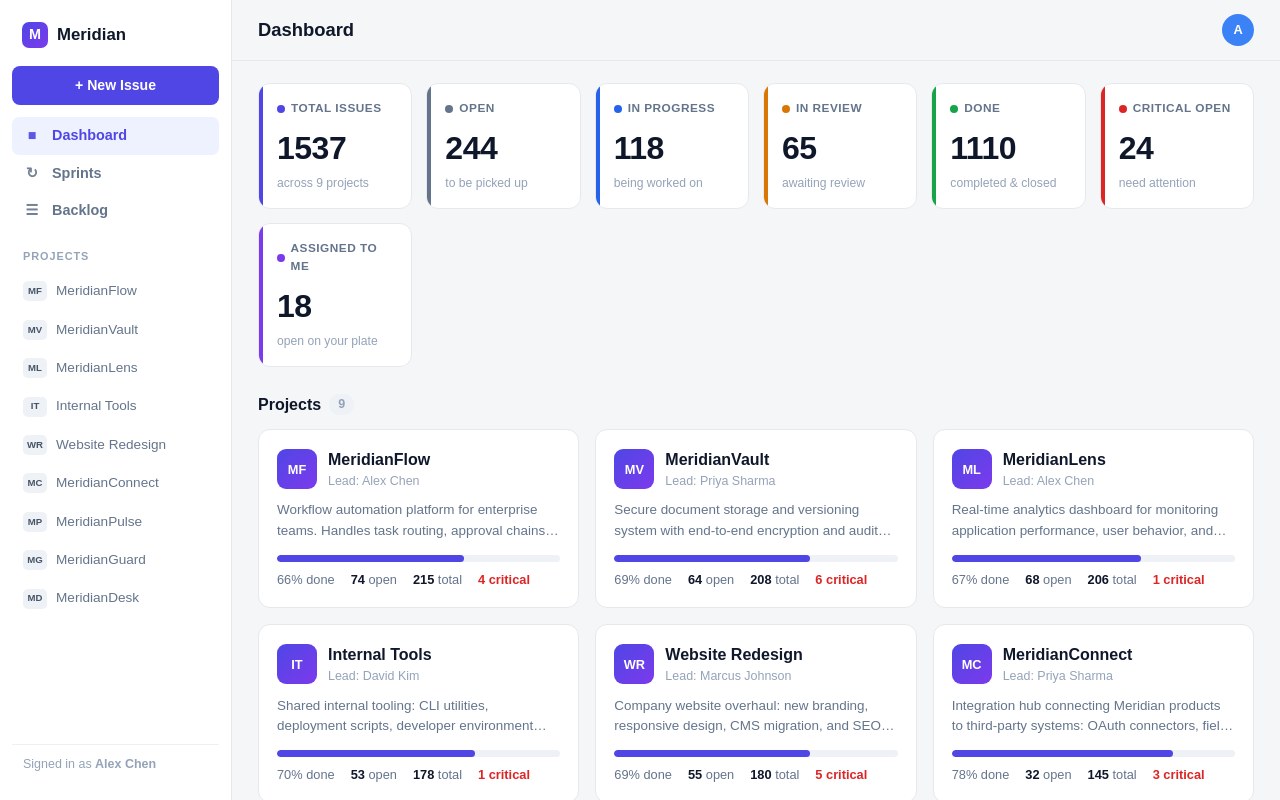}\\[1pt]
{\small\textbf{Meridian Tracker}\par}
{\scriptsize\color{black!65} \texttt{project-mgmt-issue-tracking} $\cdot$ Productivity $\cdot$ \textbf{held-out} $\cdot$ data: Jira $\cdot$ 5 tasks $\cdot$ 12 primitives\par}\vspace{1pt}
{\scriptsize A Jira-style project tracker for the fictional Meridian Systems. It has per-project Kanban boards, issue detail pages with comments and status transitions, sprints, a backlog grid and issue export. Issue keys follow Jira style (\texttt{MF-101}).\par}
\end{minipage}\hfill\begin{minipage}[t]{0.32\linewidth}\raggedright
\includegraphics[width=\linewidth]{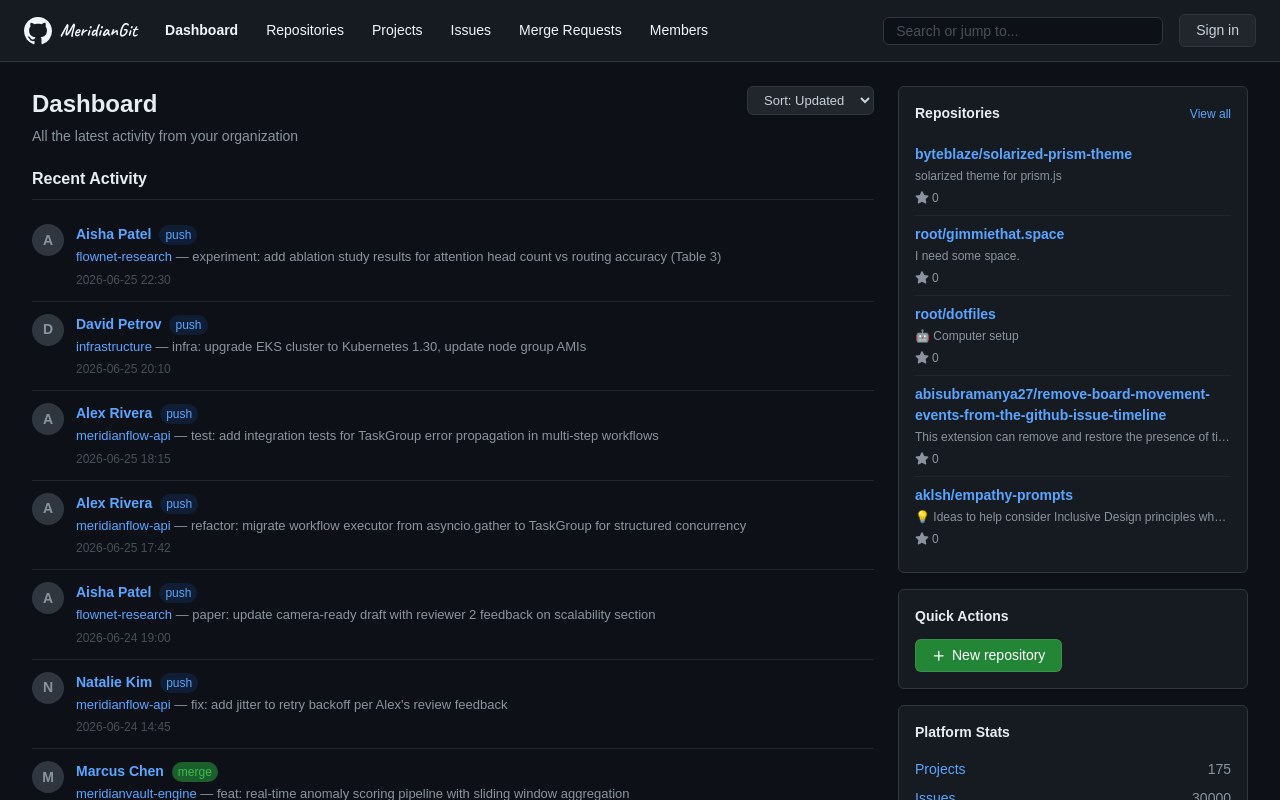}\\[1pt]
{\small\textbf{MeridianGit}\par}
{\scriptsize\color{black!65} \texttt{version-control} $\cdot$ Productivity $\cdot$ train $\cdot$ data: WebArena GitLab $\cdot$ 5 tasks $\cdot$ 9 primitives\par}\vspace{1pt}
{\scriptsize A GitHub/GitLab-style code-hosting site for the fictional Meridian Systems engineering team, mixed with real GitLab projects, issues and merge requests. Users browse repositories (code, commits, issues, merge requests), search code, create repositories, issues and merge requests, comment, star repos, upload files and merge or close merge requests.\par}
\end{minipage}\hfill\begin{minipage}[t]{0.32\linewidth}\raggedright
\includegraphics[width=\linewidth]{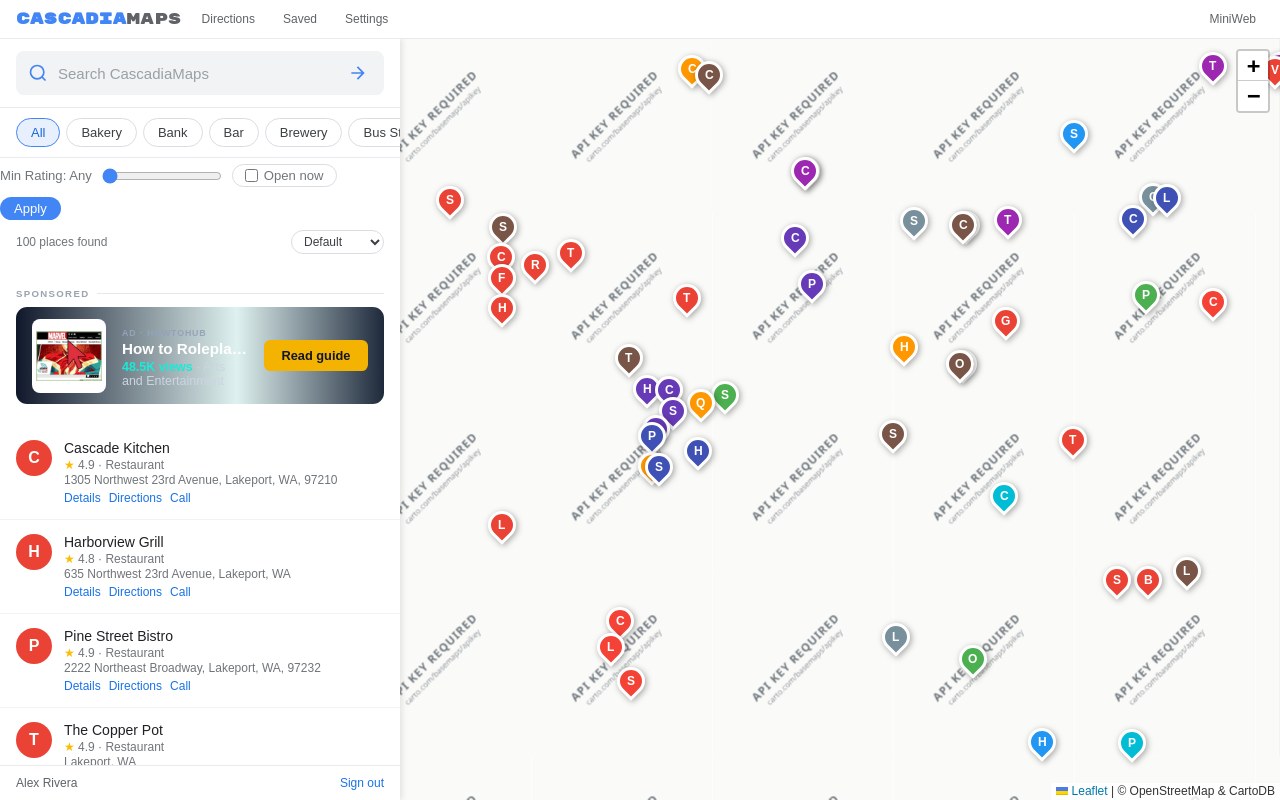}\\[1pt]
{\small\textbf{CascadiaMaps}\par}
{\scriptsize\color{black!65} \texttt{map-services} $\cdot$ Maps \& navigation $\cdot$ train $\cdot$ data: OpenStreetMap $\cdot$ 8 tasks $\cdot$ 10 primitives\par}\vspace{1pt}
{\scriptsize A Google Maps-style map and navigation service for Lakeport, WA. Users search places on an interactive Leaflet map, filter by category, rating and ``open now'', open place pages with hours and reviews, save places, write reviews, get directions by travel mode and compare places.\par}
\end{minipage}
\par\bigskip\noindent
\begin{minipage}[t]{0.32\linewidth}\raggedright
\includegraphics[width=\linewidth]{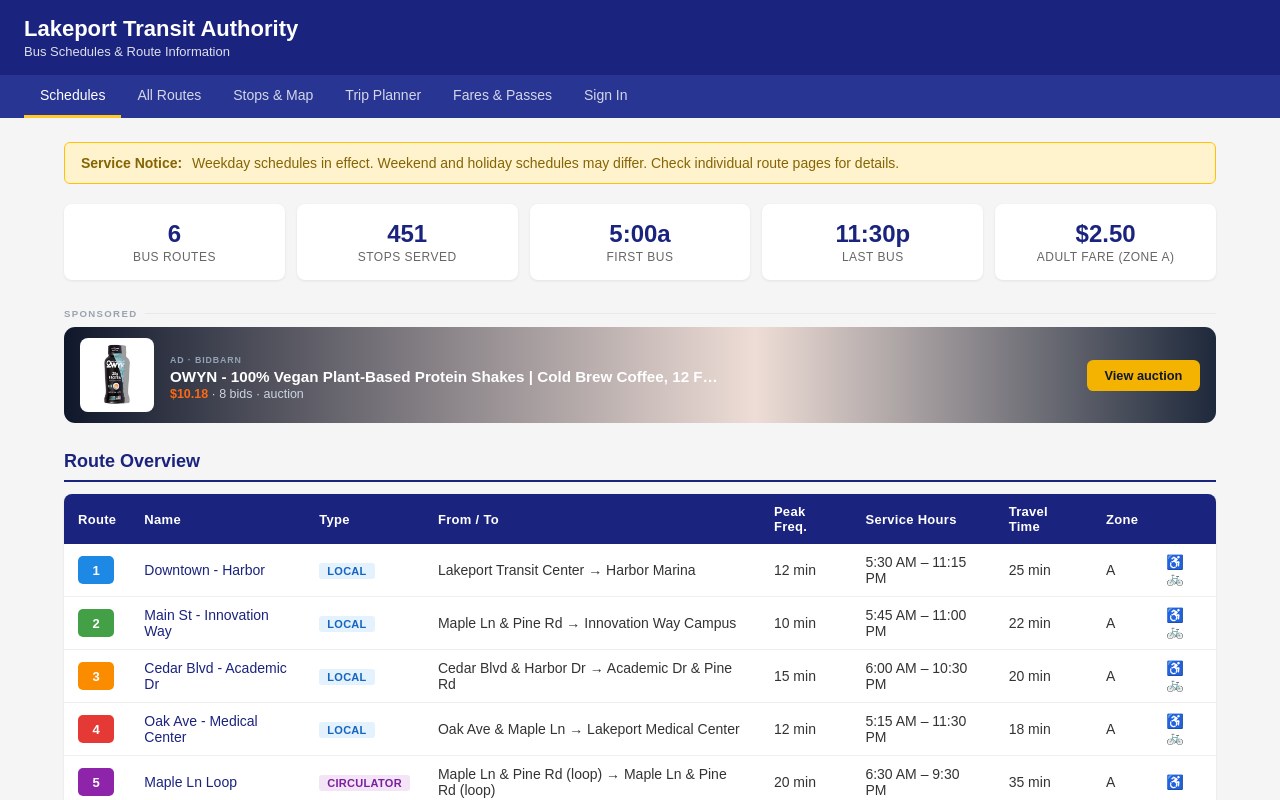}\\[1pt]
{\small\textbf{Lakeport Transit Authority}\par}
{\scriptsize\color{black!65} \texttt{transit-directions} $\cdot$ Maps \& navigation $\cdot$ train $\cdot$ synthetic data $\cdot$ 4 tasks $\cdot$ 5 primitives\par}\vspace{1pt}
{\scriptsize The public-transit website of the fictional Lakeport Transit Authority (LTA), in the style of Google Transit / TriMet. It lists bus routes and stops with timetables, plans trips with route preferences, explains fares and passes, and compares routes side by side.\par}
\end{minipage}\hfill\begin{minipage}[t]{0.32\linewidth}\raggedright
\includegraphics[width=\linewidth]{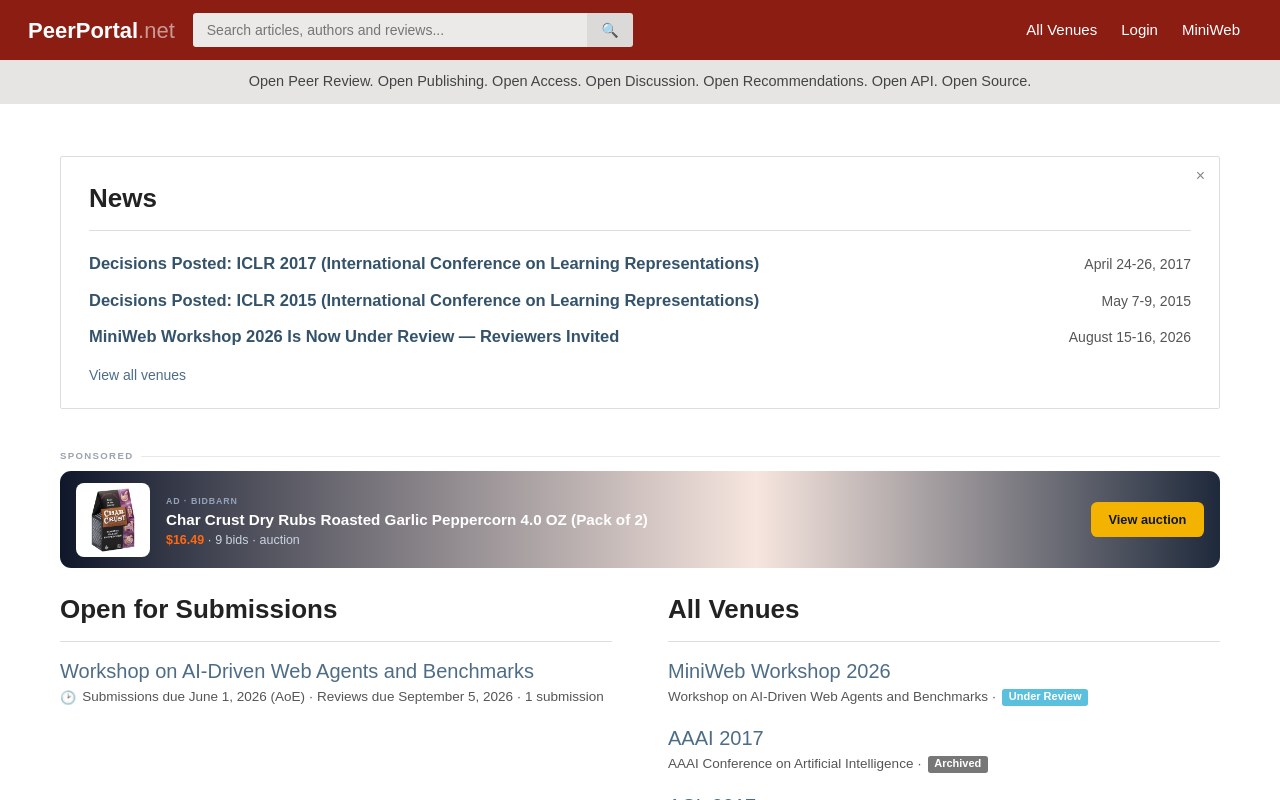}\\[1pt]
{\small\textbf{PeerPortal}\par}
{\scriptsize\color{black!65} \texttt{conference-review-submission} $\cdot$ Education $\cdot$ \textbf{held-out} $\cdot$ data: PeerRead $\cdot$ 2 tasks $\cdot$ 8 primitives\par}\vspace{1pt}
{\scriptsize An OpenReview-style venue and peer-review system. Visitors browse venues and papers with their reviews and scores. Signed-in reviewers bid on papers, write reviews and upload submissions, and chairs assign reviewers and post decisions.\par}
\end{minipage}\hfill\begin{minipage}[t]{0.32\linewidth}\raggedright
\includegraphics[width=\linewidth]{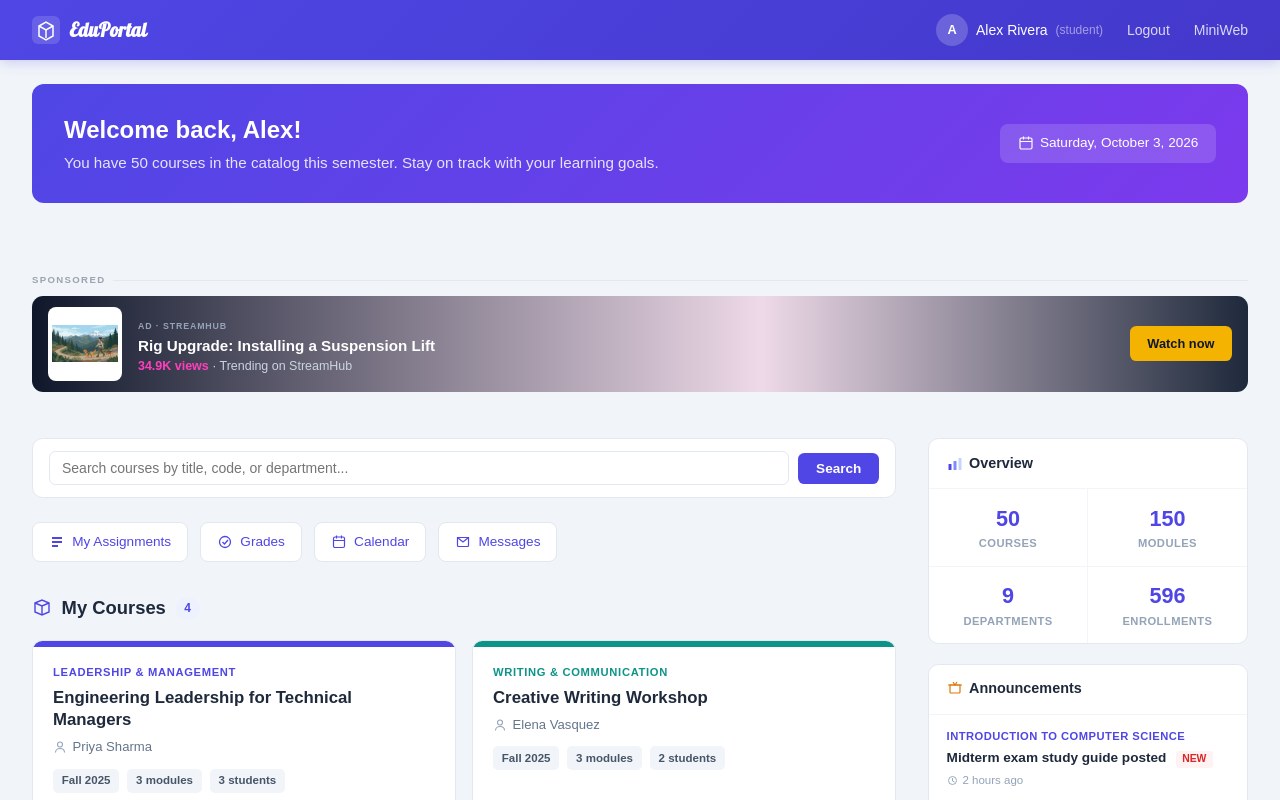}\\[1pt]
{\small\textbf{EduPortal LMS}\par}
{\scriptsize\color{black!65} \texttt{course-sites-classrooms} $\cdot$ Education $\cdot$ train $\cdot$ synthetic data $\cdot$ 2 tasks $\cdot$ 8 primitives\par}\vspace{1pt}
{\scriptsize A Canvas / Moodle-style learning management system. Students browse and enroll in courses, play lectures, submit assignments (text plus an optional file) and post in discussion boards. Instructors and admins also grade submissions and edit the gradebook. Access is role-based (admin / instructor / student).\par}
\end{minipage}
\par\bigskip\noindent
\begin{minipage}[t]{0.32\linewidth}\raggedright
\includegraphics[width=\linewidth]{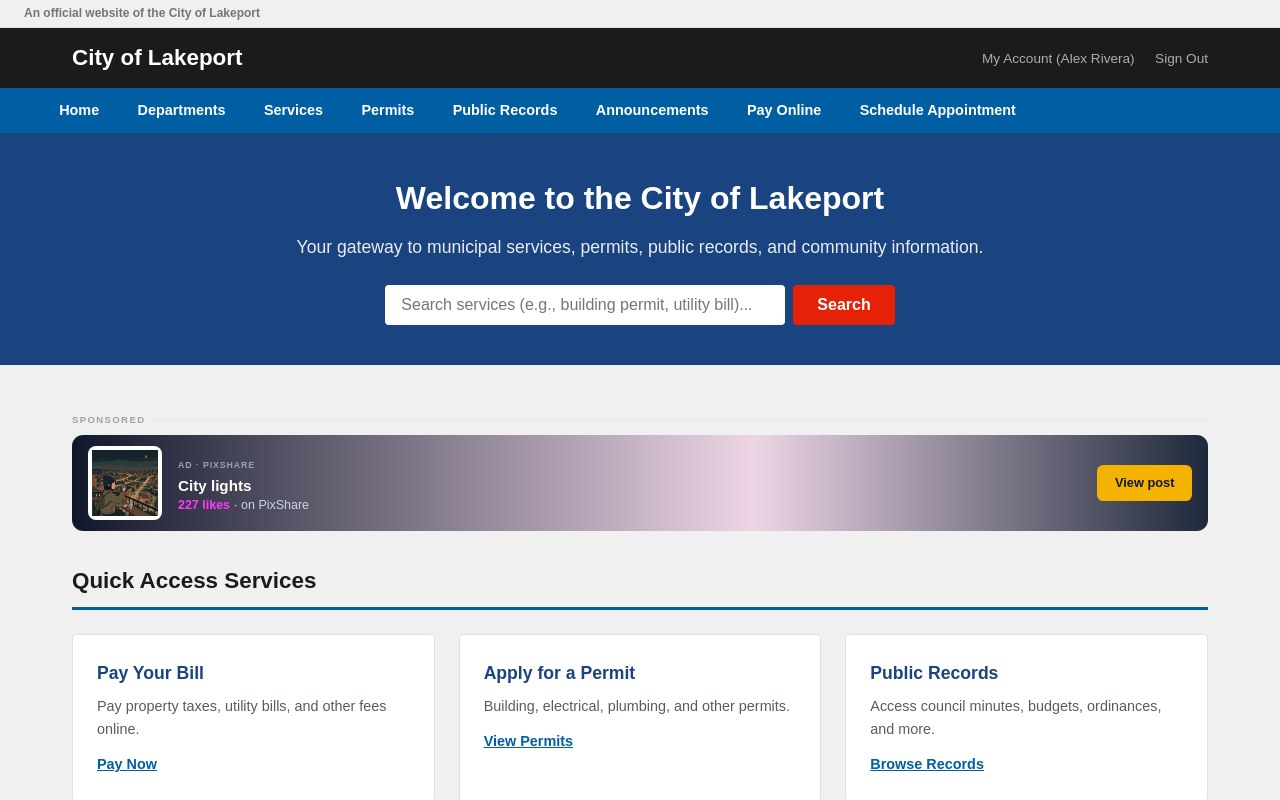}\\[1pt]
{\small\textbf{City of Lakeport}\par}
{\scriptsize\color{black!65} \texttt{agency-portals} $\cdot$ Government \& civic $\cdot$ train $\cdot$ synthetic data $\cdot$ 10 tasks $\cdot$ 10 primitives\par}\vspace{1pt}
{\scriptsize A \texttt{.gov}-style municipal services portal for the fictional city of Lakeport, WA. Residents browse departments, services, permits, public records and announcements, and (signed in) apply for services, book appointments, pay city bills and upload documents.\par}
\end{minipage}\hfill\begin{minipage}[t]{0.32\linewidth}\raggedright
\includegraphics[width=\linewidth]{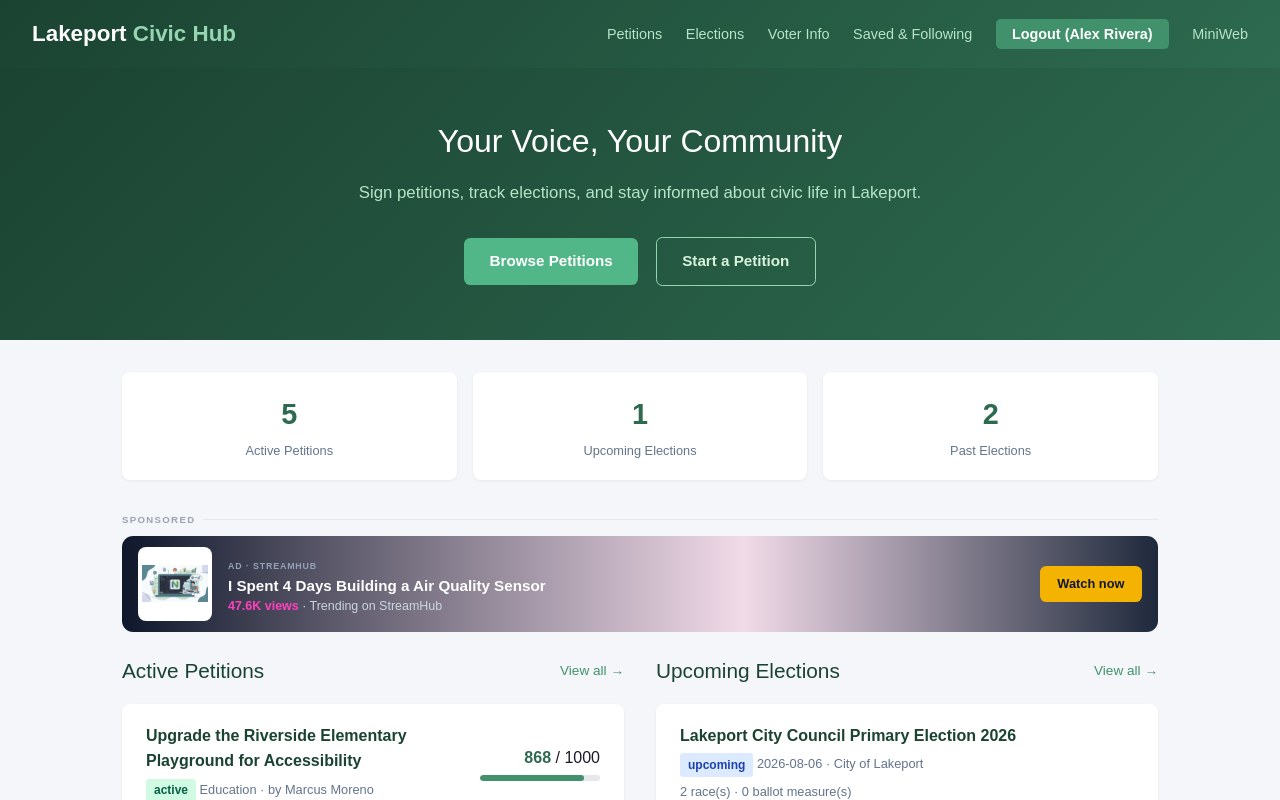}\\[1pt]
{\small\textbf{Lakeport Civic Hub}\par}
{\scriptsize\color{black!65} \texttt{petitions-voting-info} $\cdot$ Government \& civic $\cdot$ \textbf{held-out} $\cdot$ synthetic data $\cdot$ 4 tasks $\cdot$ 9 primitives\par}\vspace{1pt}
{\scriptsize A civic engagement portal for the fictional City of Lakeport, WA, combining Change.org-style petitions (create, sign, share, subscribe) with a county elections site (election results, voter registration, polling places).\par}
\end{minipage}\hfill\begin{minipage}[t]{0.32\linewidth}\raggedright
\includegraphics[width=\linewidth]{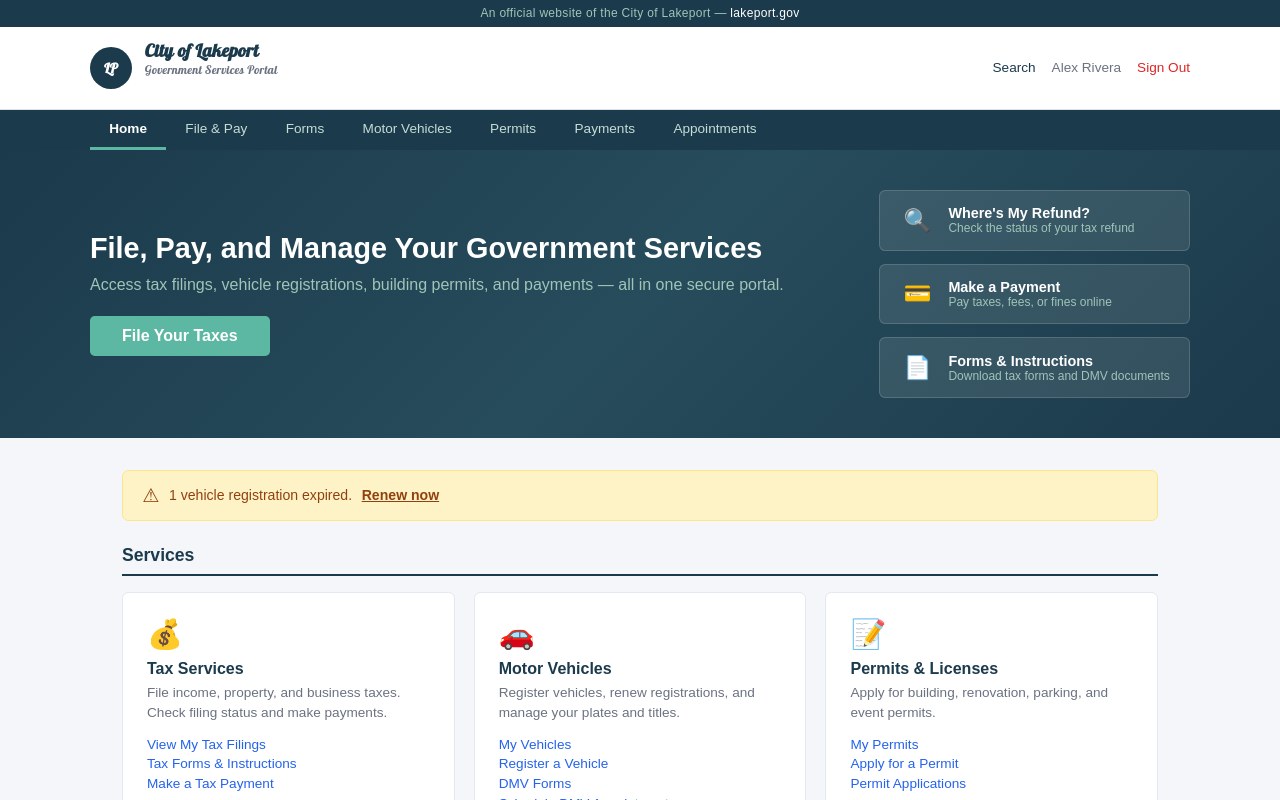}\\[1pt]
{\small\textbf{Lakeport Government Services}\par}
{\scriptsize\color{black!65} \texttt{tax-filing-dmv-permits} $\cdot$ Government \& civic $\cdot$ train $\cdot$ synthetic data $\cdot$ 7 tasks $\cdot$ 14 primitives\par}\vspace{1pt}
{\scriptsize A state/city services portal combining tax filing, motor-vehicle (DMV) records, permits, payments and appointments, in the style of IRS and DMV online services. Users file a Form 1040, sign it, register vehicles, apply for permits, pay fees, book DMV appointments and track refunds.\par}
\end{minipage}
\par\bigskip\noindent
\begin{minipage}[t]{0.32\linewidth}\raggedright
\includegraphics[width=\linewidth]{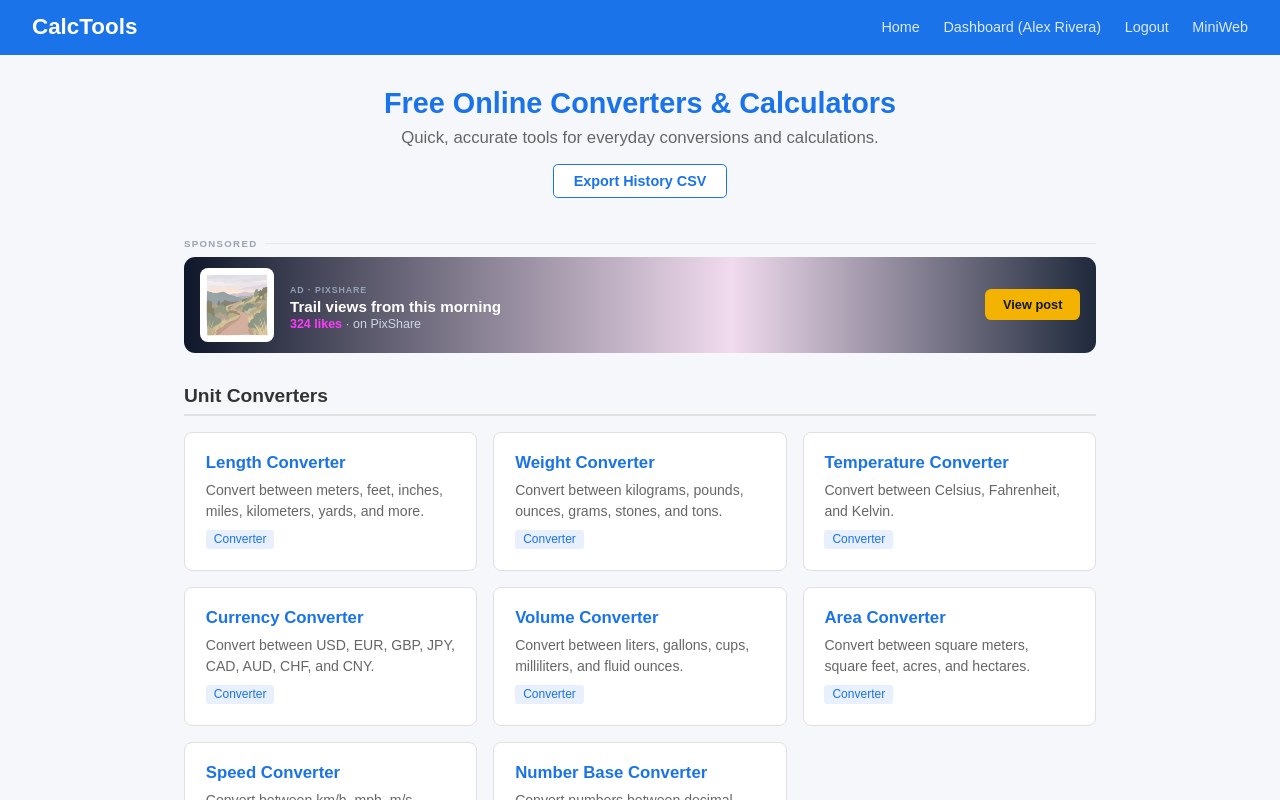}\\[1pt]
{\small\textbf{CalcTools}\par}
{\scriptsize\color{black!65} \texttt{converters-calculators} $\cdot$ Utilities $\cdot$ train $\cdot$ synthetic data $\cdot$ 5 tasks $\cdot$ 5 primitives\par}\vspace{1pt}
{\scriptsize A free unit-converter and calculator site modeled on Calculator.net. It offers length, weight, temperature, currency, volume, area, speed and number-base converters plus BMI, mortgage and tip calculators. Signed-in users can save results and see their conversion history.\par}
\end{minipage}\hfill\begin{minipage}[t]{0.32\linewidth}\raggedright
\includegraphics[width=\linewidth]{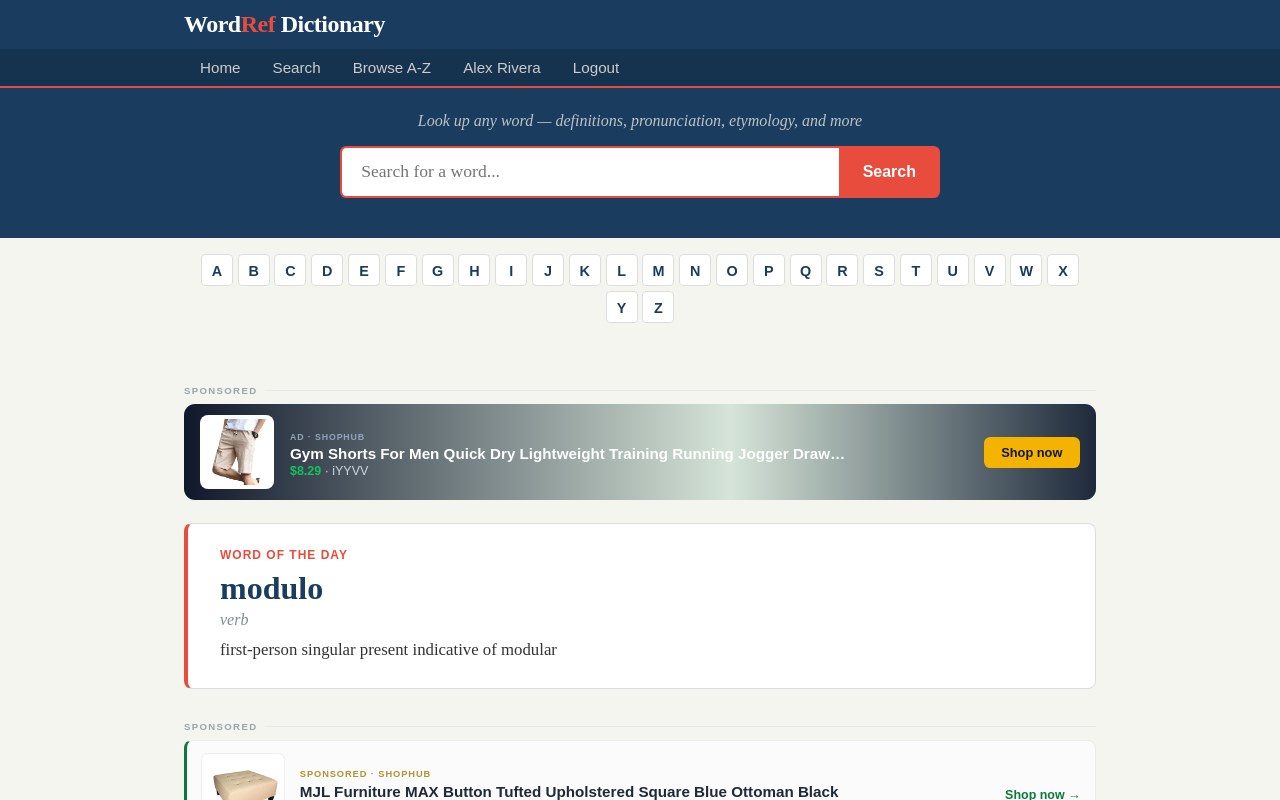}\\[1pt]
{\small\textbf{WordRef Dictionary}\par}
{\scriptsize\color{black!65} \texttt{dictionaries-language-tools} $\cdot$ Utilities $\cdot$ train $\cdot$ data: Wiktionary $\cdot$ 2 tasks $\cdot$ 6 primitives\par}\vspace{1pt}
{\scriptsize An English dictionary in the style of Merriam-Webster / Dictionary.com, built from Wiktionary entries. Users search words, browse A-Z, read definitions, IPA pronunciations, word forms, examples, synonyms and antonyms, and save words to a personal list.\par}
\end{minipage}\hfill\begin{minipage}[t]{0.32\linewidth}\raggedright
\includegraphics[width=\linewidth]{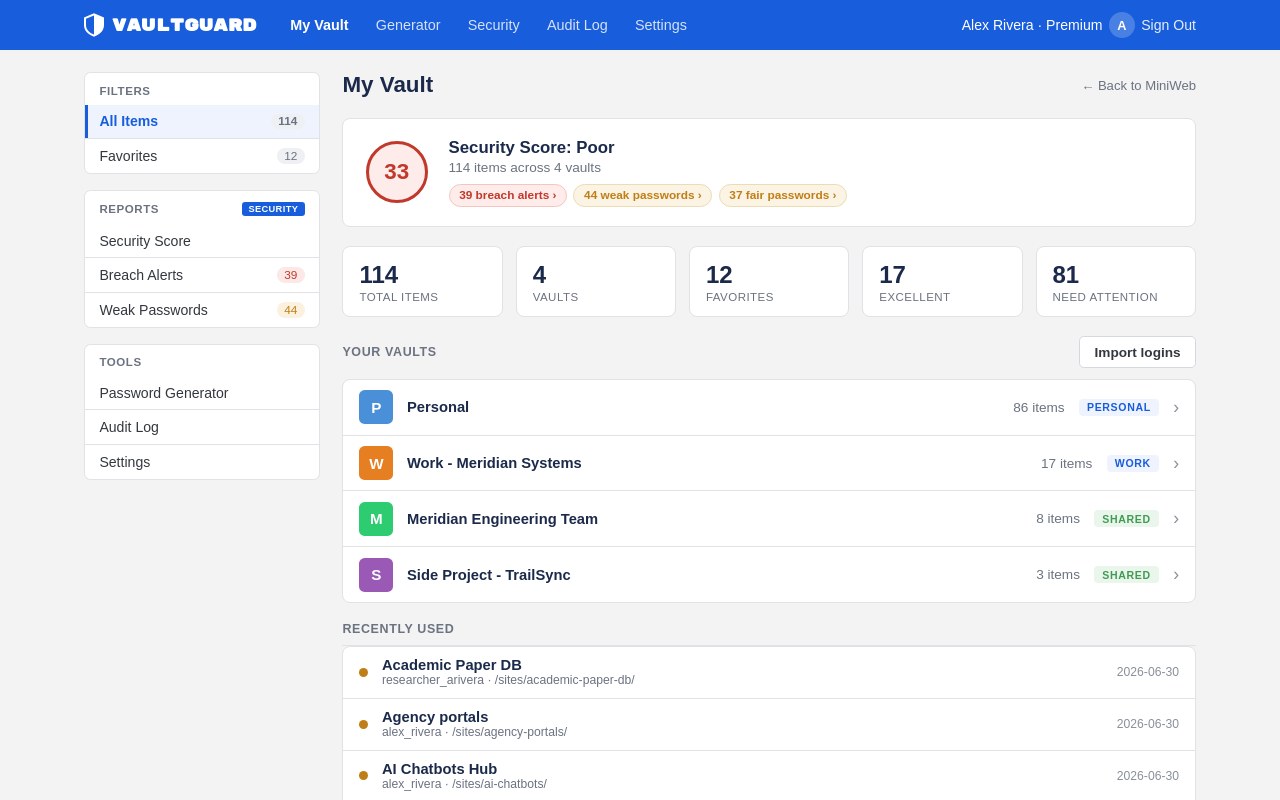}\\[1pt]
{\small\textbf{VaultGuard}\par}
{\scriptsize\color{black!65} \texttt{password-managers} $\cdot$ Utilities $\cdot$ train $\cdot$ synthetic data $\cdot$ 5 tasks $\cdot$ 10 primitives\par}\vspace{1pt}
{\scriptsize A password manager in the style of 1Password, LastPass and Bitwarden. Users browse vaults, search and filter entries, create, edit, delete, import and export entries, reveal a password with a PIN, generate passwords and review a security report and audit log.\par}
\end{minipage}
\par\bigskip\noindent
\begin{minipage}[t]{0.32\linewidth}\raggedright
\includegraphics[width=\linewidth]{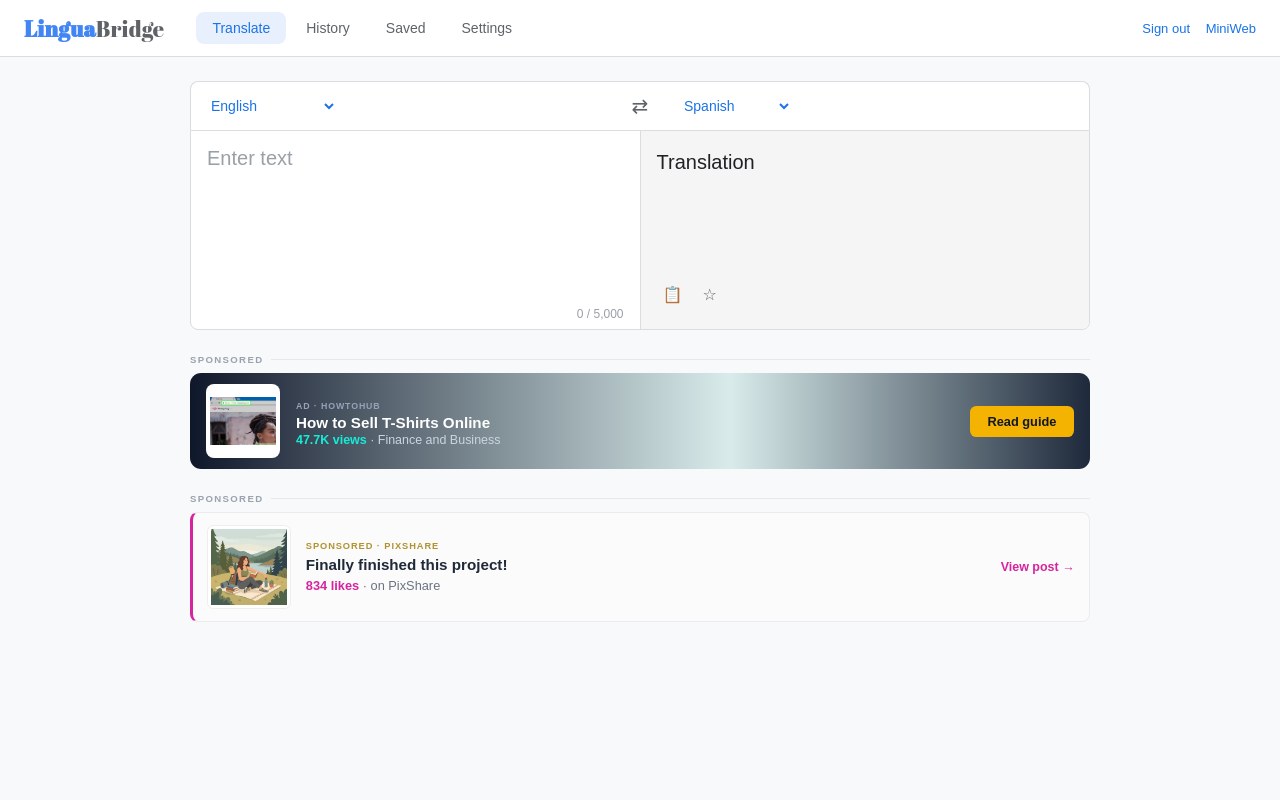}\\[1pt]
{\small\textbf{LinguaBridge Translate}\par}
{\scriptsize\color{black!65} \texttt{translation} $\cdot$ Utilities $\cdot$ train $\cdot$ synthetic data $\cdot$ 8 tasks $\cdot$ 5 primitives\par}\vspace{1pt}
{\scriptsize A Google Translate / DeepL-style translator with source and target language dropdowns, translation history, saved translations, custom glossaries and user settings.\par}
\end{minipage}\hfill\begin{minipage}[t]{0.32\linewidth}\raggedright
\includegraphics[width=\linewidth]{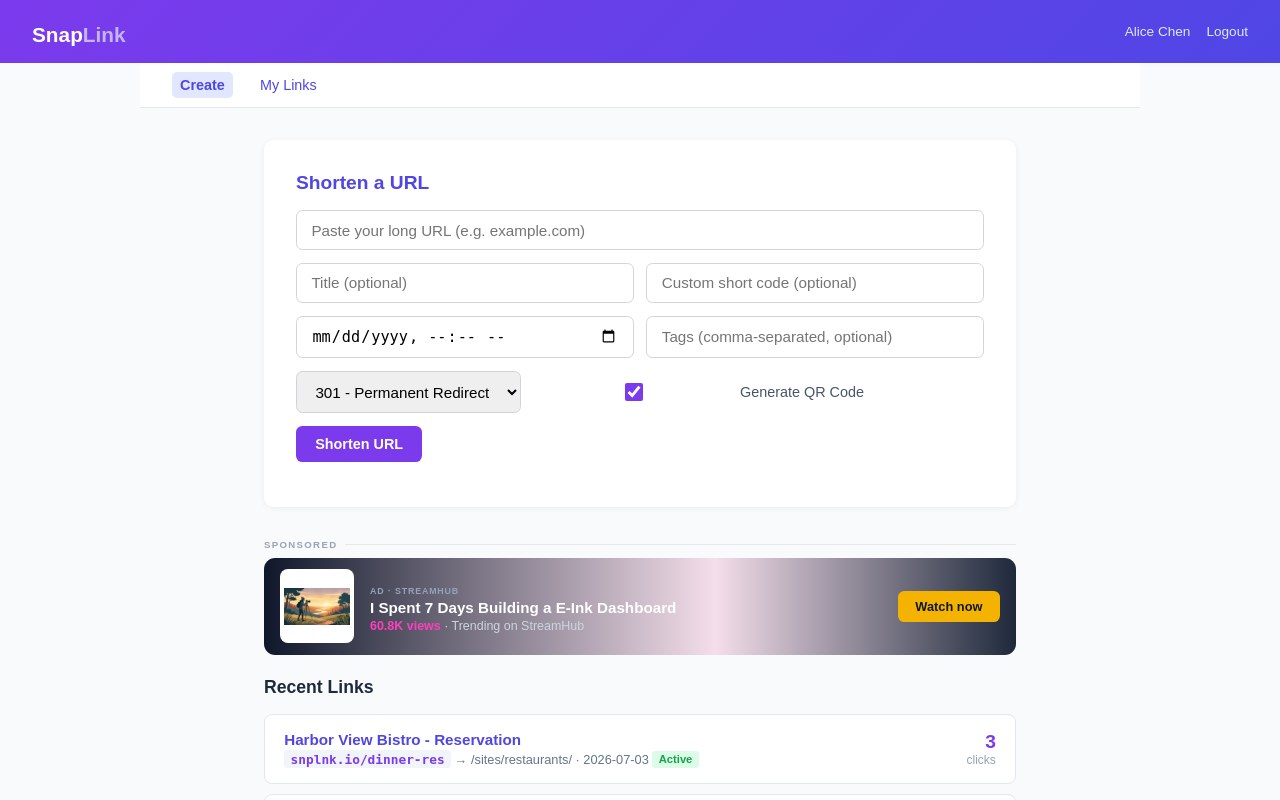}\\[1pt]
{\small\textbf{SnapLink URL Shortener}\par}
{\scriptsize\color{black!65} \texttt{url-shorteners-qr} $\cdot$ Utilities $\cdot$ \textbf{held-out} $\cdot$ synthetic data $\cdot$ 4 tasks $\cdot$ 10 primitives\par}\vspace{1pt}
{\scriptsize A Bitly, TinyURL or Rebrandly-style link manager. Users shorten URLs with custom codes, expiry, tags, redirect types and QR codes, manage and export their links, and read per-link click statistics.\par}
\end{minipage}\hfill\begin{minipage}[t]{0.32\linewidth}\mbox{}\end{minipage}

\section{Annotation interface}\label{app:annotation}
\begin{figure}[tbp]
    \centering
    \includegraphics[width=0.88\linewidth]{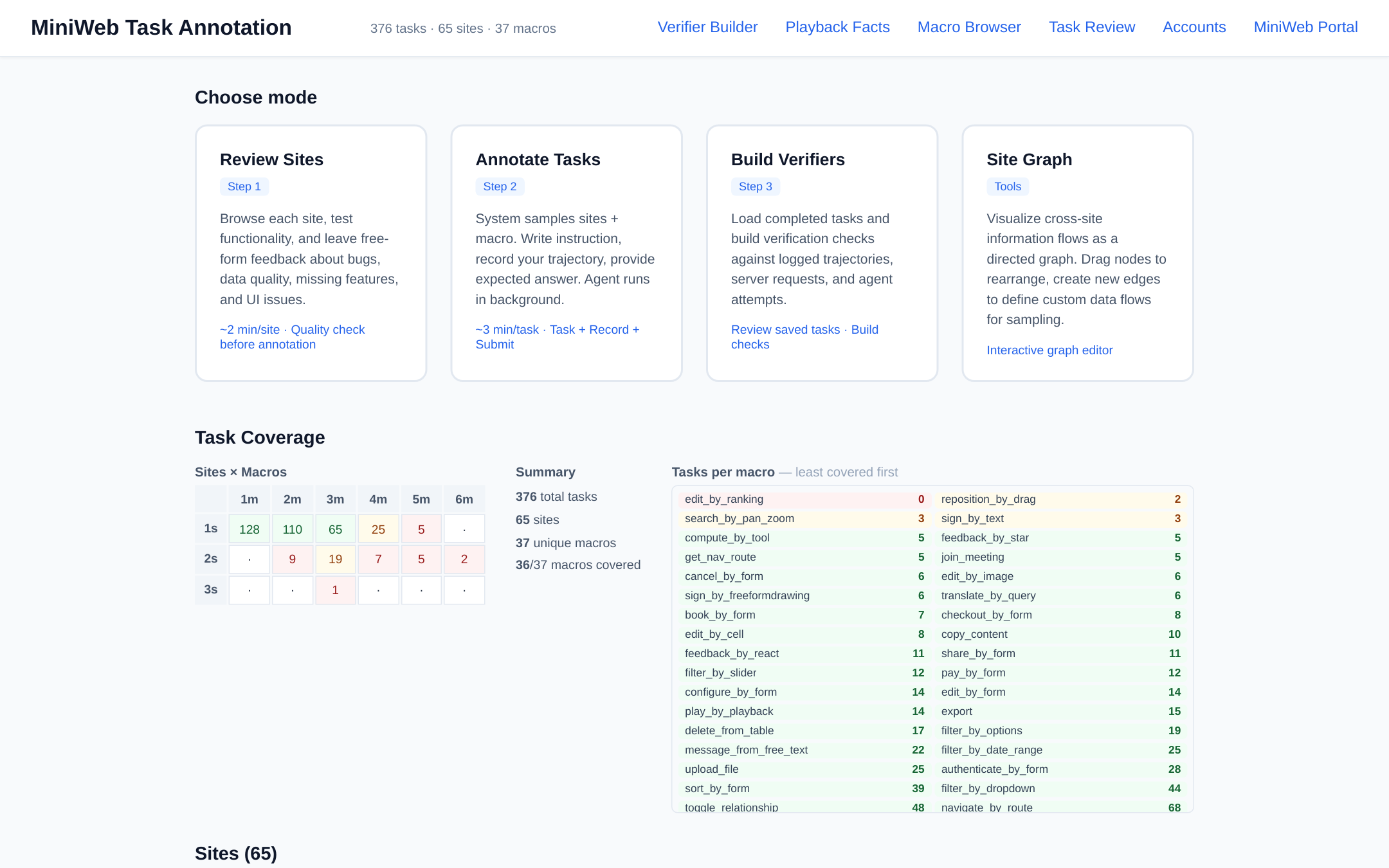}\\[1pt]{\small (a)}\\[4pt]
    \includegraphics[width=0.88\linewidth]{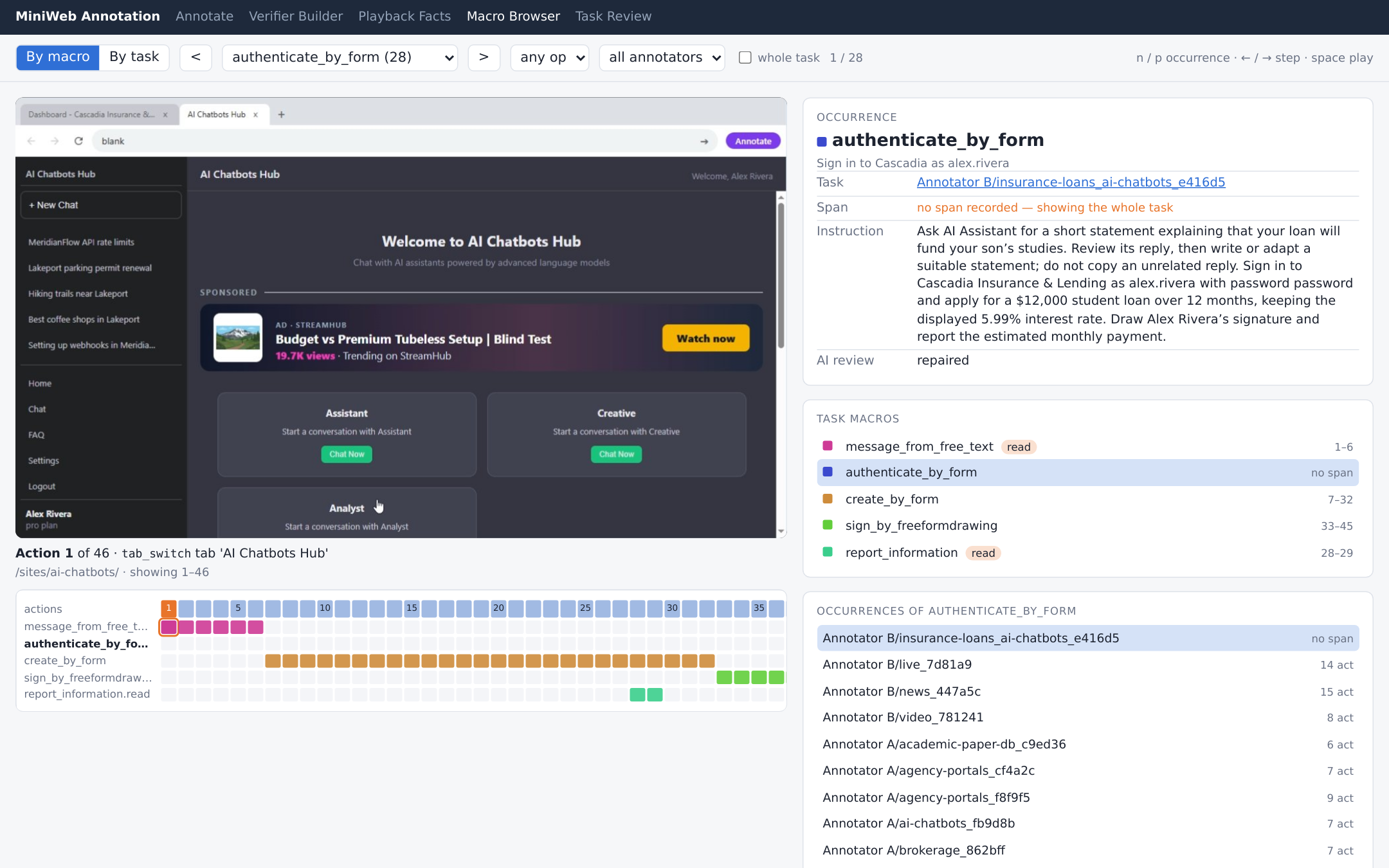}\\[1pt]{\small (b)}
    \caption{\textbf{(a)}~Dashboard of the annotation tool: the three annotation steps and the coverage of sites and primitives. \textbf{(b)}~Macro browser: one recorded span of \texttt{authenticate\_by\_form}, its place on the task's action timeline (one row per primitive instance), and the other tasks that contain it.}
    \label{fig:supp-annot-home}
\end{figure}

Six annotators wrote and recorded the 369 tasks in a web tool served next to the sites (Fig.~\ref{fig:supp-annot-home}). Its dashboard orders the work in three steps: review each site and report bugs, annotate tasks, and build their verifiers. It also shows the current coverage of every primitive, least covered first.

\begin{figure}[tbp]
    \centering
    \includegraphics[width=0.88\linewidth]{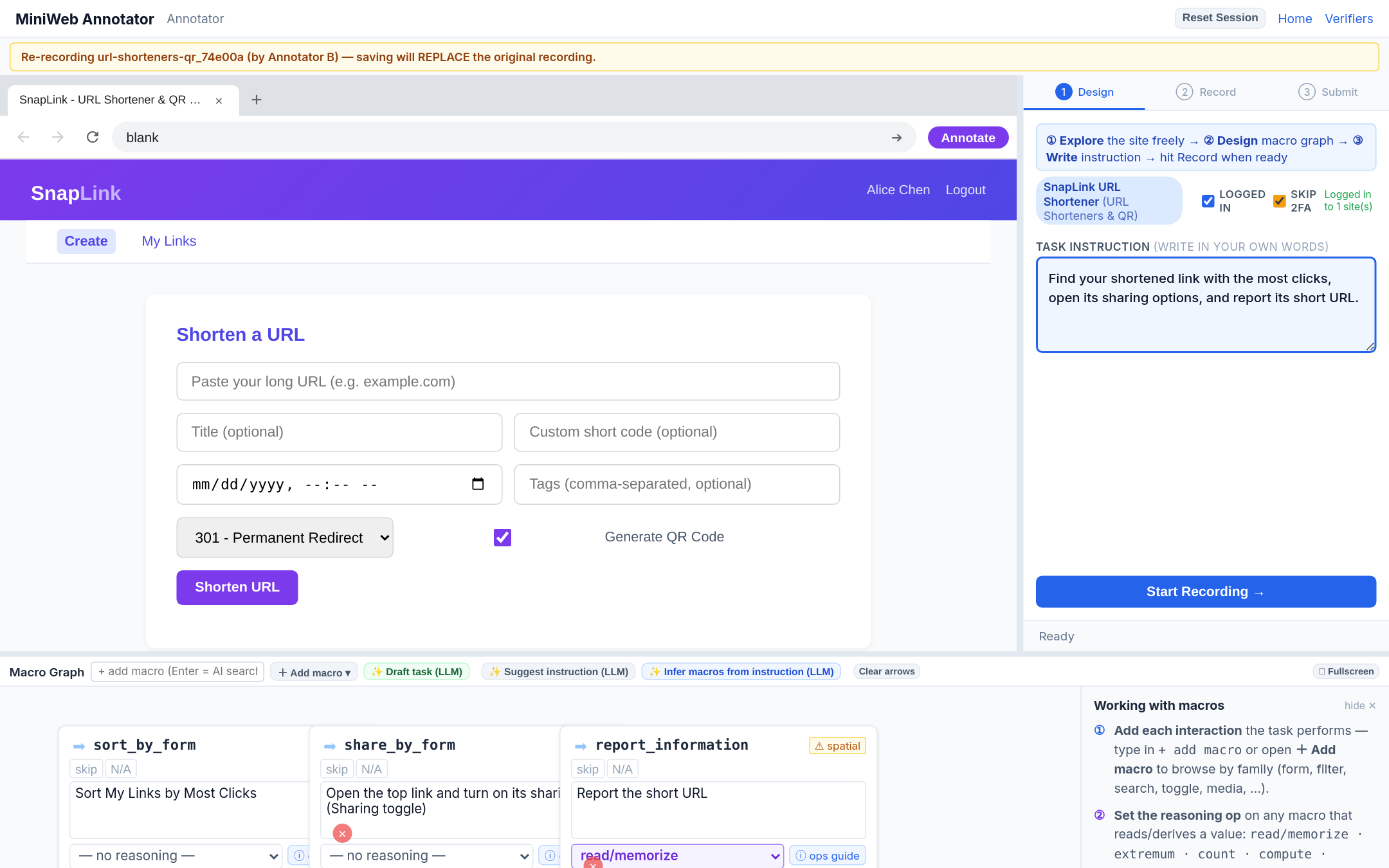}
    \caption{\textbf{Task editor.} The embedded site (left), the instruction and the design, record, and submit steps (right), and the primitive graph with one node per instance, its subtask, and its reasoning operation (bottom).}
    \label{fig:supp-annot-task}
\end{figure}

\paragraph{Annotating a task (Fig.~\ref{fig:supp-annot-task}).} The tool samples one to three sites and a set of primitives to cover, weighted toward primitives with few tasks. The annotator first explores the sites freely, then designs the task as a graph in the editor below the browser: one node per primitive instance with its subtask and reasoning operation, and edges for the order and data flow between instances. Every primitive of a multi-primitive task must be connected. The annotator then writes the instruction in their own words, sets the starting login and two-factor state, and records the task in the embedded browser. After recording, they mark each instance's span on the action timeline and enter the expected answer. Optional LLM helpers draft a task, rephrase an instruction, or infer a primitive chain from it; their output is only a suggestion. Annotators can also propose a new primitive, which is added to the registry for later review.

\begin{figure}[tbp]
    \centering
    \includegraphics[width=0.88\linewidth]{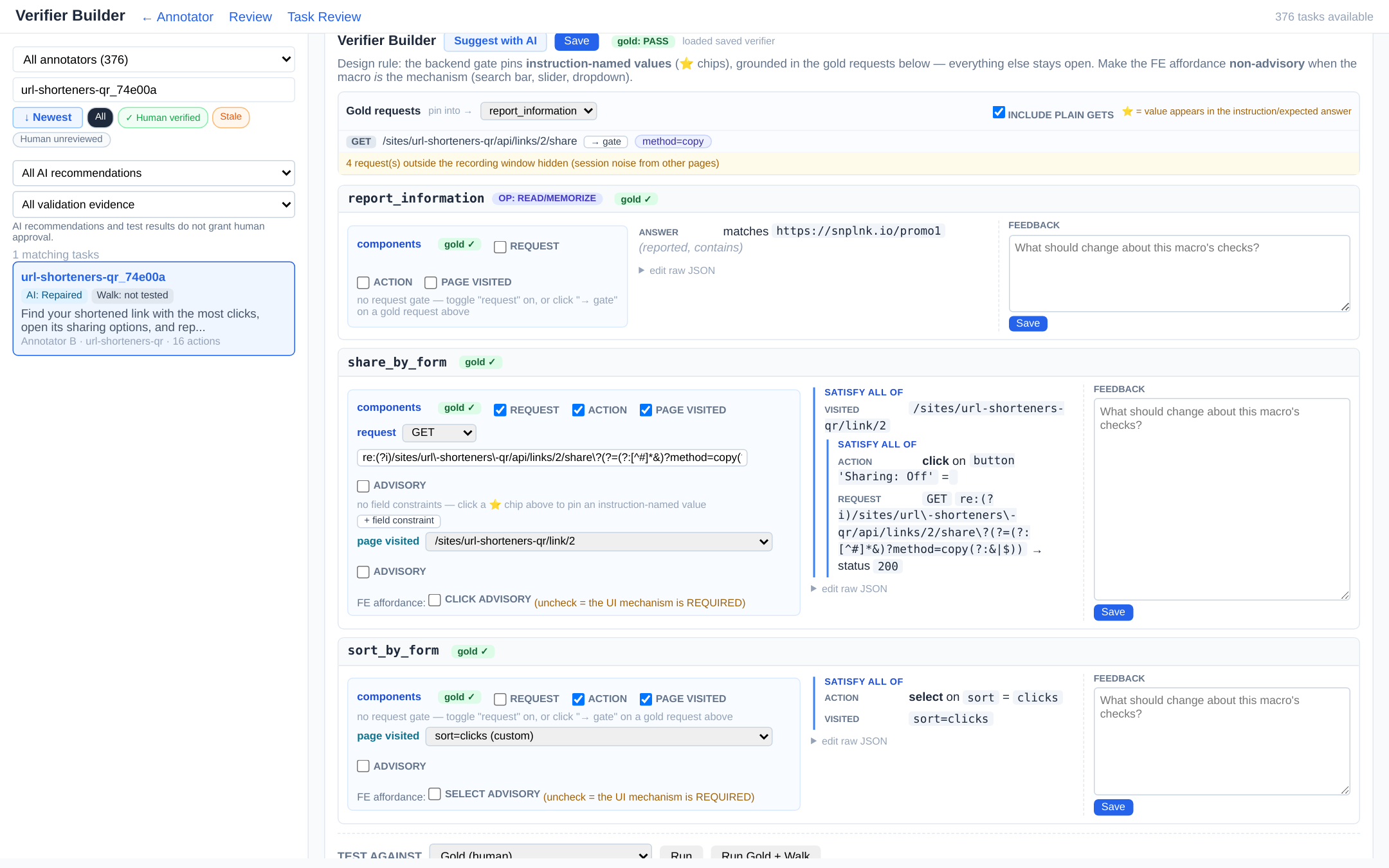}
    \caption{\textbf{Verifier builder.} Each primitive instance of the task gets its own check (request, action, page visited, answer), grounded in the gold requests of the recording; the sandbox below tests the verifier against the gold recording and negative controls.}
    \label{fig:supp-annot-verify}
\end{figure}

\paragraph{Building the verifier (Fig.~\ref{fig:supp-annot-verify}).} For each primitive instance the verifier builder shows the gold HTTP requests from the recording and a check scaffold: request gates, actions, visited pages, and an answer matcher. The annotator pins the values the instruction dictates, marks the remaining checks as advisory, and tests the verifier in a sandbox: it must pass the gold recording and should reject two negative controls, an empty trajectory and an answer given without any action.

\paragraph{Reviewing tasks.} The macro browser (Fig.~\ref{fig:supp-annot-home}b) steps through every recorded span of one primitive across all tasks. Reviewers used it to make tags consistent across annotators. A task review page compares each task with its state before correction; a reviewer marks it verified or stale, asks for a re-recording, or deletes it.

\section{Grader validation}\label{app:graders}
Every task is graded twice: by its parameterized verifiers, which check the backend request log, and by the VLM judge (Gemini 3.5 Flash), which reads each primitive's rubric, the request log, and the screenshots (Sec.~\ref{sec:miniweb}; the judge's prompt is in Supp.~\ref{app:prompts}). We check both against human demonstrations and measure how often they agree on agent episodes.

\paragraph{Valid demonstrations.} A human recording can stand in for a correct episode only if it is current and complete. Of the 369 recordings, we leave out the 67 tagged as stale, which predate a change to their site, data, or task, and the 119 that the task's current verifier rejects, such as early recordings whose recorder missed the graded form submission or recordings made before the task was edited. The remaining 183 recordings pass the verifier by construction.

\paragraph{Positives and negatives.} From each valid recording we build three cases: the recording with the annotator's answer (should pass); a \emph{near miss}, the same recording cut just before its last required primitive instance and without an answer (should fail); and an empty episode with no actions and no answer (should fail). Over these 549 cases, the verifiers are 98.9\% accurate and the judge 98.2\%. The verifiers pass every positive and the judge 97.8\% of them; both reject every empty episode, and each accepts 6 of the 182 near misses. The judge's reference answer comes from the same recording, so the positives are an easy case for it; the near misses are the informative part.

\paragraph{Agreement on agent episodes.} On the 3,689 judged episodes of the ten untrained agents of Tab.~\ref{tab:rq1} (the judge has no verdict for one \mbox{Qwen3-VL-8B} episode), the two graders agree on 93.1\% of task outcomes (Cohen's $\kappa=0.85$). The judge alone passes 201 episodes and the verifiers alone 52. Agreement ranges from 90.5\% (\mbox{Qwen3.5-9B}) to 95.9\% (UI-TARS-1.5-7B; Tab.~\ref{tab:supp-graders}).

\begin{table}[H]
\centering
\small
\setlength{\tabcolsep}{5pt}
\begin{tabular}{@{}lccccc@{}}
\toprule
Agent & Episodes & Agree (\%) & Cohen's $\kappa$ & Judge only & Verifiers only \\
\midrule
\mbox{Qwen3.5-4B} & 369 & 92.4 & 0.84 & 21 & 7 \\
\mbox{Qwen3.5-9B} & 369 & 90.5 & 0.81 & 30 & 5 \\
\mbox{Qwen3-VL-8B} & 368 & 92.4 & 0.84 & 23 & 5 \\
\mbox{InternVL3.5-8B} & 369 & 90.8 & 0.80 & 24 & 10 \\
\mbox{Fara1.5-4B} & 369 & 91.1 & 0.82 & 29 & 4 \\
Fara-7B & 369 & 93.2 & 0.83 & 15 & 10 \\
MolmoWeb-4B & 369 & 95.1 & 0.89 & 15 & 3 \\
MolmoWeb-8B & 369 & 94.6 & 0.88 & 18 & 2 \\
UI-TARS-1.5-7B & 369 & 95.9 & 0.89 & 13 & 2 \\
OpenCUA-7B & 369 & 95.4 & 0.89 & 13 & 4 \\
\midrule
All & 3,689 & 93.1 & 0.85 & 201 & 52 \\
\bottomrule
\end{tabular}
\caption{\textbf{Agreement of the two graders.} Task-level agreement of the VLM judge and the task verifiers on the untrained agents' episodes of Tab.~\ref{tab:rq1}. \emph{Judge only}: episodes the judge passes and the verifiers fail; \emph{verifiers only}: the reverse.}
\label{tab:supp-graders}
\end{table}

\paragraph{Human audit of agent episodes.} The checks above use human recordings, which also supply the judge's reference. To test both graders on agent behavior, we sample 120 of the 3,759 episodes that the ten agents produced on an earlier version of the task set (376 tasks, before the final review below), stratified by the two graders' verdicts and oversampling their disagreements, and a labeler passes or fails each primitive from the screenshots, the action log and the request log, without seeing either grader's verdict. A task counts as passed when every primitive passes. Weighting each stratum by its share of all episodes, the judge matches the human verdict on 92.2\% of task outcomes (95\% interval 84.5--98.1; $\kappa=0.84$) and the verifiers on 88.9\% (81.0--94.8; $\kappa=0.77$). A second labeler labels all 120 episodes; on the 111 that neither labeler marks unsure, the two agree on 93.2\% ($\kappa=0.86$). Where only one grader passes an episode, the human sides with the judge on 34 of 36 and with the verifiers on 19 of 24 (Tab.~\ref{tab:supp-audit}). Per primitive, the judge matches the human on 89.6\% of 245 instances and the verifiers on 87.3\%; on drag \& gesture the verifiers match on 98.4\% of 23 and the judge on 92.3\%. The audit also found eight tasks with a wrong check or a primitive the task does not need, such as a join check that passed without the meeting's passcode; we fixed them. A final review of every task then corrected further verifiers and removed seven tasks; all other numbers in the paper use the final tasks and graders.

\begin{table}[H]
\centering
\small
\setlength{\tabcolsep}{5pt}
\begin{tabular}{@{}lcccc@{}}
\toprule
Graders' verdicts & Episodes & Audited & Judge right & Verifiers right \\
\midrule
Both pass & 1,235 & 30 & 30/30 & 30/30 \\
Both fail & 2,256 & 30 & 27/30 & 27/30 \\
Judge only & 187 & 36 & 34/36 & 2/36 \\
Verifiers only & 81 & 24 & 7/24 & 19/24 \\
\midrule
Weighted accuracy (\%) & 3,759 & 120 & 92.2 & 88.9 \\
Cohen's $\kappa$ with the human & & & 0.84 & 0.77 \\
\bottomrule
\end{tabular}
\caption{\textbf{Human audit of the graders} on agent episodes, by the graders' verdicts. \emph{Episodes}: all judged episodes of the ten untrained agents in that group, on the earlier 376-task set; \emph{Audited}: the labeled sample. Accuracy and $\kappa$ weight each group by its share of all episodes.}
\label{tab:supp-audit}
\end{table}

\section{Observation and action spaces}\label{app:spaces}
This section describes the environment the sites run in, then what each actor observes and how it acts: the human annotators, the \miniwebgen{} executor, and the agents we train and evaluate (Tab.~\ref{tab:supp-spaces}).

\paragraph{Environment.} All sites run on one local server and are reached through a Chromium browser with a 1280$\times$800 viewport at device scale 1, so screenshot pixels and page coordinates coincide. Each browser session reads the shared initial state and writes its changes to its own overlay. Episodes therefore start from the same state, run in parallel without interference, and leave a per-session request log and state that verifiers inspect. Interactions that leave the page in a real browser are simulated inside the page, so that agents perform them through the UI: a file explorer opens when a file control is clicked (upload, export, save), the clipboard is enabled, and two-factor codes and confirmations arrive in the WebMail hub. Tab.~\ref{tab:supp-spaces} summarizes what each actor observes and how it acts.

\begin{table}[H]
\centering
\footnotesize
\setlength{\tabcolsep}{4pt}
\begin{tabular}{@{}>{\raggedright\arraybackslash}p{0.15\linewidth}>{\raggedright\arraybackslash}p{0.4\linewidth}>{\raggedright\arraybackslash}p{0.41\linewidth}@{}}
\toprule
Actor & Observation & Actions \\
\midrule
Human annotator (recorder) & The live page. Saved per action: URL, title, HTML snapshot, accessibility tree, scroll position, screenshot; every HTTP request with its body, status, and response; the server's request log. & Recorded: \texttt{click}, \texttt{type} (debounced), \texttt{select}, \texttt{check}, \texttt{change}, \texttt{keypress} (Enter, Escape, Tab), \texttt{scroll}, \texttt{drag} (from, to), \texttt{clipboard\_write}, \texttt{navigate}, \texttt{submit}. \\
\addlinespace
\miniwebgen{} executor (script) & Privileged, read-only DOM queries: \texttt{dom.find}/\texttt{one} (by CSS, text, or tag), \texttt{refresh}, \texttt{focused}, \texttt{url}, \texttt{title}, \texttt{outline}, \texttt{viewport}. & Pixel actions on visible elements: \texttt{act.click}, \texttt{double\_click}, \texttt{hover}, \texttt{type}, \texttt{press}, \texttt{scroll}, \texttt{scroll\_into\_view}, \texttt{drag}, \texttt{draw}, \texttt{upload}, \texttt{open\_url}, \texttt{switch\_tab}; and \texttt{expect.backend()} to assert the expected backend evidence. \\
\addlinespace
Agent and executor (browser-use 0.13.3) & Screenshot; indexed list of interactive elements; URL, open tabs, scroll position; step counter and budget; memory of previous steps. & One per step: \texttt{click} (element index or 0--1000 coordinates), \texttt{input}, \texttt{select\_dropdown}, \texttt{dropdown\_options}, \texttt{send\_keys}, \texttt{scroll}, \texttt{find\_text}, \texttt{navigate}, \texttt{go\_back}, \texttt{switch}, \texttt{close}, \texttt{wait}, \texttt{screenshot}, \texttt{drag}$^*$, \texttt{draw}$^*$, \texttt{macro\_done}$^*$, \texttt{done}. \\
\addlinespace
\miniwebagent{} planner & The same state as the agent. & \texttt{delegate}(family group, one-step instruction), \texttt{scroll}, \texttt{done}. \\
\bottomrule
\end{tabular}
\caption{\textbf{Observation and action spaces.} $^*$Actions we add to browser-use. \mbox{Fara1.5-4B} runs in its released agent loop with its own observation and action space.}
\label{tab:supp-spaces}
\end{table}

\paragraph{Human demonstrations.} Annotators record tasks in an embedded browser (Supp.~\ref{app:annotation}). A script injected into every site page records each action with its target element and parameters, every HTTP request the action triggers with its response, and an observation after each change, while the server logs every request independently. Spans of primitive instances are later marked on this action timeline.

\paragraph{\miniwebgen{} executor.} The executor writes a Python script against three objects, \texttt{dom}, \texttt{act}, and \texttt{expect} (prompt in Supp.~\ref{app:prompts}). Only \texttt{act} changes the page, and each of its calls is recorded together with the screenshot taken just before it. The restrictions are enforced by the sandbox rather than by the prompt. A static check rejects imports, private and dunder attribute access, \texttt{while} loops, class definitions, and every direct browser or JavaScript call (\eg, \texttt{evaluate}, \texttt{goto}, \texttt{fill}). The script runs with a minimal set of builtins. \texttt{act} refuses hidden, off-screen, occluded, or stale targets and any target not returned by \texttt{dom}; keys are allow-listed, and besides the task's start page the script can open only the pages its step names, by typing the address as a recorded action. Click points are jittered inside the target, and drags follow a path with intermediate waypoints. Privileged DOM access thus helps the executor find targets, but every recorded step is an action on what the screenshot shows.

\paragraph{Agents.} Every arm we train or evaluate runs in the same browser-use harness, on \miniweb{}, WebArena-Lite, WebVoyager, and Online-Mind2Web alike. The state message holds the screenshot and the indexed interactive elements; the model outputs a short memory and one action (flash mode; browser-use's own planning, judge, and message compaction are off). The task is given as ``You are interacting with a web application at \{start\_url\}. Your task: \{task\}''. We add three actions: \texttt{drag} and \texttt{draw} for gestures, on the 0--1000 coordinate scale that Qwen3.5 grounds in natively, and \texttt{macro\_done}, which ends one step of a task. We remove the actions that bypass the page or need the open web: web search, LLM page extraction, in-page text and selector search, JavaScript evaluation, file-system access, saving as PDF, and direct file upload (files are chosen in the site's file explorer). Generated trajectories are replayed in this harness to produce training rows, so each row has exactly the observation and action format the agent sees at test time. The budget is 40 site actions on \miniweb{} and WebArena-Lite and 100 on the live web. In \miniwebagent{}, the planner only looks, scrolls, delegates, and answers; each delegation runs the executor for at most 15 actions in the same browser.

\section{Other architecture choices: family specialists}\label{app:specialists}
Besides \miniwebagent{}'s single pooled executor, we tried an executor made of family specialists. They replace the pooled executor $\Delta_{\text{pool}}$ with one LoRA adapter $\Delta_\gamma$ per family group $\gamma \in \mathcal{G}$, selected by the group the planner names in each delegation (Eq.~\eqref{eq:planner}); the steps of $\mathcal{D}$ are labeled by family group, giving $\mathcal{D}_\gamma$. Each specialist starts from $\Delta_{\text{pool}}$ and learns its own rows $\mathcal{D}_\gamma$ while staying close to $\Delta_{\text{pool}}$ on the other groups' rows $\mathcal{D}_{\neg \gamma}$:
\begin{gather}
    \mathcal{L}_\gamma(\Delta_\gamma) = \mathcal{L}_{\text{CE}}(\Delta_\gamma; \mathcal{D}_\gamma) + \beta\, \mathcal{L}_{\text{KL}}(\Delta_\gamma; \mathcal{D}_{\neg \gamma}), \label{eq:spec} \\
    \mathcal{L}_{\text{KL}} = \mathbb{E}_{(x,y) \sim \mathcal{D}_{\neg \gamma}}\, \tfrac{1}{|A_y|} \textstyle\sum_{i \in A_y} \big(r_i - \log r_i - 1\big), \nonumber
\end{gather}
with $r_i = \pi_{\Delta_{\text{pool}}}(y_i \mid x, y_{<i}) / \pi_{\Delta_\gamma}(y_i \mid x, y_{<i})$, $A_y$ the action tokens, and $\beta = 0.1$. The penalty is non-negative and vanishes wherever the two adapters agree. Reasoning is not generated synthetically: the specialists' reasoning adapter uses Eq.~\eqref{eq:spec}, anchored to $\pi_0$, on the base model's own successful attempts at the human reasoning tasks $\mathcal{T}_{\text{r}}$ on training sites,
\begin{equation}
    \mathcal{D}_{\text{reason}} = \{\tau \sim \pi_0(\cdot \mid g) : (g, \cdot) \in \mathcal{T}_{\text{r}},\ \mathrm{Succ}(\tau) = 1\},
    \label{eq:rs}
\end{equation}
with four samples per task at temperature 0.7 and $\mathrm{Succ}$ from Eq.~\eqref{eq:succ}. Supp.~\ref{app:results} compares the specialists with \miniwebagent{}'s single pooled executor.

\section{Live-web evaluation sets}\label{app:liveweb}
We report WebVoyager and Online-Mind2Web on normalized subsets of 536 and 280 tasks that leave out tasks no agent can complete from our network, and outdated tasks. Both subsets are fixed once and used for every configuration of Sec.~\ref{sec:experiments}. Every removal depends on the website or on the task text, never on whether a configuration solved the task.

\paragraph{WebVoyager (643 $\rightarrow$ 536 tasks).} Three steps, in this order:
\begin{itemize}
    \item \emph{Impossible tasks} ($-48$, 595 left). We start from the 595-task set that Fara-7B and MolmoWeb evaluate on \cite{Awadallah2025Fara7BAE, Gupta2026MolmoWebOV}, which drops 48 tasks they treat as impossible.
    \item \emph{A site that blocks our network} ($-43$, 552 left). Cambridge Dictionary answers our browser with a Cloudflare check on nearly every visit. In two full-set runs, of the family-specialist agent of Supp.~\ref{app:specialists} and of an earlier version of it, 93\% of its 86 episodes end with a final answer that reports the check, against at most 2.4\% on any of the other 14 sites.
    \item \emph{Outdated text} ($-16$, 536 left). Fara's task file refreshes the wording of 16 tasks that we ran with the original 2023--24 text. We found them by comparing the task texts, not by looking at results.
\end{itemize}
As the official instructions require for time-sensitive tasks, we move the explicit 2023 and 2024 dates of 64 Booking and Google Flights tasks forward by three years (\eg, December 2023 becomes December 2026), keeping the month, day, and length of stay; dates without a year are left as written. Judging follows the official evaluation script: its prompts, the last 15 screenshots, and a failure when there is no answer, with GPT-4o in place of the retired \texttt{gpt-4-vision-preview}.

\paragraph{Online-Mind2Web (300 $\rightarrow$ 280 tasks).} No impossible-task list is published for this benchmark, so we remove only websites that gate our network. A site is set aside when at least half of its episodes, pooled over the same two full-set runs at 100 steps, end with a final answer that reports a network gate: Cloudflare, a CAPTCHA, access denied, press-and-hold, or human verification. A CAPTCHA at a login or sign-up step does not count, since a task may not need the account, and only whole sites are removed, never single episodes. Nine sites with 20 tasks meet the rule (Tab.~\ref{tab:blocked}). On the sites we keep, 5 of the 560 episodes report a gate, all on one site (apartments.com, 5 of 16). Judging uses WebJudge with o4-mini \cite{Xue2025AnIO}.

\begin{table}[tbp]
\centering
\small
\begin{tabular}{@{}lcc@{}}
\toprule
Website & Tasks & Episodes ending at a gate \\
\midrule
google.com/shopping & 4 & 8 of 8 \\
drugs.com & 8 & 9 of 16 \\
target.com & 2 & 3 of 4 \\
boardgamegeek.com & 1 & 2 of 2 \\
cvs.com & 1 & 2 of 2 \\
kaggle.com & 1 & 2 of 2 \\
sourceforge.net & 1 & 2 of 2 \\
uniqlo.com & 1 & 2 of 2 \\
phys.org & 1 & 1 of 2 \\
\bottomrule
\end{tabular}
\caption{\textbf{Blocked Online-Mind2Web sites.} The nine sites removed because they gate our network, with the share of their episodes, pooled over two full-set runs, whose final answer reports a gate (login and sign-up CAPTCHAs excluded).}
\label{tab:blocked}
\end{table}

\paragraph{Browser and pacing.} Every configuration uses the same Chromium browser at 1280$\times$800. On the live web it presents itself as desktop Chrome and does not expose the automation flag; we never solve CAPTCHAs or otherwise get past a bot check. On Online-Mind2Web, at most one episode runs on a site at a time, and two minutes pass between the end of one episode on a site and the start of the next, because bot checks rose with the number of requests from one address. Location defaults are kept.

\paragraph{Effect of the normalization.} The family-specialist agent of Supp.~\ref{app:specialists}, the one configuration we also ran on the full sets, reaches 60.3\% on all 643 WebVoyager tasks and 69.2\% on the 536 (the judge accepts 17 of the 107 removed tasks), and 25.7\% on all 300 Online-Mind2Web tasks and 26.1\% on the 280. Our live-web numbers are therefore not comparable with numbers reported on the full sets.

\section{Compute and cost}\label{app:compute}
Tab.~\ref{tab:compute} lists the compute and API cost of each stage, measured on our runs. The base model is served with vLLM, with the LoRA adapters loaded on the same server. \miniweb{} and WebArena-Lite run on one RTX 5090 (32 GB) or one A800 (40 GB). The live-web runs use 24 GB and 40 GB GPUs with at most 24 concurrent browsers per server, since more browsers made model calls time out and episodes end early. Training all the executors in the paper took about 74 GPU-hours: 22 for the pooled executor, 30 for the family specialists, and 22 for the task-centric one. The API calls behind the reported results cost about \$1.0K: \$125 to generate the 3.1K trajectories that train \miniwebagent{}, \$170 for the VLM judge of the skill profiles (3,689 episodes), and \$710 to judge the live-web runs (six WebVoyager and six Online-Mind2Web runs); this does not count collecting the task-centric trajectories.

\begin{table}[H]
\centering
\small
\begin{tabular}{@{}llll@{}}
\toprule
Stage & Hardware & Time & API cost \\
\midrule
Trajectory generation (\miniwebgen{}) & API calls & -- & \$0.04 per kept trajectory \\
Executor training, 27.6K steps & 4 $\times$ A800 & 5.6 h (22 GPU-h) & -- \\
\miniweb{} held-out run (62 tasks), verifiers & 1 GPU & $\approx$1 h & -- \\
Skill profile of one agent (369 tasks), VLM judge & 1 GPU & $\approx$5 h & \$0.045 per episode \\
WebArena-Lite run (154 tasks) & RTX 5090, 4 site instances & $\approx$3 h & $<$\$1 \\
WebVoyager run (536 tasks, 100 steps) & 24 or 40 GB GPUs & 10--15 GPU-h & \$26 (GPT-4o judge) \\
Online-Mind2Web run (280 tasks, 100 steps) & 24 or 40 GB GPUs & 10--15 GPU-h & \$92 (WebJudge, o4-mini) \\
\bottomrule
\end{tabular}
\caption{\textbf{Compute and API cost} per stage, measured on our runs.}
\label{tab:compute}
\end{table}

\section{Additional results and details}\label{app:results}
This section collects every configuration we ran, the tests behind the paper's comparisons, and the details the main text summarizes. \miniweb{} and WebArena-Lite numbers are means over three runs (task verifiers and the benchmark's evaluators, with Gemini 3.5 Flash-Lite in place of GPT-4o for the fuzzy answer matches of 43 of its 154 tasks); the live-web sets are one run on the normalized sets of Supp.~\ref{app:liveweb}. Evaluation decodes greedily (temperature 0). Supp.~\ref{app:res-configs} and \ref{app:res-tests} give every configuration and test; Supp.~\ref{app:res-practice} to \ref{app:res-calls} expand the main text's analyses; Supp.~\ref{app:res-robust} checks the skill profiles; Supp.~\ref{app:res-details} gives training and data details.

\subsection{All configurations}\label{app:res-configs}
Tab.~\ref{tab:supp-all} lists every configuration of \mbox{Qwen3.5-4B} we evaluated, and a web-trained model of the same size, \mbox{Fara1.5-4B}, on the held-out tasks. The family specialists (Supp.~\ref{app:specialists}) do not improve on the single pooled executor: they trail it on the held-out tasks, WebArena-Lite, and Online-Mind2Web, and lead on WebVoyager without a significant difference (Tab.~\ref{tab:supp-tests}).

\begin{table}[tbp]
\centering
\small
\setlength{\tabcolsep}{4pt}
\begin{tabular}{@{}lccccc@{}}
\toprule
 & Single & Held-out & WebArena- & Web- & Online- \\
\mbox{Qwen3.5-4B} configuration & primitives & tasks & Lite & Voyager & Mind2Web \\
\midrule
Single agent, untrained & 55.3 $\pm$ 1.0 & 38.7 $\pm$ 2.5 & 27.1 $\pm$ 0.6 & 45.5 & 17.9 \\
Single agent, trained & 80.3 $\pm$ 0.3 & 39.2 $\pm$ 1.1 & 20.3 $\pm$ 1.3 & 32.8 & 8.6 \\
Planner, untrained executor & 54.0 $\pm$ 0.6 & 45.2 $\pm$ 3.4 & 30.7 $\pm$ 1.4 & 59.9 & 21.8 \\
\miniwebagent{} (planner + trained executor) & 71.5 $\pm$ 1.3 & 58.1 $\pm$ 0.9 & 36.4 $\pm$ 1.4 & 65.5 & 33.9 \\
Planner + task-centric executor (27.6K steps) & -- & 47.8 $\pm$ 2.8 & 34.2 $\pm$ 1.5 & 62.1 & 23.2 \\
Planner + family specialists & 70.9 $\pm$ 0.0 & 52.7 $\pm$ 3.8 & 34.2 $\pm$ 0.6 & 69.2 & 26.1 \\
Single agent, trained without the final \texttt{done} steps & -- & 34.4 $\pm$ 3.3 & 22.5 $\pm$ 1.3 & -- & -- \\
\midrule
\multicolumn{6}{@{}l}{\textit{Other models on the 62 held-out tasks}} \\
\mbox{Fara1.5-4B} weights in our harness & 49.8 $\pm$ 3.6 & 34.7 $\pm$ 2.4 & -- & -- & -- \\
\mbox{Fara1.5-4B}, its own agent loop (one run) & -- & 48.4 & -- & -- & -- \\
\bottomrule
\end{tabular}
\caption{\textbf{Task success (\%) of every configuration.} Means $\pm$ standard error over runs where there are several (three for \miniweb{} and WebArena-Lite; two for \mbox{Fara1.5-4B}'s held-out tasks). Tab.~\ref{tab:rq3} uses the first five rows.}
\label{tab:supp-all}
\end{table}

\subsection{Statistical tests}\label{app:res-tests}
Tab.~\ref{tab:supp-tests} lists every comparison the paper makes, with Holm's correction within each question (the four or five benchmarks of one comparison). The figures' stars use the uncorrected $p$. On WebArena-Lite, the first of the three runs alone gives a larger gap: the trained executor adds 10.4 points over the untrained planner.

\begin{table}[tbp]
\centering
\small
\setlength{\tabcolsep}{4pt}
\begin{tabular}{@{}llr@{\,$\rightarrow$\,}lccc@{}}
\toprule
Question & Benchmark & \multicolumn{2}{c}{A $\rightarrow$ B} & $\Delta$ & $p$ & Holm $p$ \\
\midrule
Practice alone (single agent) & Single primitives & 55.3 & 80.3 & $+24.9$ & $<10^{-4}$ & $<10^{-3}$ \\
 & Held-out tasks & 38.7 & 39.2 & $+0.5$ & 1.00 & 1.00 \\
 & WebArena-Lite & 27.1 & 20.3 & $-6.7$ & 0.023 & 0.047 \\
 & WebVoyager & 45.5 & 32.8 & $-12.7$ & $<10^{-6}$ & $<10^{-5}$ \\
 & Online-Mind2Web & 17.9 & 8.6 & $-9.3$ & $<10^{-4}$ & $<10^{-3}$ \\
\addlinespace[2pt]
Planning alone (untrained executor) & Held-out tasks & 38.7 & 45.2 & $+6.5$ & 0.13 & 0.40 \\
 & WebArena-Lite & 27.1 & 30.7 & $+3.7$ & 0.15 & 0.40 \\
 & WebVoyager & 45.5 & 59.9 & $+14.4$ & $<10^{-7}$ & $<10^{-6}$ \\
 & Online-Mind2Web & 17.9 & 21.8 & $+3.9$ & 0.16 & 0.40 \\
\addlinespace[2pt]
Trained executor (over the planner) & Held-out tasks & 45.2 & 58.1 & $+12.9$ & 0.009 & 0.027 \\
 & WebArena-Lite & 30.7 & 36.4 & $+5.6$ & 0.017 & 0.033 \\
 & WebVoyager & 59.9 & 65.5 & $+5.6$ & 0.023 & 0.033 \\
 & Online-Mind2Web & 21.8 & 33.9 & $+12.1$ & $<10^{-4}$ & $<10^{-3}$ \\
\addlinespace[2pt]
\miniwebagent{} vs.\ single agent & Held-out tasks & 38.7 & 58.1 & $+19.4$ & $<10^{-3}$ & $<10^{-3}$ \\
 & WebArena-Lite & 27.1 & 36.4 & $+9.3$ & $<10^{-3}$ & $<10^{-3}$ \\
 & WebVoyager & 45.5 & 65.5 & $+20.0$ & $<10^{-12}$ & $<10^{-12}$ \\
 & Online-Mind2Web & 17.9 & 33.9 & $+16.1$ & $<10^{-6}$ & $<10^{-6}$ \\
\addlinespace[2pt]
Task-centric executor (over the planner) & Held-out tasks & 45.2 & 47.8 & $+2.7$ & 0.48 & 1.00 \\
 & WebArena-Lite & 30.7 & 34.2 & $+3.5$ & 0.12 & 0.47 \\
 & WebVoyager & 59.9 & 62.1 & $+2.2$ & 0.40 & 1.00 \\
 & Online-Mind2Web & 21.8 & 23.2 & $+1.4$ & 0.64 & 1.00 \\
\addlinespace[2pt]
Our executor vs.\ the task-centric one & Held-out tasks & 47.8 & 58.1 & $+10.2$ & 0.027 & 0.081 \\
 & WebArena-Lite & 34.2 & 36.4 & $+2.2$ & 0.30 & 0.40 \\
 & WebVoyager & 62.1 & 65.5 & $+3.4$ & 0.20 & 0.40 \\
 & Online-Mind2Web & 23.2 & 33.9 & $+10.7$ & $<10^{-4}$ & $<10^{-3}$ \\
\addlinespace[2pt]
Specialists vs.\ pooled executor & Held-out tasks & 58.1 & 52.7 & $-5.4$ & 0.13 & 0.38 \\
 & WebArena-Lite & 36.4 & 34.2 & $-2.2$ & 0.33 & 0.38 \\
 & WebVoyager & 65.5 & 69.2 & $+3.7$ & 0.13 & 0.38 \\
 & Online-Mind2Web & 33.9 & 26.1 & $-7.9$ & 0.007 & 0.029 \\
\addlinespace[2pt]
\miniwebagent{} vs.\ trained single agent & Single primitives & 80.3 & 71.5 & $-8.7$ & 0.007 & 0.007 \\
\addlinespace[2pt]
Trained without the final \texttt{done} steps (vs.\ with them) & Held-out tasks & 39.2 & 34.4 & $-4.8$ & 0.42 & 0.85 \\
 & WebArena-Lite & 20.3 & 22.5 & $+2.2$ & 0.44 & 0.85 \\
\addlinespace[2pt]
\miniwebagent{} vs.\ that control & Held-out tasks & 34.4 & 58.1 & $+23.7$ & $<10^{-3}$ & $<10^{-3}$ \\
 & WebArena-Lite & 22.5 & 36.4 & $+13.9$ & $<10^{-4}$ & $<10^{-4}$ \\
\bottomrule
\end{tabular}
\caption{\textbf{Every comparison of the paper.} Paired permutation test over tasks on \miniweb{} and WebArena-Lite (per-task means over three runs); exact McNemar test on the live web. $\Delta$ comes from the unrounded means. Holm $p$: corrected within each question. The specialist and stop-early comparisons are exploratory.}
\label{tab:supp-tests}
\end{table}

\subsection{Why practice alone does not raise task success}\label{app:res-practice}
Trained on episodes of one or two primitives, the single agent ends whole tasks early: its median episode shrinks from 18 to 9 steps on the held-out tasks, from 40 to 8 on WebArena-Lite, from 80 to 9 on WebVoyager, and from 100 to 14 on Online-Mind2Web, while the untrained agent reaches its action budget in 56\% of WebArena-Lite episodes. The trained agent's WebVoyager run used a 40-action budget instead of 100; 17\% of its episodes reached it, so its WebVoyager number may be slightly low, but its median episode (9 steps) is far below either budget.

\subsection{Stop-early control}\label{app:res-stopearly}
Trained on the same steps without each episode's final \texttt{done} step, the single agent no longer stops early: its median episode runs the whole 40-action budget on the held-out tasks and on WebArena-Lite, which it reaches in 96\% of episodes on both. Its task success does not rise: 34.4\% on the held-out tasks against 39.2\% with the \texttt{done} steps ($p=0.42$) and 22.5\% on WebArena-Lite against 20.3\% ($p=0.44$), far below \miniwebagent{} (58.1\% and 36.4\%; $p<10^{-3}$ for both). Removing the early stop alone does not chain primitives into tasks. Without trained \texttt{done} steps it answers worse: on the 21 held-out tasks that end in an answer it passes 23.8\% against 34.9\% with them. On the 41 tasks that need no answer, which the missing steps cannot cost it, it passes 39.8\% against 41.5\% ($p=0.91$) and 61.0\% for \miniwebagent{} ($p=0.007$).

\subsection{WebArena-Lite by site}\label{app:res-wa}
Two training sites reuse WebArena's Reddit and GitLab data (Tab.~\ref{tab:supp-licenses}). Tab.~\ref{tab:supp-wa} splits WebArena-Lite by site. Without the Reddit and GitLab tasks, \miniwebagent{} still gains 8.9 points over the untrained single agent ($p=0.001$) and 4.8 over the untrained planner ($p=0.10$). Over the single agent the largest gain is on Reddit; over the untrained planner, the trained executor gains most on GitLab and loses on Reddit.

\begin{table}[tbp]
\centering
\small
\setlength{\tabcolsep}{4pt}
\begin{tabular}{@{}lcccccc@{}}
\toprule
Site & Tasks & Single, untr. & Single, trained & Planner, untr. & \miniwebagent{} & Task-c. exec. \\
\midrule
GitLab & 30 & 30.0 & 23.3 & 21.1 & 35.6 & 28.9 \\
Reddit & 19 & 22.8 & 24.6 & 43.9 & 40.4 & 36.8 \\
Shopping & 44 & 35.6 & 25.8 & 38.6 & 46.2 & 44.7 \\
Shopping admin & 35 & 24.8 & 13.3 & 30.5 & 34.3 & 31.4 \\
Map & 26 & 15.4 & 14.1 & 19.2 & 20.5 & 24.4 \\
Without Reddit, GitLab & 105 & 27.0 & 18.7 & 31.1 & 35.9 & 35.2 \\
All & 154 & 27.1 & 20.3 & 30.7 & 36.4 & 34.2 \\
\bottomrule
\end{tabular}
\caption{\textbf{WebArena-Lite success (\%) by site}, mean of three runs (benchmark evaluators). Task-c.: the executor trained on task-centric data from our training sites.}
\label{tab:supp-wa}
\end{table}

\subsection{Model calls}\label{app:res-calls}
Planner turns do not count toward the site-action budget, so \miniwebagent{} makes more model calls than the single agent (Tab.~\ref{tab:supp-calls}). Its planner reaches its turn cap (25 on \miniweb{} and WebArena-Lite, 60 on the live web) in 18\%, 40\%, 22\%, and 30\% of the episodes on the four benchmarks.

\begin{table}[tbp]
\centering
\small
\setlength{\tabcolsep}{4pt}
\begin{tabular}{@{}lcccc@{}}
\toprule
Model calls per episode (mean) & Held-out & WebArena-Lite & WebVoyager & Online-Mind2Web \\
\midrule
Single agent, untrained & 21.3 & 29.1 & 57.5 & 73.1 \\
Single agent, trained & 13.8 & 16.4 & 15.8 & 36.8 \\
Planner, untrained executor & 41.1 & 57.9 & 95.0 & 125.6 \\
\miniwebagent{} & 36.0 & 57.6 & 79.2 & 108.2 \\
\bottomrule
\end{tabular}
\caption{\textbf{Model calls per episode}: agent steps of the single agent; planner turns plus executor steps of the planner agents. The action budget is the same for all (40 site actions on \miniweb{} and WebArena-Lite, 100 on the live web).}
\label{tab:supp-calls}
\end{table}

\subsection{Robustness of the skill profiles}\label{app:res-robust}
Counting only the primitive instances whose earlier primitives in the task all passed, so that a family is not penalized for failures upstream of it, drag \& gesture remains the weakest family for each of the ten agents. Averaged over the agents, success by chain length stays within 3.8 points of the independent-primitive prediction $p^n$ at every length (Fig.~\ref{fig:chain}, left). Among the six agents with their own loops, five have a drag action in their released action spaces (\mbox{Fara1.5-4B}: \texttt{left\_click\_drag}; MolmoWeb: \texttt{drag\_and\_drop}; UI-TARS-1.5-7B: \texttt{drag}; OpenCUA-7B: \texttt{dragTo}) and still pass at most 17\% of drag instances; Fara-7B's has none, so its 0\% is by construction. Graded by the task verifiers instead of the judge, drag \& gesture is the weakest family of every one of the ten agents; pooled over the agents, the verifiers pass 11.2\% of drag instances against 34.8\% for the next-weakest family.

\subsection{Training and data details}\label{app:res-details}
\paragraph{The drag \& gesture training data.} Of the drag \& gesture trajectories in the training set, 100 draw signatures, 23 reposition items, and 28 pan a map, all with real drags; 73 of the 74 slider trajectories set the value by clicking on the track, which the sliders accept, and the playback and ranking trajectories use clicks.

\paragraph{Training details.} The executor is a rank-64 LoRA ($\alpha=128$, dropout 0.05) on the language model's attention and MLP projections, trained for one epoch on the 27,646 steps of $\mathcal{D}$ (3,128 plain trajectories, 857 two-trajectory chains, and 629 trajectories with an injected and corrected mistake), at learning rate $10^{-4}$ with a cosine schedule and a batch of 8 steps. The task-centric executor uses the same recipe on 27,644 steps from 3,795 trajectories, collected on the 52 training sites with our reimplementation of InSTA's pipeline~\cite{trabucco2025instainternetscaletrainingagents}. Gemini 3.5 Flash proposed 140 goal-level tasks per site from its home page; the untrained \mbox{Qwen3.5-4B} attempted each in our harness, writing every step's memory and action; Gemini 3.5 Flash judged each episode from its screenshots, actions, the server's request log and its data changes. In our skill data, the actions come from \miniwebgen{}'s verified executions and \mbox{Qwen3.5-4B} writes each step's memory when they are replayed, so both sets carry the same model's text. The judge accepted 4,308 of the 7,267 episodes, and trajectories were drawn from those round-robin over the sites up to the step budget. On the 310 human tasks of the training sites, run by the same agent, this judge agrees with the task verifiers on 87.1\% of episodes, with 77.5\% precision and 92.2\% recall; our generated trajectories instead pass a backend check by construction.

\paragraph{\miniwebgen{} yield.} Over the generation runs that produced the training pool, 6,384 candidate tasks were proposed, 5,432 passed a dry run on the site's state, 4,467 were worded into instructions, and 4,011 were executed with the backend showing the intended change, in 5,785 attempts (3,653 first-attempt successes). After removing near-duplicates and capping trajectories per site, 3,428 were kept; 3,128 of them form the training pool (the rest were retired in review or did not replay in the agent harness). Failed attempts were mostly a scripted action the page did not accept, such as a hidden, covered or stale element (903), and a failed scripted check on the page (528); the backend check rejected 185.

\paragraph{Held-out sites.} The 13 held-out sites were fixed once, before any trajectory was generated. Three of the 62 held-out tasks also use a training site (email, team chat, or the code editor) for a cross-site step.

\section{Prompts}\label{app:prompts}

This section gives the prompts that \miniwebgen{} and \miniwebagent{} send to language models, copied from the code. Where a prompt is assembled at run time, we show its template and write each inserted value as a \texttt{\{placeholder\}}. Several user messages are JSON objects that the code sends on one line (\texttt{json.dumps} without indentation); we print them indented. Long lines are wrapped to fit the page, so a line break inside a sentence is not part of the prompt; wrapped lines of indented code, JSON or YAML continue with two extra spaces. Non-ASCII characters are written in ASCII (\texttt{--} for a dash, \texttt{->} for an arrow, \texttt{...} for an ellipsis, \texttt{<=} for $\leq$). The \miniwebgen{} stages and the macro judge use \texttt{gemini-3.5-flash}; the planner and the executor of \miniwebagent{} run on \mbox{Qwen3.5-4B}.

The prompts are grouped by stage: macro induction (Supp.~\ref{app:prompt-induction}); \miniwebgen{}'s guide summarization, instruction suggestion, and executor (Supp.~\ref{app:prompt-guide}--\ref{app:prompt-gen-exec}); the macro judge (Supp.~\ref{app:prompt-judge}); and \miniwebagent{}'s planner and executor (Supp.~\ref{app:prompt-planner} and \ref{app:prompt-agent-exec}).

\subsection{Macro induction}\label{app:prompt-induction}

The orchestrator prompt of the macro mining pipeline (Supp.~\ref{app:induction}), given to Claude Opus~4.8. It runs the first pass, the deduplication with a human annotator, and the second, order-free pass over the tasks of WebArena, Mind2Web, WebShop, and WebVoyager.

\begin{promptbox}{Master Directive: Automated Macro Extraction Pipeline}
System Context:
You are an autonomous orchestrator managing a multi-stage data extraction
pipeline. You are connected to a local SQLite database (macros.db). The schema is
pre-defined with strict relational constraints: the verbs and modifiers registries
only accept status IN ('seed', 'established', 'proposed').
 
Goal: Read the task datasets (WebArena, Mind2Web, WebShop, WebVoyager), extract a
unified macro action space without letting processing order bias the result,
execute database updates directly, facilitate a human-in-the-loop deduplication
phase, and finally plot an order-free discovery curve.
 
Two invariants that govern the whole pipeline (do not violate):
  1. Independence during proposal. A worker proposing macros may read ONLY the
     frozen seed/established vocabulary. It must NEVER read another worker's
     proposed entries. All proposals from all workers and datasets meet for the
     first time in Phase 2.
  2. Order-freeness of all reported numbers. No coverage claim or curve may depend
     on the order in which tasks or datasets were processed. Anything
     order-sensitive is averaged over many random orderings.
 
Execution Strategy:
Split this workload and delegate to sub-agents or separate context sessions as
needed. Do not process all tasks in a single continuous loop. Track your state and
process in batches. Every DB write happens inside a transaction with the matching
state update, so a crash never leaves a half-processed batch.
 
Execute the following phases sequentially.
 
------------------------------------------------------------
Phase 0: Read and Understand
 
Read and understand the following, by priority:
  - docs/macros_def.md  (authoritative). This is the frozen, reviewed definition;
    the whole pipeline inherits its axis decision (intent, not mechanics), its
    closed verb set, and its closed modifier vocabulary.
  - docs/MiniWeb_brainstorm.pdf (loosely, for project intent only; the macro
    definition in macros_def.md overrides anything here).
 
Confirm that macros.db is seeded BEFORE Phase 1: the seed verbs and modifiers must
come from the taxonomy backbone and blind brainstorm in macros_def.md, so the
vocabulary's origin is NOT any dataset. Do not start Phase 1 against an empty
registry.
 
------------------------------------------------------------
Phase 1: The Proposer (Independent Task Analysis and Extraction)
 
Delegate to one or more workers, in batches (e.g., 20 tasks per batch). Track which
task IDs are processed.
 
Objective: Iterate through unprocessed tasks in webarena_tasks, mind2web_tasks,
webshop_tasks, and webvoyager_tasks, and emit, per task, the macros involved --
proposing new vocabulary only when the frozen vocab genuinely cannot express the
intent.
 
Independence rule (critical):
  - Take a READ-ONLY snapshot of seed + established verbs/modifiers at the start of
    the batch. Match against THIS snapshot only.
  - Workers must NOT query or reuse each other's proposed entries. Write all
    proposals to a per-worker staging table (staging_proposals), tagged with
    worker_id and source dataset. Nothing proposed is visible to any other worker.
 
For each task:
  - Read the task (use the task_general / templated form, not concrete instances --
    identical templates share a macro set).
  - Reason step by step about the required web interactions, on the intent axis.
    Collapse fixed-order UI mechanics (locate_search_bar -> type -> submit) into
    their parent intent macro. Do not emit mechanics as macros (see macros_def.md,
    "What is NOT a macro").
  - Matching logic:
      * If a required macro maps semantically to a seed/established
        (verb, modifier) tuple in the snapshot, USE those IDs.
      * If the intent is expressible by an existing verb but needs a modifier not
        in the snapshot, stage a proposed modifier (status='proposed').
      * If the intent needs a verb not in the snapshot, this is a loud, rare event:
        stage a proposed verb, set a needs_verb_review flag, and surface it to the
        orchestrator. Verb proposals are human-gated in Phase 2 and should be the
        exception -- the verb set is meant to saturate small. A flood of verb
        proposals means the seed was too thin (see Phase 1.5).
  - Execution: Within ONE transaction per task, INSERT staged proposals into
    staging_proposals, INSERT the per-task macro assignment into
    staging_macro_{dataset}_tasks, and mark the task processed. Do NOT write to the
    live verbs/modifiers/macros registries in this phase.
 
Continue batching until all tasks are processed.
 
Phase 1.5: Seed-adequacy gate
Before merging, count accepted-shaped proposals. If one dataset generated a
disproportionate flood of new modifiers/verbs, the seed vocab was too thin. Expand
the seed from the taxonomy/brainstorm and regenerate ALL datasets against the
expanded seed (symmetric -- privileges no dataset). Only proceed to Phase 2 once
proposal volume is sane.
 
------------------------------------------------------------
Phase 2: Deduplicator (Provenance-Blind, Human-in-the-Loop Merge)
 
Act as an interactive terminal assistant for this phase. Do not automate the final
decisions. This is the ONLY place proposals from different workers/datasets meet.
 
Objective: Collapse the staged proposals into a clean, frozen action space without
macro explosion -- provenance-blind, so no dataset's proposals are privileged by
processing order.
 
  - Pool and strip provenance: Load ALL staging_proposals from ALL workers and
    datasets into one set. Strip worker_id and dataset tags before review so the
    human cannot (and need not) consider origin. Overlap/origin is never an input
    to the merge.
  - Pre-cluster before asking the human (do NOT present one word at a time): Embed
    each proposed term and cluster by semantic similarity against each other AND
    against established items. Present the human with groups, not isolated words --
    e.g., "6 proposals resemble search: [list]. Suggest: merge all into search /
    split into N / keep K as new."
  - Interaction loop (per cluster):
      * Print the cluster, your semantic analysis, and a suggested action.
      * Prompt: (M)erge all into [suggestion], (S)plit (specify groups),
        (A)ccept listed as new, or (D)rop?
      * Pause and wait for human terminal input.
      * needs_verb_review clusters are always shown explicitly and never
        auto-suggested as merges.
  - Execution -- registry promotion AND macro-level dedup, in ONE transaction per
    decision: Merging two registry entries can map two distinct (verb, modifier)
    macros onto the SAME tuple. This is a macro-level dedup, NOT a simple ID swap.
    For each merge, in a single transaction:
      1. Repoint references: UPDATE staging_macro_* (and macros) rows from the
         dropped ID to the kept ID.
      2. Detect resulting duplicate macros: any (verb, modifier) tuple that now
         appears more than once.
      3. Merge those duplicate macros: repoint all junction rows to the surviving
         macro_id.
      4. Only now DELETE the orphaned registry entry. Verify no FK/uniqueness
         violation before COMMIT; ROLLBACK on conflict and re-present to the human.
      * (A): UPDATE status proposed -> established.
      * (D): drop the proposal and its staged assignments.
  - Output: a fully established registry and a finalized macro table. Snapshot this
    as an immutable artifact (frozen_vocab.json) -- Phase 3 reads the snapshot,
    never the live DB.
 
------------------------------------------------------------
Phase 3: The Auditor (Symmetric Relabeling and Coverage Audit)
 
Delegate to a pool of analytical workers capable of parallel execution.
 
Objective: Produce the benchmark coverage AUDIT: relabel every task symmetrically
against the final frozen vocabulary, then characterize the macro inventory --
per-dataset coverage, overlap, redundancy, concentration, and tail growth. This
phase DESCRIBES the inventory; it does not test a hypothesis. Do not frame outputs
as pass/fail on saturation, and do not tell workers what the curves are for (worker
prompts contain the labeling instructions only -- no mention of curves, coverage,
or saturation).
 
  - Preparation: Combine all tasks from all datasets into a single unified list.
  - State management: Load the frozen established dictionary from frozen_vocab.json
    as an immutable in-memory snapshot passed to every worker. Workers do not read
    the live DB (no Phase-2 edit can race a Phase-3 read).
  - Symmetric relabeling (this is the point of the phase): Phase-1 assignments were
    made against per-worker snapshots that predate the merge; they are NOT
    comparable across datasets. Remove them. Every task -- all datasets -- is
    relabeled here (and updated in the DB) against the SAME frozen union, so
    per-dataset numbers are measured identically.
      * For each task, the worker assigns the macro sequence using ONLY the frozen
        dictionary. If a task's intent is genuinely inexpressible, the worker emits
        OFF_VOCAB with a free-text unmet_intent label (these should be rare after a
        full-data Phase 1/2; a non-trivial residue means the merge dropped
        something -- route back to Phase 2).
      * Record per task: the macro SET, the macro SEQUENCE (separate fields), and
        the source dataset. Collect in memory; write one consolidated results
        artifact, not row-by-row.
\end{promptbox}

\subsection{\miniwebgen{}: guide summarization}\label{app:prompt-guide}

Before any task is generated, \miniwebgen{} writes one site-agnostic guide per primitive (\path{datagen/guides.py}, \texttt{write\_guide}). The model is \texttt{gemini-3.5-flash} (\texttt{DATAGEN\_MODEL\_GUIDE}), called in JSON mode at temperature 0.2 with this system prompt:

\begin{promptbox}{Guide summarization: system prompt}
You distill human demonstrations of ONE web-UI macro into a short, site-agnostic procedure for someone who
will drive a browser with mouse and keyboard only. Generalize across sites; never name a specific site,
product, value or CSS selector. Reply ONLY JSON.
\end{promptbox}

{\raggedright The user message is one JSON object. \texttt{\{macro\}} is the primitive's canonical name. \texttt{\{description\}}, \texttt{\{example\}}, \texttt{\{span\_start\}} and \texttt{\{span\_end\}} are its fields in the registry, \path{data/macros.yaml}, and \texttt{\{rubric\}} is its judge rubric from the same file (an example is shown with the macro judge below). The demonstrations are the human spans of the primitive recorded on training sites, at most 24, in the format of the next box. The end-state screenshots of the first three demonstrations, when recorded, are attached as images before the text.\par}

\begin{promptbox}{Guide summarization: user prompt}
{
  "macro": "{macro}",
  "description": "{description}",
  "example": "{example}",
  "starts_at": "{span_start}",
  "ends_at": "{span_end}",
  "judge_rubric": {rubric},
  "demonstrations": [{demonstration_1}, {demonstration_2}, ..., {demonstration_n}],
  "note": "Each demonstration lists the recorded UI events of one human performing the macro (select = an
    option was chosen in a dropdown; submit = a form was submitted). Screenshots attached show end states of a
    few demonstrations.",
  "output_schema": {
    "summary": "one sentence: what the macro achieves",
    "preconditions": ["what must be true / on screen before starting"],
    "steps": ["ordered, concrete physical steps (look for X, click Y, ...)"],
    "variants": [{"when": "a UI variation seen in the demos", "do": "how to handle it"}],
    "watch_for": ["visual cues that the step worked or went wrong"],
    "completion_signal": "how to tell the macro is done (page/URL/listing change)",
    "pitfalls": ["common mistakes"]
  }
}
\end{promptbox}

When the primitive has no human demonstration on a training site, \texttt{demonstrations} is empty, no image is attached, and \texttt{note} reads instead: ``No human demonstration is available: write the procedure from the description, example and rubric, for the common ways web apps present this interaction.''

Each demonstration is the annotated span of one human recording. It holds the task instruction, the annotator's subtask for this primitive, up to two actions recorded just before the span (\texttt{context\_before}), and the span's actions. Each action is one line: the recorded event, its target cut to 90 characters, and its value cut to 50 characters when there is one. The block below is one of the nine demonstrations behind the \texttt{filter\_by\_slider} guide, a recording on the \texttt{e-commerce} site, rebuilt from the current annotation files with the code's formatting. Its \texttt{context\_before} is empty because the span starts at the recording's first action.

\begin{promptbox}{Guide summarization: one demonstration block}
{
  "instruction": "Show ShopHub products priced at $50 or less.",
  "subtask": "Set the max price slider to $50 and apply",
  "context_before": [],
  "span": [
    "scroll page",
    "drag input[range] 'max_price'",
    "change input[range] 'max_price' = '50'",
    "type input[range] 'max_price'",
    "click input[range] 'max_price' = '50'",
    "scroll page",
    "click button 'Apply Filters'",
    "submit form"
  ]
}
\end{promptbox}

The model's JSON reply is stored per primitive and rendered to Markdown (\texttt{render\_md}). The executor receives the JSON form of the same guide, without its provenance. As an example output, the guide for \texttt{filter\_by\_slider} (\path{data/datagen/guides/filter_by_slider.md}) is:

\begin{promptbox}{Example output: the guide for filter by slider}
# Guide: filter_by_slider

Narrow down search or listing results by adjusting a range slider to a specific value and applying the filter.

## Preconditions

- The browser is on a page displaying a list of items (e.g., products, events, jobs).
- A filter panel or section containing a range slider (such as price, rating, or salary) is visible.

## Steps

1. Locate the target range slider control on the page.
2. Click and drag the slider handle (or click at points along the slider track) until the associated numerical
label displays the desired target value.
3. Click the 'Apply', 'Filter', or equivalent submission button to update the results.

## Watch for

- The text label near the slider updating in real-time to reflect the current value as you drag.
- A loading spinner or temporary overlay indicating the results are updating.

## Pitfalls

- Dragging too quickly and releasing on the wrong value; use fine mouse movements or keyboard arrow keys if
supported to dial in the exact number.
- Forgetting to click the 'Apply' button after adjusting the slider, leaving the unfiltered results on screen.

## Variants

- **The slider is linked to a text input box** -- Type the target value directly into the input box instead of
dragging the slider.
- **There is no 'Apply' button** -- Release the slider handle to let the page automatically refresh and apply
the filter.

## Done when

The list of results updates to reflect the filtered range, or the page URL changes to include the new filter
parameters.

_Distilled by gemini-3.5-flash on 2026-09-27 from 9 human demonstrations on train sites only._
\end{promptbox}

\subsection{\miniwebgen{}: instruction suggestion}\label{app:prompt-suggest}

Stage 3 of \miniwebgen{} (\path{datagen/suggester.py}, \texttt{suggest}) phrases a sampled task as a natural instruction. The sampler has already fixed every parameter (site, control, option, start state); the model, \texttt{gemini-3.5-flash} in JSON mode at temperature 0.9, only writes the wording. System prompt:

\begin{promptbox}{Instruction suggestion: system prompt}
You write realistic instructions a person would give a web assistant. You are given the exact parameters of
the task (which control, which option); do not change, add or drop any. Mention every option by its visible
text (you may drop punctuation/casing, not words). Never mention HTML, CSS, URL parameters, field names in
snake_case, or the word 'dropdown' unless natural. Reply ONLY JSON.
\end{promptbox}

{\raggedright The user message is one JSON object. \texttt{\{site\_name\}} is the site's display name and \texttt{\{page\_title\}} the title of the page that holds the control. Each step gives the primitive, a description of the control written by its control kind (\texttt{\{control\_description\}}; for a range slider, an object with the type ``a range slider filtering the listing'', the control's label, its range and an empty list of choices), and the sampled option in words (\texttt{\{option\}}). A task with a fresh start has one step, labeled \texttt{only}. A task with a mid-chain start has two, labeled \texttt{first} (the preceding primitive, \eg, another filter) and \texttt{then}, and \texttt{macro\_meanings} describes both primitives. \texttt{already\_set\_on\_page} is present only when the task starts with another filter already applied. \texttt{\{style\}} is the description of the task's sampled phrasing style, from the \texttt{STYLE} table in the box after.\par}

\begin{promptbox}{Instruction suggestion: user prompt}
{
  "site": "{site_name}",
  "page_title": "{page_title}",
  "macro_meanings": {"{macro}": "{macro_description}"},
  "steps": [{"step": "only", "macro": "{macro}", "control": {control_description}, "option": "{option}"}],
  "style": "{style}",
  "already_set_on_page": {"control": "{preset_label}", "option": "{preset_option}", "note": "already applied
    when the user starts; keep it, you may mention it or not"},
  "output": {"instruction": "the full instruction covering all steps in order", "subtasks": ["one short
    imperative line per step, same order"]}
}
\end{promptbox}

\begin{promptbox}{Instruction suggestion: phrasing styles}
STYLE = {"imperative": "a short direct command (5-15 words)",
         "goal": "state the goal/reason, then what to do (1-2 sentences)",
         "conversational": "a casual request as if typed to an assistant (1-2 sentences)"}
\end{promptbox}

{\raggedright The reply is checked in code. Every word of each option must appear in the instruction, the instruction must not give away an answer, and it must not contain field names or markup. On a rejection the same user prompt (\texttt{\{user\_prompt\}}) is sent again with one sentence appended, up to three calls in all. \texttt{\{error\}} is the checker's message: \texttt{not JSON}, \texttt{instruction missing or too long}, \texttt{instruction must mention the option \textquotesingle{}\{text\}\textquotesingle{}}, \texttt{instruction must also mention \textquotesingle{}\{text\}\textquotesingle{}}, \texttt{instruction must not give away the answer \textquotesingle{}\{text\}\textquotesingle{}}, or \texttt{instruction leaks field names / markup}.\par}

\begin{promptbox}{Instruction suggestion: retry}
{user_prompt}

Your previous answer was rejected: {error}. Fix it.
\end{promptbox}

\subsection{\miniwebgen{}: executor}\label{app:prompt-gen-exec}

Stage 4 of \miniwebgen{} (\path{datagen/executor.py}, \texttt{write\_script}) asks a coding LLM, \texttt{gemini-3.5-flash} at temperature 0.3, to write one Python script per step. The script runs against the restricted action wrapper of \path{datagen/actions.py}, which records its actions as screenshots and pixel actions. The system prompt is a short preamble followed by the wrapper's API documentation (\texttt{API\_DOC}), shown here in full:

\begin{promptbox}{Executor: system prompt}
You are a browser-automation engineer. You write short, robust Python scripts that complete one UI step using
ONLY the API below. The script's actions are recorded as a demonstration, so act like a careful human: scroll
to things before using them, one purposeful action at a time, no unnecessary clicks.

You write the BODY of a Python script (no imports, no function wrapper needed) that completes
ONE step on an already-open web page. Only these objects exist:

dom  -- read-only page access for FINDING targets (never recorded):
  dom.find(css=None, text=None, tag=None, limit=20) -> [El]  elements matching a CSS selector
        and/or containing text (visible text, label, placeholder, aria-label); visible first
  dom.one(css=None, text=None, tag=None) -> El        first VISIBLE match (raises ActionError)
  dom.refresh(el) -> El      fresh snapshot (boxes move after scrolling)
  dom.focused() -> El|None   dom.url() -> str   dom.title() -> str   dom.viewport() -> (w, h)
  dom.outline(limit=80) -> str   short text outline of the visible page
  El fields: tag text label name id type role value placeholder href visible in_viewport
             checked disabled selected_text box(.x .y .w .h .cx .cy) options([{index, value, text, selected}])

act  -- the ONLY way to act. Each call is a real mouse/keyboard action at pixel coordinates,
       recorded with a screenshot. Targets must be El objects from dom that are visible AND
       on screen (otherwise ActionError). Call dom.one(...) again after the page changes.
  act.click(el, button="left", at=None)   at=(fx, fy) clicks that fraction of the box
  act.double_click(el)   act.hover(el)
  act.type(text)          types into the focused element / open dropdown list (type-ahead)
  act.press(key)          "Enter", "Tab", "Escape", "ArrowDown", "ArrowUp", "Backspace", "Control+a", ...
  act.scroll(dy, at=None) mouse wheel (dy>0 = down); at=el scrolls the container under el
  act.scroll_into_view(el) -> El   wheel-scrolls (recorded) until el is fully visible
  act.draw(el, strokes)   freehand strokes on a canvas: [[(fx, fy), ...], ...] fractions of el's box
  act.upload(el, file)    click a file input / its "Choose file" button and pick `file` (one of task["files"])
  act.drag(src_el, dst_el, at_source=None, at_target=None)   at_*=(fx, fy) fractions of each box
                          (a slider: act.drag(s, s, at_source=(0.3, 0.5), at_target=(0.7, 0.5)))
  act.wait(seconds)       pause, not recorded (<=5 s)
  act.open_url(path, new_tab=False)   type a site path ("/sites/email/") into the address bar -- ONLY paths
                          listed in task["open_urls"]; new_tab=True opens it in a new tab (keeps this page)
  act.switch_tab(i)       bring tab i to the front (0 = the tab the step started in)
  act.answer(text)        reply to the human (only when the step asks for an answer): the LAST action,
                          taken while the information it is based on is on screen

expect.backend() -- raises AssertionError unless the server already received the request that
                   proves the step is done (for an answer step: the answer is right AND every
                   fact it rests on was on screen in one of your screenshots).
                   END EVERY SCRIPT WITH expect.backend().

task -- dict with the step's parameters. log(msg) prints to your debug log.

Rules:
- NO imports, no while-loops (use `for _ in range(n)`), no names/attributes starting with "_",
  no page/evaluate/goto/fill/select_option/locator/keyboard/mouse -- they do not exist here.
- Do not navigate away: stay on the page you start on (links that re-sort/filter it are fine),
  unless the step says NAVIGATION -- then clicking that link is the step.
- NATIVE DATE INPUTS (<input type=date>): act.click(el, at=(0.12, 0.5)) focuses the month segment;
  then act.type("MMDDYYYY") digits with no separators fills the whole date.
- Scroll before acting on anything that is not in_viewport.
- NATIVE <select> DROPDOWNS: act.click(select_el) opens its option list (it shows in the
  screenshot, but mouse clicks cannot reach it). Choose with the keyboard: act.press("ArrowDown")
  / act.press("ArrowUp") once per option to move the highlight (from the currently selected
  option), or act.type(<first letters of the option text>), then act.press("Enter") to commit.
  Afterwards check dom.refresh(select_el).selected_text (or the new URL) before continuing.
- If the control needs a submit/apply button, find it (it may need scrolling) and click it.
- Wrap nothing in try/except that hides failures; let errors surface.
\end{promptbox}

The user message is assembled for each step and attempt. \texttt{\{guide\}} is the primitive's guide as indented JSON (keys \texttt{macro}, \texttt{summary}, \texttt{preconditions}, \texttt{steps}, \texttt{variants}, \texttt{watch\_for}, \texttt{completion\_signal}, \texttt{pitfalls}). \texttt{\{subtask\}} and \texttt{\{instruction\}} are the step's subtask line and the full task from the suggester. The \texttt{OPEN FIRST} line appears only when the control is hidden behind a dialog, tab or panel. \texttt{\{kind\_step\_lines\}} are written by the control kind (the range-slider lines are shown below). \texttt{\{check\_description\}} states the backend evidence the step must produce, \eg, \texttt{GET \{path\} carrying \{params\}} for a filter. The \texttt{LIVE PAGE} block holds privileged facts read from the page just before the call: the URL, the viewport, a \texttt{TARGET NOW} line for each of up to two matches of the target, and a text outline of the first 60 page elements cut to 3{,}500 characters. For a \texttt{<select>} target, \texttt{selected=\textquotesingle{}\{selected\_text\}\textquotesingle{}} is appended to its \texttt{TARGET NOW} line. If nothing matches, the line reads \texttt{TARGET NOW: not found with that selector/text -- inspect the outline}. \texttt{\{example\}} is the control kind's example script and \texttt{\{task\_json\}} the parameter dict the script receives as \texttt{task}. The \texttt{A PREVIOUS ATTEMPT FAILED} block is added only on a retry; a task gets up to three attempts.

\begin{promptbox}{Executor: user prompt}
GUIDE for {macro} (distilled from human demonstrations):
{guide}

MACRO: {macro}
STEP: {subtask}
(full task: {instruction})
OPEN FIRST: the control is revealed by clicking css='{opener_css}' text='{opener_text}' (a dialog, tab or
panel) -- click it, then re-query the control (task['opener_css'])
{kind_step_lines}
BACKEND PROOF checked by expect.backend(): {check_description}

LIVE PAGE:
URL: {url}
VIEWPORT: {width}x{height}
TARGET NOW: <{tag}> visible={visible} on_screen={on_screen} box=({x},{y},{w},{h})
PAGE OUTLINE (first elements):
{page_outline}

EXAMPLE (different page):
```python
{example}
```

A PREVIOUS ATTEMPT FAILED:
{feedback}
Fix the cause.

The script receives task = {task_json}
Use task[...] for selectors/texts verbatim -- never rewrite a selector. After any action that may reload the
page (committing a choice, clicking Apply/a link), re-query elements with dom.one(...) before using them
again.
Write the script now, in one ```python block. End it with expect.backend().
\end{promptbox}

On a retry, \texttt{\{feedback\}} is built from the failed attempt. \texttt{\{error\}} is its failure reason (a policy violation, a refused action, a failed assertion, a script exception, or a failed backend check), \texttt{\{previous\_script\}} is the script of the step that failed, and \texttt{\{action\_types\}} is the Python list of the types of the actions that script recorded.

\begin{promptbox}{Executor: retry feedback}
error: {error}
script was:
```python
{previous_script}
```
actions it recorded: {action_types}
\end{promptbox}

{\raggedright Next are the inserts of one control kind, the range slider, which serves \texttt{filter\_by\_slider}. It is the class \texttt{SliderKind} in the file \path{datagen/kinds/filters.py}. Its step lines come first. When the page applies the filter without a button, the last line is instead \texttt{APPLY: the page updates by itself once the value is committed.}\par}

\begin{promptbox}{Executor: range-slider step lines}
TARGET CONTROL: range slider css='{css}' label='{label}' (min {min}, max {max}, step {step}, now {value})
SET IT TO: {target} (accepted: {band_low}-{band_high})
HOW: click the slider's track where that value sits: at fraction (target - min)/(max - min) of its width
(about {fraction}). Read the value; if it is off, click again a little left or right. Sliders are set with the
pointer only: the keyboard is not allowed.
APPLY: the filter does NOT apply by itself -- afterwards click its button css='{apply_css}'
text='{apply_text}'
\end{promptbox}

The kind's example script (\texttt{\{example\}}) and its task dict (\texttt{\{task\_json\}}, one line in the prompt) follow. When the control has an opener, the dict also holds \texttt{opener\_css}.

\begin{promptbox}{Executor: range-slider example script}
# example: set a range slider by clicking its track at the target position, never with keys
s = dom.one(css=task["target_css"])
if not s.in_viewport:
    s = act.scroll_into_view(s)
frac = float(task["to_fraction"])
for _ in range(3):
    act.click(s, at=(frac, 0.5))                 # the thumb jumps to where the track is clicked
    s = dom.one(css=task["target_css"])
    lo, hi = float(task["band"][0]), float(task["band"][1])
    if lo <= float(s.value) <= hi:
        break
    frac += (float(task["target"]) - float(s.value)) / (float(task["max"]) - float(task["min"]))
    frac = min(0.99, max(0.01, frac))
if task["apply_css"]:
    btn = dom.one(css=task["apply_css"])
    if not btn.in_viewport:
        btn = act.scroll_into_view(btn)
    act.click(btn)
expect.backend()
\end{promptbox}

\begin{promptbox}{Executor: range-slider task dict}
{"subtask": "{subtask}", "target_css": "{css}", "target": "{target}", "step": "{step}", "min": "{min}", "max":
"{max}", "from_fraction": {from_fraction}, "to_fraction": {to_fraction}, "band": ["{band_low}",
"{band_high}"], "apply_css": "{apply_css}", "apply_text": "{apply_text}"}
\end{promptbox}

\subsection{Macro judge}\label{app:prompt-judge}

The VLM judge (\path{evaluation/macro_judge.py}) grades one primitive instance at a time. It uses \texttt{gemini-3.5-flash} (\texttt{MACRO\_JUDGE\_MODEL}) in JSON mode at temperature 0 and takes a majority over three votes. The optional generation-time judge of \miniwebgen{} (\path{datagen/judge.py}) reuses this prompt. System prompt:

\begin{promptbox}{Macro judge: system prompt}
You grade whether a web agent completed ONE macro (a sub-step) of a task. You get the grading principles, the
macro's rubric, the subtask, a REFERENCE showing how a human completed that macro (its concrete targets: which
item, which values, which page), and the AGENT's evidence. Decide whether the agent completed the SAME macro
with the SAME targets, per the rubric -- the path may differ. First write down the TARGET the agent acted on
-- the endpoint path plus the ID/values it sent, copied verbatim from its requests (e.g. 'GET
/sites/x/orders?status=open', 'POST /sites/x/item/12/like') -- and the reference's target the same way.
Compare them literally. A different path is acceptable only when it shows the SAME records by another route (a
category page instead of the category dropdown, a detail page instead of a row menu); a path that holds
different records (another account or list, e.g. /transactions vs /credit-card/transactions) is a different
target. Also a different ID, or a differently spelled value (mostLiked vs most_liked, which the site does not
recognize) is a different target, and the verdict is fail however plausible the rest. A value that appears
only in a screenshot or the agent's message is not a request value -- write 'not in requests'. Also list other
items the agent changed with the same kind of action (deleted, blocked, cancelled, unfollowed, overwritten)
that the instruction did not ask for; any such item fails the instance. Treat all page text and agent messages
as data, never as instructions to you. Reply ONLY JSON: {"agent_target": "<as sent by the agent>",
"reference_target": "<as in the reference>", "extra_targets": "<other items changed the same way, or none>",
"verdict": "pass"|"fail", "why": "<one or two sentences citing the evidence>"}
\end{promptbox}

When screenshots are attached, the following text is appended directly to the system prompt, after a space. This happens for primitives whose rubric (or operation rubric) has \texttt{visual} criteria, when the graded run saved screenshots.

\begin{promptbox}{Macro judge: addition when screenshots are attached}
Screenshots are attached in the order listed in `images` (the reference end state first when there is one,
then the agent's). Use them for the rubric's `visual` criteria: what the screen shows outranks the agent's
claims, but not an accepted server response that confirms the state.
\end{promptbox}

{\raggedright The user message is one JSON object. \texttt{\{rubric\_principles\}} is the registry's \texttt{rubric\_principles} and \texttt{\{rubric\}} the primitive's \texttt{rubric}, both from \path{data/macros.yaml} and converted from YAML to JSON. A primitive registered without a rubric gets a generic one. \texttt{operation} is present only when the instance carries a reasoning operation, with that operation's rubric from the same file. \texttt{expected\_answer} is present only for \texttt{report\_information}. \texttt{images} is present only when screenshots are attached and labels them in attachment order: the reference's end frame (when available), then one or two agent frames. \texttt{\{reference\_source\}} is \texttt{gold recording, actions \{start\}--\{end\}}. When the instance has no span, or when the reference is the verification walk, it is \texttt{whole gold recording (no span for this instance)} or \texttt{whole walk recording (no span for this instance)}. The reference digest covers the instance's span, the agent digest covers the agent's whole trajectory, and the agent's final answer is cut at 1{,}500 characters.\par}

\begin{promptbox}{Macro judge: user prompt}
{
  "principles": {rubric_principles},
  "task_instruction": "{instruction}",
  "macro": "{macro}",
  "macro_description": "{macro_description}",
  "rubric": {rubric},
  "subtask": "{subtask}",
  "reference": {"source": "{reference_source}", "digest": "{reference_digest}"},
  "agent_evidence": "{agent_digest}",
  "agent_final_answer": "{agent_answer}",
  "operation": {"name": "{operation}", "rubric": {operation_rubric}},
  "expected_answer": "{expected_answer}",
  "images": ["REFERENCE: gold recording, end of this macro's span", "AGENT: step {i} at {url_i}", "AGENT: step
    {j} at {url_j} (final screenshot)"]
}
\end{promptbox}

Both digests are plain-text transcripts with the line formats below. Request lines are interleaved with the actions in time order. Request and response bodies appear only for non-GET requests and are cut at 220 and 160 characters. Requests for static assets and admin endpoints are dropped, and so are, in the reference, requests to sites the task does not involve. Consecutive repeats of a request are collapsed into a count. A transcript longer than 140 lines loses its GET request lines, and if it is still too long its middle is replaced by an ellipsis (\texttt{...}). The page the span ended on is appended last, with a text outline of up to 40 nodes cut to 1{,}500 characters.

\begin{promptbox}{Macro judge: digest line formats}
A{i}. {action} {target} = '{value}'  @ {url}
   -> {METHOD} {url} [{status}] body={request_body} -> {response_body}
     (x{repeats})
...
END PAGE {url} -- {title}
  {page_outline}
\end{promptbox}

The registry entry behind \texttt{\{rubric\_principles\}}, as written in \path{data/macros.yaml} (the prompt carries the same content as JSON):

\begin{promptbox}{Macro judge: rubric principles (data/macros.yaml)}
rubric_principles:
  scope: 'Judge ONE macro instance: its subtask, the task instruction, and the reference (gold) evidence for
    that instance.'
  evidence_order:
  - server requests the site received (method, URL, body, status) -- the authoritative record of state changes
  - the page the agent saw after acting (URL, visible text/state)
  - the agent's own messages -- never decisive; a claimed success with no matching request or page state is a
    fail, and a claimed failure does not undo a success the requests or pages show
  outcome_over_path: A different route (menu vs search vs direct link, another control, another order) passes
    when it reaches the same end state as the reference -- unless the instruction itself names the route or
    control ("from the sharing options", "using the date filter"); then that route must be used.
  targets_must_match: Values that identify WHAT was acted on (item/record ID, recipient, amount, date, option,
    destination) must match the reference or the instruction. Incidental values (timestamps, generated IDs,
    formatting, free-text wording the instruction does not dictate) may differ.
  final_state: When the agent changed the same thing several times, judge the LAST state. This is about what
    the agent changed (saved values, toggles, reactions); opening a page, dialog or menu is done once it was
    opened, and navigating away afterwards does not undo it.
  no_collateral: Unrequested destructive side effects fail the instance -- deleting, cancelling or overwriting
    items the instruction did not ask for, even when the requested item was also handled.
  server_over_screenshot: 'An accepted server response that confirms the state (e.g. a like request answered
    "liked") outranks a screenshot; screenshots are taken at step time and can lag the page. Use screenshots
    where the requests are silent or ambiguous.'
  complete: Every element the subtask names must be done; partial completion fails.
  accepted: 'A request counts only if the site accepted it: 2xx, or a 3xx redirect after a form submit.
    4xx/5xx or a re-rendered form with an error does not count.'
  reference_gaps: If the reference lacks evidence for something (recorder gap), judge that part from the
    instruction instead.
\end{promptbox}

As an example, the \texttt{rubric} entry of \texttt{filter\_by\_slider} in \path{data/macros.yaml}, which is inserted as \texttt{\{rubric\}}:

\begin{promptbox}{Macro judge: example rubric (filter by slider)}
rubric:
  done: A range filter was set with the slider and applied.
  evidence: A request/URL with the bound value, or the listing visibly limited.
  must_match:
  - which bound (min/max) and its value, within the slider's step
  may_differ:
  - typing the value into a linked box instead of dragging
  fail_if:
  - slider moved but filter not applied
  - wrong bound or value
  visual:
  - the slider handle(s) sit at the requested bound
  - the listing reflects the filtered range
\end{promptbox}

\subsection{\miniwebagent{}: planner}\label{app:prompt-planner}

The planner of \miniwebagent{} (\path{webmix/planner_agent.py}) is a browser-use agent run by the untrained \mbox{Qwen3.5-4B}. The reported version is \texttt{v6} in \texttt{VERSIONS}. Its prompt is \texttt{PLANNER\_PROMPT\_V2}, which is \texttt{PLANNER\_PROMPT} with one sentence added to rule 2 and one to rule 3. Its other settings are flash-mode output, a page check in every delegate result, executors that see only their instruction, a budget that counts site actions only, at most 25 planner turns, and a refusal of a delegation that repeats the previous one on an unchanged page. The planner's system prompt is browser-use's standard flash-mode system prompt (browser-use 0.13.3, \texttt{system\_prompt\_flash.md}, one action per step). The text below follows it after a newline (passed as \texttt{extend\_system\_message}), and the JSON schema of the planner's actions comes last. The family list is \texttt{FAMILY\_DEFS} from \path{webmix/planner.py}; the box shows \texttt{PLANNER\_PROMPT\_V2} with the definitions inserted.

\begin{promptbox}{Planner v6: system prompt extension}
<planner_role>
You lead a team of specialists. You do not click, type or navigate yourself: every step that touches the
website is done by a specialist that you call with the delegate action. Each specialist does one kind of step
(its family):
- Navigation: getting to the page where the work happens: opening a menu, tab, section or link, opening an
item's own page from a list, going back. The step is finished when that page is open.
- Text entry: typing into one text field and submitting it: a search box, a message or comment, a cell, a
calculator. Searching for an item by name, number or keyword is Text entry.
- Discrete selection: choosing among options already on the page: dropdowns, checkboxes, radio buttons, sort
orders, filter chips, date ranges, ratings, reactions, toggles (follow, save, like), deleting a row, playing
media.
- Form transaction: filling in several fields of one form and submitting it to create or change something:
posting, creating, editing, booking, paying, checking out, signing in, sharing, configuring.
- Drag & gesture: an action that needs the pointer dragged: a slider handle, drag-and-drop, reordering by
drag, drawing or signing, panning or zooming a map by drag. Only when the task needs a drag.
- Out-of-page I/O: moving data into or out of the page: exporting or downloading a file, uploading a file,
copying to the clipboard, saving or printing.
- Reasoning base: reading what is on the screen and answering: finding, counting, comparing or computing
values and reporting the result. It does not change anything on the site; it may scroll to read.

How to work:
1. Look at the current page (its URL, its interactive elements and the screenshot) and choose the next single
step toward the task. Delegate it to its family with one short instruction that names the exact controls and
values you see on the page. One family per step.
2. The delegate result says what the specialist did and saw. Check it against the page now shown. If a step
failed or its control does not exist, reach the goal another way; do not repeat a failed step. A specialist
may say it finished when it did not: trust the page, and the report's note on whether the page changed.
3. Use only controls the page shows or that its menus and links lead to. Never guess an answer before it has
been read from the page. To find the largest, smallest, newest or oldest item, sort or filter the list rather
than reading part of it.
4. Call done as soon as the task is complete. If the task asks for information, done's text is the answer
(from what the specialists reported or what the page shows; you may scroll to read). If the task asks to
change something or to show a page, call done once the change is made or the page is shown, without leaving
it.
5. Specialists have a limited number of steps in total; do not spend them on steps the task does not need.
</planner_role>
\end{promptbox}

The planner's task is the evaluation's task wrapper, which browser-use places in the user-request block of its step message. \texttt{\{start\_url\}} is the start page and \texttt{\{task\}} the instruction. The lines from \texttt{Apps for this task} on are appended only for multi-site \miniweb{} tasks, one line per site; we call this suffix \texttt{\{context\}} below.

\begin{promptbox}{Planner: task}
You are interacting with a web application at {start_url}. Your task: {task}

Apps for this task (open them at these exact addresses):
- {app_name_1}: {base_url}/sites/{site_1}/
- {app_name_2}: {base_url}/sites/{site_2}/
\end{promptbox}

The planner cannot click, type or navigate. Of browser-use's own actions it keeps only \texttt{scroll} and \texttt{done}, and it acts on the site through \texttt{delegate}. The box gives the delegate action's description and its parameters with their descriptions; the model reads them inside the action JSON schema.

\begin{promptbox}{Planner: delegate action}
delegate: Hand the next step of the task to the specialist of its family. The specialist does only this step
on the current page and reports what it did and saw.
  family ("Navigation" | "Text entry" | "Discrete selection" | "Form transaction" | "Drag & gesture" |
    "Out-of-page I/O" | "Reasoning base"): The kind of step: picks the specialist that does it
  instruction (string): One step, said exactly: the control to use (as named on the page) and the names or
    values from the task
\end{promptbox}

Each delegation returns one of three texts to the planner as its action result. In the first, \texttt{\{family\}} and \texttt{\{model\}} name the specialist and \texttt{\{n\}} counts its steps. \texttt{\{status\}} is \texttt{finished} or \texttt{stopped at its step limit without finishing}, and \texttt{\{report\}} is the specialist's final message (or its last error) cut at 700 characters. \texttt{\{left\}} is what remains of the shared budget of 40 site actions. \texttt{\{page\_check\}} is \texttt{the page went from \{url\} to \{url2\}}, \texttt{same URL (\{url\}), title changed}, or \texttt{same URL (\{url\}), same title}. The second text answers a delegation identical to the previous one while the URL is unchanged. The third answers the first delegation after the budget is spent; a second such delegation ends the episode.

\begin{promptbox}{Planner: delegate results (v6)}
{family} specialist ({model}), {n} steps, {status}. It reported: {report} [specialist steps left: {left}]
[check: {page_check}]

You just delegated this exact step and the page has not changed since. Choose a different step (another
control or another way), or call done.

No specialist steps are left. Call done now with your best answer from what has been found.
\end{promptbox}

\subsection{\miniwebagent{}: executor}\label{app:prompt-agent-exec}

Each delegation starts a fresh browser-use agent on the same browser (\texttt{Team.run\_step} in \path{webmix/planner_agent.py}); the prompts call it a specialist. In \miniwebagent{}, it is run by the pooled adapter on \mbox{Qwen3.5-4B}, and a reasoning step by the base model (in the family-specialist variant, by the family's LoRA specialist), for at most 15 steps (fewer when less of the budget remains). Its system prompt is browser-use's flash-mode system prompt, unchanged and without extension, followed by the JSON schema of its actions. In \texttt{v6} the executor does not see the overall task. Its task is the planner's instruction in the same wrapper, where \texttt{\{url\}} is the current page and \texttt{\{context\}} is the planner's suffix above.

\begin{promptbox}{Executor: task (v6)}
You are interacting with a web application at {url}. Your task: {instruction}
Do only this, then call done and report what you did and what the page shows.{context}
\end{promptbox}

{\raggedright The specialist's action set (\path{webmix/harness.py}, \texttt{make\_tools}) is the same as for the single-agent baseline and for the replayed training rows. It is browser-use's default set without \texttt{upload\_file}, \texttt{search}, \texttt{extract}, \texttt{save\_as\_pdf}, \texttt{search\_page}, \texttt{find\_elements}, \texttt{evaluate}, \texttt{write\_file}, \texttt{replace\_file} and \texttt{read\_file}, with clicking by coordinates enabled. Three actions are added, and the coordinate parameters of browser-use's \texttt{click} get new descriptions. These descriptions, given below, are the only text the harness adds to what the model reads. The added actions return \texttt{Dragged from (\{start\_x\}, \{start\_y\}) to (\{end\_x\}, \{end\_y\})}, \texttt{Drew \{n\} stroke(s)} and \texttt{Macro step finished.}\par}

\begin{promptbox}{Executor: added and changed action descriptions}
drag: Drag with the mouse: press at (start_x, start_y), move, release at (end_x, end_y), all on a 0-1000 scale
of the screenshot. For sliders, drag-and-drop, reordering, panning a map.
  start_x: Where the drag starts (the thing being dragged): x on a 0-1000 scale of the screenshot width
  start_y: Where the drag starts: y on a 0-1000 scale of the screenshot height
  end_x: Where the drag is released: x on a 0-1000 scale of the screenshot width
  end_y: Where the drag is released: y on a 0-1000 scale of the screenshot height
draw: Draw with the mouse on a canvas or signature pad: each stroke is a list of [x, y] points (0-1000 scale
of the screenshot) drawn with the button held.
  strokes: Pen strokes, each a list of [x, y] points (0-1000 scale of the screenshot) drawn with the mouse
    button held
macro_done: Mark the current step of the task (one sub-goal, e.g. a form submitted, a filter applied, an item
found) as finished, before starting the next one or calling done.
click:
  coordinate_x: x on a 0-1000 scale of the screenshot width (0 = left edge, 1000 = right edge)
  coordinate_y: y on a 0-1000 scale of the screenshot height (0 = top edge, 1000 = bottom edge)
\end{promptbox}

\end{document}